\documentclass[letterpaper]{article} 
\usepackage[preprint]{aaai2027}
\usepackage{times} 
\usepackage{helvet} 
\usepackage{courier} 
\usepackage[hyphens]{url} 
\usepackage{graphicx} 
\usepackage{natbib} 
\usepackage{caption} 
\usepackage{algorithm}
\usepackage{algorithmic}
\usepackage{amsmath}
\usepackage{amssymb}
\usepackage{amsthm}
\usepackage{booktabs}
\usepackage{newunicodechar}

\newcommand{\system}{\textsc{MECA}}

\theoremstyle{definition}
\newtheorem*{problem}{Problem}

\newunicodechar{²}{\ensuremath{^2}}
\newunicodechar{·}{\ensuremath{\cdot}}
\newunicodechar{¹}{\ensuremath{^1}}
\newunicodechar{½}{\ensuremath{\tfrac12}}
\newunicodechar{×}{\ensuremath{\times}}
\newunicodechar{̂}{\llap{\ensuremath{\hat{\phantom{x}}}}}
\newunicodechar{̄}{\llap{\ensuremath{\bar{\phantom{x}}}}}
\newunicodechar{Α}{\ensuremath{\mathrm A}}
\newunicodechar{Γ}{\ensuremath{\Gamma}}
\newunicodechar{Δ}{\ensuremath{\Delta}}
\newunicodechar{Θ}{\ensuremath{\Theta}}
\newunicodechar{Λ}{\ensuremath{\Lambda}}
\newunicodechar{Π}{\ensuremath{\Pi}}
\newunicodechar{Σ}{\ensuremath{\Sigma}}
\newunicodechar{Φ}{\ensuremath{\Phi}}
\newunicodechar{Ψ}{\ensuremath{\Psi}}
\newunicodechar{Ω}{\ensuremath{\Omega}}
\newunicodechar{α}{\ensuremath{\alpha}}
\newunicodechar{β}{\ensuremath{\beta}}
\newunicodechar{γ}{\ensuremath{\gamma}}
\newunicodechar{δ}{\ensuremath{\delta}}
\newunicodechar{ε}{\ensuremath{\varepsilon}}
\newunicodechar{η}{\ensuremath{\eta}}
\newunicodechar{θ}{\ensuremath{\theta}}
\newunicodechar{κ}{\ensuremath{\kappa}}
\newunicodechar{λ}{\ensuremath{\lambda}}
\newunicodechar{μ}{\ensuremath{\mu}}
\newunicodechar{ν}{\ensuremath{\nu}}
\newunicodechar{ξ}{\ensuremath{\xi}}
\newunicodechar{π}{\ensuremath{\pi}}
\newunicodechar{ρ}{\ensuremath{\rho}}
\newunicodechar{σ}{\ensuremath{\sigma}}
\newunicodechar{τ}{\ensuremath{\tau}}
\newunicodechar{φ}{\ensuremath{\phi}}
\newunicodechar{χ}{\ensuremath{\chi}}
\newunicodechar{ψ}{\ensuremath{\psi}}
\newunicodechar{ω}{\ensuremath{\omega}}
\newunicodechar{ᵀ}{\ensuremath{^{\mathsf T}}}
\newunicodechar{‖}{\ensuremath{\lVert}}
\newunicodechar{⁺}{\ensuremath{^+}}
\newunicodechar{⁻}{\ensuremath{^-}}
\newunicodechar{ⁿ}{\ensuremath{^n}}
\newunicodechar{₀}{\ensuremath{_0}}
\newunicodechar{₁}{\ensuremath{_1}}
\newunicodechar{₂}{\ensuremath{_2}}
\newunicodechar{₃}{\ensuremath{_3}}
\newunicodechar{₄}{\ensuremath{_4}}
\newunicodechar{₅}{\ensuremath{_5}}
\newunicodechar{ℕ}{\ensuremath{\mathbb N}}
\newunicodechar{ℝ}{\ensuremath{\mathbb R}}
\newunicodechar{→}{\ensuremath{\to}}
\newunicodechar{↓}{\ensuremath{\downarrow}}
\newunicodechar{↦}{\ensuremath{\mapsto}}
\newunicodechar{⇀}{\ensuremath{\rightharpoonup}}
\newunicodechar{⇉}{\ensuremath{\rightrightarrows}}
\newunicodechar{⇒}{\ensuremath{\Rightarrow}}
\newunicodechar{∂}{\ensuremath{\partial}}
\newunicodechar{∇}{\ensuremath{\nabla}}
\newunicodechar{∈}{\ensuremath{\in}}
\newunicodechar{∑}{\ensuremath{\sum}}
\newunicodechar{−}{\ensuremath{-}}
\newunicodechar{√}{\ensuremath{\sqrt{\vphantom{x}}}}
\newunicodechar{∞}{\ensuremath{\infty}}
\newunicodechar{∥}{\ensuremath{\lVert}}
\newunicodechar{∧}{\ensuremath{\wedge}}
\newunicodechar{∩}{\ensuremath{\cap}}
\newunicodechar{∪}{\ensuremath{\cup}}
\newunicodechar{∫}{\ensuremath{\int}}
\newunicodechar{≠}{\ensuremath{\ne}}
\newunicodechar{≤}{\ensuremath{\leq}}
\newunicodechar{≥}{\ensuremath{\geq}}
\newunicodechar{⊂}{\ensuremath{\subset}}
\newunicodechar{⊆}{\ensuremath{\subseteq}}
\newunicodechar{⊕}{\ensuremath{\oplus}}
\newunicodechar{⊗}{\ensuremath{\otimes}}
\newunicodechar{★}{\ensuremath{\star}}
\newunicodechar{⟨}{\ensuremath{\langle}}
\newunicodechar{⟩}{\ensuremath{\rangle}}

\makeatletter
\def\input@path{{appendix/problems/appendix/}{appendix/problems/appendix/entries/}}
\makeatother

\title{MECA: A Mechanism-Centered Agent for Constructing Well-Specified and Valuable
Mathematical Conjectures}

\author{
Wentao Long\textsuperscript{\rm 1},
Yunfei Zhang\textsuperscript{\rm 2},
Chenyi Li\textsuperscript{\rm 3},
Zaiwen Wen\textsuperscript{\rm 3}\thanks{Corresponding author: \texttt{wenzw@pku.edu.cn}.}
}

\affiliations{
\textsuperscript{\rm 1}Fudan University
\textsuperscript{\rm 2}Qingdao University
\textsuperscript{\rm 3}Peking University
}
\begin{document}

\maketitle

\begin{abstract}
Automatically constructing well-specified and valuable mathematical conjectures
remains a central challenge in AI-assisted mathematical discovery. Many existing
open problems and conjectures are often too broad, underspecified, or
difficult to connect to plausible proof or refutation strategies. We view a
mathematical mechanism as a structure or reasoning principle that connects the
assumptions of a candidate problem to its target conclusion, such as an inequality, invariant, decomposition, or reduction to an intermediate claim. We present
MECA (\emph{ME}chanism-centered \emph{C}onjecture \emph{A}gent), a multi-agent
framework that constructs conjectures by jointly developing candidate
statements and their supporting mechanisms.
Explorer agents propose mechanisms, test how they apply, and revise the
candidate conjecture accordingly, while critic agents assess their mathematical
validity and research value. Their feedback guides changes to the assumptions,
scope, and conclusion. Through this process, MECA transforms broad research
directions into precise conjectures with substantive mathematical support while
retaining a clearly identified unresolved core.
We evaluate MECA in two complementary settings. First, we compare it with a
generate-and-revise baseline on reconstructing preselected target-paper
conclusions from target-conditioned but article-blind source materials. Second,
we construct 100 semi-open problems from literature-derived seeds and existing
open problems and evaluate them through independent proof and refutation
attempts by automated provers. Our results indicate that mechanism-centered
refinement produces well-specified and research-worthy conjectures that remain
challenging for current automated provers.
\end{abstract}

\section{Introduction}
\label{sec:introduction}

Recent work in AI for mathematics has largely focused on solving or
formalizing targets supplied by humans, benchmarks, or existing mathematical
texts. Large language models solve competition and quantitative reasoning
problems \citep{hendrycks2021math,lewkowycz2022minerva}. Neural and
LLM-based theorem provers construct machine-checkable proofs through tactic
generation, premise selection, and proof search
\citep{polu2020generative,han2022proof,lample2022hypertree,
jiang2022thor,jiang2023draft,yang2023leandojo,
xin2024deepseekprover,li2026optprover}. Autoformalization systems translate
informal mathematical statements and larger bodies of mathematical text into
proof-assistant languages
\citep{wu2022autoformalization,azerbayev2023proofnet,
ying2024leanworkbook,poiroux2024improving,
li2025sita,wang2026m2f}. Some formalization tasks also require an implicit
conjecturing step \citep{sivakumar2025conjecturing}. Nevertheless, the
intended mathematical targets are typically determined by the input in
advance.

Automated conjecturing has been studied through heuristic search, testing, and
numerical experimentation. Early systems invented concepts and proposed
conjectures \citep{lenat1976am,colton2002automated}. Later methods generated
equations, auxiliary lemmas, inequalities, numerical identities, or symbolic
expressions from examples and observations
\citep{claessen2010quickspec,johansson2014hipster,
davila2024txgraffiti,raayoni2021ramanujan,udrescu2020aifeynman}. Recent
LLM-based systems generate formal conjectures and lemmas for prover training or
library development
\citep{dong2025stp,onda2025leanconjecturer,
alhessi2025lemmanaid,liu2026mathliblemma}. Other systems formulate discovery
as search under an automatic evaluator
\citep{fawzi2022alphatensor,romeraparedes2024funsearch,
trinh2024alphageometry,novikov2025alphaevolve,
georgiev2025mathematical}. These approaches typically search over a predefined class of mathematical
objects, such as equations, symbolic expressions, programs, or formal lemmas,
under a fixed theory context or evaluation objective.

Scientific ideation systems generate research proposals from literature or
connect idea generation with experimentation
\citep{baek2025researchagent,lu2024aiscientist,si2025novelideas}. Recent
mathematical systems predict theorem-like claims, generate research problems,
and assess creativity or interestingness
\citep{busbib2026compose,chen2026researchproblems,
chen2026moonshine,chen2025deepmathcreative,mishra2026interest}. Mathematical
research agents also address supplied open problems through decomposition,
retrieval, criticism, and verification
\citep{an2026qed,feng2026aletheia,cao2026mechmath,
schmitt2026proofcouncil}. However, these systems generally either operate
within predefined search spaces or begin from an already selected research
target. A general method is still needed to construct bounded conjectures from
broad mathematical sources, jointly refine them with supporting mechanisms,
exclude candidates already resolved during construction, and retain a precise
unresolved core.

The remaining challenge is to construct research-worthy conjectures from broad
and partially specified mathematical sources. Such conjectures must be precise,
supported by plausible proof or refutation routes, and valuable under either
outcome, while retaining a genuine unresolved core. This requires the candidate
statement and its supporting reasoning to be refined jointly: fixing the
statement too early may preserve an ill-posed target, whereas adapting it too
freely to available arguments may make it trivial or already resolved.


We organize this refinement around mathematical \emph{mechanisms}. A mechanism is a reusable mode of reasoning, typically grounded in the structural properties of the problem, that links a set of assumptions to a corresponding conclusion. Although such a mechanism may initially arise in the analysis of a particular theorem, it can often be extended to related objects, alternative assumptions. We characterize each mechanism in terms of the conditions required for its validity, the portion of the candidate statement that it supports, and the circumstances under which it fails. This characterization clarifies the progress already established, isolates the unresolved obstruction, and may further indicate whether the remaining difficulty calls for an additional argument or a revision of the statement itself. Accordingly, the identification of a valid mechanism may constitute substantive mathematical progress even when it does not lead to a complete resolution of the conjecture.

Motivated by this view, we introduce \system{}
(\emph{ME}chanism-centered \emph{C}onjecture \emph{A}gent), a multi-agent
framework that treats candidate statements and mechanisms as coupled search
objects. Explorer agents retrieve, transfer, and adapt proof and refutation
mechanisms, while critic agents audit whether their assumptions, conclusions,
and claimed applicability are valid. Candidates lacking meaningful support are
revised or rejected, and candidates already resolved by their mechanisms are
excluded. Each accepted conjecture is paired with a mechanism bundle recording
the supported steps and the explicit unresolved obligation, yielding a bounded
problem suitable for independent proof or refutation.

We evaluate \system{} in two settings. First, we compare it with a
generate-and-revise baseline on recovering paper-level conclusions from
target-conditioned but article-blind source materials. Second, we construct 100 semi-open problems
from literature-derived seeds and existing open problems. We evaluate their
quality and difficulty through blinded assessment and independent proof and
refutation attempts. We further compare prover performance with and without
the mechanism bundle.

Our main contributions are as follows:
\begin{itemize}

\item We formulate conjecture construction as the joint refinement of a
candidate statement and its supporting mechanisms, guided by audited
partial support and explicit unresolved obstacles.

\item We introduce \system{}, a mechanism-centered multi-agent framework
for constructing precise conjectures from mathematical
literature and broad open problems.

\item We contribute 100 prover-facing open problems with explicit assumptions
and mathematical motivation, exposing limitations of automated proof and
refutation systems.

\end{itemize}

\section{Mechanism-Centered Conjecture Construction}
\label{sec:formulation}

We formulate conjecture construction as the joint refinement of a candidate
statement and its supporting mathematical mechanisms. Rather than generating
an isolated statement and later attaching a proof sketch, the candidate's
assumptions, scope, and conclusion are shaped by audited mathematical progress
with obstacles that remain unresolved.

At iteration \(t\), the construction state is
\[
\mathcal C_t=(A,p_t,G_t),
\]
where \(A\) fixes the admissible research boundary, \(p_t\) is the current
candidate statement, and \(G_t\) is its audited mechanism graph. The anchor
remains fixed, while the candidate and graph are refined jointly,
\(p_t \longleftrightarrow G_t\).

\begin{figure*}[t]
    \centering
    \includegraphics[width=0.98\textwidth]{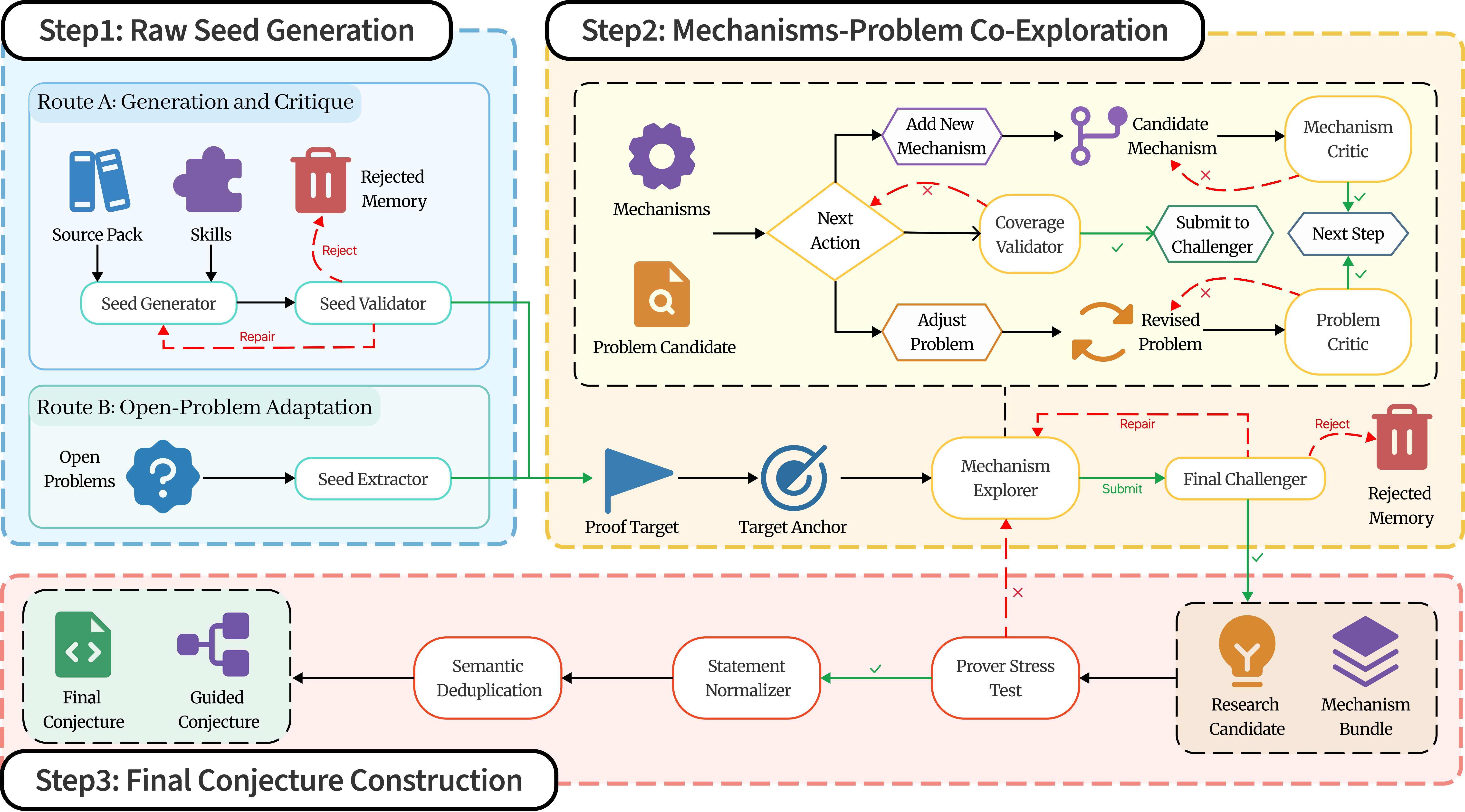}
    \caption{Overview of \system{}.}
    \label{fig:pipeline-overview}
\end{figure*}

A reviewed research direction determines a target anchor
\(A=(o,\mathcal P_A,\mathcal B_A,\mathcal F_A)\),
where \(o\) is the central mathematical object, \(\mathcal P_A\) contains the
admissible assumptions, \(\mathcal B_A\) fixes the source and research
boundary, and \(\mathcal F_A\) excludes simplifications that would remove the
intended difficulty. The candidate \(p_t\) must satisfy
\(\operatorname{Asm}(p_t)\subseteq\mathcal P_A\).
It may be strengthened, weakened, localized, or split, but it may not change
the central object, introduce assumptions outside the anchor, or collapse the
task into a forbidden easier problem.

\subsection{Mechanisms}
\label{subsec:mechanisms}

A \emph{mechanism} is a mathematical argument or structural principle that
establishes a concrete step toward proving or refuting the current candidate.
Examples include inequalities, invariants, localization arguments,
decompositions, stability principles, extremal constructions, and reductions
to intermediate claims.

We represent a mechanism \(m\) by
\[
m=
\bigl(
\operatorname{Req}(m),
\operatorname{Prov}(m),
\operatorname{Gap}(m)
\bigr),
\]
where \(\operatorname{Req}(m)\) contains the assumptions or previously
established facts required by the argument.
\(\operatorname{Prov}(m)\) contains the consequences established by it, and
\(\operatorname{Gap}(m)\) records the nonroutine conditions or downstream
obligations that remain unresolved in the current mechanism graph. Thus, a
condition may appear in \(\operatorname{Req}(m)\) because the argument uses it
and in \(\operatorname{Gap}(m)\) when it has not yet been derived from the
anchor assumptions or an upstream mechanism.

A mechanism is therefore more than the name of a proof technique. It must
contain an independently reviewable derivation, construction, or
counterexample argument. It is retained as an audited conditional step only
when its claimed consequences follow under the stated requirements and its use
for the current candidate is valid and non-circular. Any requirement not yet
available in the current graph must remain explicit in its gap and be supplied
by an upstream mechanism before composition.

\paragraph{Example.}
Consider two Bregman proximal updates generated by kernels \(h\) and
\(\widetilde h\), with
\[
x_h^{+}
=
\arg\min_y
\left\{
\langle g,y-x\rangle+r(y)
+\frac{1}{\eta}D_h(y,x)
\right\},
\]
and \(x_{\widetilde h}^{+}\) defined analogously. The goal is to show that if
the two kernels have similar local curvature, then the corresponding proximal
updates are also close. More precisely, suppose that \(r\) is convex and that,
for every \(z\in\mathcal R\),
\[
\begin{aligned}
\nabla^2 h(z)&\succeq \mu I,\\
\|\nabla^2 h(z)-\nabla^2\widetilde h(z)\|&\leq \varepsilon.
\end{aligned}
\]
We would like to derive a bound on
\(\|x_h^{+}-x_{\widetilde h}^{+}\|\).

The relevant mechanism is a comparison of the optimality conditions for the
two proximal subproblems. Let \(d=x_h^{+}-x_{\widetilde h}^{+}\).
Subtracting the two optimality conditions, using the monotonicity of
\(\partial r\), and applying the curvature assumptions gives
\[
\mu\|d\|^2
\leq
\varepsilon
\|x_{\widetilde h}^{+}-x\|
\|d\|,
\]
provided that
\(x,x_h^{+},x_{\widetilde h}^{+}\in\mathcal R\). Consequently,
\[
\|x_h^{+}-x_{\widetilde h}^{+}\|
\leq
\frac{\varepsilon}{\mu}
\|x_{\widetilde h}^{+}-x\|.
\]

This comparison argument is the mechanism \(m_{\mathrm{cmp}}\). Its role is to
turn two local properties of the kernels---strong curvature of \(h\) and
closeness between the Hessians of \(h\) and \(\widetilde h\)---into a
quantitative stability bound for their proximal updates. Its audited
requirements and consequence are
\[
\operatorname{Req}(m_{\mathrm{cmp}})
=
\left\{
\begin{array}{l}
r\text{ is convex},\\
\nabla^2 h(z)\succeq\mu I
    \text{ for }z\in\mathcal R,\\
\|\nabla^2 h(z)-\nabla^2\widetilde h(z)\|
    \leq\varepsilon
    \text{ for }z\in\mathcal R,\\
x,x_h^{+},x_{\widetilde h}^{+}\in\mathcal R
\end{array}
\right\}
\]
and
\[
\operatorname{Prov}(m_{\mathrm{cmp}})
=
\left\{
\|x_h^{+}-x_{\widetilde h}^{+}\|
\leq
\frac{\varepsilon}{\mu}
\|x_{\widetilde h}^{+}-x\|
\right\}.
\]

The argument also reveals exactly what remains to be proved. The curvature
assumptions are available only inside \(\mathcal R\), so the comparison is
valid only after establishing that both updated points remain in this region.
This unresolved localization step is recorded as
\[
\operatorname{Gap}(m_{\mathrm{cmp}})
=
\left\{
x_h^{+},x_{\widetilde h}^{+}\in\mathcal R
\right\}.
\]
Thus, a mechanism is not merely the vague suggestion to ``compare the two
updates.'' It is a concrete conditional argument that identifies the
assumptions it uses, the mathematical conclusion it establishes, and the
additional fact that must still be proved before the argument can be applied.

\subsection{Mechanism Graphs}
\label{subsec:mechanism-graphs}

Accepted mechanisms are organized into a directed graph \(G_t=(V_t,E_t)\),
where an edge \(m_i\longrightarrow m_j\) means that a conclusion of \(m_i\)
supplies a condition required by \(m_j\), namely
\(\operatorname{Prov}(m_i)\cap\operatorname{Req}(m_j)\neq\varnothing\).

In the Bregman proximal-update example above, the comparison mechanism
\(m_{\mathrm{cmp}}\) requires
\(x_h^{+},x_{\widetilde h}^{+}\in\mathcal R\).
A localization mechanism \(m_{\mathrm{loc}}\) may establish this condition,
for example from a sufficiently small step size and a bound on the update
magnitude, so that
\(\operatorname{Prov}(m_{\mathrm{loc}})
\supseteq\{x_h^{+},x_{\widetilde h}^{+}\in\mathcal R\}\).
This produces the dependency
\(m_{\mathrm{loc}}\longrightarrow m_{\mathrm{cmp}}\).
The localization mechanism first keeps both updates inside the region where
the curvature assumptions hold, after which the comparison mechanism derives
the stability bound. Additional mechanisms may establish the required
step-length bound, extend the valid region, or construct a counterexample
outside it.

The graph may contain chains of arguments, merges of several supporting
mechanisms, and parallel proof or refutation branches. It therefore records
how mathematical arguments compose, rather than collecting an unordered set
of proof ideas.
Retained mechanisms must be valid, applicable, mutually compatible, and
non-circular. Restating the target, assuming its central difficulty, or
delegating a step to an unspecified argument does not constitute valid support.

\begin{figure}[!htbp]
    \centering
    \includegraphics[width=\columnwidth]{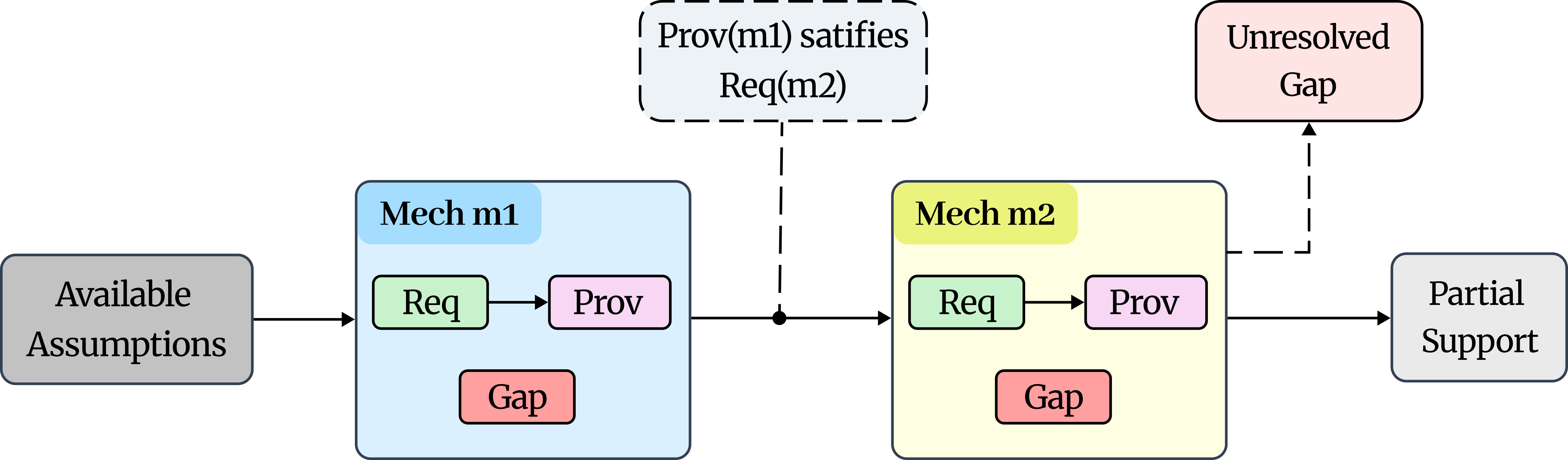}
    \caption{Mechanism illustration.}
    \label{fig:mechanism-illustration}
\end{figure}

The candidate and graph are refined jointly: the candidate determines which
mechanisms are relevant, while the graph may reveal missing assumptions, an
overly strong conclusion, or a useful decomposition. Substantial revisions
therefore require the retained mechanisms to be rechecked.
Figure~\ref{fig:mechanism-illustration} illustrates how mechanisms compose
through requirement--consequence interfaces to construct partial support for the problem while leaving the unresolved gaps
explicit.

Through joint refinement, the candidate and mechanism graph form a mutually
consistent proof problem. Its assumptions must support concrete mechanisms,
and the scope of its conclusion must remain consistent with what those
mechanisms establish. The graph thereby explains both the mathematical support
for the statement and the specific obstacle that remains unresolved, without
certifying that the statement is true or genuinely open.

\section{The MECA Framework}
\label{sec:framework}

Figure~\ref{fig:pipeline-overview} summarizes how \system{} operationalizes the
mechanism-centered formulation in Section~\ref{sec:formulation}. The
formulation defines the mathematical objects and validity conditions, whereas
the framework specifies how source-grounded candidates are constructed,
reviewed, refined, and filtered.

The pipeline consists of three stages: source-grounded seed construction,
mechanism--problem co-exploration, and prover-facing problem construction. It
retains candidates with audited nontrivial support and explicit unresolved
cores, while rejecting unsupported speculation and removing possible
resolutions from the open-problem path. Supplementary Sections A--C give the
implementation-level action sets, record schemas,
receipt semantics, and stopping rules.

\subsection{Source-grounded seed construction}
\label{subsec:seed-construction}

\system{} begins with either structured mathematical literature or a manually
curated broad open problem. In the literature-driven route, retrieved records
provide definitions, established results, limitations, and source locators. In
the open-problem route, the supplied description provides the initial direction,
and supporting literature clarifies known regimes and barriers.

\textbf{The seed generator} converts these materials into a seed recording the
closest known result, a proposed source-relative advance, a preliminary
prove-or-refute target, plausible counterexample regimes, and an anticipated
mathematical bottleneck. Claims inherited from the sources retain their
locators, whereas new claims are marked as hypotheses.

\textbf{The seed validator} reviews each seed for research value and mathematical
validity. The research-value review rejects restatements, direct corollaries,
and superficial parameter variations, while the mathematical review checks
definitions, assumptions, quantifiers, boundary cases, and simple degenerate
examples. A local defect may be returned for bounded repair without changing
the intended research direction. Invalid, resolved, or insufficiently
meaningful seeds are rejected. Only validated seeds enter mechanism--problem
co-exploration. Supplementary Section A specifies the seed
schema, repair contracts, and replacement rules.

Before Stage~2, reviewed source records are consolidated into a frozen source
pack, and the research boundary is encoded in the target anchor. Later agents
may use these records and computational tools but cannot enlarge the source or
assumption boundary.

\subsection{Mechanism--problem co-exploration}
\label{subsec:co-refinement}

For each validated seed, \system{} instantiates and freezes the target anchor
\(A\) defined in Section~\ref{sec:formulation}. The anchor fixes the
research object, admissible assumptions, source boundary, and forbidden
trivializations. Candidate revisions may change the local statement, but they
must remain semantically consistent with this anchor.

\paragraph{Candidate initialization and revision.}

The Mechanism--problem explorer initializes the candidate \(p\) and jointly
refines it with the mechanism graph. At each iteration, it may propose
a proof or refutation mechanism, revise or split the statement in response to
audited evidence, or flag a possible resolution. The candidate critic
independently reviews each candidate version for well-posedness, consistency
with the anchor, nontriviality, and alignment among its assumptions, proof and
refutation targets, and current mathematical support. Its feedback returns to
the explorer for revision or splitting, while the validity of individual
mechanisms is assessed separately by the mechanism critic. Every revision must
remain within the frozen anchor \(A\), and mechanisms accepted for an earlier
version are rechecked after a substantive change. The complete action set and
revision rules are given in Supplementary Section B.

\paragraph{Mechanism exploration.}

The explorer searches forward from the assumptions, backward from the target,
or between accepted mechanisms for an argument establishing part of the proof
or refutation route. Each proposal states its requirements, consequences,
applicability conditions, and first unresolved nonroutine step. Explicit failed
attempts may guide later exploration, but they are not treated as evidence that
an entire class of arguments must fail.

\paragraph{Mechanism audit.}

The mechanism critic independently checks the central derivation, the claimed
requirement-consequence interface, compatibility with the anchor and current
candidate, boundary cases, and freedom from circular reasoning. Accepted
mechanisms enter the graph, repairable proposals are returned for revision, and
invalid ones are excluded. Candidate defects or complete proofs
and counterexamples are escalated to the corresponding revision or resolution
path. These audit decisions authorize composition but do not constitute formal
proof certificates. Their exact contracts are given in
Supplementary Section B.

\paragraph{Compositional diagnosis.}

Individually accepted mechanisms may still fail to form a coherent route toward
the candidate. The coverage judge therefore evaluates declared composition
steps in a selected mechanism subgraph \(G'\subseteq G\) and a declared dependency order
\[
\pi=(\pi_1,\ldots,\pi_k).
\]
Let \(\mathcal A_0\) denote the capabilities supplied by the candidate
assumptions. A mechanism may be applied only when
\[
\operatorname{Req}(m_{\pi_i})
\subseteq
\mathcal A_{i-1},
\]
after which
\[
\mathcal A_i
=
\mathcal A_{i-1}
\cup
\operatorname{Prov}(m_{\pi_i}).
\]
This test determines whether the declared interfaces compose. It does not
independently recertify the underlying mathematical arguments.

Accepted mechanisms need not merge into a single terminal node. Independent
mechanisms may support different assumptions, intermediate reductions, positive
targets, or refutation targets. The relevant question is whether their declared
interfaces are jointly compatible and whether they establish a genuine part of
the route toward the candidate.

\paragraph{Coverage decisions and routing.}

We distinguish three mathematical support decisions:
\begin{itemize}
    \item \textsc{InsufficientSupport}: the accepted mechanisms do not yet
    establish a meaningful nonroutine part of the proof or refutation route.
    \item \textsc{SubstantivePartial}: the accepted mechanisms compose and
    establish a genuine nonroutine part of the route, while one to three precise
    unresolved cores remain.
    \item \textsc{FullCoverage}: the accepted mechanisms appear to cover the
    complete positive or refutation target.
\end{itemize}

\textsc{InsufficientSupport} returns the candidate to exploration with the
missing capability identified. \textsc{SubstantivePartial} passes a candidate
with meaningful support and explicit unresolved cores to the Final Challenger.
\textsc{FullCoverage} is isolated for independent verification rather than
presented as an unresolved conjecture. Detailed routing and stopping rules are
given in Supplementary Section B. Algorithm~\ref{alg:co-refinement} summarizes this co-exploration process.

\paragraph{Final challenge.}

To reduce confirmation bias, the Final Challenger receives the candidate,
anchor, seed, and source pack but not the mechanism graph or exploration
history. It checks completeness, consistency, source-relative value, and
nontriviality; attempts a direct proof or counterexample; and tests a concrete
or boundary case. Passing candidates proceed to prover-facing construction;
defects return for revision or rejection, and credible resolutions are
quarantined for verification. Passing the challenge
does not certify truth or historical openness. The exact audit contract is
given in Supplementary Section C.

\begin{algorithm}[t]
\caption{Mechanism--Problem Co-Exploration}
\label{alg:co-refinement}
\begin{algorithmic}[1]
\REQUIRE Validated seed \(s\), frozen anchor \(A\), source pack \(D_A\)
\ENSURE Accepted pair \((p,G)\), \textsc{Split}, \textsc{Reject}, or
\textsc{Quarantine}

\STATE \(p\leftarrow\textsc{InitializeCandidate}(s,A)\), \(G\leftarrow\varnothing\)

\WHILE{\textsc{BudgetAvailable}()}
    \STATE \((\rho,u)\leftarrow\textsc{CoverageJudge}(p,G,A)\)

    \IF{\(\rho=\textsc{SubstantivePartial}\)}
        \STATE \((\delta,v)\leftarrow\textsc{FinalChallenge}(s,A,p,D_A)\)
        \IF{\(\delta=\textsc{Accept}\)}
            \RETURN \((p,G)\)
        \ENDIF
        \STATE \((p,G,\omega)\leftarrow
        \textsc{ApplyRoute}(\delta,v,p,G,A)\)

    \ELSIF{\(\rho=\textsc{InsufficientSupport}\)}
        \STATE \(\alpha\leftarrow\textsc{SelectNextAction}(p,G,u)\)
        \IF{\(\alpha=\textsc{Explore}\)}
            \STATE \((\kappa,w)\leftarrow
            \textsc{ExploreCritique}(p,u,A,G,D_A)\)
            \STATE \((p,G,\omega)\leftarrow
            \textsc{ApplyRoute}(\kappa,w,p,G,A)\)
        \ELSE
            \STATE \((p,G,\omega)\leftarrow
            \textsc{ApplyRoute}(\alpha,u,p,G,A)\)
        \ENDIF

    \ELSE
        \STATE \((p,G,\omega)\leftarrow
        \textsc{ApplyRoute}(\rho,u,p,G,A)\)
    \ENDIF

    \IF{\(\omega\neq\textsc{Continue}\)}
        \RETURN \(\omega\)
    \ENDIF
\ENDWHILE

\RETURN \textsc{Reject}
\end{algorithmic}
\end{algorithm}

\subsection{Prover-facing problem construction}
\label{subsec:problem-construction}

An accepted candidate and selected mechanism subgraph are deterministically
projected into a self-contained problem
\[
q=\Theta(A,p,G').
\]
The projection preserves the accepted assumptions, quantifiers, scope, positive
target, and refutation target, while turning each unresolved core into a
concrete proof obligation. It may normalize notation or presentation, but
cannot add assumptions, weaken the target, or change the accepted mathematical
content.

The statement is evaluated without its mechanism bundle. A proof probe
attempts the positive target or a checkable first nonroutine step, while a
refutation probe searches for a counterexample satisfying all assumptions.
Neither receives the mechanism graph or exploration history.
Complete solutions and valid counterexamples are sent to independent
verification, whereas failure of either probe is recorded only as a difficulty
signal and does not establish falsity, truth, or openness. The exact probe and
hardening rules are given in Supplementary Section C.

Only presentation-preserving normalization is allowed after probing; any
semantic change returns to Stage~2 for renewed review. Surviving statements are
deduplicated by mathematical content rather than wording or mechanism route,
then admitted under the domain quotas and retained with their mechanism
bundles for downstream comparison.

\section{Experiments}
\label{sec:experiments}

Since no established protocol directly evaluates the full end-to-end process of mathematical conjecture construction, we assess MECA in two complementary settings: target-conditioned, article-blind recovery of hidden mathematical conclusions, which provides a clear and objective reference point, and open-ended conjecture construction followed by independent downstream verification.

\subsection{Target-Conditioned Conclusion Recovery}
\label{subsec:conclusion-recovery}

\paragraph{Setup.}
We select and freeze 20 post-2023 target papers, each reporting a substantive
mathematical advance. For each target, we build an article-blind pack from seven
pre-2024 papers covering the relevant setting, prior results, proof mechanisms,
and known limitations. These papers are converted into anonymized source cards
that preserve the mathematical evidence while removing identifying information
and target-revealing statements. A representative benchmark case is shown in
Supplementary Section F.1.

The target paper is used only to define the reconstruction task. During
generation, neither system can access the target paper, its conclusions, or any
post-2023 evidence. All outputs are frozen before evaluation, so performance
reflects reconstruction from historically available evidence rather than
retrieval or target-specific adaptation.

We compare MECA with a Draft--Audit--Revise baseline that receives the same
task and source pack. The baseline has no access to hidden targets,
target-derived evaluation records, or external retrieval during generation.
Both systems may return up to three theorem-level conjectures per case. For
each case, the baseline performs five audit steps and five revision steps
following its initial draft.

\paragraph{Evaluation.}
For each system and target paper, an independent evaluator selects the
generated conjecture with the strongest overall correspondence to the target
result. All evaluation dimensions are scored on this same conjecture, preventing
evidence from being combined across different outputs. The evaluator compares
the selected conjecture with the target paper using a 100-point rubric.
\begin{itemize}
    \item \textbf{Setting (20 points):} alignment of the mathematical problem,
    objects, and algorithms with the target result.
    \item \textbf{Assumptions (15 points):} consistency of the hypotheses and
    parameter regimes with those in the target paper.
    \item \textbf{Conclusion (30 points):} recovery of the central
    theorem-level claim.
    \item \textbf{Quantitative (20 points):} accuracy of rates, complexity
    bounds, constants, and thresholds.
    \item \textbf{Boundary (15 points):} coverage of sharpness, necessity,
    failure regimes, and relevant counterexamples.
\end{itemize}
The component scores are summed for each case and averaged over the benchmark.

\paragraph{Results.}
Table~\ref{tab:discovery-results} reports the target-alignment scores. MECA
improves the average score from 48.0 to 69.0 and outperforms the baseline on
every dimension. The largest gains occur in Conclusion (+6.0) and Quantitative
Accuracy (+5.5), indicating that mechanism-centered construction improves not
only broad problem alignment but also recovery of the central claim and its
precise strength. The gain in Boundary (+3.5) further suggests better recovery
of sharpness and failure regimes.

Because all five dimensions are scored on one selected conjecture, these gains
cannot be obtained by combining a well-matched setting from one output with a
strong conclusion from another. Conclusion and Quantitative Accuracy account
for 11.5 of the 21.0-point improvement, while the simultaneous gains in
Assumptions and Boundary indicate better recovery of the target's validity
regime rather than only its topic or theorem shape.

\begin{table}[!htbp]
    \centering
    \caption{Target-alignment scores on the target-conditioned article-blind
    benchmark.}
    \label{tab:discovery-results}
    \resizebox{\columnwidth}{!}{%
    \begin{tabular}{lrrrrrr}
        \toprule
        \textbf{Method} & \textbf{Setting} & \textbf{Assumptions} &
        \textbf{Conclusion} & \textbf{Quantitative} & \textbf{Boundary} &
        \textbf{Score} \\
        \midrule
        MECA       & \textbf{18.0} & \textbf{11.5} & \textbf{21.0} & \textbf{12.5} & \textbf{6.0} & \textbf{69.0} \\
        Baseline   & 15.5 &  8.0 & 15.0 &  7.0 & 2.5 & 48.0 \\
        \midrule
        Difference & +2.5 & +3.5 & +6.0 & +5.5 & +3.5 & +21.0 \\
        \bottomrule
    \end{tabular}%
    }
\end{table}

\subsection{Conjecture Generation and Verification}
\label{subsec:conjecture-generation}

\begin{figure}[t]
    \centering
    \includegraphics[width=0.9\columnwidth]
    {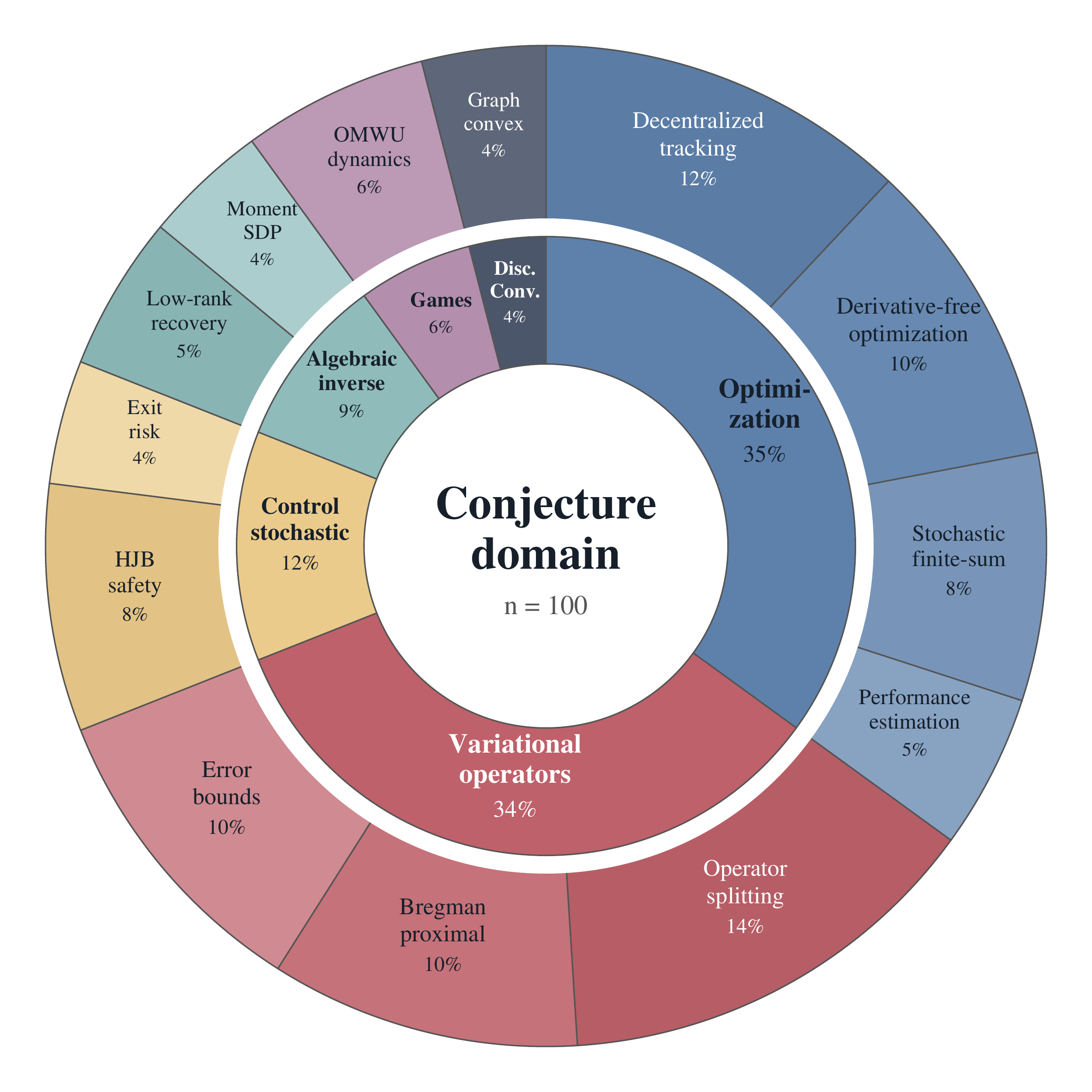}
    \caption{Distribution of the 100 generated conjectures across mathematical
    domains.}
    \label{fig:conjecture-domain-distribution}
\end{figure}

\paragraph{Setup.}
We evaluate 100 final conjectures constructed through the two routes described
in Section~\ref{sec:framework}. Route~A contributes 92 conjectures constructed
from literature collections retrieved from a private repository of mathematical
papers, while Route~B contributes 8 conjectures obtained by refining broad
descriptions from a privately curated collection of mathematical open problems.
Both collections were assembled before this study and frozen before the
experiments. The mathematical-domain distribution of the resulting conjectures
is shown in Figure~\ref{fig:conjecture-domain-distribution}.

Each conjecture is submitted in an isolated run to QED, an open-source
multi-agent system for research-level mathematical proof generation
\citep{an2026qed}, with a five-hour wall-clock limit. QED receives only the
final conjecture statement, without access to MECA's mechanisms, critic
reports, or intermediate reasoning. This statement-only evaluation tests
whether the conjectures are sufficiently precise, self-contained, and
actionable to serve as standalone theorem-proving tasks.

\paragraph{Evaluation.}
Prover success alone does not establish conjecture quality: easy or nearly
resolved statements may be proved quickly, while other statements may lack
necessary assumptions, contain ambiguities, or provide too little mathematical
foothold for proof or refutation to proceed. We therefore analyze both QED
outcomes and the causes of unresolved cases.

Each run is classified as \emph{proved}, \emph{refuted}, or
\emph{unresolved}, depending on whether QED produces a verified proof, a
verified counterexample or contradiction, or neither within five hours. Proof
and refutation are both treated as successful downstream adjudications because
they demonstrate that the final statement supports an independently checkable
mathematical conclusion.

\begin{figure}[!htbp]
    \centering
    \includegraphics[width=\columnwidth]{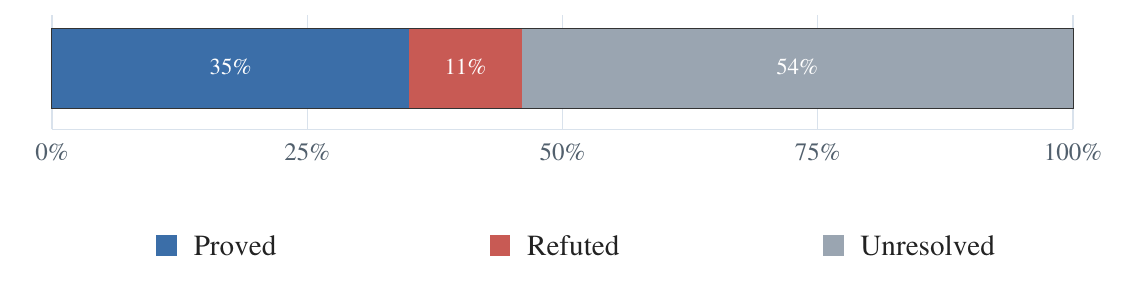}
    \caption{QED outcomes.}
    \label{fig:qed-outcomes}
\end{figure}

For the 54 unresolved conjectures, we assign a single primary failure cause:
\emph{technical proof difficulty}, \emph{missing assumptions},
\emph{insufficient foothold}, or \emph{ambiguous formulation}. Missing
assumptions indicate that an additional condition is needed to support the
conclusion. Ambiguous formulation indicates that the objects, quantifiers, or
scope are not uniquely specified. Insufficient foothold indicates that the
statement is precise but does not expose a sufficiently concrete route toward
proof or refutation.

Technical proof difficulty is assigned only when the statement is well posed,
its assumptions are adequate, and a plausible proof or refutation route can be
identified, but completing the required estimate, construction, reduction, or
case analysis exceeds QED's search capability or time budget. This category
therefore represents an unresolved mathematical obstacle rather than a defect
in the conjecture statement.

\begin{figure}[!htbp]
    \centering
    \includegraphics[width=\columnwidth]{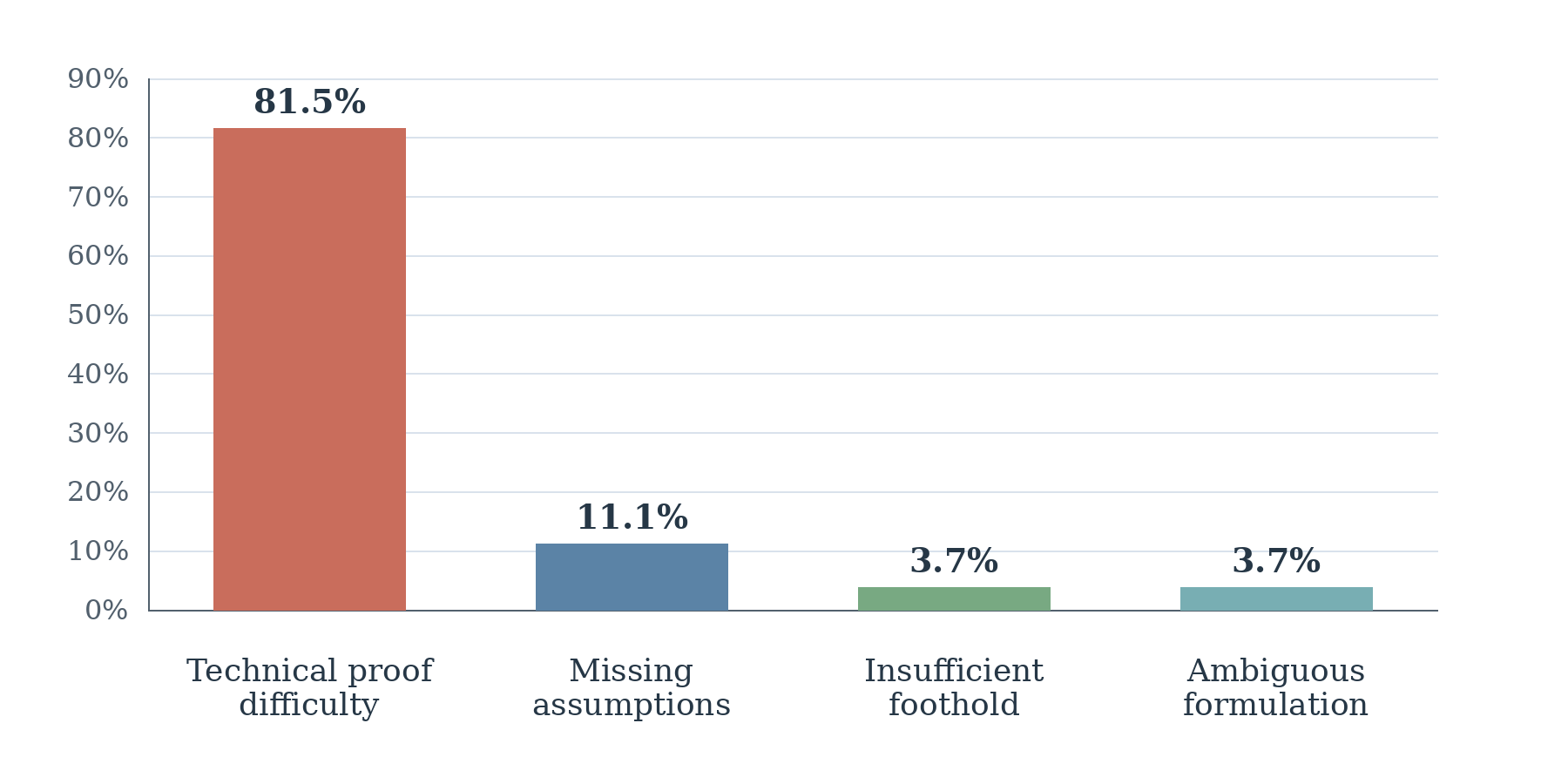}
    \caption{Causes of the 54 unresolved QED outcomes.}
    \label{fig:err-anal}
\end{figure}

\paragraph{Results.}
Given only the final conjecture statements, QED proves 35 conjectures
(\(35\%\)), refutes 11 (\(11\%\)), and leaves 54 (\(54\%\)) unresolved.
Thus, 46\% receive an independently verified positive or negative resolution.

Among the unresolved cases, 44 (\(81.5\%\)) are attributed to technical proof
difficulty, while 6 involve missing assumptions, 2 lack a sufficient
mathematical foothold, and 2 contain ambiguous formulations. Consequently, only
10\% of all generated conjectures exhibit an identifiable formulation or
proof-readiness defect. The remaining 90\% are either independently resolved or
remain well-posed and mathematically actionable despite exceeding QED's proof
budget. These results suggest that MECA produces conjectures that are usable as
standalone theorem-proving tasks while retaining substantive mathematical
difficulty.

The 11 verified refutations are not construction failures: they show that the
statements expose checkable negative branches and can support decisive
counterexamples. Likewise, the 44 technical-difficulty outcomes should not be
conflated with the 10 cases exhibiting formulation or proof-readiness defects:
they retain a bounded mathematical core after the statement, assumption, and
foothold checks pass. Supplementary Section F.3 illustrates this
contrast: one case retains a specific global certificate as its unresolved
core, whereas the other reaches a verified refutation after a local numerical
error is repaired.

\section{Conclusion}
\label{sec:conclusion}

MECA frames conjecture construction as the coordinated development of a
mathematical statement and the mechanisms that support or challenge it. Rather
than generating a conjecture first and attaching a proof sketch afterward, the
framework jointly refines the candidate, its assumptions, and its proof or
refutation routes through mechanism exploration, independent criticism, and
coverage diagnosis. This process turns broad research directions into
prover-facing conjectures with explicit objects, assumptions, and unresolved
cores. We evaluate MECA through article-blind conclusion recovery and isolated
proof and refutation probes.

\bibliography{aaai2027}

\clearpage
\appendix

\section{Details of Step 1: Source-Grounded Seed Construction}
\label{app:source-grounded-seeds}

This section expands Step~1 of the pipeline.  Its purpose is to construct a
small portfolio of research directions from a fixed body of mathematical
evidence.  A Raw Seed is deliberately less specific than a theorem statement:
it identifies a source-relative gap, the mathematical object to be controlled,
the assumptions that may be used, and concrete ways in which the direction can
fail.  It does not prescribe the mechanism that should solve the gap.

The phrase \emph{source grounded} refers to this evidence discipline.  It does
not impose a universal publication-year cutoff on conjecture generation.  In a
normal run, the evidence boundary may be a user-supplied library, a retrieval
snapshot, or a curated open-problem dossier.  Once materialized, that boundary
is frozen so that the closest-result comparison and later revisions are
reproducible.

\subsection{Input Routes and Frozen Evidence Boundary}

\system{} supports two source routes.  In the literature-driven route, the
source pack contains structured paper cards, definitions, established results,
known limitations, and source locators selected for a research focus.  In the
open-problem route, the pack begins from a manually curated broad problem and
adds records describing known regimes, barriers, and partial results.  Both
routes are converted into the same versioned source-pack representation before
seed generation begins.  Each source card records its identity, mathematical
setting, relevant definitions, established conclusions, known limitations, and
the evidence supporting those fields.  Thus, the routes differ in how evidence
is collected, but later agents receive the same categories of information.

The complete evidence snapshot, retrieval policy, prompt version, model
configuration, and source-card contents are assigned immutable identifiers.
Every generated seed and review decision records the identifiers of the inputs
on which it depends.  A resumed run is valid only when those identifiers still
match, so newly retrieved material cannot silently alter an earlier decision.
If retrieval reveals a closer result, the proposed gap may be narrowed or
rejected, but the change creates a new evidence snapshot and generation record;
it may not modify a frozen benchmark case or temporal cutoff in place.

For each proposed direction, source subtraction is performed before novelty is
claimed.  If the closest established result proves conclusion $X$ under
assumptions $A$, the seed must identify a bounded delta $\Delta$ beyond $X$:
for example, a removed assumption, a changed geometric regime, a sharper
complexity boundary, or a failure classification.  Terminology changes,
arbitrary extra parameters, and routine corollaries are not valid deltas.  A
formulation, unification, application bridge, diagnostic, or adaptivity claim
is retained only if resolving it changes a genuine mathematical boundary,
rate, obstruction, complexity, or theorem class.

\subsection{Portfolio Planning}

Several independent generations from one broad focus often produce cosmetic
variants.  We therefore construct the portfolio jointly before generating any
individual seed.  For a requested portfolio size $m$, the planner returns
exactly the slots $1,\ldots,m$.  Each slot specifies a source focus, route type,
novelty axis, and explicit nonoverlap constraints relative to the other slots.
It deliberately omits a theorem statement and proof mechanism, leaving those
choices to later stages.

The plan is accepted only when every slot is present, uniquely numbered, tied
to the frozen evidence snapshot, and mathematically distinguishable from the
others.  Each Seed Scout receives one accepted brief and must remain within its
source focus and novelty axis even if another direction appears easier.  If a
seed is rejected, its replacement must address the same brief rather than
migrating to a different slot.  The article-blind configuration uses $m=3$;
this is an experimental diversity budget rather than a restriction of the
general method.

\begin{center}
\small
\begin{tabular}{@{}p{0.28\linewidth}p{0.64\linewidth}@{}}
\toprule
Planner field & Meaning \\
\midrule
\texttt{slot} & Exact index in $1,\ldots,m$. \\
\texttt{source\_focus} & Source result, limitation, or object from which the direction must originate. \\
\texttt{route\_type} & One of the six source-relative routes in Table~\ref{tab:seed-routes}. \\
\texttt{novelty\_axis} & The intended boundary change after subtracting known results. \\
\texttt{nonoverlap} & Other slots whose objects, gaps, or contribution types must not be duplicated. \\
\bottomrule
\end{tabular}
\end{center}

\begin{table}[t]
\centering
\small
\setlength{\tabcolsep}{4pt}
\begin{tabular}{p{0.36\linewidth}p{0.54\linewidth}}
\toprule
Route & Source-relative question \\
\midrule
Limitation-driven & Can an explicit assumption or scope limitation be crossed? \\
Bottleneck-gap-driven & Can a missing comparison, obstruction, or nonroutine step be isolated? \\
Assumption-transfer-driven & Does a conclusion transfer across changed geometry, regularity, noise, or operator assumptions? \\
Complexity-gap-driven & Can a rate, dimension, sample, oracle, memory, or iteration gap be closed or shown sharp? \\
Geometry-mismatch-driven & What fails between local analytic conditions and global geometric structure? \\
Failure-case-driven & Can a boundary failure or counterexample family be characterized? \\
\bottomrule
\end{tabular}
\caption{Raw Seed route types.}
\label{tab:seed-routes}
\end{table}

\subsection{Raw Seed Record}

The Seed Scout returns one JSON object.  The record must be sufficiently
concrete for independent attack while leaving mechanism discovery to Step~2.
Its mathematical surface is summarized in Table~\ref{tab:raw-seed-fields}.

\begin{table*}[t]
\centering
\small
\setlength{\tabcolsep}{5pt}
\begin{tabular}{p{0.22\textwidth}p{0.68\textwidth}}
\toprule
Field & Required content \\
\midrule
\texttt{raw\_conjecture} & A preliminary research direction rather than a final quantified theorem or proof plan. \\
\texttt{source\_boundary} & Closest results, the minimal delta after subtraction, why the delta remains unresolved, and explicit overlap or source-coverage risks. \\
\texttt{claim\_shape} & Central mathematical object, controlled quantity, assumption-to-conclusion direction, and the complete list of authorized premises. \\
\texttt{attack\_surface} & Decisive bottleneck, a smallest informative toy case, and at least two concrete failure tests or counterexample knobs. \\
\bottomrule
\end{tabular}
\caption{Required mathematical content of a Raw Seed.}
\label{tab:raw-seed-fields}
\end{table*}

Every authorized premise is stored as an \texttt{\{id, statement\}} record.
An empty premise list is legal only for a genuinely unconditional direction.
The desired conclusion, a certificate, or the output of a hoped-for mechanism
may not be inserted as an assumption.  Likewise, the Seed Scout may name a
bottleneck and falsification tests, but it may not include a
\texttt{mechanism\_handle}, a bridge-lemma recipe, or a named proof mechanism.
This separation prevents Step~1 from appearing successful merely because it
assumed the principal object that Step~2 is intended to discover.

Before review, each Raw Seed is converted into a canonical versioned record.
This normalization may assign administrative identifiers and standardize list
or field representations, but it may not add an assumption, claim, example, or
other mathematical content.  A deterministic precheck then requires a
nonempty direction, a complete source boundary, unique premise identifiers, a
decisive bottleneck, an informative toy case, and at least two failure tests.
Passing this check establishes only that the record is complete and
well-formed; it does not establish correctness, novelty, or value.  The exact
revision identifier is fixed before model-based review, and every review
decision names the revision it evaluates.

\subsection{Independent Review, Repair, and Replacement}

The review stage uses two independent critics.  The mathematical critic audits
correctness and counterexample pressure, whereas the research critic audits
source-relative novelty, value, and readiness for mechanism exploration.  Each
dimension returns one of \texttt{survives\_deep\_attack},
\texttt{repair\_needed}, \texttt{reject\_needed}, or
\texttt{audit\_incomplete}, together with the attack actually attempted, the
first failed derivation or test, a minimal repair, confidence, and warnings to
carry forward.

The mathematical critic identifies the weakest implied claim and tries its
first nontrivial derivation.  It also instantiates an in-scope toy, endpoint,
boundary, or limiting family, verifies the assumptions, and calculates the
target observable.  A deterministic substance check downgrades an apparent pass to
\texttt{audit\_incomplete} unless the response contains a specified instance,
an assumption check, a target, and an explicit calculation.

The research critic retrieves or identifies the closest result inside the
allowed evidence boundary, subtracts routine consequences, and asks whether a
bounded nonroutine delta remains.  It rejects direct coverage, superficial
reparameterization, and low-value repackaging.  Readiness passes only when a
downstream Explorer can identify the object, quantity, implication,
bottleneck, and falsification test without already being handed a solution.
When the mathematical critic finds a decisive rejection, the research call may
be skipped; otherwise all four dimensions are retained so that one passing
dimension cannot conceal a blocker in another.

The review loop is as follows.
\begin{enumerate}
    \item Freeze the current Raw Seed as a uniquely identified revision.
    \item Run the mathematical and, when legal, research reviews.
    \item Promote only if every completed dimension survives and no blocker
    remains.
    \item If the defect is bounded, write one revision-bound repair contract
    containing all blocking findings and request a full replacement record
    under the same seed ID.
    \item If the direction is false, source-covered, duplicate, or low value,
    record the negative lesson and request a new realization within the same
    portfolio slot.
    \item Abandon the slot if its repair or replacement budget is exhausted;
    budget exhaustion never weakens the acceptance gate.
\end{enumerate}

A repair may narrow assumptions, sharpen the closest-result contrast, replace
an inappropriate residual by a metric-native one, or make the toy test
concrete.  It may not broaden the novelty delta, migrate to another portfolio
slot, introduce a mechanism, or turn the seed into a final theorem.  The
article-blind run permits four repair iterations and three replacement attempts
per slot.  A promoted seed is stored together with its exact revision
identifier, source-snapshot identifier, and completed critic decisions.
Changing any mathematical field creates a new revision and invalidates the
target anchor and every downstream candidate, mechanism, coverage assessment,
and challenge decision derived from the previous version.

\section{Details of Step 2: Mechanism--Problem Co-Exploration}
\label{app:step-two}
\label{app:agents-and-schemas}

This section specifies the runtime state, legal transitions, audit contracts,
and stopping rules used during mechanism--problem co-exploration. The design
separates mutable candidate content from the immutable research boundary and
separates mechanism acceptance from candidate acceptance. Every transition is
recorded against explicit artifact revisions so that a resumed run cannot
silently reuse evidence produced for an earlier candidate.

\subsection{Immutable Target Anchor and Candidate State}

For each reviewed Raw Seed, the protocol instantiates an immutable target
anchor \(A\). The anchor records the mathematical object under study, the source
boundary, admissible assumptions, required target direction, excluded
trivializations, and source-relative value criteria. Agents may refine the
candidate statement within this boundary, but no transition may mutate \(A\).
An apparent need to enlarge the source pack or assumption boundary therefore
terminates the current route rather than changing the benchmark instance.

The mutable Research Candidate is stored as a revisioned record \(p^{(r)}\).
Each revision contains:
\begin{itemize}
    \item a self-contained statement with explicit domains and quantifiers;
    \item the current assumption set and scope restrictions;
    \item a positive target and a corresponding refutation target;
    \item one to three named unresolved mathematical cores;
    \item validation checks, boundary cases, and the next proposed experiment;
    \item provenance linking every semantic field to the seed and anchor; and
    \item a content hash used to bind mechanisms and receipts to this revision.
\end{itemize}

The mechanism state is a directed acyclic graph \(G\). A node is an exact
mechanism revision, and an edge declares that one mechanism consumes a
capability supplied by another. Each node records its requirements,
consequences, derivation, applicability conditions, support boundary, positive
test, obstruction test, first nonroutine step, unresolved bottleneck, and
dependencies. Failed and rejected proposals may remain as search memory, but
only independently promoted nodes with current acceptance receipts belong to
the auditable graph used by the Coverage Judge.

The construction state additionally records the action history, Critic
decisions, coverage decisions, probe outputs, and the exact revisions of all
inputs needed to resume. Its identifier binds the candidate, mechanisms,
anchor, source snapshot, prompt version, and effective configuration. A
resumed run is valid only when these identifiers agree with the saved state.

The Explorer is the only role permitted to propose a semantic candidate
revision. The Mechanism Critic may request such a revision, and the Coverage
Judge or Final Challenger may route the state back to the Explorer, but these
roles do not directly edit the candidate. The Final Challenger described in
Section~\ref{app:step-three} receives an isolated view without \(G\), receipts,
or the exploration trace.

\subsection{Explorer Transitions}

At each iteration the Explorer reads the complete current state and emits
exactly one legal action. Actions are validated by the transition protocol
rather than executed implicitly from free-form text. This prevents a response
from both changing the candidate and claiming support from receipts tied to the
previous revision.

\begin{table*}[t]
\centering
\small
\begin{tabular}{@{}p{0.24\textwidth}p{0.72\textwidth}@{}}
\toprule
Action & State transition \\
\midrule
\texttt{propose\_mechanism} &
Create one mechanism revision and send it to independent audit. \\
\texttt{add\_support} &
Propose a nonredundant mechanism after insufficient coverage. \\
\texttt{request\_refute} &
Run the one-shot candidate-scoped refutation probe when permitted. \\
\texttt{revise\_frontier} &
Create a new semantic candidate revision and invalidate current receipts. \\
\texttt{harden} &
Strengthen a permitted statement field after a recorded trigger. \\
\texttt{split} &
Replace the route with exactly two nonoverlapping child candidates. \\
\texttt{emit} &
Send a stable substantively supported candidate to the Final Challenger. \\
\texttt{reject} &
Terminate an invalid, redundant, or insufficiently valuable route. \\
\texttt{resolution\_candidate} &
Quarantine a possible complete proof or counterexample for verification. \\
\bottomrule
\end{tabular}
\caption{Explorer actions and their state transitions.}
\label{tab:explorer-actions}
\end{table*}

A mechanism proposal may reason forward from the assumptions, backward from a
target, or by connecting accepted intermediate capabilities. These are search
strategies rather than distinct action types. Every proposal must state the
first nonroutine mathematical step and identify what remains unresolved; a
schema-complete restatement of the target is not a mechanism.

A frontier revision must identify every semantic change relative to
\(p^{(r)}\), give a reason tied to an audit or experiment, and update both the
positive and refutation targets. The transition is invalid if it alters the
immutable anchor, introduces an unsupported assumption, removes a known
counterexample through artificial scope restriction, or eliminates the
unresolved core without reporting a possible resolution.

The \texttt{split} action creates exactly two children whose scopes or claims
are independently meaningful and nonoverlapping. A split is not a device for
returning multiple loosely related conjectures. Each child receives its own
candidate identity and begins with no current mechanism receipts.

\subsection{Mechanism Proposal and Independent Audit}

The Mechanism Critic receives the exact mechanism revision, current candidate,
anchor, declared dependencies, and any coverage brief that motivated the
proposal. It replays the central derivation and checks correctness,
noncircularity, typed-interface honesty, applicability, provenance, and
candidate relevance. At least one positive cell and one obstruction or
boundary cell must be examined concretely.

The Critic returns one of the following routes:
\begin{itemize}
    \item \texttt{promote}: the mechanism is correct, closed, noncircular, and
    contributes a concrete capability relevant to the candidate;
    \item \texttt{repair}: a bounded local defect can be corrected without
    changing the candidate frontier;
    \item \texttt{reject}: the mechanism is incorrect, circular, inapplicable,
    redundant, or fails to provide its declared consequence;
    \item \texttt{revise\_frontier}: the audit exposes a candidate-level defect;
    \item \texttt{resolution\_candidate}: the audit yields a credible complete
    proof or counterexample requiring independent verification.
\end{itemize}

Promotion creates an acceptance receipt containing the candidate revision
hash, mechanism revision hash, dependency hashes, Critic decision, tested
cells, and support boundary. The receipt permits the mechanism to participate
in coverage composition. It is not a formal proof certificate, a novelty
judgment, or evidence that the full candidate is true.

\subsection{Coverage Composition and Stopping Rule}

The Coverage Judge receives the exact candidate revision, immutable anchor,
and all promoted mechanism revisions with current receipts. It selects a
nonredundant subgraph \(G'\subseteq G\) and a dependency order
\(\pi=(\pi_1,\ldots,\pi_k)\). Starting from the capabilities supplied by the
candidate assumptions, composition is valid only when
\begin{equation}
\begin{aligned}
\mathcal A_0
    &= \operatorname{Cap}(p), \\
\operatorname{Req}(m_{\pi_i})
    &\subseteq \mathcal A_{i-1},
    \qquad i=1,\ldots,k, \\
\mathcal A_i
    &= \mathcal A_{i-1}
       \cup \operatorname{Prov}(m_{\pi_i}).
\end{aligned}
\label{eq:mechanism-composition}
\end{equation}
The Judge replays the declared interfaces in order and verifies that the
composition supports the candidate rather than merely connecting notation.

The Judge returns exactly one of seven decisions:
\texttt{insufficient\_support}, \texttt{substantive\_partial},
\texttt{full\_coverage}, \texttt{split}, \texttt{revise\_frontier},
\texttt{resolution\_candidate}, or \texttt{reject}. The first three describe
mathematical support; the remaining four are direct control-flow routes.

\texttt{Insufficient\_support} names the missing capability and gives a
lightweight success condition without prescribing the key equation or
mechanism. \texttt{Substantive\_partial} requires a genuine nonroutine
intermediate result while leaving one to three precise mathematical cores
open. This is the intended co-exploration stopping state.
\texttt{Full\_coverage} is not emitted as an open conjecture: it is quarantined
for resolution checking, as is any credible complete proof or counterexample.

Emission additionally requires a stable candidate revision, current receipts,
a valid dependency order, completed positive and obstruction checks, and no
pending repair or refutation result. The candidate is then passed to the
isolated Final Challenger. Challenger repair returns to co-exploration;
acceptance advances to prover-facing construction.

\subsection{Receipt Invalidation and Checkpoint Semantics}
\label{app:transition-rules}

Receipts are candidate-version-bound. Any semantic change to assumptions,
domains, quantifiers, scope, positive target, refutation target, or unresolved
cores creates a new candidate revision and invalidates every current receipt.
Formatting-only normalization may preserve the revision only when the
normalized abstract statement and its content hash are unchanged.

A repaired mechanism receives a new mechanism revision. Its previous receipt
and all receipts that depend on it become stale. Removing a graph node likewise
invalidates dependent compositions. This invalidation is applied transitively;
an agent cannot declare a receipt from an earlier revision to be current.

Each accepted transition is recorded before the next agent call. A resumable
state is committed only after the candidate, graph, receipts, and transition
history agree. On resume, all revision identifiers are verified; a state with
a missing input, stale receipt, changed source snapshot, different effective
configuration, or incomplete transition is rejected.

These rules make retries idempotent. Repeating an agent call may produce a new
proposal, but it cannot overwrite an accepted revision or attach evidence to a
different candidate. Terminal routes---rejection, split, possible resolution,
and accepted emission---are recorded explicitly so that downstream stages can
distinguish mathematical outcomes from infrastructure failures.
\section{Details of Step 3: Prover-Facing Construction}
\label{app:step-three}

Step~3 separates candidate acceptance from mechanism acceptance.  It performs
an isolated candidate-level challenge, projects the surviving mathematical
content into a self-contained prove-or-refute task, and then tests only that
statement for triviality and concrete refutation.

\subsection{Final Challenger}

The Final Challenger receives the exact Research Candidate, immutable target
anchor, reviewed Raw Seed, and frozen source pack.  It intentionally does not
receive the mechanism graph, coverage composition, critic decisions, or
acceptance receipts.  This visibility restriction prevents a plausible proof
route from substituting for a coherent and valuable statement.

The Challenger performs four independent audits:
\begin{enumerate}
    \item \textbf{Statement audit:} close definitions, domains, quantifiers,
    assumptions, and scope; verify that the positive and refutation targets
    are logical alternatives under identical premises.
    \item \textbf{Counterexample audit:} calculate an in-scope toy, boundary,
    endpoint, degenerate, or scaling case and attack the candidate itself.
    Failure of one possible mechanism is not a counterexample.
    \item \textbf{Value audit:} subtract the closest frozen-source result and
    state the value of proof, valid refutation, and meaningful partial progress.
    \item \textbf{Difficulty audit:} identify the remaining nonroutine
    bottleneck and a finite progress certificate; reject routine tasks and
    unbounded research programs.
\end{enumerate}

The decisions are \texttt{accept}, bounded \texttt{repair}, exactly-two-way
\texttt{split}, \texttt{reject}, or \texttt{resolution\_candidate}.  A repair
returns to Step~2, where a semantic candidate change invalidates all receipts.
Acceptance means only that the candidate passes this isolated audit; it does
not certify truth or historical openness.

\subsection{Deterministic Projection}

An accepted pair consisting of anchor $A$, candidate $p$, and selected audited
mechanisms $G'$ is deterministically projected to
\begin{equation}
q=\Theta(A,p,G').
\end{equation}
The projector preserves assumptions, quantifiers, scope, positive target, and
refutation target.  It closes notation, states all objects locally, converts
each unresolved core into a checkable obligation, and separates the public
problem statement from the optional mechanism bundle.  It may not add an
assumption, weaken a target, or silently replace the accepted mathematical
content.  The result is a \texttt{semi\_open\_problem} record.

For experiment-facing projection, the projection procedure considers accepted paper-level
outputs, semi-open tasks, resolution records, and research candidates in that
order, normalizes them to a common identity-free surface, removes exact
duplicates, and retains at most three documents.  This broader benchmark
projector measures whether a run produced a theorem-level proposal; it does
not mean every projected record was admitted as a semi-open task.

\subsection{Statement-Only Probes and Final Filtering}

The final prover stress gate schedules one Blind Prover and one Refute Prover.  Both see
only the prover-facing statement.  The Blind Prover attempts a solution or a
checkable first nonroutine step.  A first complete solution triggers one
hardening opportunity; a second complete solution routes the task to
resolution quarantine.  Failure to identify a first step is acceptable
evidence of difficulty and is not itself evidence of ambiguity.

The Refute Prover must supply a concrete witness, verify every shared
assumption, and calculate violation of the positive predicate.  An unverified
objection is not a counterexample.  A valid counterexample is quarantined for
independent verification.  After the probes, notation-only normalization is
permitted, but any semantic statement change must return to Step~2.

Finally, the scheduler deduplicates globally using mathematical objects,
assumptions, scope, constraints, positive target, and refutation contract,
rather than surface wording or mechanism route.  It then applies any frozen
domain quota.  Rejected candidates and possible resolutions remain in explicit
artifact collections instead of being silently counted as open problems.

\providecommand{\system}{\textsc{MECA}}

\section{Experimental Settings}
\label{app:experimental-settings}

This section provides the implementation details omitted from the Experiments
section of the main paper: model assignments, information boundaries,
frozen-data construction, per-case budgets, and stopping rules.  Evaluation-only
information is never returned to a generation system, and a run enters the
reported summaries only after producing a complete case-level record.

\subsection{Models, Roles, and Information Boundaries}
\label{app:models-tools}

\begin{table*}[t]
\centering
\small
\setlength{\tabcolsep}{5pt}
\begin{tabular}{p{0.20\textwidth}p{0.14\textwidth}p{0.10\textwidth}p{0.46\textwidth}}
\toprule
Role group & Model & Effort & Information and tools available \\
\midrule
\system{} and baseline generation & \texttt{gpt-5.5} & \texttt{xhigh} &
The same case description and seven anonymized pre-2024 source cards.
Neither system receives the target paper, target-result decomposition,
evaluation record, or unrestricted external retrieval. \\
Target-alignment evaluator & \texttt{gpt-5.5} & \texttt{xhigh} &
One frozen system output set and the preselected target-result record.
Evaluation begins only after generation is complete and cannot trigger output
revision. \\
QED prover and verifiers & \texttt{gpt-5.6-sol} & \texttt{xhigh} &
Only the final self-contained prove-or-refute statement.  MECA mechanisms,
critic reports, source cards, and intermediate construction traces are
withheld. \\
\bottomrule
\end{tabular}
\caption{Models, reasoning effort, and information boundaries used in the
experiments.}
\label{tab:models-tools}
\end{table*}

All roles run in the same versioned execution environment with fixed
role-specific instructions.  Cases are independent.  Within a case, records
are passed only along the protocol-defined edges, so parallel scheduling
changes throughput but not the information available to a role or the order of
case-level decisions.

\subsection{Target-Conditioned Article-Blind Recovery}
\label{app:article-blind-benchmark}

\paragraph{Benchmark unit.}
Each case \(D_i\) consists of three frozen objects: a broad mathematical case
description \(d_i\), an article-blind source pack
\(\mathcal S_i=\{s_{i1},\ldots,s_{i7}\}\), and a target-result record \(T_i\).
The record \(T_i\) decomposes the central result of one later paper into the
five evaluation fields used in the first experiment of the main paper:
setting, assumptions, conclusion, quantitative regime, and boundary behavior.
During generation, both systems receive only \((d_i,\mathcal S_i)\).  After
their conjecture sets are frozen, the evaluator receives one frozen output set
and \(T_i\).  The task is therefore theorem-level reconstruction from an
earlier mathematical neighborhood, not prediction of a paper title, identifier,
or free-form topic.

\paragraph{Frozen target set.}
The recovery benchmark contains 20 cases, denoted D001--D020.  Each case is
anchored to one target paper first made public after December 31, 2023 and
reporting a substantive theorem-level mathematical advance.  The complete set
of target papers is selected and frozen before either system is run.  Thus,
``target-conditioned'' means that the benchmark designer uses a fixed target
to define the mathematical neighborhood and the later evaluation reference;
it does not mean that the target conclusion is exposed to the generation
systems.

\paragraph{Source-pack construction.}
For each target, seven papers available by December 31, 2023 are selected to
cover the relevant setting, established results, proof mechanisms, and known
limitations.  This yields 140 source cards in total.  Each card preserves four
classes of mathematical evidence: the problem setting, conclusions already
established, mechanisms supporting those conclusions, and limitations or
failure regimes stated by the source.

The cards remove titles, authors, venues, URLs, persistent identifiers,
filenames, theorem numbers, and references that reveal the target paper.
Target-paper text, target conclusions, and post-2023 evidence are not included.
The cards are therefore intended to preserve the mathematical premises needed
for source-grounded reasoning while preventing direct article lookup or title
completion.  This protocol controls information supplied at inference time; it
does not claim parametric unlearning of all post-2023 mathematics.

\paragraph{Task presented to both systems.}
Both systems receive the same broad case description and the same seven source
cards.  They are asked to propose theorem-level advances supported by that
evidence and may return at most three conjectures for the case.  A valid output
must state the mathematical objects, assumptions, quantified claim, relevant
quantitative regime, and any important boundary or failure condition.  Neither
system may retrieve the target paper or use evaluator feedback during
generation.

\paragraph{\system{} protocol.}
\system{} follows the source-grounded construction process detailed in
Sections~\ref{app:source-grounded-seeds}, \ref{app:step-two}, and
\ref{app:step-three}.  It first forms research seeds from gaps and tensions in
the source cards, then co-refines candidate statements and candidate proof or
refutation mechanisms, and finally converts surviving candidates into
self-contained theorem-level outputs.  Mechanisms are treated as partial
mathematical support, not as extra assumptions or completed proofs.
Independent critics may reject, repair, split, or narrow a candidate before it
is retained.

\begin{table*}[t]
\centering
\small
\setlength{\tabcolsep}{5pt}
\begin{tabular}{p{0.31\textwidth}r@{\qquad}p{0.31\textwidth}r}
\toprule
Item & Limit & Item & Limit \\
\midrule
Raw Seed directions & 3 & Seed repairs/direction & 4 \\
Seed replacements/direction & 3 & Joint search rounds & 4 \\
Coverage assessments & 4 & Active candidates & 2 \\
Splits/candidate & 1 & Candidate attempts/seed & 3 \\
Mechanisms/candidate & 3 & Repairs/mechanism & 3 \\
Critic steps & 30 & Stable confirmations & 2 \\
Final statement repairs & 2 & Blind probes & 1 \\
Refutation probes & 1 & Outputs/case & 3 \\
\bottomrule
\end{tabular}
\caption{Per-case \system{} ceilings in target-conditioned recovery.  These are
stopping limits, not required numbers of calls.}
\label{tab:generation-budgets}
\end{table*}

The limits in Table~\ref{tab:generation-budgets} are nested ceilings.  A case
may terminate earlier when all directions are rejected, a candidate is already
implied by the sources, a valid counterexample is found, or no candidate
passes the review gates.  Exhausting a budget does not turn an incomplete
candidate into an accepted output.

\paragraph{Draft--Audit--Revise baseline.}
The baseline starts from an initial set of theorem-level drafts, performs five
audit steps, and then performs five revision steps.  Each audit checks the
drafts against the supplied sources for mathematical coherence, unsupported
assumptions, overclaiming, and missing quantitative or boundary conditions.
Each revision may repair or replace a draft in response to the preceding audit.
The baseline has the same three-output cap and the same information boundary
as \system{}, but it does not maintain explicit candidate-mechanism
co-evolution, mechanism-coverage records, or isolated proof and refutation
probes.  The comparison therefore holds the task, evidence, output cap, and
target blindness fixed while varying the construction procedure.

\paragraph{Freeze and target-based evaluation.}
All outputs are frozen before the evaluator receives any target information.
Evaluation then follows the same-candidate 100-point rubric in the first
experiment of the main paper.  Reversed implications, target
conclusions inserted as assumptions, incompatible quantifiers, conflicting
parameter regimes, and mismatched rates are treated as substantive errors
rather than repaired across outputs.  Scores are computed per case and averaged
over the 20 frozen targets.

\paragraph{Interpretation.}
This experiment measures target alignment under an article-blind input
protocol.  It does not test whether the generated conjecture is historically
novel, and it does not use post-hoc retrieval to choose a favorable target.
Because the target is fixed before generation, the reported score always
refers to reconstruction of the same hidden result for both systems.

\subsection{Open-Ended Conjecture Generation and Verification}
\label{app:open-ended-setting}

\paragraph{Conjecture sample.}
The second experiment uses 100 final conjectures constructed through the two
routes in the Framework section of the main paper.  Route~A contributes 92 conjectures from
literature collections in a private repository of mathematical papers, and
Route~B contributes 8 conjectures obtained by refining broad descriptions from
a privately curated collection of mathematical open problems.  Both
collections were assembled before the study and frozen before generation.
Unlike the recovery benchmark, these tasks have no preselected hidden target
paper and are not scored by similarity to a later result.

\paragraph{Statement freeze and QED input.}
Once a conjecture passes MECA's final construction checks, its prover-facing
statement is frozen.  Each statement is submitted to QED in a fresh isolated
run with a five-hour wall-clock limit.  QED receives definitions, assumptions,
the positive conclusion, and the admissible refutation objective contained in
the final statement, but no MECA mechanism, critic report, source card, or
intermediate reasoning trace.

\paragraph{Terminal outcomes.}
Terminal outcomes and unresolved-case diagnoses follow the verification
criteria in Section~\ref{app:qed-verification}.  Timeout or exhausted search is
not treated as evidence that the conjecture is true, false, or ill posed.

\paragraph{Relation between the two experiments.}
QED is not used to rescore the target-recovery outputs, and target alignment is
not used to judge the 100 open-ended conjectures.  Recovery evaluates whether
the construction procedure approaches a fixed hidden conclusion; QED
evaluation tests whether independently generated statements support a
checkable proof-or-refutation attempt.  Keeping these outcome spaces separate
avoids treating similarity to an existing theorem as proof of conjecture
quality or treating prover failure as evidence of poor target recovery.

\section{Evaluation Criteria}
\label{app:evaluation-rubrics}

\subsection{Complete Target-Recovery Output}
\label{app:projectable-output}

A target-recovery case is complete when deterministic normalization finds at
least one retained theorem-level conjecture.  The frozen record exposes the
formal setting, assumptions, conclusion, quantitative regime, boundary
conditions, source-relative motivation, and source-card identifiers.
Completeness is a structural criterion; it does not establish correctness or
agreement with the hidden target.  For frozen output sets $\mathcal P_i$,
\begin{equation}
R_{\mathrm{valid}}=\frac{1}{20}\sum_{i=1}^{20}
\mathbb{1}\{\mathcal P_i\neq\varnothing\}.
\end{equation}

\subsection{Target Alignment and Evaluator Scoring}
\label{app:match-criteria}

The evaluator applies the rubric from the first experiment of the main paper to
one selected conjecture per case.  Reversed implications, assumed conclusions,
incompatible quantifiers, conflicting regimes, and mismatched rates count as
substantive mismatches rather than being repaired by combining different
outputs.  This same-candidate rule prevents topical agreement from obscuring
errors in theorem direction, quantitative strength, or validity regime.

Three independent evaluator replicas each select and score a conjecture.  The
case-level selection is the majority choice, with total target-alignment score
used only to break a three-way tie; component scores are averaged across the
replicas.
\label{app:judge-agreement}

\label{app:quality-rubric}
The critic-aligned conjecture-quality rubric is conjunctive: a weighted score
cannot compensate for a fatal mathematical or source-value failure.

\begin{table*}[t]
\centering
\small
\setlength{\tabcolsep}{4pt}
\begin{tabular}{p{0.23\textwidth}ccp{0.51\textwidth}}
\toprule
Dimension & Weight & Minimum & Passing interpretation \\
\midrule
Correctness and falsification resistance & 0.30 & 1.0 & Definitions, domains, quantifiers, assumptions, and target duality close; no fatal toy, sign, scaling, circularity, or counterexample defect. \\
Source-subtracted value & 0.25 & 1.0 & Both proof and valid refutation advance beyond the closest result and routine consequences. \\
Bounded nonroutine difficulty & 0.20 & 0.5 & One to three mathematical cores remain and admit finite progress certificates. \\
Direct statement quality & 0.15 & 1.0 & The prover-facing statement is self-contained and checkable. \\
Substantive partial mechanism & 0.10 & 0.5 & A correct candidate-relevant intermediate is established with an honest support boundary. \\
\bottomrule
\end{tabular}
\caption{Critic-aligned quality rubric.}
\label{tab:quality-rubric}
\end{table*}

\subsection{QED Verification and Error Taxonomy}
\label{app:qed-verification}

QED adjudication requires both structural and detailed verification.
\textbf{Proved} requires a complete positive proof passing both layers, while
\textbf{Refuted} requires an in-scope counterexample or negative argument
passing the same standard.  All other terminal outcomes are
\textbf{Unresolved}.

\label{app:error-taxonomy}
Each unresolved case receives one primary label: (i) technical proof
difficulty after statement and foothold checks pass; (ii) missing assumptions;
(iii) insufficient mathematical foothold; or (iv) ambiguous formulation.
Engineering failures, if any, are recorded separately rather than treated as
mathematical outcomes.

\providecommand{\system}{\textsc{MECA}}
\section{Illustrative Examples}
\label{app:illustrative-examples}

This section first gives a schematic example of the target-conditioned recovery
benchmark and then two technical case studies from the open-ended setting of
Section~\ref{app:open-ended-setting}.  The schematic example explains the
information boundary and scoring unit but is not one of the 20 scored benchmark
cases.  The two open-ended cases respectively preserve an explicit unresolved
core and reach a complete refutation after verifier-requested repair.  We
distinguish exact reductions, numerical reconnaissance, incomplete
certificates, and terminally verified claims throughout.


\subsection{Anatomy and Rationale of the Recovery Benchmark}
\label{app:ill-recovery-benchmark}

\paragraph{Why a constructed benchmark is needed.}
The object of evaluation is not theorem proving from a supplied statement, but
the recovery of a plausible theorem-level advance from the mathematical
boundary that preceded it.  A purely open-ended task cannot provide a stable
reference for this purpose: two systems may select different topics, and
judgments of novelty or importance then become inseparable from topic choice.
Conversely, revealing a later theorem would reduce the task to restatement or
proof completion.  We therefore construct a retrospective benchmark in which a
later paper supplies an external evaluation anchor while remaining completely
withheld from generation.  This design fixes what counts as successful
recovery without making the answer part of the input.

\paragraph{What one benchmark case contains.}
As defined in Section~\ref{app:article-blind-benchmark}, a case contains a broad
description \(d_i\), seven pre-cutoff source cards \(\mathcal S_i\), and a
hidden target record \(T_i\).  The first two objects are visible to both
systems; \(T_i\) is opened only by the evaluator after outputs are frozen.  The
example below is an abbreviated, format-level rendering of such a case in
decentralized optimization.  It illustrates what the benchmark looks like and
how it is scored, but it is not an additional observation in the D001--D020
aggregate and none of its illustrative details enter the reported averages.

\paragraph{Visible generation-side input.}
The broad case description asks for at most three complete theorem-level
advances beyond a supplied pre-2024 boundary in decentralized first-order
optimization and gradient tracking.  A candidate must close its mathematical
setting and assumptions, define any new algorithmic object, state the
direction and scope of its conclusion, report a quantitative regime when the
evidence supports one, and name a direct failure or falsification condition.
The description does not request flexible gradient tracking, a Pareto
frontier, or any particular rate.

Both generation systems receive this description together with exactly the
same seven-card public pack.  Table~\ref{tab:flexgt-public-pack} records the
mathematical content visible at generation time.  Card identifiers are public
handles, not bibliographic identifiers.

\begin{table*}[t]
\centering
\scriptsize
\setlength{\tabcolsep}{4pt}
\begin{tabular}{p{0.06\textwidth}p{0.34\textwidth}p{0.50\textwidth}}
\toprule
Card & Setting & Resolved result and exposed mechanism \\
\midrule
S01 &
Smooth strongly convex decentralized empirical-risk minimization on a fixed
connected network. &
A tracked global-gradient surrogate can be used in regularized local
subproblems.  With \(K\) averaging rounds, the card records an effective
mixing factor \(\beta_{\mathrm{mix}}=\beta_0^K\) and separate local-computation
and communication dependence. \\

S02 &
Smooth convex decentralized optimization over time-varying graphs. &
Nesterov-style acceleration can be coupled to gradient tracking.
Multiple-consensus and accelerated-gossip variants expose explicit
gradient-versus-communication complexity. \\

S03 &
Smooth strongly convex optimization on a fixed directed graph with
deterministically quantized messages. &
Quantized primal and tracking innovations retain exact linear convergence
when their scales decay compatibly with a lifted error recursion. \\

S04 &
Smooth strongly convex decentralized optimization over time-varying networks
under first-order oracle models. &
Communication and local-gradient complexities have separate lower bounds;
multi-consensus primal-dual constructions attain corresponding upper bounds. \\

S05 &
Smooth optimization on a fixed general directed network with compressed
messages. &
Compressed push--pull gradient tracking is controlled through a joint
spectral-radius condition for optimality, disagreement, tracking, and
compression errors. \\

S06 &
Smooth strongly convex optimization over time-varying strongly connected
directed graphs. &
Row-stochastic primal mixing and column-stochastic tracking yield global
linear convergence by closing a three-error comparison recursion. \\

S07 &
Distributed stochastic optimization with gradient tracking over time-varying
directed networks. &
Moving weighted contractions give linear convergence to a
variance-dependent neighborhood under smooth strong convexity. \\
\bottomrule
\end{tabular}
\caption{The seven anonymous source cards visible to both systems in the
representative benchmark rendering.  Each card also states limitations of its
own result; none states what the hidden target contributes.}
\label{tab:flexgt-public-pack}
\end{table*}

The source cards contain only four fields: setting, resolved conclusions,
mechanisms, and limitations.  They omit titles, authors, venues, URLs,
persistent identifiers, filenames, theorem numbers, and post-cutoff
citations.  The target paper and any source that directly discloses its central
construction are excluded.  Thus the input supplies ingredients such as
repeated consensus, accelerated gossip, stochastic neighborhoods, and separate
resource lower bounds without supplying their later theorem-level assembly.

\paragraph{Evaluator-only target record.}
Neither \system{} nor the Draft--Audit--Revise baseline receives the following
information during generation.  After their output sets are frozen, the
target-alignment evaluator opens the corresponding record.  Under the five-field
evaluation used in the first experiment of the main paper, a representative
record requires the following reconstruction.

\begin{itemize}
\item \textbf{Setting.}
There are \(n\) nodes with possibly heterogeneous, non-i.i.d.\ local
distributions \(\mathcal D_i\), local objectives
\[
f_i(x)=\mathbb E_{\xi_i\sim\mathcal D_i}[f_i(x;\xi_i)],
\qquad
f(x)=\frac1n\sum_{i=1}^n f_i(x),
\]
and a connected static mixing network.  The result covers strongly convex,
convex, and smooth nonconvex objectives.

\item \textbf{Assumptions.}
Each \(f_i\) is \(L\)-smooth and the global objective is bounded below.  The
stochastic gradients have uniformly bounded variance,
\[
\mathbb E\!\left[
 \left\|\nabla f_i(x;\xi_i)-\nabla f_i(x)\right\|^2
\right]\leq \sigma^2.
\]
The mixing matrix \(W\) is doubly stochastic and satisfies
\[
\begin{aligned}
W\mathbf 1&=\mathbf 1,&
\mathbf 1^\top W&=\mathbf 1^\top,\\
\rho_W&:=\|W-J\|_2^2<1,&
J&=\frac1n\mathbf 1\mathbf 1^\top.
\end{aligned}
\]
Convexity or \(\mu\)-strong convexity is added only for the corresponding
convex or strongly convex theorem branch.

\item \textbf{Conclusion.}
The new object is FlexGT, a snapshot gradient-tracking method that
independently chooses \(\beta\) local stochastic updates and \(\alpha\)
communication steps in each round.  Gradients within the round are evaluated
at the round snapshot rather than at every changing local iterate.
Acc-FlexGT replaces direct repeated communication by accelerated gossip when
graph information is available.  The main theorem gives: a linear
strongly-convex Lyapunov recursion with separate centralized stochastic-noise
and topology-effect terms; an averaged sublinear objective-gap bound in the
convex regime; and an averaged sublinear stationarity bound
\[
\frac1K\sum_{k=0}^{K-1}
\mathbb E\!\left[\|\nabla f(\bar x_{\beta k})\|^2\right]
\]
in the nonconvex regime.

\item \textbf{Quantitative regime.}
For the nonconvex Acc-FlexGT corollary, the target record includes the
graph-dependent communication schedule
\[
\alpha=
\left\lceil
\frac{
 \max\{\ln 2,\frac12\ln(\beta\max\{n,R_0\})\}
}{
 \sqrt{1-\sqrt{\rho_W}}
}
\right\rceil
\]
and the round complexity
\[
\widehat K=
O\!\left(
 \left[
  \frac{L}{\epsilon}
  +\frac{L\sigma^2}{n\beta\epsilon^2}
 \right]
 \mathbb E[f(\bar x_0)-f^\star]
\right),
\]
subject to the theorem's stepsize condition.

\item \textbf{Boundary behavior.}
The pair \((\alpha,\beta)\) is a genuine
communication--computation trade-off: increasing one resource need not
improve the other, and the nondominated choices form a Pareto frontier.
After resource tuning, the target summarizes the Acc-FlexGT nonconvex costs as
\[
\widetilde O\!\left(
 \frac{L^2\sigma^2}{n\epsilon^2}
 +\frac{L}{\epsilon(1-\rho_W)}
\right)
\quad\text{iterations}
\]
and
\[
\widetilde O\!\left(
 \frac{L}{\epsilon(1-\rho_W)}
\right)
\quad\text{communications},
\]
matching the referenced lower bounds up to logarithmic factors.  A statement
of only one empirically preferred weighted cost, without the Pareto claim or
the lower-bound comparison, does not recover this boundary field.
\end{itemize}

In the benchmark implementation, each field is stored at finer atomic
resolution so that the evaluator can distinguish recovering the theorem's
direction from recovering its assumptions, quantitative strength, or validity
boundary.  This decomposition is important: topical similarity alone cannot
receive a high score, and a stronger-looking claim is penalized when it removes
an assumption needed by the target result.

\paragraph{How the source cards support, but do not reveal, the target.}
The target can be approached only by composing evidence across cards.  S01
suggests treating the number of consensus rounds as a resource; S02 supplies
multi-consensus and accelerated-gossip mechanisms; S04 supplies the
communication-versus-local-oracle lower-bound perspective; and S07 supplies
the stochastic-gradient-tracking error decomposition.  At the same time, the
cards leave genuine reconstruction work.  S01 is fixed-network and
strongly-convex or quadratic, S02 is not a heterogeneous stochastic snapshot
method, S04 does not identify an algorithm-specific Pareto frontier, and S07
fixes one communication exchange per update.  No individual card therefore
states FlexGT, the independent \((\alpha,\beta)\) design, the three-regime
theorem, or the target complexity law.

\paragraph{Same-candidate evaluation example.}
Suppose a frozen output defines snapshot gradient tracking with independently
tunable local-update and communication counts and recovers the nonconvex
complexity dependence on \(n\), \(\beta\), \(\sigma^2\), and \(\epsilon\).
However, suppose that the same output replaces the static doubly-stochastic
network by an unsupported time-varying directed model, omits bounded stochastic
variance, and claims exact fixed-stepsize convergence in the strongly convex
case despite nonzero variance.

The evaluator may give this document strong credit for the new object and
quantitative regime, and partial credit for the nonconvex conclusion.  It
loses setting and assumptions credit because the target theorem is not stated
under a compatible network and oracle model.  Its exact-convergence claim
conflicts with the target's stochastic residual and is a central
contradiction, rather than a harmless omission.  If the document also omits
the Pareto frontier and lower-bound comparison, another output cannot provide
those missing boundary claims: the evaluator scores one complete candidate
against the target and never splices fields across the system's other
conjectures.

\paragraph{Information-boundary interpretation.}
This format is target-conditioned because the frozen target record determines
which mathematical facts receive evaluation credit.  It remains
article-blind because generation sees only the broad task and the seven
earlier anonymous cards; the title, identifier, target decomposition, and
evaluation feedback appear only after the outputs are immutable.  The
protocol therefore tests reconstruction of a hidden theorem-level
contribution from a bounded earlier neighborhood.  It does not test title
completion, and it does not claim that the model has parametrically forgotten
all post-2023 mathematics.

\paragraph{Why this is a reasonable evaluation benchmark.}
The benchmark is constructed rather than sampled from an existing task suite
because the desired construct requires both an earlier evidence boundary and a
later theorem-level reference.  Its validity does not rest on treating our
annotations as a substitute for mathematical truth.  The central result of a
later paper provides the external anchor; benchmark construction only controls
which earlier evidence is visible and decomposes that result into auditable
fields.  Target conditioning removes variation caused by systems choosing
different topics, while article blindness prevents direct lookup or copying.
The seven-card packs preserve established facts, proof mechanisms, and known
failure regimes, so success still requires combining distributed evidence
rather than guessing from a title or reproducing a supplied theorem.

The comparison is further controlled by freezing targets and source packs
before generation, enforcing the same temporal cutoff, giving both systems
identical inputs and output caps, freezing all candidate texts before
evaluator access to \(T_i\), and scoring every dimension on one candidate
rather than assembling the best fields from several outputs.  These controls
make the benchmark suitable for the paper's comparative question: whether a
mechanism-centered construction process more reliably recovers a hidden
mathematical advance from the same bounded evidence.

\paragraph{Scope.}
This benchmark does not measure unrestricted topic discovery, historical
priority, or parameter-level forgetting of post-cutoff work.  Nor does it claim
that the 20 targets represent all of mathematics.  It measures the narrower
and directly auditable capability needed here: recovering the setting,
assumptions, conclusion, quantitative regime, and boundary behavior of a
withheld theorem-level advance from a controlled snapshot of earlier
mathematical evidence.

\subsection{Candidate--Mechanism Co-Evolution with an Open Technical Core}
\label{app:ill-corner-switch}

The case ``Memory-dependent switch of the most stable directed-weight
corner'' concerns the sharp stepsize boundary of a two-agent directed gradient
tracker with reconstruction memory and one stale communication phase.  Its
archived trace contains the anchor, candidate, mechanism, coverage, challenge,
and projection records.  The case remains unresolved because detailed
verification did not close the final global certificate.

\subsubsection{Source boundary, Raw Seed, and frozen anchor}

The Raw Seed records three closest-result shapes: linear convergence of
compressed gradient tracking, sufficient convergence conditions for directed
compressed push--pull tracking, and directed asynchronous tracking with
bounded delays.  Those result shapes do not provide a necessary-and-sufficient
boundary for the simultaneous memory-and-delay recurrence below or classify
the weights that maximize such a boundary.  The archived example retains this
derived source-boundary record rather than reproducing the underlying private
source documents.  It therefore illustrates the construction trace, not an
independently complete literature audit.

For $a,b\in[1/4,3/4]$, communication alternates between
\begin{equation}
\begin{aligned}
A_0&=\begin{bmatrix}1&0\\a&1-a\end{bmatrix},&
B_0&=\begin{bmatrix}1-b&0\\b&1\end{bmatrix},\\
A_1&=\begin{bmatrix}1-a&a\\0&1\end{bmatrix},&
B_1&=\begin{bmatrix}1&b\\0&1-b\end{bmatrix}.
\end{aligned}
\label{eq:ill-corner-mixing}
\end{equation}
The local objectives are $f_i(z)=\tfrac12(z-r_i)^2$.  For
$q\in[1/4,1/2]$, the memories are
\begin{equation}
H_x=(1-q)h_x+qx,\qquad H_y=(1-q)h_y+qy.
\end{equation}
Phase zero mixes the preceding memories, phase one mixes the new memories,
and the constant-stepsize recurrence is
\begin{equation}
\begin{split}
x^{k+1}&=x^k+(A_k-I)h_x^{k-d_k}-\alpha y^k,\\
y^{k+1}&=y^k+(B_k-I)h_y^{k-d_k}
 +(x^{k+1}-r)-(x^k-r),
\end{split}
\label{eq:ill-corner-recurrence}
\end{equation}
where $(d_0,d_1)=(1,0)$, $y^0=x^0-r$, and the missing memory history is
zero.  The Raw Seed asks whether the initialized reachable dynamics have a
sharp connected-from-zero stability endpoint and whether memory and delay
affect it nonseparably.

The immutable anchor can be summarized as
\begin{equation}
A=(o,\mathcal P_A,\mathcal B_A,\mathcal F_A).
\label{eq:ill-anchor}
\end{equation}
Here $o$ is the initialized reachable quotient of the two-period tracker,
$\mathcal P_A$ contains the recurrence and assumptions above,
$\mathcal B_A$ is the recorded closest-result boundary, and $\mathcal F_A$
forbids replacing the recurrence, interpreting $q$ as lossy compression,
discarding reachable modes, or defining away a missing endpoint proof.  The
tracker invariant excludes a stable biased-consensus limit, so such a limit
was not treated as the proposed new failure mode.

\subsubsection{From a fixed slice to a bifurcation candidate}

An accepted candidate in the anchored portfolio fixed $q=1/2$ and included
the claim
\begin{equation}
p_0:\qquad
\operatorname*{arg\,max}_{(a,b)\in[1/4,3/4]^2}
\alpha_\star(1/2,a,b)=\{(1/4,1/4)\}.
\label{eq:ill-p0}
\end{equation}
Its accepted mechanism, denoted here by $m_{\mathrm{slice}}$, exposed the
following interface:
\begin{equation}
\begin{aligned}
\operatorname{Req}(m_{\mathrm{slice}})
 &=\{\mathcal P_A,q=1/2,\text{ stale cell}\},\\
\operatorname{Prov}(m_{\mathrm{slice}})
 &=\{\text{quintic and Cayley reduction}\},\\
\operatorname{Gap}(m_{\mathrm{slice}})
 &=\{\text{first contact and global extrema}\}.
\end{aligned}
\label{eq:ill-slice-interface}
\end{equation}
If $M_d$ is the exact two-period monodromy, direct block multiplication gives
\begin{equation}
\det(tI_8-M_d)=(t-1)(t-1/4)^2p(t;a,b,\alpha),
\label{eq:ill-slice-factor}
\end{equation}
where $p$ is monic of degree five.  The optimizer direction and a strictly
Schur double memory root can therefore be removed before studying the
remaining contacts.  With the Cayley substitution,
\begin{equation}
(1-iu)^5p\!\left(\frac{1+iu}{1-iu}\right)=R(u)+iI(u),
\label{eq:ill-slice-cayley}
\end{equation}
every nonreal unit-circle contact satisfies $R=I=0$.  This is a checkable
reduction, but it does not prove the maximizer in~\eqref{eq:ill-p0}.

The next exploration varied $q$.  Its numerical attack data were
\begin{equation}
\begin{array}{c|cc}
q & \alpha_\star(q,3/4,1/4) & \alpha_\star(q,1/4,1/4)\\ \hline
1/4 & 1.966716 & 1.963166\\
1/2 & 1.927152 & 1.931347
\end{array}
\label{eq:ill-corner-reconnaissance}
\end{equation}
and a provisional crossing was observed near $q=0.370460332$.  These values
are reconnaissance, not a proof of endpoint identity, global maximality, or
crossing uniqueness.  They do refute the attempted $q$-uniform extrapolation
of one preferred corner and motivate a sharper candidate $p_1$: the global
maximizer switches once as $q$ varies.

The implementation stores $p_0$ and $p_1$ as sibling candidates under the
same immutable anchor, rather than as revisions of one candidate record.
Accordingly, the arrow $p_0\rightsquigarrow_A p_1$ below denotes
portfolio-level candidate--mechanism co-evolution, not an in-place edit.

\begin{table*}[t]
\centering
\scriptsize
\setlength{\tabcolsep}{4pt}
\begin{tabular}{p{0.13\textwidth}p{0.23\textwidth}p{0.27\textwidth}p{0.27\textwidth}}
\toprule
Field & $p_0$ & $p_1$ & Mathematical trigger \\
\midrule
Parameter scope & Fixed slice $q=1/2$ & Joint prism $q\in[1/4,1/2]$ & The preferred corner reverses between two admissible slices. \\
Conclusion & One fixed maximizing corner & One unique corner switch, with both corners tied at the switch & A $q$-uniform monotonicity sign in $a$ is incompatible with the attack data. \\
Switch localization & Not applicable & Unique $q_c\in(7/20,3/8)$ & The numerical crossing is used only to bound a theorem-level target. \\
Refutation target & Failure of the fixed-slice endpoint or extremizer clauses & Endpoint failure, a competing maximizer, or zero/multiple switches & The joint claim must have direct, in-scope falsifiers. \\
Required certificate & Weight-box boundary analysis & First-contact and global-extremizer analysis on a three-parameter prism & Corner samples cannot exclude interior, edge, or degenerate cells. \\
\bottomrule
\end{tabular}
\caption{Portfolio-level evolution from $p_0$ to $p_1$ under mechanism support and counterexample pressure.}
\label{tab:ill-candidate-evolution}
\end{table*}

Thus the update is substantive: the mathematical object remains fixed by
$A$, while the candidate scope and the required mechanism both change in
response to an identified failure of extrapolation.

\subsubsection{Joint mechanism and honest support boundary}

The accepted joint mechanism extends the spectral reduction symbolically in
$q$.  It is stored as one mechanism record; for exposition only, we display
two derived nodes $m_{\mathrm{spec}}\to m_{\mathrm{ext}}$.  The first gives
\begin{equation}
\det(tI_8-M_d)
=(t-1)\bigl(t-(1-q)^2\bigr)^2p(t;q,a,b,\alpha),
\label{eq:ill-joint-factor}
\end{equation}
where $p$ is monic quintic and $(1-q)^2\in[1/4,9/16]$.  For
$0<\alpha\leq2$,
\begin{equation}
p(1)=2ab\alpha q^2
\left((2-q)^2(\alpha-2)^2+\alpha bq^2\right)>0,
\label{eq:ill-p-plus-one}
\end{equation}
and the accepted exact certificate gives $p(-1)<0$ on the full prism.
Consequently, any remaining unit-circle contact in this slab is nonreal.

For such a contact, put
\begin{equation}
H=(1-iu)^5p\!\left(\frac{1+iu}{1-iu}\right)=R+iI
\end{equation}
and define
\begin{equation}
\begin{split}
\Delta&=R_uI_\alpha-R_\alpha I_u,\\
N_a&=I_uR_a-R_uI_a,
\qquad N_b=I_uR_b-R_uI_b.
\end{split}
\end{equation}
At a simple contact with $\Delta\ne0$, implicit differentiation yields
\begin{equation}
\partial_a\alpha=\frac{N_a}{\Delta},
\qquad
\partial_b\alpha=\frac{N_b}{\Delta}.
\label{eq:ill-endpoint-derivatives}
\end{equation}
Every smooth interior extremum is therefore contained in the finite algebraic
system $R=I=N_a=N_b=0$.  Edge extrema use the corresponding tangential
numerator; the degeneracy system $R=I=\Delta=0$ is retained separately.

The two expository interfaces are
\begin{equation}
\begin{aligned}
\operatorname{Req}(m_{\mathrm{spec}})
 &=\mathcal P_A\cup\{0<\alpha\leq2\},\\
\operatorname{Prov}(m_{\mathrm{spec}})
 &=\{\text{quintic factorization},\ t\ne\pm1\},\\
\operatorname{Gap}(m_{\mathrm{spec}})
 &=\{\text{reachability, first contact, simplicity}\},\\
\operatorname{Req}(m_{\mathrm{ext}})
 &=\operatorname{Prov}(m_{\mathrm{spec}})
   \cup\{\Delta\ne0,\text{ smooth branch}\},\\
\operatorname{Prov}(m_{\mathrm{ext}})
 &=\{R=I=N_a=N_b=0\},\\
\operatorname{Gap}(m_{\mathrm{ext}})
 &=\{\text{cell isolation, switch uniqueness}\}.
\end{aligned}
\label{eq:ill-joint-interface}
\end{equation}
They compose as
\begin{equation}
\begin{aligned}
\mathcal P_A&\xrightarrow{m_{\mathrm{spec}}}
\text{nonreal quintic contacts}\\
&\xrightarrow{m_{\mathrm{ext}}}
\text{finite extremum cells}.
\end{aligned}
\label{eq:ill-mechanism-graph}
\end{equation}

The Coverage Judge returned
\begin{equation}
\rho=\text{\textsc{SubstantivePartial}}.
\end{equation}
The support is substantive because an eight-dimensional spectral search is
reduced to an exact quintic contact problem and finite algebraic extremum
systems.  It remains partial for two precise reasons:
\begin{enumerate}
    \item prove connected-from-zero first contact, reachable rank, finite
    positivity, and continuity of $\alpha_\star$ throughout the prism,
    including $\alpha=2-q$ and $\Delta=0$;
    \item isolate every interior, edge, corner, and degenerate cell and prove
    that the two proposed corner branches cross exactly once, with no third
    maximizer at the crossing.
\end{enumerate}
A root of the Cayley system is therefore only a candidate contact.  The
mechanism has not shown that the root is reachable, first, simple, or outward
transverse.  This is the support boundary that prevents a plausible reduction
from being counted as a solution.

\subsubsection{Statement-blind challenge and prover-facing projection}

The Final Challenger received the candidate statement without the mechanism
graph or exploration history.  It tested
$(q,a,b,\alpha)=(1/4,3/4,1/4,1/10)$ under the exact recurrence.  The
initialized Krylov space had rank eight, the optimizer vector was fixed, and
after removing only that optimizer root the approximate moduli were
\begin{equation}
0.740272^{(2)},\quad0.807797^{(2)},\quad0.810479,
\quad0.5625^{(2)}.
\end{equation}
This attack found a stable neighborhood rather than a counterexample.  The
Challenger also passed definition closure, target duality, source-relative
value, and bounded difficulty, and returned \texttt{Accept} for prover
handoff.  This acceptance does not establish the corner switch.

To state the projection, let $M_d=T_1T_0$ act on
$s=(x,y,h_x,h_y)\in\mathbb R^8$ and define
\begin{equation}
\begin{split}
\mathcal I&=\{(x^0,x^0-r,0,0):x^0,r\in\mathbb R^2\},\\
K(M_d)&=\operatorname{span}\{M_d^jv:v\in\mathcal I,\ 0\leq j\leq7\},\\
O&=\operatorname{span}\{(\mathbf1,0,\mathbf1,0)\}.
\end{split}
\label{eq:ill-reachable-defs}
\end{equation}
Let $C(q,a,b)$ be the tracker-reachably Schur-stable component adjacent to
$\alpha=0$, set $\alpha_\star(q,a,b)=\sup C(q,a,b)$, and write
\begin{equation}
\operatorname{Max}(q)=
\operatorname*{arg\,max}_{(a,b)\in[1/4,3/4]^2}
\alpha_\star(q,a,b).
\end{equation}

\par\smallskip\noindent
\textbf{Final prover-facing conjecture $q_1$.}
Under~\eqref{eq:ill-corner-mixing}--\eqref{eq:ill-corner-recurrence}, for
every $(q,a,b)\in[1/4,1/2]\times[1/4,3/4]^2$, the component $C(q,a,b)$
exists, its endpoint is finite and positive, and $\alpha_\star$ is continuous
on the closed prism.  There is a unique $q_c\in(7/20,3/8)$ such that
\begin{equation}
\operatorname{Max}(q)=
\begin{cases}
\{(3/4,1/4)\},&1/4\leq q<q_c,\\
\{(3/4,1/4),(1/4,1/4)\},&q=q_c,\\
\{(1/4,1/4)\},&q_c<q\leq1/2.
\end{cases}
\label{eq:ill-final-corner-conjecture}
\end{equation}
The refutation branch permits any admissible witness to a missing,
nonpositive, infinite, or discontinuous endpoint; an interior, edge, or
different-corner maximizer; no switch or multiple switches; or an additional
maximizer at $q_c$.  A floating-point comparison or failure of one proposed
mechanism is not a refutation.
\par\smallskip

In the pipeline notation,
\begin{equation}
q_1=\Theta(A,p_1,G').
\end{equation}
The projection closes notation and makes the two logical targets explicit.  It
does not add assumptions, remove parameter cells, or weaken the maximizer law.

\subsubsection{QED handoff and audited outcome}

Blind QED independently recovered a useful reachable-rank reduction.  If
$E:\mathbb R^4\to\mathbb R^8$ injects the allowed initialization coordinates,
then exact block algebra gives
\begin{equation}
\det[E,M_dE]=q^4(q-2)^2(\alpha+q-2)^2.
\label{eq:ill-krylov-minor}
\end{equation}
Thus $K(M_d)=\mathbb R^8$ away from $\alpha=2-q$.  A proof candidate then
identified a seven-dimensional invariant reachable space on the exceptional
locus, recovered the quintic factorization in~\eqref{eq:ill-joint-factor},
and supplied finite certificate artifacts for its proposed global sign and
switch argument.

This is substantive, independently checkable partial progress, not a verified
proof of the conjecture.  The structural review completed, but the detailed
review stopped while independently rerunning the exact global certificate and
never issued a terminal verdict.  The remaining audit obligations were the
defining polynomials, complete rational leaf data, sign controls, equality
cells, and resultant isolation.

The subsequent unresolved-problem audit classified the case as
\begin{equation}
\boxed{\text{\textsc{Unresolved--TechnicalDifficulty}}}.
\label{eq:ill-corner-outcome}
\end{equation}
No formulation ambiguity, missing assumption, or direct counterexample had
been established.  The unresolved core is the exact global certification
connecting the accepted spectral reductions to first-contact order,
continuity, global-extremizer exclusion, and uniqueness of the switch.

\subsection{Terminal QED Closure by Verified Refutation}
\label{app:ill-omwu-refutation}

The conjecture ``Punctured global OMWU robustness radius for perturbed
rock--paper--scissors'' provides the complementary case: QED found and
verified a refutation after repairing a local error in the first proof.

\subsubsection{Prover-facing statement and refutation target}

Let
\begin{equation}
R=\begin{pmatrix}
0&-1&1\\
1&0&-1\\
-1&1&0
\end{pmatrix},
\qquad
C=\begin{pmatrix}
2&-1&-1\\
-1&2&-1\\
-1&-1&2
\end{pmatrix},
\label{eq:ill-omwu-matrices}
\end{equation}
and give the column player payoff matrix $M_\epsilon=-R+\epsilon C$.
Fix $\eta=1/4$.  From a two-step history
$h=(x(-1),x(0),y(-1),y(0))\in(\operatorname{int}\Delta_3)^4$, exact OMWU
uses the payoff vectors
\begin{equation}
\begin{aligned}
g^x_t&=2Ry(t)-Ry(t-1),\\
g^y_t&=2M_\epsilon^Tx(t)-M_\epsilon^Tx(t-1),
\end{aligned}
\end{equation}
and the normalized updates
\begin{equation}
\begin{aligned}
x_i(t+1)&=
\frac{x_i(t)e^{g^x_{t,i}/4}}{\sum_kx_k(t)e^{g^x_{t,k}/4}},\\
y_j(t+1)&=
\frac{y_j(t)e^{g^y_{t,j}/4}}{\sum_ky_k(t)e^{g^y_{t,k}/4}}.
\end{aligned}
\label{eq:ill-exact-omwu}
\end{equation}
Let $u=(1/3,1/3,1/3)$ and
\begin{equation}
d_h^\epsilon(t)=
\left(\|x(t)-u\|_2^2+\|y(t)-u\|_2^2\right)^{1/2}.
\end{equation}
For $|\epsilon|\leq1/10$, define
\begin{equation}
\begin{split}
\mathcal G&=\{\epsilon:\ d_h^\epsilon(t)\to0
\text{ for every }h\in(\operatorname{int}\Delta_3)^4\},\\
r_\star&=\sup\{r\in[0,1/10]:
0<|\epsilon|<r\Longrightarrow\epsilon\in\mathcal G\}.
\end{split}
\label{eq:ill-omwu-radius}
\end{equation}
The positive branch asks for an $r_0>0$ giving convergence from every
strictly interior history whenever $0<|\epsilon|<r_0$.  The explicit
refutation branch asks for distinct nonzero $\epsilon_n\to0$ and strictly
interior histories $h_n$ satisfying
\begin{equation}
\limsup_{t\to\infty}d_{h_n}^{\epsilon_n}(t)>0
\qquad\text{for every }n.
\label{eq:ill-omwu-refutation-target}
\end{equation}
Boundary histories, local instability, finite escape, and diverging transient
times do not meet this target.

\subsubsection{Exact refutation mechanism}

Write $\epsilon=-\delta$ with $\delta>0$, set
$P=I-\tfrac13\mathbf1\mathbf1^T$, and let $\sigma$ be softmax.  Centered
logits $z_t=P\log x(t)$ and $w_t=P\log y(t)$ satisfy the exact lift
\begin{equation}
\begin{aligned}
z_{t+1}
 &=z_t+\frac14R\bigl(2\sigma(w_t)-\sigma(w_{t-1})\bigr),\\
w_{t+1}
 &=w_t+\frac14(R+\epsilon C)
 \bigl(2\sigma(z_t)-\sigma(z_{t-1})\bigr).
\end{aligned}
\label{eq:ill-centered-logits}
\end{equation}
This change of variables is exact in both directions for finite logits, so it
does not introduce boundary histories or alter the OMWU convention.

At large logit scale, softmax approaches the maximizing action and the lifted
dynamics approach a piecewise-linear tropical system.  QED found the six-switch
itinerary
\begin{equation}
\begin{aligned}
(1,3)&\to(1,2)\to(3,2)\to(3,1)\\
&\to(2,1)\to(2,3)\to(1,3).
\end{aligned}
\label{eq:ill-six-switch}
\end{equation}
Exact composition of the six constant-velocity chambers gives a Poincar\'e
return $p\mapsto M_\delta p$.  Its exact algebra establishes
\begin{equation}
\begin{aligned}
\det M_\delta&=1,\\
\chi_\delta(1)&=-\frac{243}{32}\delta(1+3\delta^2)<0,\\
\lambda_\delta&=1+\frac{18}{5}\delta+O(\delta^2)>1.
\end{aligned}
\label{eq:ill-return-expansion}
\end{equation}
The expanding eigenvalue is simple for all sufficiently small $\delta>0$;
the other two multipliers have modulus $\lambda_\delta^{-1/2}<1$.  At
$\delta=0$, every intended switch is transverse with crossing speed $1/2$,
and the nonswitching action is separated from the tie by the exact margin
$3$.  Hence a projective neighborhood of the expanding eigenray preserves the
same itinerary for small positive $\delta$.

The optimistic two-step term does not accumulate over a circuit.  Summing
\eqref{eq:ill-centered-logits} gives the exact telescoping identity
\begin{equation}
z_n-z_0
=\frac14\sum_{t=0}^{n-1}R\sigma(w_t)
+\frac14R\bigl(\sigma(w_{n-1})-\sigma(w_{-1})\bigr),
\label{eq:ill-telescoping}
\end{equation}
with an analogous identity for $w$.  After division by a logit radius
$A\to\infty$, the endpoint term vanishes.  Away from the six transverse tie
layers, the softmax error is exponentially small; the tie layers occupy
vanishing scaled time and have no competing switch because of the strict
third-action margin.  The exact one-circuit return is therefore
\begin{equation}
\mathcal R_\delta(p)=M_\delta p+o(\|p\|)
\label{eq:ill-exact-return}
\end{equation}
uniformly on a compact cone around the expanding eigenray.  The spectral gap
then yields a forward-invariant subcone on which the exact return expands the
radial coordinate by a factor greater than one.

Starting at any sufficiently large but finite radius in that cone gives one
fixed exact orbit, not a family of increasingly long transients.  Finite
logits yield a strictly interior four-profile history.  Along successive
circuits the logit radius diverges, and at the designated phase midpoint the
profiles tend to $(e_1,e_3)$.  Since
\begin{equation}
\left(\|e_1-u\|_2^2+\|e_3-u\|_2^2\right)^{1/2}
=\sqrt{4/3}>1,
\end{equation}
the resulting exact orbit satisfies
\begin{equation}
\limsup_{t\to\infty}d_h^{-\delta}(t)\geq1.
\label{eq:ill-persistent-orbit}
\end{equation}
Choose a sufficiently small $\bar\delta>0$ for which the cone construction
holds and set
\begin{equation}
\epsilon_n=-\frac{\bar\delta}{n+1}.
\end{equation}
For every $n$, the preceding construction supplies a strictly interior
history with the persistent limsup in~\eqref{eq:ill-persistent-orbit}.
The bad parameters are distinct and converge to zero, so every positive
candidate radius contains one.  Therefore
\begin{equation}
\boxed{r_\star=0}.
\label{eq:ill-omwu-result}
\end{equation}

\subsubsection{Verifier-requested repair}

The first proof attempt contained one local numerical error: it described the
nonswitching margin in the tropical itinerary as $6$, while exact
reconstruction of the displayed switching endpoints gives $3$.  The regulator
returned \texttt{REVISE\_PROOF}.  It did not accept the otherwise plausible
chain merely because the argument needed only a positive margin.

The revised proof replaced $6$ by $3$ and made the robustness step explicit:
by continuity, a margin greater than $3/2$ persists in a sufficiently small
neighborhood.  Independent reconstruction then confirmed the six hitting
times, return matrix, and final goal.  Thus \eqref{eq:ill-omwu-result} is a
verified refutation, unlike the incomplete certificate in
Section~\ref{app:ill-corner-switch}.

\subsection{What the Two Cases Establish}
\label{app:ill-two-case-summary}

\begin{table*}[t]
\centering
\small
\setlength{\tabcolsep}{5pt}
\begin{tabular}{p{0.19\textwidth}p{0.34\textwidth}p{0.37\textwidth}}
\toprule
Property & Directed-weight corner switch & OMWU robustness radius \\
\midrule
Primary role & Candidate--mechanism co-evolution under a fixed anchor & Repair and verification of an explicit refutation \\
Established support & Exact spectral reduction with a stated support boundary & Complete centered-logit, tropical-return, and invariant-cone argument \\
Remaining core & First-contact/continuity certification and unique global-maximizer switch & None in the accepted proof \\
Final status & Unresolved--TechnicalDifficulty & Verified refutation with $r_\star=0$ \\
\bottomrule
\end{tabular}
\caption{The two examples serve complementary evidentiary roles.}
\label{tab:ill-two-case-summary}
\end{table*}

The comparison isolates the substantive difference between useful partial
support and a completed mathematical resolution: the first trace narrows the
problem to a specific global certificate, whereas the second closes its stated
refutation branch.


\providecommand{\system}{\textsc{MECA}}

\section{Prompt and Skill Design}
\label{app:prompt-skill-design}

\system{} does not use a single monolithic prompt or an unconstrained running
conversation.  Each model call combines a stage-specific \emph{role prompt}, a
possibly empty set of reusable \emph{mathematical skills}, and a frozen snapshot
of the current research state.  These components have deliberately different
responsibilities.  The role prompt states what decision the agent is authorized
to make; the skills specify how the relevant mathematical reasoning should be
carried out; the run artifacts contain the particular sources, candidate, and
mechanisms to be examined; and the host controller decides whether the returned
record is valid enough to change the pipeline state.  In short,

\begin{center}
\small
\begin{tabular}{c@{\;}c@{\;}c@{\;}c@{\;}c}
role prompt & $+$ & skill references & $+$ & frozen artifacts \\
$P_r$ && $\Sigma(r,t)$ && $X_t$
\end{tabular}
\end{center}

defines the model-visible call for role $r$ at stage $t$, while promotion,
repair, rejection, and checkpoint updates remain host-owned operations.  This
separation is important: a well-formed JSON response is only a proposal or an
audit result, not evidence that the mathematics is correct.

\subsection{Four-Layer Instruction Architecture}

Table~\ref{tab:meca-instruction-layers} summarizes the instruction boundary.
A \emph{prompt} is role-specific and action-oriented.  A \emph{skill} is a
versioned procedural reference such as a source-subtraction test, a
counterexample protocol, or a mechanism-composition audit.  A skill is not an
additional agent, a learned module, or a source of mathematical facts.  It
does not update state and it cannot authorize an assumption.  Finally, the
host controller enforces the output schema, stage transition, lineage, and
budget independently of the language model.

\begin{table*}[t]
\centering
\small
\setlength{\tabcolsep}{5pt}
\begin{tabular}{p{0.14\textwidth}p{0.20\textwidth}p{0.36\textwidth}p{0.20\textwidth}}
\toprule
Layer & Main question & Contents & Delivery and authority \\
\midrule
Role prompt & What must this agent decide? & Role, visible inputs, permitted actions, prohibited moves, and the exact response contract. & One prompt file is selected for each call; it may propose a state transition but cannot commit it. \\
Skill bundle & How should the mathematics be tested? & Reusable procedures for retrieval, source subtraction, seed construction, falsification, mechanism discovery, interface replay, composition, and theoremization. & Only role-relevant skill paths are supplied.  Skills advise reasoning but do not provide evidence or state authority. \\
Run artifacts & What mathematical instance is under review? & Frozen source pack, exact candidate revision, reviewed mechanisms, critic records, and optional memory. & Artifacts stay on disk and are named by absolute paths; their bodies are not duplicated in the prompt. \\
Host controller & May the result change the run? & Parsing, schema checks, revision hashes, legal decisions, repair budgets, invalidation, and checkpoint commits. & Deterministic code validates the response and applies or rejects the transition. \\
\bottomrule
\end{tabular}
\caption{The four layers used to construct a \system{} model call.  The
division prevents procedural advice, run memory, and schema compliance from
being mistaken for mathematical evidence.}
\label{tab:meca-instruction-layers}
\end{table*}

The distinction also makes the design easier to audit.  Changing the output
contract of the Mechanism--Frontier Explorer changes one role prompt; improving
the boundary-case protocol changes one skill shared by several critics; and
changing a candidate changes only the referenced artifact.  The three changes
therefore have separate identities in the run manifest.

\subsection{Prompt Design Principles}
\label{app:prompt-principles}

All production prompts follow five principles.

\paragraph{Stage-local authority.}
Each role has a finite action space.  A Seed Scout may generate, repair, or
replace one seed, but it may not prescribe the proof mechanism.  The
Mechanism--Frontier Explorer may propose one mechanism or revise the frontier,
but it may not certify its own proposal.  A Coverage Judge may select and
compose reviewed mechanisms, but it may not rewrite them.  Restricting each
call to one state transition makes failures attributable and prevents a single
response from silently generating, judging, and accepting its own argument.

\paragraph{Mathematical work before labels.}
Prompts ask for the first nonroutine derivation, an explicit positive toy case,
and a hostile in-scope test before asking for a verdict.  Phrases such as
``plausible by standard arguments'' or ``no counterexample was found'' do not
support promotion.  A surviving counterexample audit must identify a concrete
instance, verify its assumptions, calculate the target observable, and state
what the calculation does or does not falsify.

\paragraph{Explicit epistemic boundaries.}
Prompts distinguish primitive assumptions, established source results,
candidate mechanism outputs, and unresolved obligations.  A desired
conclusion, certificate, or downstream mechanism output may not be inserted as
an assumption.  Likewise, an acceptance receipt records that a mechanism
passed the configured review; it does not turn the mechanism into a theorem.

\paragraph{Bounded repair.}
A repair prompt receives the exact rejected revision and a revision-bound
critic contract.  It may correct the named mathematical or statement defect,
but it may not change portfolio slots, broaden the novelty claim, replace the
research object, or weaken the frontier merely to fit the current mechanism.
If the defect requires a different direction, the controller routes to
replacement or rejection instead of calling the change a repair.

\paragraph{Deliberate information asymmetry.}
Roles receive only the information required by their audit.  In particular,
the Final Challenger reads the candidate, target anchor, reviewed seed, and
source pack, but not the mechanism graph, coverage decision, or critic
receipts.  The independent prover probes receive the prover-facing statement
rather than the construction trace.  These restrictions test the statement
itself instead of inviting later agents to repeat the pipeline's preferred
route.

\subsection{Runtime Assembly and File-Referenced Context}
\label{app:path-only-payload}

Executable role prompts are loaded from
\path{src/prompts/single_mechanism/multi_mechanism/}.  Mathematical skills are
stored separately under \path{src/prompts/common/skills/}.  The runtime does
not concatenate skill contents, source packs, memory, branch blackboards, or
artifact bodies into the prompt.  Instead, it sends a compact control envelope
containing routing scalars and absolute file references.  A representative
envelope is:

\noindent\begin{minipage}{\columnwidth}
{\scriptsize
\begin{verbatim}
{
  "stage": "joint_frontier_mechanism",
  "record_id": "candidate-id",
  "input_revision_hash": "<sha256>",
  "source_pack_file": "/abs/run/source_pack.json",
  "artifact_files": {
    "candidate": "/abs/run/candidate.json",
    "mechanisms": "/abs/run/mechanisms.jsonl"
  },
  "skill_reference_files": [
    "/abs/repo/skills/mechanism.md"
  ],
  "memory_files": {
    "run_memory": "/abs/run/memory.jsonl"
  }
}
\end{verbatim}
}
\end{minipage}

The role process reads the referenced files when needed.  This design keeps
the stable instruction text distinct from large and evolving state, binds a
call to an exact revision, and avoids producing different effective prompts
merely because a JSONL history has grown.  It also makes the evidence boundary
inspectable: run memory can discourage repetition, but it is process context,
not literature evidence, and it cannot enlarge the frozen source pack.

Because local files are part of the interface, live runs use a
local-file-capable model process.  The asset registry fails closed when a required prompt
or skill is missing or when an unregistered instruction file appears in the
production directories.

\subsection{Prompt Families Across the Pipeline}
\label{app:prompt-families}

Table~\ref{tab:meca-role-prompts} gives the principal prompt families.  Repair
and replacement prompts are separate files because their authority is narrower
than that of the corresponding generation prompt.  Mathematical and research
critics are also separate calls: the former attacks correctness and
counterexamples, whereas the latter attacks source-relative novelty, value,
and readiness.  A passing result on one axis cannot conceal a blocker on the
other.

\begin{table*}[t]
\centering
\small
\setlength{\tabcolsep}{4pt}
\begin{tabular}{p{0.12\textwidth}p{0.19\textwidth}p{0.26\textwidth}p{0.22\textwidth}p{0.13\textwidth}}
\toprule
Stage & Prompt family & Mathematical task & Allowed result & Main skill family \\
\midrule
Seed planning & Portfolio Planner & Partition a broad focus into distinct source-relative directions without proposing the theorem or mechanism. & Exactly $m$ fixed slot briefs and overlap exclusions. & Source-gap discovery, route taxonomy, contribution boundary. \\
Seed construction & Seed Scout & Subtract the closest result, expose a bounded delta, list authorized premises, and give a bottleneck and falsification surface. & One new, repaired, or replacement Raw Seed. & Evidence ledger, claim-shape contract, failure tests, memory-aware selection. \\
Seed review & Mathematical and Research Critics & Independently test derivability, counterexamples, source coverage, research value, and handoff readiness. & Per-dimension survive, repair, reject, or incomplete decisions. & Correctness, counterexample, novelty/value, and readiness attacks. \\
Co-exploration & Mechanism--Frontier Explorer & Jointly revise the candidate and propose one mechanism with an explicit proof object, interface, calculation, and residual gap. & Propose/add mechanism, revise, split, emit, reject, or quarantine a resolution. & Joint frontier--mechanism discovery, evidence, typed-interface discipline. \\
Independent review & Mechanism Critic and Coverage Judge & Replay local derivations, then compose only accepted interfaces and locate the exact remaining gap. & Promote/repair/reject a mechanism; or request support, revise, emit, split, reject, or quarantine. & Mechanism audit, bundle composition, closure and chain audits. \\
Final construction & Final Challenger and prover probes & Audit statement closure, direct counterexamples, value, difficulty, and independent solvability without relying on the preferred mechanism route. & Accept, bounded repair, split, reject, resolution quarantine, and probe outcome. & Statement audit and theoremization constraints; mechanism skills are withheld from isolated checks. \\
\bottomrule
\end{tabular}
\caption{Principal \system{} prompt families and their authority.  The table
groups prompts by mathematical function; the released prompt files contain the
exact JSON response schemas.}
\label{tab:meca-role-prompts}
\end{table*}

\subsection{Condensed Operative Prompt Templates}
\label{app:condensed-prompts}

For readability, the templates below omit repeated JSON field declarations and
runtime bookkeeping.  They preserve the operative mathematical instructions
and decision boundaries.  The released Markdown files are authoritative for
the exact prompts used in the experiments.

\noindent\begin{minipage}{\columnwidth}
\paragraph{Seed Scout.}
\begin{quote}
\small
Read the frozen source pack, the assigned portfolio brief, and the permitted
run memory.  Produce exactly one research-grade direction, not a final theorem
or proof plan.  Identify the closest result and subtract what it already
establishes.  State the remaining bounded, nonroutine delta and any unresolved
source-coverage risk.  Name the central object, controlled quantity,
assumption-to-conclusion direction, and every authorized premise.  Give the
decisive bottleneck, one informative toy case, and at least two concrete
failure tests.  Do not include a named mechanism, bridge recipe, desired
conclusion as an assumption, or prose outside the required JSON object.
\end{quote}
\end{minipage}

\noindent\begin{minipage}{\columnwidth}
\paragraph{Raw-seed critics.}
\begin{quote}
\small
Read the exact seed revision.  The mathematical critic must locate the weakest
implied step, attempt its first nontrivial derivation, and instantiate an
in-scope toy or boundary family; it must verify assumptions and calculate the
target observable.  The research critic must identify the closest result,
subtract routine consequences, and decide whether a meaningful delta and a
usable next-stage problem remain.  Return all requested dimensions even when
one fails.  Diagnose but do not repair the artifact.
\end{quote}
\end{minipage}

\noindent\begin{minipage}{\columnwidth}
\paragraph{Mechanism--Frontier Explorer.}
\begin{quote}
\small
Review the complete candidate state and choose exactly one legal transition.
When proposing a mechanism, state an actual proof object, its key move, the
first nonroutine calculation, ordered derived steps, exactly one residual
bottleneck, and a typed requirement--consequence interface.  Run both a
positive calculation and a hostile in-scope test.  Revise the frontier only
when the mathematics exposes a false, vague, trivial, source-covered, or
unbounded target; never weaken it merely to make the current route succeed.
Emit only after reviewed mechanisms give substantive support and a finite
nonroutine core remains.  Quarantine a credible proof or assumption-checked
counterexample rather than presenting a solved claim as open.
\end{quote}
\end{minipage}

\noindent\begin{minipage}{\columnwidth}
\paragraph{Mechanism Critic and Coverage Judge.}
\begin{quote}
\small
Replay the mechanism from its declared inputs.  Check domains, quantifiers,
signs, constants, limiting regimes, circularity, source subtraction, a worked
positive cell, and a worked obstruction cell.  Promotion requires a correct,
closed, candidate-relevant contribution, not merely a complete schema.  For
coverage, ignore proposal order, choose a minimal nonredundant set, and replay
interfaces in topological order.  An output becomes available only after a
selected predecessor establishes the identical mathematical quantity.  If a
hidden lemma or semantic jump remains, request one missing kind of support
without prescribing its solution.
\end{quote}
\end{minipage}

\noindent\begin{minipage}{\columnwidth}
\paragraph{Final Challenger.}
\begin{quote}
\small
Read only the candidate, immutable anchor, reviewed seed, and frozen source
pack.  Do not inspect mechanisms or their reviews.  Close every definition,
domain, quantifier, and assumption; verify that the positive and refutation
targets use the same scope; calculate a direct in-scope counterexample test;
subtract the closest source result; and identify both a finite progress
certificate and the remaining nonroutine bottleneck.  A failed proof route is
not a refutation of the candidate.  Accept, request bounded statement repair,
split, reject, or quarantine a genuine resolution.
\end{quote}
\end{minipage}

These templates intentionally combine positive construction requirements with
explicit prohibitions.  The prohibitions encode recurrent failure modes:
assumption smuggling, novelty inflation, mechanism-shaped seeds, self-review,
frontier weakening, circular dependency, and the misclassification of a failed
route as a counterexample to the target.

\subsection{The Mathematical Skill Library}
\label{app:skill-library}

Skills turn general role instructions into repeatable mathematical procedures.
They are attached selectively rather than globally: the Seed Scout needs
source-gap and attack-surface skills, the Mechanism Critic needs derivation and
interface audits, and the Final Challenger is intentionally denied the
mechanism bundle.  Table~\ref{tab:meca-skill-families} summarizes the main
families and representative files.

\begin{table*}[t]
\centering
\small
\setlength{\tabcolsep}{4pt}
\begin{tabular}{p{0.16\textwidth}p{0.31\textwidth}p{0.43\textwidth}}
\toprule
Skill family & Representative files & Operational content \\
\midrule
Evidence and source grounding & {\raggedright \path{literature_web_search}, \path{evidence_ledger}, \path{source_gap_seed_discovery}\par} & Locate the closest result inside the allowed boundary, distinguish cited fact from hypothesis, and retain source locators and coverage risks. \\
Seed construction & {\raggedright \path{seed_route_taxonomy}, \path{claim_shape_contract}, \path{attack_surface_failure_tests}, \path{contribution_boundary}\par} & Choose a legitimate source-relative route, expose the central object and implication, forbid hidden target assumptions, and require concrete failure pressure. \\
Adversarial review & {\raggedright \path{attack_correctness}, \path{attack_counterexamples}, \path{attack_novelty_value}, \path{attack_readiness_for_next_stage}\par} & Separate four review dimensions and require an attempted derivation, an assumption-checked calculation, source subtraction, and a usable but not over-solved handoff. \\
Mechanism discovery & {\raggedright \path{expert_mechanism_discovery}, \path{proof_pattern_mechanism_extraction}, \path{joint_frontier_mechanism_discovery}\par} & Compare competing proof objects, transfer only the mathematical core of a source argument, expose the first nonroutine step, and let calculations drive frontier revision. \\
Interfaces and composition & {\raggedright \path{joint_mechanism_interface}, \path{mechanism_audit}, \path{bundle_composition}, \path{bundle_closure_audit}\par} & Check exactly what each mechanism consumes and produces, reject circular or unavailable inputs, compose dependencies topologically, and recompute missing support independently. \\
End-to-end task audit & {\raggedright \path{joint_chain_frontier_audit}, \path{theoremization}, \path{semi_open_audit}\par} & Replay the whole chain from anchor assumptions, close the final statement, preserve positive/refutation duality and provenance, and verify that a finite unresolved core remains. \\
\bottomrule
\end{tabular}
\caption{Mathematical skill families used by \system{}.  For compactness, the
table shows file stems without the \texttt{.md} suffix and omits a directory
prefix when the family makes it unambiguous.  Only the registered subset is
exposed to a role.}
\label{tab:meca-skill-families}
\end{table*}

\subsubsection{Source subtraction and contribution boundary}

The source-grounding skills operationalize novelty as a theorem-level
difference rather than a topic-level similarity judgment.  Given a proposed
claim $C$ and the closest established result $S$, an agent first removes the
consequences of $S$ that follow by routine specialization, notation change, or
standard closure.  The residue
\[
    \Delta(C;S)
    = \operatorname{Claims}(C)
      \setminus \operatorname{RoutineClosure}(S)
\]
must correspond to a concrete change in assumptions, geometry, regime, rate,
complexity, obstruction, or theorem class.  Formulation, unification,
application, adaptivity, sharpness, or diagnostic value counts only when it
changes such a mathematical boundary.  Uncertain coverage is recorded as a
risk; it is not converted into a confident novelty claim.

\subsubsection{Concrete falsification protocol}

The counterexample skills require more than an invitation to ``look for edge
cases.''  An attack record must specify a family and its parameters, verify
that the primitive assumptions hold, identify the observable whose sign,
bound, rank, limit, or asymptotic order is at issue, and perform the relevant
calculation.  The result is classified as breaking the candidate, narrowing
its admissible scope, or surviving the named test.  When no concrete test can
be extracted from a record, the audit is incomplete rather than favorable.

\subsubsection{Mechanism interface and chain replay}

For a mechanism $m$, the interface skills distinguish
\[
    \operatorname{Req}(m), \qquad
    \operatorname{Prov}(m), \qquad
    \operatorname{Gap}(m).
\]
Every nonprimitive requirement must be matched to the exact consequence of an
earlier mechanism, including its domain, quantifiers, parameter regime, and
epistemic status.  The critic then locates the first equation or inference that
uses the dependency and replays the derivation.  Coverage is recomputed in
topological order rather than inferred from mechanism names or proposal order.
Thus, several locally plausible arguments can still fail global composition
when their constants, scopes, or quantifiers do not align.

\subsubsection{Theoremization without closing the problem}

The theoremization skills convert an audited partial chain into a
self-contained prove-or-refute statement.  They may expand definitions,
instantiate authorized primitives, align notation, and expose routine
projection steps.  They may not add a new lemma, certificate, invariant,
transfer principle, or assumption.  The positive target and refutation witness
must live under the same scope, and failure of a suggested mechanism must not
be presented as refutation of the target.  The final task is retained only when
the chain provides substantive support while leaving a named, finite, and
genuinely nonroutine obligation.

\subsection{Memory Is Not a Skill}
\label{app:memory-versus-skills}

Run memory records compact negative lessons such as a failed inequality, an
invalid transfer, a source-covered direction, or a counterexample family.  It
helps later agents avoid repeating the same failed route.  A skill, by
contrast, is a stable procedure for performing an attack or construction.  The
two are never merged: memory may suggest which test to run, but only a fresh
assumption check and calculation can support the current decision.  Branch
blackboards and critic summaries are governed by separate reading and
compaction contracts, and they are passed by path only when memory is enabled.

This distinction prevents an accumulation of model-generated claims from
becoming an informal proof.  It also permits ablations of memory while holding
the prompts and mathematical skill library fixed.

\subsection{Reproducibility}
\label{app:prompt-reproducibility}

The implementation maintains an explicit registry from each role, and where
necessary each stage--role pair, to one prompt file and a fixed tuple of skill
files.  At startup, the registry checks that every required file exists and
that no unregistered Markdown file is silently active.  Experimental manifests
record the effective prompt, skill, runtime, source-pack, and configuration
identities.  Resuming a run is permitted only against the expected frozen
state; changing a parent revision invalidates dependent reviews and
compositions.

The prose templates in this section explain the design and retain the
decision-relevant instructions, but they are not a substitute for the released
runtime assets.  Exact reproduction should use the archived prompt and skill
files associated with the reported manifest, together with the same output
schemas and host-side gates.  This qualification is material because prompts,
skills, and controller semantics jointly determine the effective procedure;
holding only the natural-language prompt fixed does not fully specify a
\system{} run.

\section{Blind Problem Collection}
\label{app:blind-problem-collection}

This section reports the 100 statement-only mathematical problems used in the
open-ended evaluation.  The collection is grouped by the frozen audit outcome:
35 proved, 11 refuted, and 54 unresolved.  The labels summarize the independent
verification outcome and are not exposed to the system during conjecture
construction.  Only problem statements are included; proofs, refutations,
audit transcripts, and source identities are omitted.

\begingroup
\let\problemcollectionsection\section
\let\clearpage\relax
\renewcommand{\section}[1]{\subsubsection{#1}}

\subsection{Proved Problems}
\clearpage
\section{P001: Degree-two corruption threshold}
\label{problem:P001}

\begin{problem}
For integers \(r,n\), let \(\mathcal C_{r,n}\) denote the space of
\(n\times r\) real matrices of rank \(r\) whose rows have unit Euclidean norm,
and let \(\mathcal D_s\) denote the symmetric zero-diagonal matrices whose
off-diagonal support graphs have maximum degree at most \(s\).  Two factors
which differ by a common right orthogonal transformation represent the same
correlation matrix, so uniqueness below concerns the Gram matrix and the
sparse term rather than the factor itself.

Fix \(r=5\), \(n=13\), and \(s=2\), and consider a noiseless equality
\[
\begin{aligned}
M=UU^{\mathsf T}+S_\star=VV^{\mathsf T}+T,\\
\qquad,\\
U,V\in\mathcal C_{5,13},\quad S_\star,T\in\mathcal D_2 .
\end{aligned}
\]
The degree bounds on \(S_\star\) and \(T\) are separate; merely requiring
their union to have degree at most four is not sufficient.  If \(F\) is the
union of their support graphs and \(G=K_{13}\setminus F\), then
\(UU^{\mathsf T}\) and \(VV^{\mathsf T}\) agree on the diagonal and on every
edge of \(G\).  Call such a quadruple a degree-two ambiguity when
\(UU^{\mathsf T}\ne VV^{\mathsf T}\).

Genericity is imposed only on \(U\).  More precisely, in every relevant real
irreducible component \(X\) of the unit-row configuration space, the
exceptional set must be a proper real algebraic subset of the exact-rank-five
locus \(X^{(5)}\).  It must be fixed independently of the two support graphs
and of the competitor \(V\); no genericity or proximity assumption is allowed
for \(V\).  For reference, let \(\mathcal A_X\subseteq X^{(5)}\) be the set of
clean factors which occur in a degree-two ambiguity.  A factor is called
full spark if every five of its rows are linearly independent.

Decide whether, for every relevant component \(X\), there is a nonempty
relative real Zariski-open set \(\mathcal O\subseteq X^{(5)}\), chosen
independently of all supports and competitors, such that
\[
\begin{aligned}
U\in\mathcal O,\quad S_\star,T\in\mathcal D_2,\quad,\\
V\in\mathcal C_{5,13},\quad,\\
UU^{\mathsf T}+S_\star=VV^{\mathsf T}+T
\end{aligned}
\]
always imply
\[
 UU^{\mathsf T}=VV^{\mathsf T}
 \qquad\text{and}\qquad S_\star=T .
\]
The required conclusion is global Gram uniqueness, not merely infinitesimal
rigidity.

Finally,
\[
 s_c=\left\lceil\frac{n-r+1}{4}\right\rceil
     =\left\lceil\frac{13-5+1}{4}\right\rceil=3 .
\]
At degree three, four independent separator rows can be chosen and the
remaining nine rows split into groups of sizes four and five, with one group
reflected across the separator span, producing generic nonuniqueness.
Thus the degree-three construction supplies the upper side of the exact
threshold in this fixed rank-five cell.
\end{problem}

\clearpage
\section{P002: Universal mixed-term lower bound}
\label{problem:P002}

\begin{problem}
Consider finite-dimensional centralized composite minimization
\(\Phi=F+r\).  The function \(F\) is differentiable, possibly nonconvex, and
has \(L\)-Lipschitz gradient; \(r\) is proper, closed, and convex, and its
proximal mapping is available exactly without stochastic-gradient cost.
Assume that \(\Phi\) is bounded below, that an initial point \(x_0\) is given,
and that
\[
 \Phi(x_0)-\inf\Phi\leq\Delta .
\]
The regularizer must be genuinely nonsmooth: there must be a
\(z\in\operatorname{dom}\partial r\) and distinct
\(u,v\in\partial r(z)\).

Fix
\[
 p=\frac32,\qquad c=L=\sigma=\rho=\Delta=1,\qquad
 \eta=\frac cL=1 .
\]
At any point \(x\), a one-point oracle query draws a fresh seed \(\xi\),
returns \(g(x,\xi)\), and costs one response.  A paired query at \(x,y\)
draws one fresh seed and returns both \(g(x,\xi)\) and \(g(y,\xi)\), at a
cost of two responses.  Conditionally on the transcript, distinct fresh
seeds are independent, seed reuse is permitted only within such a declared
pair, and
\[
\begin{aligned}
\mathbb E[g(x,\xi)\mid\text{transcript}]=\nabla F(x),\\
\sup_x\mathbb E\|g(x,\xi)-\nabla F(x)\|^{3/2}\leq1
\end{aligned}
\]
\[
\begin{aligned}
\mathbb E\!\left[
{}\right.&\|g(x,\xi)-g(y,\xi)\\
&\left.{}-(\nabla F(x)-\nabla F(y))\|^2
\,\middle|\,\text{transcript}\right]
\leq\|x-y\|^2 .
\end{aligned}
\]
A finite marginal second moment at a single point is not assumed.

An algorithm may be any adaptive randomized centralized algorithm.  It is
given \(\varepsilon,x_0\), the declared parameter bounds, and exact proximal
access to \(r\), but it is not given the hidden instance, \(F\), or the noise
law.  No zero-respecting, span, or analogous restriction is imposed.  After a
finite charged transcript it declares a random output \(X\), whose
stationarity is measured by
\[
 G_\eta(x)=\eta^{-1}
 \bigl(x-\operatorname{prox}_{\eta r}(x-\eta\nabla F(x))\bigr).
\]
For \(0<\varepsilon\leq1/16\), let \(N_{\rm all}(\varepsilon)\) be the
infimum, over all such algorithms that satisfy
\(\mathbb E\|G_\eta(X)\|\leq\varepsilon\) uniformly on the complete normalized
instance class, of their smallest almost-sure charged-response cap.

Decide whether there are constants \(\kappa>0\) and
\(\varepsilon_0\in(0,1/16]\) such that
\[
 N_{\rm all}(\varepsilon)\geq
 \kappa\,\varepsilon^{-7/2}
 \qquad(0<\varepsilon\leq\varepsilon_0).
\]
\end{problem}

\clearpage
\section{P003: Rank-two attainment on the BDCA fiber}
\label{problem:P003}

\begin{problem}
Let \((a,b,c)\) be the unique positive real triple satisfying
\[
\begin{aligned}
\frac{1414213}{10^6}<a<\frac{1414214}{10^6},\\
\frac{1601231}{10^6}<b<\frac{1601232}{10^6},\\
\frac{3005144}{10^6}<c<\frac{3005145}{10^6}
\end{aligned}
\]
and
\[
a^2=2,\qquad b^2+ab-2-2a=0,\qquad c^2=2a+2b+3.
\]
Fix the ordered four-step schedule
\[
h_{\rm bdca}=(b,3/2,c,a).
\]

For any positive finite integer \(d\), any \(L,D>0\), and any convex,
differentiable, \(L\)-smooth function
\(f:\mathbb R^d\to\mathbb R\) having a minimizer \(x_\star\), start from a
point satisfying \(\|x_0-x_\star\|\leq D\) and perform
\[
x_{t+1}=x_t-\frac{h_t}{L}\nabla f(x_t),\qquad 0\leq t<4.
\]
Write
\[
E_4=\frac{f(x_4)-f(x_\star)}{LD^2},
\qquad
q_{\rm bdca}=W_4(h_{\rm bdca})=\sup E_4,
\]
where the supremum is over every admissible instance in every positive finite
dimension; attainment is not assumed in this definition.

Decide whether both
\[
\frac{977}{25000}<q_{\rm bdca}<\frac{391}{10000}
\]
and the following attainment assertion hold: there is an admissible instance
of dimension \(d\leq2\) for which \(E_4=q_{\rm bdca}\).

Only this fixed four-step interpolation fiber is in scope. The problem does not ask for the global four-step minimax value, comparison with shorter horizons, the scalar worst-case value, or a classification of all extremizers.
\end{problem}

\clearpage
\section{P004: One-coordinate memory under binary delays}
\label{problem:P004}

\begin{problem}
Consider two agents, two communication phases, and delays in
\(\{0,1\}\).  At round \(k\), the row-stochastic matrix \(A(k)\) and the
column-stochastic matrix \(B(k)\) are two-periodic (not necessarily of
fundamental period two), have positive diagonal entries, and have every
positive entry at least \(1/2\).  The unions over a period of the supports of
\(A\) and \(B\) are separately strongly connected.  A directed nonself edge
\((j,i)\) is active exactly when \(A_{ij}(k)>0\) or \(B_{ij}(k)>0\).
Each active edge delivers, before the state update, exactly one packet
\[
\begin{aligned}
p_j(k-d_{ji}(k)),\\
p_j(t)=(x_j(t),y_j(t)),\\
d_{ji}(k)\in\{0,1\}
\end{aligned}
\]
with a common age when both weights are positive.  The ages are two-periodic,
the receiver does not observe them, distinct-edge deliveries commute, and an
inactive edge retains its previous packet.

Agent \(i\) minimizes
\[
f_i(x)=\frac12(x-r_i)^2 ,
\]
so the common target is \(x_\star=(r_1+r_2)/2\).  Its persistent node state is
exactly \(x_i(k),y_i(k)\).  For each possible incoming nonself edge it also has
one two-dimensional overwrite register \(b_{ji}(k)\), containing the most
recent delivered packet, and one auxiliary scalar \(u_{ji}(k)\).  No other
persistent node, edge, global, clock, counter, or controller state is allowed,
and packets contain no timestamp, generation index, delay label, or metadata.
The phase, current incident weights, and current edge activity may be observed
transiently.

A protocol \(\pi_1\) is chosen before the schedule, centers, initialization,
and stepsize are known.  In each phase, \(x_i(k+1)\) may be any deterministic
linear function of \(x_i(k),y_i(k)\) and the current incoming
\(b_{ji}(k),u_{ji}(k)\).  After
\(\Delta g_i(k)=x_i(k+1)-x_i(k)\) is formed, the joint update of
\(y_i(k+1)\) and the incoming auxiliary coordinates may be any deterministic
local linear function of
\[
\begin{aligned}
x_i(k),\ y_i(k),\ x_i(k+1),\ \Delta g_i(k),\\
\quad\text{and all current incoming }b_{ji}(k),u_{ji}(k).
\end{aligned}
\]
Coefficients may depend on the agent, edge, phase, stepsize, and current
incident entries of \(A(k),B(k)\), but not on the centers, realized delay
labels, undeclared history, future data, or the complete schedule.  On
zero-delay schedules the law must preserve
\(\sum_i y_i(k)=\sum_i\nabla f_i(x_i(k))\) from tracker-consistent
initialization and must preserve every common-optimizer fixed trajectory.

For \(k=-1,0\), initialize
\[
x_i(k)=x_i^0,\qquad y_i(k)=x_i^0-r_i,\qquad
p_i(k)=(x_i^0,y_i(k)),
\]
and set \(b_{ji}(0)=p_j(0)\) and \(u_{ji}(0)=0\).  Let \(z(k)\) collect every
declared persistent coordinate immediately after round-\(k\) deliveries and
before the update.  Say that \(\pi_1\) has the required property if there is
\(\bar\alpha>0\) such that, for each
\(\alpha\in(0,\bar\alpha)\), constants \(C_\alpha<\infty\) and
\(\rho_\alpha\in(0,1)\) exist which are independent of the schedule, centers,
initialization, and time, and every admissible execution has an exact
two-periodic limiting orbit \(z_\infty\) with all primal coordinates equal to
\(x_\star\), all tracker coordinates zero, and
\[
\begin{aligned}
\|z(k)-z_\infty(k\bmod2)\|_2,\\
\leq C_\alpha\rho_\alpha^k,\\
\|z(0)-z_\infty(0)\|_2,\\
k\geq0.
\end{aligned}
\]
This estimate applies to all overwrite and auxiliary registers as well.

Decide whether such a protocol \(\pi_1\) exists for the complete union-active
schedule class, including schedules of fundamental period one.  Equivalently,
is one auxiliary real coordinate per possible directed nonself edge sufficient
for uniform exact linear convergence?
\end{problem}

\clearpage
\section{P005: Harmonic metric drift in split FRB}
\label{problem:P005}

\begin{problem}
Let \(H\) be a separable infinite-dimensional real Hilbert space.  Let
\(A:H\rightrightarrows H\) be a nonzero linear maximally monotone operator and
\(B:H\to H\) a nonzero bounded linear monotone \(L\)-Lipschitz operator, with
\(\operatorname{zer}(A+B)=\{0\}\).  Let \(M_k\) be bounded self-adjoint
operators satisfying
\[
mI\preccurlyeq M_k\preccurlyeq MI
\]
for fixed \(0<m\leq M<\infty\), and suppose \(M_kx\to M_\infty x\) for every
\(x\in H\).  Define the canonical two-sided metric drift by
\[
\delta_k=\inf\left\{\delta\geq0:
(1+\delta)^{-1}M_k\preccurlyeq M_{k+1}
\preccurlyeq(1+\delta)M_k\right\}.
\]
It is essential that this be the least admissible drift, rather than a freely
chosen Loewner envelope.  Assume that for some \(0<c\leq C<\infty\) and all
sufficiently large \(k\),
\[
\frac{c}{k+1}\leq\delta_k\leq\frac{C}{k+1}.
\]
Thus \((\delta_k)\in\ell^p\) for every \(p>1\), but
\((\delta_k)\notin\ell^1\).

Choose \(0<\lambda<m/(2L)\).  An inexact variable-metric
forward-reflected-backward orbit consists of an initial pair and sequences
\((x_k),(a_{k+1}),(r_{k+1})\) satisfying
\[
\begin{aligned}
a_{k+1}\in Ax_{k+1},\\
r_{k+1}=M_k(x_k-x_{k+1}),\\
-\lambda\bigl(a_{k+1}+2Bx_k-Bx_{k-1}\bigr).
\end{aligned}
\]
When \(x_{k+1}\ne x_k\), define the realized relative residual
\[
\varepsilon_k=
\frac{\|r_{k+1}\|_{M_k^{-1}}}
     {\|x_{k+1}-x_k\|_{M_k}},
\]
and when \(x_{k+1}=x_k\), require \(r_{k+1}=0\) and set
\(\varepsilon_k=0\).  Assume
\(\sum_k\varepsilon_k<\infty\), while \(r_{k+1}\ne0\) for infinitely many
\(k\).

The same datum must satisfy two global controls.  First, for the same
\(A,B,\lambda,M_\infty\), the exact fixed-\(M_\infty\) FRB orbit converges
strongly to \(0\) for every initial pair.  Second, for the same \(A\), metrics,
and \(\lambda\), deleting \(B\) and setting every residual to zero produces a
variable-metric proximal orbit which converges strongly to \(0\) for every
initial pair.  Along the selected split orbit there must additionally be
\(c_B>0\) such that
\[
\|Bx_k\|\geq c_B\|x_k\|
\]
eventually, and the selected \(a_{k+1}\) must be nonzero infinitely often.
These clauses prevent either operator from being only nominally present.

Decide whether one complete datum satisfying all of the preceding conditions
can have
\[
x_k\rightharpoonup0
\qquad\text{but}\qquad
\limsup_{k\to\infty}\|x_k\|>0.
\]
\end{problem}

\clearpage
\section{P006: Rank-six graph comparison cells}
\label{problem:P006}

\begin{problem}
Let \(G\) be a finite simple graph with at most eight vertices and
\(\operatorname{tw}(G)\leq5\), equivalently, a partial \(5\)-tree.  Its
Euclidean dimension \(\operatorname{ed}(G)\) is the least \(k\geq1\) such that
every finite-dimensional real vector configuration indexed by \(V(G)\) can be
replaced by one in \(\mathbb R^k\) preserving every edge distance.  Its Gram
dimension \(\operatorname{gd}(G)\) is defined similarly, but the replacement
must preserve every squared norm and every edge inner product.

Let \(\mathcal C(G)\) be the relevant cone of graph-supported positive
semidefinite matrices.  A matrix \(M\in\mathcal C(G)\) has the
\emph{full-annihilation property} if the only \(X\in\mathbb S^{|V(G)|}\)
satisfying
\[
\begin{aligned}
MX=0,\\
X_{ii}=0\quad(i\in V(G)),\\
X_{ij}=0\quad(ij\in E(G))
\end{aligned}
\]
is \(X=0\).  Define \(\widehat\nu(G)\) to be the maximum corank of a matrix in
\(\mathcal C(G)\) having this property.

Decide whether every such graph satisfying \(\operatorname{gd}(G)=6\) must
satisfy both
\[
\operatorname{ed}(G)=5
\qquad\text{and}\qquad
\widehat\nu(G)=6.
\]
The two equalities are independent obligations and may not be inferred from
one another.
\end{problem}

\clearpage
\section{P007: Critical-channel localization boundary}
\label{problem:P007}

\begin{problem}
Fix \(u\in(0,1]\).  On \(\mathbb R\), with base point \(0\), parameter
\(\lambda=1\), and centers \(y\in[-u,u]\), consider the Legendre kernels
\[
\begin{aligned}
\phi_0(x)=\frac{x^2}{2},\\
\phi_k(x)=\frac{k^{-3}x^2}{2},\\
+(1-k^{-3})k^2\log\!\cosh(x/k),\quad k\geq1
\end{aligned}
\]
and their Bregman divergences
\[
D_{\phi_k}(x,y)=\phi_k(x)-\phi_k(y)-\phi_k'(y)(x-y).
\]
For an objective \(f\), put
\[
H_{k,y}(x)=f(x)+D_{\phi_k}(x,y).
\]
The kernels are understood after affine normalization at the base point; they
and their gradients converge uniformly on compact sets to \(\phi_0\) and
\(\phi_0'\).  Each kernel is supercoercive, and the scoped objectives must be
proper, lower semicontinuous, finite and relatively prox-regular at \(0\) for
\(0\), with \(0\in\partial f(0)\) and relative prox-boundedness threshold
strictly larger than \(1\) for every \(k\).

Define
\[
\begin{aligned}
\alpha=(3/2)^{1/4},\\
c_0=\frac{4\alpha}{3},\\
b_c(u)=\sqrt{\frac23}\left(1-\frac{u^3}{3}\right),\\
+\sqrt{\frac32}\log2 .
\end{aligned}
\]
For \(d\in\mathbb R\), let \(\mathcal F_{\rm crit}(u,d)\) consist of all even
\(C^2\) functions \(f:\mathbb R\to\mathbb R\) such that
\[
\begin{aligned}
f(x)=0\quad(|x|\leq1),\\
f(x)=-c_0|x|^{5/4}+u|x|+b_c(u)|x|^{1/2}+d,\\
\quad(|x|\geq2).
\end{aligned}
\]
No condition is imposed on the interpolation region \(1<|x|<2\) beyond
membership in this class.

For \(f\in\mathcal F_{\rm crit}(u,d)\), set
\[
\begin{aligned}
m_u(f)&=\min_{x\in\mathbb R}\left(f(x)+\frac{(x-u)^2}{2}\right),\\
\Gamma(u,d)&=d+\frac{u^2}{2}-\frac{4u^5}{45},\\
\Delta(u,d,f)&=\Gamma(u,d)-m_u(f),\\
q(u)&=\frac32\log2-1+\frac{u^3}{3},\\
L(u)&=\log2-\frac{2u^3}{9}-\frac{q(u)^2}{3},
\end{aligned}
\]
and define the positive tail-channel energy
\[
T_k(u)=\inf_{x\geq k^3}H_{k,u}(x).
\]
For \(\tau>0\), write
\[
\mathcal N_{k,y}(\tau)=
\left\{x:H_{k,y}(x)\leq\inf_wH_{k,y}(w)+\tau\right\}.
\]
Uniform boundedness of either exact minimizers or
\(\tau\)-near-minimizers always means one radius valid simultaneously for
every integer \(k\geq1\) and every \(y\in[-u,u]\).  Its failure means that
every proposed radius is exceeded by a member of this joint family.

Decide whether the following complete critical-channel classification holds
for every \(u\in(0,1]\), \(d\in\mathbb R\), and
\(f\in\mathcal F_{\rm crit}(u,d)\).

First, \(L(u)>0\), and as \(k\to\infty\) through positive integers,
\[
T_k(u)=\Gamma(u,d)+\frac{L(u)}{k}+o(k^{-1}).
\]
Moreover, every choice
\[
x_k\in\operatorname*{argmin}_{x\geq k^3}H_{k,u}(x)
\]
satisfies
\[
x_k=\frac23k^4-\frac43q(u)k+o(k).
\]

Second, if \(\Delta(u,d,f)<0\), then every sequence
\(z_k\in\operatorname*{argmin}H_{k,u}\) satisfies
\(|z_k|\to\infty\), and
\[
\inf_xH_{k,u}(x)\longrightarrow\Gamma(u,d).
\]
If \(\Delta(u,d,f)\geq0\), there is instead a radius
\(R_{\rm exact}=R_{\rm exact}(u,d,f)\) such that
\[
\begin{aligned}
|z|\leq R_{\rm exact},\\
\quad\text{for every }k\geq1,\ y\in[-u,u],\\
z\in\operatorname*{argmin}H_{k,y}.
\end{aligned}
\]

Third, when \(\Delta(u,d,f)>0\), every fixed
\(0<\tau<\Delta(u,d,f)\) admits a joint bound
\[
\mathcal N_{k,y}(\tau)\subseteq[-R_\tau,R_\tau]
\quad(k\geq1,\ y\in[-u,u])
\]
for some \(R_\tau=R_\tau(u,d,f,\tau)\), whereas for every
\(\tau>\Delta(u,d,f)\) and every \(R>0\), some
\(k\geq1\), \(y\in[-u,u]\), and
\(x\in\mathcal N_{k,y}(\tau)\) satisfy \(|x|>R\).
No claim is made at the tolerance endpoint
\(\tau=\Delta(u,d,f)\).  When \(\Delta(u,d,f)=0\), exact minimizers retain the
joint bound above, but for every \(\tau>0\) the
\(\tau\)-near-minimizer family is not uniformly bounded.

Finally, the kernel family has no positive uniform lower-chord comparison
with \(\phi_0\): for every \(c>0\), there are
\(k\geq1\), \(x\in\mathbb R\), and \(y\in[-u,u]\) such that
\[
D_{\phi_k}(x,y)<cD_{\phi_0}(x,y).
\]
\end{problem}

\clearpage
\section{P008: Universal two-partition floor}
\label{problem:P008}

\begin{problem}
Let \(x_1,\ldots,x_{10}\in\mathbb R^2\) be distinct points, in the displayed
cyclic order, which are all vertices of their convex hull, and assume that no
three are collinear.  Partition \([10]\) into color classes
\(C_1,C_2,C_3,C_4\) of sizes \(3,3,3,1\), respectively.

A candidate partition is an unordered collection of four nonempty, pairwise
disjoint blocks whose union is \([10]\).  It is \emph{rainbow} if each block
contains at most one index from each color class.  Let
\(\mathcal R(X,\mathcal C)\) be the set of rainbow candidates for which the
four closed convex hulls of the corresponding point blocks have a nonempty
common intersection, and put
\[
T(X,\mathcal C)=|\mathcal R(X,\mathcal C)|.
\]
The blocks are unordered: permuting them or choosing a different common point
does not create a new partition.  The convex hulls are closed, so the common
point may lie on a boundary and barycentric coefficients may vanish.

Decide whether
\[
T(X,\mathcal C)\geq2
\]
for every configuration and coloring satisfying these conditions, including
diagonal-concurrence and boundary-intersection cases.

Sharpness is supplied by the fixed rational configuration
\[
\begin{aligned}
(t_1^\star,\ldots,t_{10}^\star),\\
=(-8,-5,-3,-1,0,2,4,7,11,16),\\
x_j^\star=(t_j^\star,(t_j^\star)^2)
\end{aligned}
\]
with
\[
\begin{aligned}
C_1^\star=\{2,4,9\},\quad C_2^\star=\{3,5,8\},\quad,\\
C_3^\star=\{6,7,10\},\quad C_4^\star=\{1\}.
\end{aligned}
\]
For this input an independently checked exact certificate gives precisely the
two feasible partitions
\[
\begin{aligned}
\mathcal A&=\{\{1,5,9\},\{2,6,8\},\{3,7\},\{4,10\}\},\\
\mathcal B&=\{\{1,4,8\},\{2,7\},\{3,6,9\},\{5,10\}\}.
\end{aligned}
\]
This equality is evidence only for sharpness and may not be used as a premise
for the universal lower bound.  Together with that lower bound, the
certificate would determine the universal minimum as two.
\end{problem}

\clearpage
\section{P009: Sampled-information exit exponent}
\label{problem:P009}

\begin{problem}
Fix \(T>0\), \(\sigma>0\), and \(U>\sigma^2\).  Starting from \(X_0=0\),
consider the stopped controlled Jacobi diffusion
\[
\begin{aligned}
dX_t=u_t\,dt+\sigma\sqrt{1-X_t^2}\,dW_t,\\
\tau=\inf\{t\geq0:|X_t|=1\}.
\end{aligned}
\]
The unique-in-law weak solution is absorbed once \(\tau\) occurs, and safety
failure is the endpoint-inclusive event \(\{\tau\leq T\}\).

The continuous-information comparator permits arbitrary progressively
measurable controls with \(|u_t|\leq U\), including the Markov feedback
\(u_t=-UX_t\).  For a mesh \(\Delta=T/N\), \(N\in\mathbb N\), a sampled
policy observes the exact state only at \(k\Delta\) and holds
\[
u_t=a_k\quad\text{on }[k\Delta,(k+1)\Delta).
\]
Here \(a_k\in[-U,U]\) may be any Borel function of the entire sampled-state
history and of private randomization independent of the Brownian motion.
Let
\[
p_\Delta=\inf_{\text{all such sampled policies}}\mathbb P(\tau\leq T).
\]
The infimum is over the full history-dependent randomized class, not merely
Markov or sign policies.  The constrained cost is the unit exit penalty, so
whenever the continuous feedback has zero exit probability the
continuous-versus-sampled value gap equals \(p_\Delta\).

Decide whether, for every choice of the fixed parameters above, all three
claims
\[
\mathbb P_{\,u_t=-UX_t}(\tau\leq T)=0,\qquad
p_\Delta>0\quad(N\geq1),
\]
and
\[
\lim_{N\to\infty}
\left[-\frac{T}{N}\log p_{T/N}\right]
=\frac{\pi^2}{8\sigma^2}
\]
hold.  The limit is required along the full integer sequence, not a selected
subsequence.
\end{problem}

\clearpage
\section{P010: Dyadic extremal coloring}
\label{problem:P010}

\begin{problem}
Let
\[
\mathcal D_2=\left\{t\in\mathbb R^{10}:
t_1>0,\quad t_{k+1}\geq2t_k\ (1\leq k\leq9)\right\},
\]
and for \(t\in\mathcal D_2\) set
\[
x_j(t)=(t_j,t_j^2),\qquad
X(t)=(x_1(t),\ldots,x_{10}(t)).
\]
These are ten distinct vertices of a strictly convex planar decagon in the
displayed order.  Fix the coloring
\[
\begin{aligned}
C_1^\star=\{2,5,8\},\quad,\\
C_2^\star=\{3,6,9\},\quad,\\
C_3^\star=\{4,7,10\},\quad,\\
C_4^\star=\{1\}.
\end{aligned}
\]

A partition is an unordered collection of four nonempty disjoint blocks
covering \([10]\).  It is rainbow if each block meets each color class in at
most one index, and it is a Tverberg partition if the four closed convex hulls
of its point blocks have a nonempty common intersection.  Zero barycentric
coefficients are allowed, and neither a permutation of the blocks nor another
choice of common point creates a new partition.  Let
\(\mathcal R_\star(t)\) denote the family of all such rainbow Tverberg
partitions.

Consider
\[
\begin{aligned}
\mathcal A&=\{\{1,5,9\},\{2,6,10\},\{3,7\},\{4,8\}\},\\
\mathcal B&=\{\{1,6,8\},\{2,7\},\{3,5,10\},\{4,9\}\}.
\end{aligned}
\]
Decide whether the exact family identity
\[
\mathcal R_\star(t)=\{\mathcal A,\mathcal B\}
\]
holds for every \(t\in\mathcal D_2\), equivalently whether the number of
unordered rainbow Tverberg partitions is uniformly equal to two throughout
the closed dyadic cone.

The problem is restricted to the stated coloring and the closed dyadic cone; it makes no claim about other colorings, ratios below two, or arbitrary strictly convex decagons.
\end{problem}

\clearpage
\section{P011: Sharp permutation margin}
\label{problem:P011}

\begin{problem}
For a fixed four-step schedule \(h=(h_0,h_1,h_2,h_3)\), chosen before any
gradients are observed, apply gradient descent to an arbitrary convex,
differentiable, \(L\)-smooth function \(f:\mathbb R^m\to\mathbb R\):
\[
x_{t+1}=x_t-\frac{h_t}{L}\nabla f(x_t),\qquad 0\leq t<4.
\]
Here \(m\) is any positive finite integer, \(f\) has a minimizer \(x_\star\),
and \(\|x_0-x_\star\|\leq D\).  No momentum, memory, line search, or
adaptivity is allowed.  Define
\[
W_4(h)=\sup\frac{f(x_4)-f(x_\star)}{LD^2},
\]
where the supremum is over all admissible data and all positive finite
dimensions, without assuming attainment.

Let \((a,b,c,\widehat p)\) be the unique positive real tuple in the rational
box
\[
\begin{aligned}
\begin{gathered},\\
\frac{1414213}{10^6}<a<\frac{1414214}{10^6},\\
\frac{1601231}{10^6}<b<\frac{1601232}{10^6},\,\\
\frac{3005144}{10^6}<c<\frac{3005145}{10^6},\\
\frac{31169778960345}{10^{15}}<\widehat p<,\\
\frac{31169778960346}{10^{15}},\\
\end{gathered}
\end{aligned}
\]
satisfying
\[
\begin{aligned}
a^2=2,\\
b^2+ab-2-2a=0,\\
c^2=2a+2b+3,\\
2\widehat p(c+1)^2=1.
\end{aligned}
\]
Let \(\mathcal W\) be the 24 words obtained by permuting the distinct letters
\(a,b,c,d\), and associate with each word the schedule obtained by replacing
\(d\) with \(3/2\).  Thus \(h^{bdca}=(b,3/2,c,a)\).  Remove the two words
\[
E_0=\{abcd,bacd\},\qquad
\mathcal O_{\rm perm}=\mathcal W\setminus E_0,
\]
and define
\[
\gamma(w)=W_4(h^w)-\widehat p,\qquad
\Delta_{\rm perm}=\min_{w\in\mathcal O_{\rm perm}}\gamma(w),
\]
\[
\mathcal A_{\rm perm}=
\{w\in\mathcal O_{\rm perm}:\gamma(w)=\Delta_{\rm perm}\}.
\]

Decide whether
\[
\frac{39}{5000}<\Delta_{\rm perm}<\frac1{125}
\qquad\text{and}\qquad
\mathcal A_{\rm perm}=\{bdca\}.
\]

The question is restricted to this 24-schedule orbit and makes no claim about the global minimax value over \(\mathbb R^4\).
\end{problem}

\clearpage
\section{P012: Short-period supercritical orbits}
\label{problem:P012}

\begin{problem}
For \(\delta\in(8,41/5]\), consider the two-player game on
\([-1,1]^2\) with payoffs
\[
u_1(x,y)=-\frac{x^2}{2}-(1+\delta)xy,\qquad
u_2(x,y)=-\frac{y^2}{2}+xy.
\]
Writing \(z=(x,y)\), define
\[
F_\delta(z)=(x+(1+\delta)y,\ y-x)
\]
and let \(\Pi_K\) be coordinatewise clipping onto \(K=[-1,1]^2\).  From an
arbitrary history \(z(-1),z(0)\in K\), the simultaneous exact-gradient
recurrence is
\[
z(t+1)=\Pi_K\!\left(
z(t)-\frac25F_\delta(z(t))+\frac15F_\delta(z(t-1))
\right).
\]
Equivalently, on the compact history space
\(\mathcal H=K\times K\),
\[
\Phi_\delta(z_0,z_{-1})=
\left(\Pi_K\!\left(z_0-\frac25F_\delta(z_0)
+\frac15F_\delta(z_{-1})\right),z_0\right).
\]
A history has least period \(p\) if
\(\Phi_\delta^p(h)=h\) and \(\Phi_\delta^j(h)\ne h\) for
\(1\leq j<p\).

Define the equilibrium orbit
\[
Q_0=\{((0,0),(0,0))\}.
\]
For the fully clipped cycle, extend periodically
\[
\begin{aligned}
a_0=(1,-1/3),\quad a_1=(1,1/3),\quad,\\
a_2=(-1,1/3),\quad a_3=(-1,-1/3)
\end{aligned}
\]
and set \(Q_A=\{(a_r,a_{r-1}):r\in\mathbb Z_4\}\).  For the mixed
clipped-interior cycle, extend periodically
\[
\begin{aligned}
s_0=(1,-1/6),\quad s_1=(1/2,1/3),\quad,\\
s_2=(-1,1/6),\quad s_3=(-1/2,-1/3)
\end{aligned}
\]
and set \(Q_S=\{(s_r,s_{r-1}):r\in\mathbb Z_4\}\).

Decide whether, for every \(\delta\in(8,41/5]\), every periodic point of
\(\Phi_\delta\) with least period at most eight belongs to
\[
Q_0\cup Q_A\cup Q_S.
\]
If so, the only least periods at most eight are one and four.  The
classification must use exact coordinatewise clipping, including equality on
clipping boundaries.  It must cover every feasible lower-clipped, interior,
and upper-clipped word of length at most eight, solve its affine periodicity
equations, check the corresponding inequalities uniformly throughout the
parameter interval, and remove repetitions and phase shifts before assigning
least period.  Local multipliers of the three displayed orbits alone are not
sufficient.

No claim is made about periods above eight, nonperiodic recurrence, basins, empirical measures, or regret convergence.
\end{problem}

\clearpage
\section{P013: Subdyadic two-partition threshold}
\label{problem:P013}

\begin{problem}
For \(\rho>1\), let
\[
\mathcal D_\rho=
\{t\in\mathbb R^{10}:t_1>0,\ t_{k+1}\geq\rho t_k
\text{ for }1\leq k\leq9\},
\]
and set \(x_j(t)=(t_j,t_j^2)\) and
\(X(t)=(x_1(t),\ldots,x_{10}(t))\).  Partition \([10]\) into four color
classes of sizes \(3,3,3,1\).  A candidate is an unordered collection of four
nonempty pairwise disjoint blocks covering \([10]\); it is rainbow when each
block meets each color class at most once.  Let \(T(X(t),\mathcal C)\) be the
number of such candidates whose four closed point-block convex hulls have a
nonempty common intersection.  Block permutations are identified and zero
barycentric coefficients are allowed.

Let \(P_T(\rho)\) mean that
\[
T(X(t),\mathcal C)\geq2
\]
for every \(t\in\mathcal D_\rho\) and every permitted coloring, and define
\[
\rho_T=\inf\{\rho>1:P_T(\rho)\}.
\]
The domains are nested:
\(\mathcal D_{\rho_2}\subseteq\mathcal D_{\rho_1}\) whenever
\(\rho_2\geq\rho_1\).  An independently checked exact certificate establishes
the dyadic endpoint \(P_T(2)\).

Decide whether
\[
\rho_T<2.
\]

The problem concerns only positive parabola sequences in the nested stretch domains. It does not extend to arbitrary strictly convex decagons; boundary ratios, boundary hull intersections, and counting modulo block permutations remain in scope.
\end{problem}

\clearpage
\section{P014: Robust stepsize boundary for push--pull tracking}
\label{problem:P014}

\begin{problem}
Fix \(0<a\leq1/2\) and constant matrices
\[
\begin{aligned}
A=\begin{pmatrix}1-u&u\\ v&1-v\end{pmatrix},\\
C=\begin{pmatrix}1-r&s\\ r&1-s\end{pmatrix},\\
\qquad u,v,r,s\in[a,1-a].
\end{aligned}
\]
Thus \(A\) is row-stochastic, \(C\) is column-stochastic, and both have
complete positive off-diagonal support.  Consider two scalar agents with
\[
f_i(x)=\frac{x^2}{2}+b_ix+c_i,\qquad b_i,c_i\in\mathbb R,
\]
whose optimizer is \(x_\star=-(b_1+b_2)/2\).  Communication is immediate and
uncompressed.  Starting from arbitrary \(x^0\in\mathbb R^2\), initialize
\(y^0=x^0+b\) and run constant-stepsize push--pull tracking.  With
\(z_k=(x_k,y_k)\in\mathbb R^4\), its linear state equation is
\[
z_{k+1}=Tz_k,\qquad
T=\begin{pmatrix}A&-\alpha I\\ A-I&C-\alpha I\end{pmatrix}.
\]

Define
\[
E_k=\sum_{i=1}^2|x_i^k-x_\star|^2
+\sum_{i=1}^2|y_i^k|^2
+\left|\sum_i y_i^k-\sum_i(x_i^k+b_i)\right|^2,
\]
and let \(P_{\rm box}(a,\alpha)\) mean that \(E_k\to0\) for every admissible
weight quadruple, every \(b,c\in\mathbb R^2\), and every initial \(x^0\).
The characteristic polynomial factors as
\[
\begin{aligned}
\det(\lambda I-T)=(\lambda-1)p(\lambda),\\
\qquad,\\
p(\lambda)=\lambda^3+\beta_2\lambda^2+\beta_1\lambda+\beta_0
\end{aligned}
\]
where
\[
\begin{aligned}
\beta_2={}&2\alpha+r+s+u+v-3,\\
\beta_1={}&\alpha^2+\alpha(r+s+u+v-4)+ru+rv-2r\\
&\quad+su+sv-2s-2u-2v+3,\\
\beta_0={}&-\alpha^2+\alpha(2ru-r+2sv-s-u-v+2)\\
&\quad-ru-rv+r-su-sv+s+u+v-1.
\end{aligned}
\]

Decide whether, for every \(0<a\leq1/2\) and \(0<\alpha\leq1\),
\[
P_{\rm box}(a,\alpha)
\quad\Longleftrightarrow\quad
0<\alpha<2a^2.
\]
\end{problem}

\clearpage
\section{P015: Gaussian envelope for exit-risk sensitivity}
\label{problem:P015}

\begin{problem}
Consider the open unit ball \(D\subset\mathbb R^d\), a fixed
\(x_0\in D\), identity diffusion, and a horizon
\(T=N\Delta\), where \(N\geq1\) is finite and
\(0<\Delta\leq H\leq T\).  The state coefficients and compact control set
satisfy the usual boundedness and uniform state-Lipschitz conditions ensuring
a unique strong solution up to the exit time
\(\tau_D\).  Bounded continuous running, terminal, and exit costs define a
fixed \(H\)-horizon stopped problem.  It is stipulated that this problem has a
Borel time-zero minimizing selector \(\mu_H\); deterministic tie-breaking and
a fixed boundary action extend it to \(\overline D\).

At the sampling time \(k\Delta\), an adversarial observation error \(E_k\) is
measurable with respect to the completed pre-decision history generated by
\(X_0\), the Brownian path through \(k\Delta\), and the earlier errors.
It satisfies \(\|E_k\|\leq\eta\).  The controller observes
\[
Y_k=X_{k\Delta}+E_k
\]
and holds
\[
u_t=\mu_H\!\left(\Pi_D(Y_k)\right),
\qquad t\in[k\Delta,(k+1)\Delta)\cap[0,T],
\]
until exit, where \(\Pi_D\) is Euclidean projection onto the closed unit
ball.  After exit a fixed cemetery action is used.  For exact observations,
the same held controller has the pointwise certificate
\[
\mathbb P_{x_0}(\tau_D\leq T)\leq\alpha-s,
\qquad \alpha\in(0,1),\quad s>0.
\]
No uniform safety certificate over other initial or replanning states is
assumed.

Write
\[
\bar b(x,z)=b(x,\mu_H(z)),\qquad x\in D,\ z\in\overline D,
\]
and assume that for some finite \(L_y\geq0\),
\[
\|\bar b(x,z)-\bar b(x,z')\|
\leq L_y\|z-z'\|
\]
globally in \(x,z,z'\).  Let
\[
F(\eta)=
\sup_{\substack{\text{authorized history-dependent}\\
                 \|E_k\|\leq\eta}}
\mathbb P_{x_0}(\tau_D\leq T)
\]
be the worst-case exit risk.  With \(\Phi\) the standard normal distribution
function, define
\[
G(q)=2\Phi(q/2)-1
\]
and the classwise modulus
\[
\begin{aligned}
\mathcal M(q)=,\\
\sup\bigl\{F(\eta')-F(\eta):,\\
0\leq\eta\leq\eta',\\
L_y\sqrt T\,(\eta'-\eta)\leq q\bigr\}
\end{aligned}
\]
where the outer supremum ranges over all scoped instances.  Coefficient bounds
need be finite for each instance but need not be uniform along a sharpness
sequence.  Finally set
\[
\eta_c=\sup\{\eta\geq0:F(\eta)\leq\alpha\}.
\]

Decide whether the exact classwise identity
\[
\mathcal M(q)=G(q)=2\Phi(q/2)-1,\qquad q\geq0,
\]
holds.  In particular, does every individual instance satisfy
\[
0\leq F(\eta')-F(\eta)
\leq G\!\left(L_y\sqrt T\,(\eta'-\eta)\right)?
\]
\end{problem}

\clearpage
\section{P016: Second-epi criterion for PL transfer}
\label{problem:P016}

\begin{problem}
In \(\mathbb R^2\), let \(S=[-1,1]^2\),
\(f(z)=\operatorname{dist}(z,S)^2\), and
\[
\mathcal T=\{z:\operatorname{dist}(z,S)<1/4\}.
\]
Let \(p\in\mathbb Q[x,y]\) have degree at most six and satisfy throughout
\(\mathcal T\)
\[
|p|\leq\frac1{128},\qquad
\|\nabla p\|\leq\frac1{128},\qquad
\|\nabla^2p\|_{\rm op}\leq\frac18.
\]
Set \(F_p=f+p\),
\[
\begin{aligned}
v_p=\min_{\mathcal T}F_p,\\
M_p=\{z\in\mathcal T:F_p(z)=v_p\},\\
K_p=\{z\in\mathcal T:\nabla F_p(z)=0\}.
\end{aligned}
\]
The bounds ensure that \(M_p\) is nonempty and lies in the interior of
\(\mathcal T\).  Let \(\Delta_p\) be the distance from \(v_p\) to the set of
higher stationary values
\(\{F_p(z):z\in K_p,\ F_p(z)>v_p\}\), with
\(\Delta_p=\infty\) if that set is empty.  Also put
\[
\begin{aligned}
b_p=\min_{z\in\partial\mathcal T}(F_p(z)-v_p),\\
\lambda_p=\frac14\min\{\Delta_p,b_p\},\\
\Omega_p=\{z\in\mathcal T:F_p(z)<v_p+\lambda_p\}.
\end{aligned}
\]

For \(m\in M_p\), let \(T_{M_p}(m)\) be the Bouligand tangent cone and define
\[
d^2F_p(m\mid0)(w)=
\liminf_{\substack{t\downarrow0\\w'\to w}}
\frac{2(F_p(m+tw')-F_p(m))}{t^2}.
\]
Consider the transverse second-epi condition
\[
\begin{aligned}
\exists\mu>0\quad\text{such that}\\
d^2F_p(m\mid0)(w)&\geq\\
&\mu\,\operatorname{dist}
\bigl(w,T_{M_p}(m)\bigr)^2,\\
&m\in M_p,\quad w\in\mathbb R^2.
\end{aligned}
\tag{Q}
\]
and the ambient condition that constants \(c>0\) and \(C<\infty\) exist such
that every \(z\in\Omega_p\) satisfies
\[
\begin{aligned}
c\,\operatorname{dist}(z,M_p)^2&\leq F_p(z)-v_p,\\
\operatorname{dist}(z,M_p)&\leq C\|\nabla F_p(z)\|,\\
F_p(z)-v_p&\leq C\|\nabla F_p(z)\|^2.
\end{aligned}
\tag{E}
\]

Decide whether \((Q)\) and \((E)\) are equivalent for every admissible
rational polynomial \(p\).  The stationary values, \(b_p,\lambda_p\), and
\(\Omega_p\) must be computed from the complete piecewise-polynomial function
before either predicate is tested.  Both predicates refer to the full
minimum set, including corners, boundary segments, crossings, multiple
components, and approaches across projection-cell seams.
\end{problem}

\clearpage
\section{P017: Robust-instability tongue}
\label{problem:P017}

\begin{problem}
For \(\phi\in[0,\pi/2]\), let
\[
M_0=\operatorname{diag}(1,4),\qquad
M_1(\phi)=R(\phi)\operatorname{diag}(1,4)R(\phi)^{\mathsf T},
\]
where \(R(\phi)\) is the planar rotation matrix, and define
\[
\begin{aligned}
T_i(\phi)&=\bigl(M_i(\phi)+\tfrac45I\bigr)^{-1}\\
&\quad\cdot\bigl(M_i(\phi)-\tfrac45I\bigr),\\
D_i(\phi)=I-T_i(\phi).
\end{aligned}
\]
For \(\eta\in[0,1]\), the allowed phase-\(i\) images of \(x\) are
\[
\mathcal F_i^{\phi,\eta}(x)=
\left\{
\begin{array}{l}
T_i(\phi)x+r:\\
r^{\mathsf T}M_i(\phi)r\leq
\eta^2x^{\mathsf T}D_i(\phi)^{\mathsf T}\\
\qquad\qquad\quad M_i(\phi)D_i(\phi)x
\end{array}
\right\}.
\]
The error may depend on the realized state and the entire preceding history.
For either starting phase \(\sigma\in\{0,1\}\), an admissible trajectory
satisfies
\[
x_{k+1}\in
\mathcal F_{(\sigma+k)\bmod2}^{\phi,\eta}(x_k),\qquad k\geq0.
\]
Define the worst two-step asymptotic amplification
\[
\mathcal R(\phi,\eta)=
\max_{\sigma\in\{0,1\}}
\limsup_{n\to\infty}
\left(
\sup_{\substack{x_0\ne0\\\text{admissible }(x_k)}}
\frac{\|x_{2n}\|_2}{\|x_0\|_2}
\right)^{1/n}.
\]
This is the complete adaptive trajectory family; no fixed-direction,
stationary-error, or periodic-error restriction is allowed.

Let
\[
\eta_{\rm mid}=
\sqrt{\frac{1639}{256}-\frac{375\sqrt2}{128}
-\frac3{256}\sqrt{339569-230500\sqrt2}}.
\]
Decide whether there is a unique continuous curve
\(\eta_c:[0,\pi/2]\to(0,1]\) and a unique
\(\phi_{\min}\in(3/4,39/50)\) such that \(\eta_c\) is strictly decreasing on
\([0,\phi_{\min}]\), strictly increasing on
\([\phi_{\min},\pi/2]\), and
\[
\begin{aligned}
\eta_c(0)=1,\\
\eta_c(\pi/4)=\eta_{\rm mid},\\
\frac{9449}{10000}<\eta_c(\phi_{\min})
<\frac{23623}{25000}.
\end{aligned}
\]
The curve must be the exact stability boundary:
\[
\mathcal R(\phi,\eta)
\begin{cases}
<1,&\eta<\eta_c(\phi),\\
=1,&\eta=\eta_c(\phi),\\
>1,&\eta>\eta_c(\phi).
\end{cases}
\]
For each fixed \(\phi\), the entire adaptive family at criticality must be
power bounded: some \(C_\phi<\infty\) must satisfy
\[
\|x_k\|_2\leq C_\phi\|x_0\|_2
\]
for both starting phases, every nonzero initial state, every admissible error
history, and all \(k\).  At the same time, at least one nonzero critical
trajectory must be nondecaying:
\(\inf_k\|x_k\|_2>0\).

Below the curve, the required conclusion is a per-iterate \(R\)-linear
estimate uniform over phases, states, and adaptive errors; above it, strict
instability requires one coherent infinite unbounded trajectory, not merely
a finite product of norm greater than one.  The equality set must be proved
to be exactly one strictly unimodal graph, ruling out plateaus, secondary
tongues, and active-branch switches.
\end{problem}

\clearpage
\section{P018: Ten-atom separator threshold}
\label{problem:P018}

\begin{problem}
Consider two running-intersection cliques
\[
C_L=\{z_1,z_2,z_3\}\cup U,\qquad
C_R=\{z_1,z_2,z_3\}\cup V,
\]
with disjoint finite private-variable sets \(U,V\).  Local polynomial
constraints are compact and Archimedean, and the standard sparse Moment--SOS
convention is used: at order \(t\), moments have degree at most \(2t\), while
a degree-\(d\) constraint has localizing order \(t-\lceil d/2\rceil\).
Clique moment blocks are positive semidefinite and flat, so their finite
atomic representing measures can be extracted.  Shared moments are equated,
but equality of the complete extracted separator measures is not assumed.

A ten-point set \(S\subset\mathbb R^3\) is
\emph{quadratic-poised} when evaluation
\[
\mathcal P_2(\mathbb R^3)\longrightarrow\mathbb R^S
\]
is bijective.  At relaxation order three, suppose each clique measure has
ten atoms, has an injective projection to ten distinct quadratic-poised
separator points, and satisfies
\[
\operatorname{rank}M_3=\operatorname{rank}M_2=10.
\]
The two separator marginals therefore agree on every trivariate monomial of
degree at most six.

Decide whether these hypotheses force equality of the complete separator
marginals and hence permit a finite global coupling supported on the full
feasible set whose clique marginals are the extracted measures.  For every
clique-decomposed objective, this coupling should imply equality of the
order-three sparse relaxation value \(\rho_3\) and the true optimum
\(f_3^\star\).  The argument may use only the shared degree-six moments; in
particular, it must construct degree-at-most-three polynomials separating an
alleged outside atom from a poised ten-point support, then recover the weights
using quadratic Lagrange polynomials.

The asserted threshold also includes sharpness at order two.  Decide whether
there exists a compact two-clique instance with degree-at-most-two local
constraints, a linear clique-decomposed objective, a nonempty global feasible
set, and flat blocks
\[
\operatorname{rank}M_2=\operatorname{rank}M_1=10,
\]
whose two positive ten-atomic separator marginals are disjoint,
quadratic-poised, and agree on all 35 trivariate moments through degree four,
yet satisfy the strict gap
\[
\rho_2<f_2^\star.
\]
Such a witness must give exact or certifiably algebraic nodes and positive
weights, verify disjointness and both \(10\times10\) quadratic evaluation
determinants, and compute the sparse and true values separately.

The problem asks whether both the universal order-three assertion and the
existential order-two sharpness assertion hold.
\end{problem}

\clearpage
\section{P019: Endpoint \(P_0\) diagonal faithfulness}
\label{problem:P019}

\begin{problem}
Work in \(\mathbb R^3\).  Let \(U\) be open and convex, let
\(V\Subset K\subset B\Subset U\), where \(K\) is a closed ball and \(B\) a
closed axis-aligned box, and let \(h_x(y)\) be a jointly \(C^2\) field of
supercoercive, very strictly convex Legendre kernels.  On the common
localization their \(y\)-Hessians have uniform positive lower and finite upper
bounds, and their second derivatives are locally Lipschitz.  Define
\[
D(x;y,z)=h_x(y)-h_x(z)
-\langle\nabla_yh_x(z),y-z\rangle,
\]
\[
G(x,y)=\nabla_xD(x;y,x),\qquad
H(x)=\nabla^2_{yy}h_x(x),
\]
\[
\begin{aligned}
F_x(y)=-H(x)^{-1}G(x,y),\\
A(x,y)=D_yF_x(y)=-H(x)^{-1}D_yG(x,y).
\end{aligned}
\]
Thus \(F_x(x)=0\) and \(A(x,x)=I\).

Let \(f\) be proper, lower semicontinuous, prox-bounded, relatively
prox-regular, and subdifferentially continuous on the localization, with
stationarity interpreted using the limiting subdifferential.  Choose
\(\lambda>0\) below the common thresholds.  For every \(x\in V\), assume
\[
Q_x(y)=f(y)+\lambda^{-1}D(x;y,x)
\]
has a unique exact minimizer \(p(x)\) over \(K\), lying in
\(\operatorname{int}K\), with uniform local quadratic growth and a uniform
positive objective gap on \(\partial K\).  Put
\[
M_{\rm loc}(x)=Q_x(p(x)).
\]
The neighborhoods are fixed as part of the datum and may not be shrunk later
to discard a critical center.

Assume that some \(\alpha>0\) satisfies, for every \((x,y)\in V\times B\),
that all one-by-one and two-by-two principal minors of \(A(x,y)\) are at least
\(\alpha\), while
\[
\det A(x,y)\geq0.
\]
This is the endpoint proper-\(P_0\) condition: the full determinant may
vanish.  Adding any jointly \(C^2\), \(x\)-dependent affine function of \(y\)
to \(h_x\) leaves all data and conclusions unchanged.

Decide whether these assumptions always imply that \(M_{\rm loc}\) is
\(C^1\) on \(V\), with
\[
\nabla M_{\rm loc}(x)=\lambda^{-1}G(x,p(x)),
\]
and that
\[
\nabla M_{\rm loc}(x)=0
\quad\Longrightarrow\quad
0\in\partial f(x).
\]
\end{problem}

\clearpage
\section{P020: Robust remote nonconvergence}
\label{problem:P020}

\begin{problem}
Consider three players on a triangle, each with mixed-action mean
\(x_i\in[-1,1]\), and payoff \(x_i(C_\rho x)_i\), where
\[
C_\rho=
\begin{pmatrix}
0&1+\rho&-1\\
-1&0&1\\
1&-1&0
\end{pmatrix},
\qquad |\rho|<\frac14.
\]
Starting from a finite two-state log-odds history
\((q^{-1},q^0)\in\mathbb R^6\), all players use the exact recurrence
\[
\begin{aligned}
x_i^t=\tanh(q_i^t/2),\\
q^{t+1}=q^t+\eta C_\rho(2x^t-x^{t-1}),\\
t\geq0
\end{aligned}
\]
with arbitrary constant \(\eta>0\).

For each eigenvalue \(\lambda\) of \(C_\rho\), equivalently each root of
\[
\lambda^3+(3+\rho)\lambda-\rho=0,
\]
let \(z\) range over the two roots of
\[
z^2-(1+\eta\lambda)z+\frac{\eta\lambda}{2}=0,
\]
and define \(S(\rho,\eta)\) as the maximum modulus of these six values.
This quantity describes the joint local spectral effect of \(\rho\) and
\(\eta\).  In the parameter box below, \(\rho<0\), and the Nash set in
\([-1,1]^3\) is \(\{0\}\).

Decide whether there is a single \(\delta\in(0,1/100)\) such that, whenever
\[
|\rho+1/10|\leq\delta,\qquad
|\eta-1/20|\leq\delta,
\]
and every coordinate of both \(q^{-1}\) and \(q^0\) lies within \(\delta\) of
\(-10\), the following two conclusions hold:
\[
S(\rho,\eta)<1
\qquad\text{and}\qquad
\limsup_{t\to\infty}\|x^t\|_2>0.
\]
The same \(\delta\) must work for the full two-parameter box and all six
history coordinates.  The second assertion is pointwise for every selected
trajectory; no common positive separation constant, periodicity, or
recurrence is required.
\end{problem}

\clearpage
\section{P021: Worst reset-selector survival rate}
\label{problem:P021}

\begin{problem}
Let \(K=[-1,1]\), \(U=\{-1/2,1/2\}\), \(X_0=0\), and
\[
X_{t+1}=X_t+u_t+\frac14W_{t+1},
\]
where the \(W_t\) are independent standard normal variables.  The process
exits at \(\tau=\inf\{t\geq1:X_t\notin K\}\).  For bounded Borel \(f\), write
\[
\begin{aligned}
Q_uf(x)=,\\
\mathbb E[f(X_1)\mathbf1_{\{X_1\in K\}}\mid X_0=x,u_0=u],\\
\qquad,\\
V_1(x)=\max_{u\in U}Q_u\mathbf1(x).
\end{aligned}
\]
There are two planning stages, the running and terminal costs vanish, and the
risk budget is reset after each safe transition.  Thus every feasible plan is
minimizing.

Let the inward selector be
\(\mu_{\rm in}(x)=1/2\) for \(x<0\) and \(-1/2\) for \(x\geq0\), and set
\[
\begin{aligned}
m_{\rm in}&=\min_{x\in K}Q_{\mu_{\rm in}(x)}V_1(x),\\
m_{\rm all}&=\min_{x\in K,\,u\in U}Q_uV_1(x),\\
\theta_{\rm mid}&=(m_{\rm in}+m_{\rm all})/2,\qquad
r_{\rm mid}=1-\theta_{\rm mid}.
\end{aligned}
\]
At state \(x\), the feasible first actions are
\[
\mathcal F_{\rm mid}(x)=
\{u\in U:Q_uV_1(x)\geq\theta_{\rm mid}\}.
\]
Membership means that the action can be completed by a one-step
survival-maximizing tail satisfying the reset budget.  Let
\(\mathcal S_{\rm mid}\) be all stationary Borel selectors
\(\mu(x)\in\mathcal F_{\rm mid}(x)\).  Define \(\mu_g(x)\) to minimize
\(Q_u\mathbf1(x)\) over the feasible actions, breaking ties in favor of
\(-1/2\).

For \(\mu\in\mathcal S_{\rm mid}\), put
\[
\begin{aligned}
L_-(\mu)=\liminf_{T\to\infty},\\
\mathbb P_\mu(\tau>T)^{1/T},\\
L_+(\mu)=\limsup_{T\to\infty},\\
\mathbb P_\mu(\tau>T)^{1/T}
\end{aligned}
\]
and
\(\lambda_{\rm mid}^-=\inf_{\mu\in\mathcal S_{\rm mid}}L_-(\mu)\).
Decide whether the root limit
\[
L_g=\lim_{T\to\infty}\mathbb P_{\mu_g}(\tau>T)^{1/T}
\]
exists and satisfies
\[
\lambda_{\rm mid}^-=L_g<L_-(\mu_{\rm in}).
\]
The statement includes \(m_{\rm all}<m_{\rm in}\), nonempty feasible sections
with Borel graph, extension of every selector to a feasible two-step plan, and
recursive feasibility before exit.  Extremality must be proved against every
Borel selector, accounting for its quasi-stationary distribution; pointwise
one-step leakage inequalities are insufficient.
\end{problem}

\clearpage
\section{P022: Saddle accumulation under kernel convergence}
\label{problem:P022}

\begin{problem}
Work in dimension two.  Fix compact parameter sets
\(U\Subset U_+\), \(P\Subset P_+\), a compact localization set \(W\), and a
common open neighborhood \(N\supset W\cup U_+\).  Let \(h_k,h\) be
\(C^2\) Legendre kernels satisfying, uniformly on \(N\),
\[
\begin{aligned}
mI\preccurlyeq\nabla^2h_k,\nabla^2h\preccurlyeq MI,\\
\qquad,\\
\varepsilon_k:=\sup_N\|\nabla^2h_k-\nabla^2h\|\to0.
\end{aligned}
\]
After subtracting affine functions, assume \(h_k\to h\) in \(C^1(N)\), and
define the affine-gauge-invariant error
\[
\eta_k=\inf_{a\in\mathbb R^2}
\sup_{z\in N}\|\nabla h_k(z)-\nabla h(z)-a\|.
\]

Fix one proper lower-semicontinuous, prox-bounded, uniformly prox-regular
objective \(F_\star\).  For \((x,p)\in U_+\times P_+\), let
\[
q_{k}(x,p;y)=F_\star(y)-\langle p,y\rangle
+\lambda^{-1}D_{h_k}(y,x),
\]
with \(q\) defined using \(h\).  Suppose \(q(x,p;\cdot)\) has a
\(\kappa\)-Lipschitz branch \(T(x,p)\) of strong local minimizers, and that
fixed constants \(R,c>0\) give, throughout the closed \(R\)-ball,
\[
q(x,p;y)\geq q(x,p;T(x,p))+c\|y-T(x,p)\|^2,
\]
with \(T(x,p)\) the only local minimizer there and the ball contained in
\(\operatorname{int}W\).  No uniqueness or convergence is assumed for the
perturbed minima.

For \(0<r<R\), let \(\mathcal L_k^r(x,p)\) be all strong local minimizers of
\(q_k(x,p;\cdot)\) at distance less than \(r\) from \(T(x,p)\).  Decide
whether there always exist one \(r\in(0,R)\), \(C_{\rm disp}<\infty\), and
\(K\) such that, for all \(k\geq K\) and \((x,p)\in U\times P\),
\(\mathcal L_k^r(x,p)\) is a singleton \(\{T_k(x,p)\}\) and
\[
\|T_k(x,p)-T(x,p)\|\leq C_{\rm disp}\eta_k.
\]
The radius and constants are uniform in \(x,p\), and the conclusion concerns
strong local minima, not global minimizers over \(W\).
\end{problem}

\clearpage
\section{P023: Newton formula near a minimum sphere}
\label{problem:P023}

\begin{problem}
In \(\mathbb R^4\), let
\[
\begin{aligned}
S=\{z:\|z\|=1\},\\
f(z)=(\|z\|^2-1)^2,\\
q_0=(0,0,0,1)
\end{aligned}
\]
and work in the tube \(|\|z\|-1|<1/8\).  Let
\(Q\in\mathbb Q[u,v,w]\) have degree at most ten,
\(Q(0)=0\), and \(\nabla Q(0)=0\).  With
\[
\mathcal EQ=uQ_u+vQ_v+wQ_w,
\]
choose a positive rational \(\alpha\) and define
\[
p_{\alpha,Q}(z)=\alpha\left[
Q(z_1,z_2,z_3)
-\frac{\|z\|^2-1}{2}\mathcal EQ(z_1,z_2,z_3)\right].
\]
Assume on the tube that both
\(\|\nabla p_{\alpha,Q}\|\) and
\(\|\nabla^2p_{\alpha,Q}\|_{\rm op}\) are at most \(1/6400\).
The correction makes the radial derivative vanish on \(S\), and the radial
minimum is uniquely \(r=1\); hence the north-chart reduced germ of
\(F_{\alpha,Q}=f+p_{\alpha,Q}\) is exactly \(\alpha Q\).

Assume that \(0\) is a strict isolated local minimum of \(Q\), that \(Q\) is
convenient, and that for every nonempty coordinate set
\(I\subseteq\{u,v,w\}\), every compact face polynomial of the Newton
polyhedron of the restriction \(Q_I\) has no critical point on
\((\mathbb R\setminus\{0\})^I\).
For each compact facet \(\Gamma\), write its supporting hyperplane as
\[
\sum_{j\in I}\frac{\nu_j}{a_{\Gamma,j}}=1
\]
with positive intercepts, and let \(M_{\rm str}(Q)\) be the maximum of all
\(a_{\Gamma,j}\) over all coordinate restrictions and compact facets.  Set
\[
\theta_{\rm str}(Q)=1-\frac1{M_{\rm str}(Q)}.
\]

Let \(\theta_{\rm red}(Q)\) be the optimal local exponent in
\[
Q(\xi)^\theta\leq C\|\nabla Q(\xi)\|
\]
near \(0\) on \(Q>0\), and let
\(\theta_{\rm amb}(\alpha,Q)\) be the analogous optimal exponent for
\[
(F_{\alpha,Q}(z)-F_{\alpha,Q}(q_0))^\theta
\leq C\|\nabla F_{\alpha,Q}(z)\|
\]
near \(q_0\) on the positive level set.

Decide whether every admissible pair satisfies
\[
\theta_{\rm amb}(\alpha,Q)=
\theta_{\rm red}(Q)=\theta_{\rm str}(Q).
\]
The assertion also requires a terminating exact procedure which verifies the
strict-minimum and stratified nondegeneracy conditions, computes
\(M_{\rm str}(Q)\), and returns positive rational \(C,\delta\) certifying the
ambient inequality at the predicted exponent without deleting coordinate
strata or polar arcs.
\end{problem}

\clearpage
\section{P024: One-hold diffusion crossing law}
\label{problem:P024}

\begin{problem}
In the open unit disk, start from \(x_0=(r,0)\), where
\(0<r<1/3\), and let the diffusion be \(\sigma e_2\,dW_t\) with
\(\sigma>0\).  Over the single holding interval \([0,1]\), the available
drifts are \(0\) and \(-ae_1\), with \(0<a<1+r\).  A stipulated
time-zero selector chooses \(-ae_1\) when the first coordinate of the
projected observation is negative and \(0\) otherwise, including on the
boundary.

For \(0\leq t\leq1\), define
\[
q_a(t)=r-at,\qquad h_a(t)=\sqrt{1-q_a(t)^2},
\]
and
\[
P(a)=\mathbb P\{\exists t\in[0,1]:
|\sigma W_t|\geq h_a(t)\},\qquad P_0=P(0).
\]
Let \(F_a(\eta)\) be the supremum of the disk-exit probability over all
authorized time-zero observation errors of Euclidean norm at most \(\eta\),
with Euclidean projection onto the closed disk.  Given a nominal certificate
\(P_0\leq\alpha-s\), \(0<\alpha<1\), \(s>0\), define
\[
\eta_c(a,\alpha)=\sup\{\eta\geq0:F_a(\eta)\leq\alpha\}.
\]

Decide whether the following three assertions hold for every
\(r\in(0,1/3)\) and \(\sigma>0\).  First, there is exactly one
\(a_\star\in(2r,1+r)\) such that
\[
\begin{aligned}
P(a)<P_0\ (0<a<a_\star),\quad P(a_\star)=P_0,\quad,\\
P(a)>P_0\ (a_\star<a<1+r).
\end{aligned}
\]
Second, for all \(a,\eta\),
\[
F_a(\eta)=
\begin{cases}
P_0,&0\leq\eta\leq r,\\
\max\{P_0,P(a)\},&\eta>r.
\end{cases}
\]
Third, \(\eta_c=\infty\) when \(P(a)\leq\alpha\), while when
\(P(a)>\alpha\),
\[
\begin{aligned}
\eta_c=r,\\
F_a(r)=P_0\leq\alpha-s,\\
\lim_{\eta\downarrow r,\ \eta>r}F_a(\eta)=P(a)>\alpha.
\end{aligned}
\]
\end{problem}

\clearpage
\section{P025: Small-noise feasible-horizon transition}
\label{problem:P025}

\begin{problem}
Let a hidden drift \(\Theta\in\{-a,a\}\), \(a>0\), have prior probabilities
\(p_0,1-p_0\).  Starting from \(X_0=0\),
\[
dX_t=(\Theta+u_t)\,dt,\qquad |u_t|\leq U<a,
\]
and observations satisfy
\[
dY_t=X_t\,dt+\nu\,dV_t,\qquad \nu>0.
\]
Controls are bounded nonanticipative Borel functionals of the observation
path which give a unique strong closed loop in both drift modes.  The hard
state constraint is \(K=[-1,1]\), including both endpoints in prediction
safety.

Fix \(H>\Delta>0\), \(\varepsilon\in(0,1/2)\), and
\(\alpha\in(0,1)\).  At each grid time, the receding-horizon law must select a
measurable \(H\)-window observation-feedback continuation whose conditional
mixture probability of remaining in \(K\) throughout the window is at least
\(1-\varepsilon\).  It applies the choice for \(\Delta\) time and
reoptimizes.  Let \(\tau_F\) be the first grid time at which this conditional
feasible set is empty, and let \(T_\nu\) be the supremum of horizons \(T\)
for which some such law satisfies
\[
\mathbb P\{\min(\tau_K,\tau_F)>T\}\geq1-\alpha,
\]
with the prescribed zero convention when no initial selection exists.

Assume the resonant parameter relations
\[
\begin{aligned}
\tau=\frac1{a-U},\\
\tau-H=m\Delta,\\
r=m\Delta>\frac1a,\\
\ell=r+\Delta,\\
0<\Delta<H<\frac1{a+U}
\end{aligned}
\]
where \(m\) is a positive integer.  Put \(q=\max\{p_0,1-p_0\}\) and
\[
L=
\begin{cases}
r,&1-\alpha\geq q,\\
\ell,&1-\alpha<q.
\end{cases}
\]
The equality case belongs to the short phase.

Decide whether, for every such parameter tuple,
\[
\lim_{\nu\downarrow0}T_\nu=L.
\]
The limit is taken through positive-noise information structures, using fixed
regular conditional versions on canonical observation-path space.
\end{problem}

\clearpage
\section{P026: Positive four-step minimax schedules}
\label{problem:P026}

\begin{problem}
For any positive finite dimension, let \(f\) be convex, differentiable, and
\(L\)-smooth, with minimizer \(x_\star\) and
\(\|x_0-x_\star\|\leq D\).  For a nonadaptive four-step schedule
\(h\in\mathbb R^4\), run
\[
x_{t+1}=x_t-\frac{h_t}{L}\nabla f(x_t),\qquad0\leq t<4,
\]
and define
\[
W_4(h)=\sup
\frac{f(x_4)-f(x_\star)}{LD^2},
\]
where the supremum ranges over every admissible instance in every finite
dimension.  Set
\[
V_4=\inf_{h\in\mathbb R^4}W_4(h),\qquad
\mathcal H_4=\{h:W_4(h)=V_4\},
\]
\[
\mathcal N_4=\{h:\min_{0\leq t\leq3}h_t\leq0\},\qquad
V_4^-=\inf_{h\in\mathcal N_4}W_4(h).
\]
The scalar quadratic lower barrier
\[
Q_4(h)=\max_{0\leq a\leq1}
\frac a2\prod_{t=0}^3(1-ah_t)^2
\]
may be used, but does not replace the complete smooth-convex worst case.

Decide whether
\[
V_4^->V_4.
\]
After establishing the necessary attainment and compactness facts, this is
equivalent to every \(h\in\mathcal H_4\) having all four coordinates strictly
positive.  The problem asks
only for this sign geometry, not the exact value \(V_4\), a classification of
\(\mathcal H_4\), or an arbitrary-horizon theorem.
\end{problem}

\clearpage
\section{P027: Perfect-incoherence collision constant}
\label{problem:P027}

\begin{problem}
Fix \(d\geq1\).  For each \(k\), let
\(L_{\star,k}\in\mathbb R^{n_k\times n_k}\) have exact rank \(r_k\), smallest
nonzero singular value \(\sigma_k>0\), and singular spaces \(U_k,V_k\).
Assume \(n_k\to\infty\),
\[
\frac{r_k}{n_k}=\frac1{4d},
\]
and perfect incoherence:
\[
\|U_k^{\mathsf T}e_i\|^2=
\|V_k^{\mathsf T}e_j\|^2=\frac1{4d}
\quad\text{for every }i,j.
\]
Let \(\Omega_{1,k},\Omega_{2,k}\) be two supports, each separately having at
most \(d\) entries in every row and every column.  With
\(\Omega_k=\Omega_{1,k}\cup\Omega_{2,k}\) and \(T_{\star,k}\) the tangent
space of the rank-\(r_k\) manifold, define
\[
\tau_k=
\left(1-\|P_{\Omega_k}P_{T_{\star,k}}\|_{F\to F}^2\right)^{1/2}>0
\]
and assume \(\tau_k\to0\).

Let \(R_k\) be the global infimum of
\(\|L'-L_{\star,k}\|_F\) over all distinct competitors \(L'\) of rank at most
\(r_k\), satisfying the same leverage cap, for which
\[
L_{\star,k}+S=L'+T,\qquad
\operatorname{supp}S\subseteq\Omega_{1,k},\quad
\operatorname{supp}T\subseteq\Omega_{2,k}.
\]
Rank-dropping and otherwise undisplayed competitors are included; if none
exists, \(R_k=\infty\).  Define
\[
\kappa_d=
\inf_{\text{all admissible sequences}}
\liminf_{k\to\infty}\frac{R_k}{\sigma_k\tau_k}.
\]

Decide whether
\[
\kappa_d=2\sqrt2
\qquad\text{for every }d\geq1.
\]
Scale normalization may set \(\sigma_k=1\), but no lower bound may be imposed
on any other singular value.
\end{problem}

\clearpage
\section{P028: Open-basin cyclic OMWU failure}
\label{problem:P028}

\begin{problem}
For \(0<q<1\), set
\[
J_q=\frac1{\sqrt{1+q+q^2}}
\begin{pmatrix}0&1&-q\\-q&0&1\\1&-q&0\end{pmatrix}.
\]
Consider the canonical three-player, two-action triangle game whose affine
payoff-difference map vanishes at
\((1/2,1/2,1/2)\) and has normalized Jacobian \(J_q\).  This profile is
required to be the unique Nash equilibrium over all support patterns.  From
arbitrary strictly interior profiles at times \(-1,0\), all players run exact
optimistic multiplicative weights with learning rate \(1/16\).  In logit
coordinates \(y_i=\log(x_i/(1-x_i))\) and
\(z=x-(1/2,1/2,1/2)\), the history map is
\[
\mathcal T_q(y,y')=
\left(y-\frac1{16}J_q(2z-z'),\,y\right).
\]

Let \(E_q\) be the set of strictly interior two-profile histories whose exact
profile orbit is not Cauchy, equivalently, by compactness, has at least two
distinct accumulation points.  Interior is relative to the six-dimensional
history space
\[
\mathcal H=(\operatorname{int}\Delta_2^3)^2\cong(0,1)^6.
\]
It is known that \(E_q\ne\varnothing\) throughout some punctured interval
ending at \(q=1\).

Decide whether there exists \(q_0\in(0,1)\) such that
\[
\operatorname{int}_{\mathcal H}E_q\ne\varnothing
\qquad(q_0<q<1).
\]
The exact negative alternative is that for every \(q_0<1\), some
\(q\in(q_0,1)\) has \(E_q\) of empty relative interior.  Membership always
refers to the exact infinite-time recurrence.  The problem concerns
robustness under perturbing the initial history inside this canonical family,
not perturbations of the game Jacobian or mere nonemptiness of \(E_q\).
\end{problem}

\clearpage
\section{P029: Bounded split certificates in defect one}
\label{problem:P029}

\begin{problem}
Let \(X\) be a real Banach space for which \(J_X(X)\) has codimension one in
\(X^{**}\).  Let \(A,B:X\rightrightarrows X^*\) be maximal monotone and
\[
\operatorname{dom}A\cap\operatorname{int}(\operatorname{dom}B)\ne\varnothing.
\]
Suppose \((x,x^*)\notin\operatorname{gra}(A+B)\), but
\[
\langle x-y,x^*-a^*-b^*\rangle\geq0
\]
for every \(y\in X\), \(a^*\in Ay\), and \(b^*\in By\).

For maximal monotone \(T\), let
\[
F_T(x,u^*)=\sup_{(y,y^*)\in\operatorname{gra}T}
\bigl(\langle x,y^*\rangle+\langle y,u^*\rangle
-\langle y,y^*\rangle\bigr).
\]
A finite lifted graph--epigraph system for \(T\) consists of probabilities
\(\theta_i\), graph points \((y_i,y_i^*)\), and excesses \(\delta_i\geq0\),
with barycentres
\[
\begin{aligned}
\bar y=\sum_i\theta_i y_i,\quad,\\
\bar y^*=\sum_i\theta_i y_i^*,\quad,\\
h_T=\sum_i\theta_i(\langle y_i,y_i^*\rangle+\delta_i).
\end{aligned}
\]
Pair one system for \(A\) with one for \(B\) and define its payoff
\[
\mathcal P=
\langle x,\bar a^*+\bar b^*\rangle
+\langle\bar b-x,x^*\rangle-h_A-h_B.
\]

Decide whether the positive Fitzpatrick split separation can always be
represented by a net of such paired systems for which
\(J_X(\bar a)\) and \(J_X(\bar b)\) converge weak-star in \(X^{**}\) to the
same point, the two dual barycentres converge weak-star in \(X^*\), the
heights have finite limits, and
\[
\liminf\mathcal P\geq\eta
\]
for some \(\eta>0\), while one constant \(R<\infty\) uniformly bounds
\[
\|\bar a\|+\|\bar b\|+\|\bar a^*\|+\|\bar b^*\|
+|h_A|+|h_B|.
\]
No common-primal graph synchronization or claim of maximality of \(A+B\) is
part of the target.
\end{problem}

\clearpage
\section{P030: Unit-weight elliptic moment obstructions}
\label{problem:P030}

\begin{problem}
Let
\[
E=\{(x,y)\in\mathbb R^2:
y^2=(1-x^2)(2-x),\ -1\leq x\leq1\},
\]
let \(V_d\) be the restrictions to \(E\) of real plane polynomials of total
degree at most \(d\), and let \(\mathcal U_r(E)\) consist of the probability
measures assigning weight \(1/r\) to each of exactly \(r\) distinct points of
\(E\).  Write
\[
\Phi_d(\mu)(p)=\int_Ep\,d\mu,\qquad p\in V_d.
\]

Decide whether, for every \(d,r\geq1\) with \(3d\leq2r\), there are distinct
\(\mu,\nu\in\mathcal U_r(E)\) such that
\[
\Phi_d(\mu)=\Phi_d(\nu).
\]
These are unit-coefficient obstructions, stronger than arbitrary balanced
weighted dependencies.  One possible construction starts with
\(s=\lceil3d/2\rceil\), distinct \(x_i\in(-1,1)\),
\(y_i=\sqrt{(1-x_i^2)(2-x_i)}\), and signs \(\varepsilon_i\) satisfying
\[
\sum_{i=1}^s\varepsilon_i y_ix_i^j=0,
\qquad0\leq j\leq s-2,
\]
then vertically flips the atoms and pads both measures with common new atoms
when \(s<r\).

The assertion also requires at least one ambiguous pair to admit the
following canonical sparse-moment lift.  Use clique variables
\((x,y,u)\) and \((x,y,v)\), set the private coordinates to zero, retain
moments through degree \(2r\), impose the oval equation, interval and compact
ball constraints, and require
\[
\operatorname{rank}M_r=\operatorname{rank}M_{r-1}=r
\]
in both cliques.  Couple the two independently flat sequences by equality of
exactly the separator integrals in \(V_d\), with no implicit higher-degree
identifications.  There must be no global representing measure having both
distinct separator marginals.

Nonuniform weights and prescribed support configurations are outside the
claim.
\end{problem}

\clearpage
\section{P031: Directional recovery of optimistic Clarke slopes}
\label{problem:P031}

\begin{problem}
Let
\[
\begin{aligned}
q(x,z)&=\frac12 z^{\mathsf T}Hz+(c+xa)^{\mathsf T}z,\\
S(x)&=\operatorname*{arg\,min}_{z\in P}q(x,z),\\
\phi_F(x)&=\min_{z\in S(x)}F(x,z),\\
A_F(x)&=\operatorname*{arg\,min}_{z\in S(x)}F(x,z).
\end{aligned}
\]
Let \(K(x)\) be the set of pairs \((z,\lambda)\) satisfying
\[
\begin{aligned}
Az&\leq b,&
Hz+c+xa+A^{\mathsf T}\lambda&=0,\\
\lambda&\geq0,&
\lambda_i(A_i z-b_i)&=0\quad\text{for every }i.
\end{aligned}
\]

A tuple \((\bar z,\lambda,w,\nu,d)\) is a formal directional KKT response
when \((\bar z,\lambda)\in K(0)\),
\[
Hw+da+A^{\mathsf T}\nu=0,
\]
and, for every \(i\),
\[
\begin{array}{ll}
A_i\bar z<b_i &\Longrightarrow \nu_i=0,\\
\lambda_i>0 &\Longrightarrow A_iw=0,\\
A_i\bar z=b_i,\ \lambda_i=0
&\Longrightarrow
A_iw\leq0,\ \nu_i\geq0,\ \nu_iA_iw=0.
\end{array}
\]
Let \(Z_d\) be the set of \(\bar z\in S(0)\) for which such
\(\lambda,w,\nu\) exist.

For \(\varepsilon>0\), \(B\in\mathcal M\), and \(\bar z\in S(0)\), define
\[
q_{\varepsilon,B,\bar z}(x,z)
=q(x,z)+\frac{\varepsilon}{2}
(z-\bar z)^{\mathsf T}B(z-\bar z),
\]
and let \(s_{\varepsilon,B,\bar z}(x)\) be its unique minimizer over \(P\).
For \(d\in\{-1,1\}\), set
\[
r_{d,\varepsilon,B,F}(\bar z)
=d\lim_{t\downarrow0}
\frac{F\bigl(td,s_{\varepsilon,B,\bar z}(td)\bigr)-F(0,\bar z)}{t}.
\]

Let \(\mathcal R_d(F)\) be the set of all finite limits of
\[
r_{d,\varepsilon_k,B_k,F}(z_k)
\]
along sequences \(\varepsilon_k\downarrow0\), \(B_k\in\mathcal M\), and
\(z_k\in A_F(0)\cap Z_d\).  When this set is nonempty and the indicated
value is finite, define
\[
\sigma_d(F)=d\inf\{dr:r\in\mathcal R_d(F)\}.
\]
Thus
\[
\sigma_1(F)=\inf\mathcal R_1(F),\qquad
\sigma_{-1}(F)=\sup\mathcal R_{-1}(F),
\]
and \(C_F\) is the closed convex hull of
\(\{\sigma_{-1}(F),\sigma_1(F)\}\).

For every upper-active formal response, define
\[
g=d^{-1}\left[
d\,\partial_xF(0,\bar z)
+\nabla_zF(0,\bar z)^{\mathsf T}w
\right].
\]
Let \(\Gamma_F\) be the closed convex hull of all such \(g\), over
\(d\in\{-1,1\}\), \(\bar z\in A_F(0)\), and all \(\lambda,w,\nu\)
satisfying the formal directional response system.

The Clarke set \(\partial^C\phi_F(0)\) is the closed convex hull of all
finite limits of ordinary derivatives \(\phi_F'(x_k)\) along
differentiability points \(x_k\to0\).

The parameter x is scalar, the reference parameter is 0, and all local directional statements use x=t d with d\(\in\)\{-1,1\} and t\(\downarrow\)0.

The lower problem minimizes q(x,z)=z\(^{T}\)Hz/2+(c+xa)\(^{T}\)z over the fixed nonempty compact polyhedron P=\{z:Az\(\leq\)b\}, where H is positive semidefinite.

The reference solution set S(0) is nonempty and positive-dimensional; nearby solution sets remain nonempty and compact because P is fixed and compact.

The lower polyhedral constraints satisfy Slater regularity and the constant-rank constraint qualification at every reference solution.

Reference-only transverse quadratic growth holds: there exist \(\kappa >\)0 and r\(>\)0 such that q(0,z)-min over P of q(0,\(\cdot\))\(\geq \kappa\) dist(z,S(0))\(^2\) for every z\(\in\)P with dist(z,S(0))\(<\) r. No uniform growth assertion for x≠0 is assumed.

The upper objective F is twice continuously differentiable, and the universally quantified upper-objective class consists exactly of those such F for which the optimistic value \(\phi_{F}\) is locally Lipschitz near 0.

Scalar affine-parametric convex quadratic lower problems over fixed compact polyhedra satisfying exactly A1-A6. All directional KKT equations, accessible sets, regularized slope sets, directional extrema, \(\Gamma_{F}\), and the Clarke set are defined below without external references. Auxiliary strongly convex regularizations need not satisfy A3.

A center \(\bar{z}\) contributes in direction d only when \(\bar{z}\) belongs to \(A_{F}\)(0)\(\cap Z_{d}\).

The metric B ranges over all symmetric positive-definite matrices with trace(B)=1.

For fixed \(\varepsilon\), B, \(\bar{z}\), and d, first take the derivative as t decreases to zero and only then take cluster limits as \(\varepsilon\) decreases to zero.

After forming the raw finite slope-cluster set \(\mathbb R_{d}\)(F), retain \(\sigma_{d}\)(F)=d inf\{d r:r\(\in \mathbb R_{d}\)(F)\}; equivalently \(\sigma_{1}\)=inf \(\mathbb R_{1}\) and \(\sigma_{-1}\)=sup \(\mathbb R _{-1}\).

The positive target includes proving that both raw cluster sets are nonempty and that their directionally relevant extrema are finite real numbers.

For every A1-A6 instance and authorized upper objective F, determine whether centered Tikhonov regularizations over all normalized positive-definite metrics recover the two Clarke endpoints after restricting centers to directionally accessible upper-active reference solutions and applying the optimistic directional extremum. Raw regularization slopes are not convexified directly: the positive-side endpoint is their infimum and the negative-side endpoint is their supremum. Determine also whether the resulting endpoint hull is contained in the fully explicit formal KKT slope hull \(\Gamma_{F}\).
\end{problem}

\clearpage
\section{P032: Exact stationary penalties with fixed lower normals}
\label{problem:P032}

\begin{problem}
For each \(i\), define
\[
c_i=\lVert a_i\rVert_2>0,\qquad
q_i(y)=\frac{a_i^{\mathsf T}y-b_i}{c_i},\qquad
\nabla_yq_i=\frac{a_i}{c_i}.
\]
Let \(\phi_{\mathrm{FB}}(a,b)=\sqrt{a^2+b^2}-a-b\), and define
\[
\begin{aligned}
\mathcal R_{\mathrm{KKT}}(x,y,\mu)
={}&\left\lVert\nabla_yf(x,y)
    +\sum_i\mu_i\nabla_yq_i\right\rVert_1\\
&+\sum_i\left([q_i(y)]_+ +[-\mu_i]_+\right.\\
&\hspace{3.5em}\left.
    +\left|\phi_{\mathrm{FB}}(-q_i(y),\mu_i)\right|\right).
\end{aligned}
\]
For \(\rho>0\), set
\[
\Phi_\rho(x,y,\mu)=-x+\rho\mathcal R_{\mathrm{KKT}}(x,y,\mu).
\]
A triple \((x,y,\mu)\) is ambient Clarke-stationary when \(0\) belongs
to the Clarke subdifferential of \(\Phi_\rho\) with respect to \(x,u,v\)
and every multiplier coordinate.

The upper objective F, lower objective f, and lower constraint functions \(g_{i}\) are twice continuously differentiable near the reference pair; the upper problem has no locally active constraint, and lower constraints use the sign convention \(g_{i}\)(x,y) less than or equal to zero.

For every nearby x, the lower problem is convex in y, has a nonempty locally compact solution set S(x), and its normalized KKT system characterizes S(x).

Fix a reference KKT triple (xbar,ybar,mubar). The set S(xbar) contains a positive-dimensional face H through ybar. \(\bar{\Lambda}\) denotes the complete set of normalized KKT multipliers at (xbar,ybar), contains mubar, and need not be a singleton; isolated follower solutions and strong regularity are not assumed.

Every subfamily of lower-constraint gradients active near H has locally constant rank, and every constraint included in the residual has \(c_{i}\)=\(norm_{2}\)(\(\nabla_{y} g_{i}\)(xbar,ybar)) strictly positive.

There exist neighborhoods X of xbar and Y of ybar and a constant \(\gamma >\)0 such that, for every x in X and every lower-feasible y in Y, f(x,y)-\(\min_{w}\)\{f(x,w):\(g_{i}\)(x,w) less than or equal to zero for all i\} is at least \(\gamma\) times dist(y,aff(H)) squared. This is growth only normal to the reference face, not growth in dist(y,S(x)); inner semicontinuity is not assumed because face collapse is the failure under investigation.

The residual uses \(q_{i}\)=\(g_{i}\)/\(c_{i}\), multipliers \(\mu_{i}\) for \(q_{i}\), the componentwise formula stated in the central object, and ambient stationarity in all variables (x,y,\(\mu\)). Any proposed exactness threshold must use one fixed open neighborhood U of \{xbar\} times \{ybar\} times \(\bar{\Lambda}\) that is independent of \(\rho\).

The parameter x is scalar, y=(u,v) is in \(\mathbb R^{2}\), F(x,y)=-x, and there are at most four lower constraints. Each \(g_{i}\) is affine in y and independent of x, while f is a jointly total-degree-at-most-two polynomial that is convex in y. The system satisfies A1-A6, H=\{(0,v):-1\(\leq\)v\(\leq\)1\} is a face of S(0), ybar=(0,0) lies in relint(H), and the complete normalized reference multiplier set is nonempty and compact. No residual-slope, error-bound, multiplier-uniqueness, inner-semicontinuity, or face-persistence conclusion is assumed.

For 1\(\leq\)i\(\leq\)m\(\leq\)4, \(g_{i}\)(x,y)=\(a_{i}^{T}\) y-\(b_{i}\) with constant \(a_{i}\) and \(b_{i}\); hence every normalized gradient \(\nabla_{y} q_{i}\) is independent of x and y.

The lower objective f(x,u,v) is a real polynomial of joint total degree at most two and is convex in (u,v) for every nearby x.

The reference point is xbar=0, ybar=(0,0), and H=\{(0,v):-1\(\leq\)v\(\leq\)1\} is contained in S(0) with ybar in relint(H).

The complete normalized multiplier set \(\bar{\Lambda}\) at (0,(0,0)) is nonempty and compact, but may be non-singleton.

The upper objective is F(x,u,v)=-x and has no active upper constraint.

Exactness means \(\mathbb R_{KKT}\)(x,y,\(\mu\))=0 for the full normalized KKT triple, including complementarity and multiplier correctness.

Prove or refute that parameter-independent affine lower constraints eliminate the rotating-normal cancellation responsible for jointly quadratic stationary nonexactness, so that the normalized Fischer-Burmeister lower-KKT residual is finitely exact for ambient Clarke stationarity near the relative interior of a positive-dimensional follower-solution face.
\end{problem}

\clearpage
\section{P033: Excursion-wise coercivity for inexact proximal gradients}
\label{problem:P033}

\begin{problem}
Let \(F=f+g\), and define
\[
d_k=x_{k+1}-x_k,\qquad c_k=\frac1{t_k}-\frac{L_k}{2}.
\]
For \(d_k\neq0\), set
\[
\theta_k=\frac{\lVert r_k\rVert}{\lVert d_k\rVert},
\qquad s_k=c_k-\theta_k.
\]
A zero step contributes the zero matrix.  For \(\rho>0\), let
\[
q_\rho(p)=\min\{q>p:\lVert x_q-x_p\rVert\geq\rho\},
\]
provided this set is nonempty.  For a completed excursion, define
\[
\begin{aligned}
D_{p,\rho}&=\sum_{k=p}^{q_\rho(p)-1}d_kd_k^{\mathsf T},\\
A_{p,\rho}&=\sum_{k=p}^{q_\rho(p)-1}s_kd_kd_k^{\mathsf T}.
\end{aligned}
\]

Work in a finite-dimensional Euclidean space with F=f+g, where f is continuously differentiable with globally L-Lipschitz gradient, g is proper closed convex, and F is bounded below.

The generated trajectory is bounded; for any proposed single-point convergence conclusion, F has the Kurdyka-Lojasiewicz property on the trajectory's cluster set.

For every accepted update \(d_{k}\)=\(x_{k+1}\)-\(x_{k}\), backtracking returns \(t_{k}\) in {[}\(t_{min}\),\(t_{max}\){]} and an observable \(L_{k}\) in {[}0,L̄{]} for some finite L̄ such that f(\(x_{k+1}\)) is at most f(\(x_{k}\))+\(< \nabla\) f(\(x_{k}\)),\(d_{k} >\)+(\(L_{k}\)/2)‖\(d_{k}\)‖\(^{2}\), and \(c_{k}\)=1/\(t_{k}\)-\(L_{k}\)/2 is positive.

The inner solver returns an observable residual \(r_{k}\) satisfying \(r_{k}\) in \(\nabla\) f(\(x_{k}\))+partial g(\(x_{k+1}\))+\(d_{k}\)/\(t_{k}\).

For \(d_{k}\) nonzero define \(\theta_{k}\)=‖\(r_{k}\)‖/‖\(d_{k}\)‖ and require \(\theta_{k} \leq c_{k}\)-\(m_{k}\) for an observable \(m_{k}\) in (0,\(c_{k}\){]}; if \(d_{k}\)=0 require \(r_{k}\)=0. No uniform positive lower bound, summability, eventual exactness, or prescribed decay is assumed for \(m_{k}\) or \(r_{k}\).

The choices of \(t_{k}\) and \(m_{k}\) may use all prior residuals and all current backtracking outcomes, but the accepted pair (\(x_{k+1}\),\(r_{k}\)) must satisfy A3-A5 before the next outer iteration.

Finite-dimensional Euclidean composite optimization under exactly A1-A6. The condition uses maximal observable slack and endogenous first-exit blocks rather than calendar windows. The coefficient \(\eta_{\rho}\) may depend on the fixed excursion scale and need not remain positive as \(\rho\) tends to zero. Individual slacks may vanish, excursion lengths may diverge, and no uniform relative-error gap, residual summability, prescribed schedule, finite-length premise, dyadic coercivity, strong convexity, or isolated-critical-point condition is imposed.

For \(d_{k}\) nonzero define \(\theta_{k}\)=‖\(r_{k}\)‖/‖\(d_{k}\)‖ and \(s_{k}\)=\(c_{k}\)-\(\theta_{k} >\)0; a zero step contributes the zero matrix.

For each fixed \(\rho >\)0 and starting index p, \(q_{\rho}\)(p) is the least q\(>\) p satisfying ‖\(x_{q}\)-\(x_{p}\)‖\(\geq \rho\), when such q exists.

For every fixed \(\rho >\)0 there exist \(\eta_{\rho} >\)0 and \(P_{\rho}\) such that \(A_{p,\rho} \geq \eta_{\rho} D_{p,\rho}\) in positive-semidefinite order for every completed minimal \(\rho\)-excursion starting at p\(\geq P_{\rho}\).

The constants may depend on \(\rho\), and no coercivity is required on dyadic, fixed-length, or arbitrary non-excursion windows.

Consider any bounded trajectory satisfying A1-A6. For each nonzero step set \(s_{k}\)=\(c_{k}\)-‖\(r_{k}\)‖/‖\(d_{k}\)‖. For \(\rho\)>0 and p\(\geq\)0, let \(q_{\rho}\)(p) be the first q>p with ‖\(x_{q}\)-\(x_{p}\)‖\(\geq \rho\), when it exists. On this minimal excursion define \(D_{p,\rho}\)=\(\sum_{k=p}^{q_{\rho}(p)-1} d_{k} d_{k}^{T}\) and \(A_{p,\rho}\)=\(\sum_{k=p}^{q_{\rho}(p)-1} s_{k} d_{k} d_{k}^{T}\). Determine whether convergence to one critical point follows if, for every fixed \(\rho\)>0, there are \(\eta_{\rho}\)>0 and \(P_{\rho}\) such that \(A_{p,\rho}\) is positive-semidefinite greater than or equal to \(\eta_{\rho} D_{p,\rho}\) for every p\(\geq P_{\rho}\) having a completed \(\rho\)-excursion.
\end{problem}

\clearpage
\section{P034: Discounted error budgets for proximal gradients}
\label{problem:P034}

\begin{problem}
Let \(F=f+g\), and define
\[
q_k=\frac{\lVert x_{k+1}-x_k\rVert^2}{t_k}.
\]
The proximal-gradient model and its gap are
\[
\begin{aligned}
Q_k(y)&=f(x_k)+\langle\nabla f(x_k),y-x_k\rangle
+\frac{\lVert y-x_k\rVert^2}{2t_k}+g(y),\\
\gamma_k&=Q_k(x_{k+1})-\inf_yQ_k(y).
\end{aligned}
\]
The discounted anti-hoarding rule is
\[
0\leq B_{k+1}\leq\eta B_k+\beta q_k
\qquad\text{for every }k,
\]
where \(0<\eta<1\) and \(0\leq\beta<\infty\) are fixed.  A trajectory has
finite length when
\[
\sum_{k\geq0}\lVert x_{k+1}-x_k\rVert<\infty.
\]

A point x is critical when 0 belongs to \(\nabla\) f(x)+partial g(x).

The ambient space is finite-dimensional Euclidean space, F=f+g is proper and bounded below, f is continuously differentiable with an L-Lipschitz gradient for an unknown finite L, and g is proper, closed, and convex.

The initial sublevel set is bounded, the critical set is nonempty, and F has the Kurdyka-Lojasiewicz property on the relevant sublevel set; its exponent and desingularizing function are not algorithm inputs.

At iteration k, geometric backtracking selects a positive stepsize \(t_{k}\) bounded away from zero and \(\infty\) and accepts an approximate minimizer \(x_{k+1}\) of the standard strongly convex proximal-gradient model.

The inner solver returns a computable certificate \(e_{k}\) equal to or dominating the proximal-model objective gap, and accepted iterates satisfy F(\(x_{k+1}\)) \(\leq\) F(\(x_{k}\))-c\(\star\)‖\(x_{k+1}\)-\(x_{k}\)‖\(^{2}\)/\(t_{k}\)+\(e_{k}\) for a fixed c\(>\)0.

For fixed \(\rho\) in (0,1), the account starts with \(B_{0}\)=0 and satisfies \(B_{k+1}\)=\(B_{k}\)+\(\rho\)\emph{c}‖\(x_{k+1}\)-\(x_{k}\)‖\(^{2}\)/\(t_{k}\)-\(e_{k} \geq\) 0.

No exogenous summability, decay rate, eventual exactness, or per-iteration relative upper bound is imposed on \(e_{k}\) beyond the accepted-descent test and the account constraint.

Finite-dimensional composite objectives and backtracking inexact proximal-gradient trajectories satisfying exactly A1-A6, specialized to c=1/4 and \(\rho\)=1/2. The discount η is fixed strictly between zero and one, \(\beta\) is fixed and finite, and certificates may strictly dominate their true model gaps. No exogenous accuracy schedule, eventual exactness, strong convexity, isolated-critical-point condition, or prescribed error decay is allowed. Unlike the established age-one capability \(B_{k+1} \leq \beta q_{k}\), the discounted rule permits positive balance inherited from arbitrarily old descent, with geometrically decaying influence.

Set c=1/4 and \(\rho\)=1/2, so \(B_{k+1}\)=\(B_{k}\)+\(q_{k}\)/8-\(e_{k}\) and F(\(x_{k}\))+\(B_{k}\) decreases by at least \(q_{k}\)/8 at each accepted step.

Fix η in (0,1) and \(\beta\) in {[}0,\(\infty\)), and require 0\(\leq B_{k+1} \leq\)η \(B_{k}\)+\(\beta q_{k}\) for every k.

Quantify over every certificate authorized by A4, including certificates strictly larger than the true proximal-model gap.

The proof may use the established finite age-one-cap result only on a legitimately derived age-one-capped trajectory; it may not assume that the discounted rule implies \(B_{k+1} \leq\)C \(q_{k}\) for some finite C.

Fix c=1/4 and \(\rho\)=1/2. For \(q_{k}\)=‖\(x_{k+1}\)-\(x_{k}\)‖\(^{2}\)/\(t_{k}\), fix η in (0,1) and finite \(\beta \geq\)0 and augment A5 by the computable discounted cap 0\(\leq B_{k+1} \leq\)η \(B_{k}\)+\(\beta q_{k}\). Determine whether every resulting trajectory has finite length and converges to one critical point.
\end{problem}

\clearpage
\section{P035: Weak-only convergence under harmonic metric drift}
\label{problem:P035}

\begin{problem}
\(r_{k+1}=M_{k}(x_{k}-x_{k+1})-\lambda(a_{k+1}^{x}+2Bx_{k}-Bx_{k-1}\)), with \(a_{k+1}^{x}\) in A \(x_{k+1}\).

Set \(y_{-1}=x_{-1}\) and \(y_{0}=x_{0}\). For every k greater than or equal to 0, \(y_{k+1}\) is the unique point for which there exists \(a_{k+1}^{y}\) in A \(y_{k+1}\) satisfying \(M_{k}(y_{k}-y_{k+1})=\lambda(a_{k+1}^{y}+2By_{k}-By_{k-1}\)).

\(\delta_{k}\) is the least nonnegative \(\delta\) such that \((1+\delta)^{-1} M_{k}\) is less than or equal to \(M_{k+1}\), which is less than or equal to \((1+\delta)M_{k}\) in Loewner order.

When \(x_{k+1}\) differs from \(x_{k}, \varepsilon_{k}\)=‖\(r_{k+1}\)‖\_\{\(M_{k}^{-1}\)\}/‖\(x_{k+1}-x_{k}\)‖\(_{M_{k}}\); otherwise \(\varepsilon_{k}\)=0 and \(r_{k+1}\)=0.

An orbit z has persistent strong failure when \(z_{k}\) converges weakly to 0 and \(limsup_{k}\) ‖\(z_{k}\)‖\(>\)0.

H is a separable infinite-dimensional real Hilbert space.

A:H⇉H is a nonzero linear maximally monotone operator, B:H\(\to\)H is a nonzero bounded linear monotone L-Lipschitz operator, and zer(A+B)=\{0\}.

Each \(M_{k}\) is bounded and self-adjoint with mI\(\preccurlyeq M_{k}\preccurlyeq\)MI for fixed 0\(<\) m\(\leq\)M\(< \infty, M_{kx} \to M_{\in ftyx}\) for every x\(\in\)H, and \(\delta_{k}\) is the least \(\delta \geq\)0 such that \((1+\delta)^{-1} M_{k}\preccurlyeq M_{k+1}\preccurlyeq(1+\delta)M_{k}\).

There exist constants 0\(<\) c\(\leq\)C\(< \infty\) and K such that c/(k+1)\(\leq \delta_{k} \leq\)C/(k+1) for every k\(\geq\)K; in particular, (\(\delta_{k}\)) belongs to \(\ell^{p}\) for every p\(>\)1 and does not belong to \(\ell^{1}\).

The constant stepsize satisfies 0\(< \lambda <\) m/(2L).

For each k there are \(a_{k+1} \in Ax_{k+1}\) and \(r_{k+1}=M_{k}(x_{k}-x_{k+1})-\lambda(a_{k+1}+2Bx_{k}-Bx_{k-1}\)). Set \(\varepsilon_{k}\)=‖\(r_{k+1}\)‖\_\{\(M_{k}^{-1}\)\}/‖\(x_{k+1}-x_{k}\)‖\(_{M_{k}}\) when \(x_{k+1}\)≠\(x_{k}\) and \(\varepsilon_{k}\)=0 otherwise, requiring \(r_{k+1}\)=0 in the latter case. The realized ratios satisfy \(\Sigma_{k} \varepsilon_{k} < \infty\), and \(r_{k+1}\)≠0 for infinitely many k.

For the same A, B, \(\lambda\), limiting metric \(M_{\infty}\), and every initial pair, the exact fixed-\(M_{\infty}\) forward-reflected-backward orbit converges strongly to 0.

Along the candidate orbit there are \(c_{B} >\)0 and \(K_{B}\) such that ‖\(Bx_{k}\)‖\(\geq c_{B}\)‖\(x_{k}\)‖ for all k\(\geq K_{B}\), and the selected values \(a_{k+1}\) are nonzero infinitely often. For the same A, metrics, \(\lambda\), and every initial pair, deleting B and setting all residuals to zero yields a variable-metric proximal orbit converging strongly to 0.

Separable infinite-dimensional real Hilbert-space data satisfying exactly A1--A8. The auxiliary sequence y is the exact variable-metric FRB replay using the same A, B, metrics, stepsize, and initial pair as the selected A6 orbit; it is an additional target object, not a replacement for the genuinely inexact selected orbit.

The selected orbit x and exact replay y use identical H, A, B, \(M_{k}, \lambda, x_{-1}=y_{-1}\), and \(x_{0}=y_{0}\).

The selected orbit x must still satisfy A6, including summable realized relative residual ratios and infinitely many nonzero residuals; only the auxiliary replay y has zero residual.

The harmonic bounds apply to the least two-sided relative drift \(\delta_{k}\) from A3.

The exact replay must satisfy \(y_{k}\) weakly converging to 0 and \(limsup_{k}\) ‖\(y_{k}\)‖\(>\)0.

The every-initial-pair fixed-limit control in A7 and B-deleted variable-metric control in A8 must hold for the same operators and metrics.

Determine whether the established weak-without-strong behavior can be realized as a metric-driven phenomenon: seek one datum satisfying A1–A8 whose selected inexact orbit \(x_{k}\) converges weakly but not strongly to zero and whose unique residual-free variable-metric FRB replay \(y_{k}\) from the same initial pair also converges weakly but not strongly to zero.
\end{problem}

\subsection{Refuted Problems}
\clearpage
\section{P036: Adaptivity in one-dimensional biased search}
\label{problem:P036}

\begin{problem}
Every query x returns y(x)=f(x)+b(x).

For an algorithm A, W(A)=sup over f in \(F_{L}\) and b in \(B_{\beta}\) of f(\(\widehat{x}_{A})-\min_{x \in X}\)f(x), where \(\widehat{x}_{A}\) is A's transcript-dependent output.

\(\mathbb R_{n}^{ad}\)(L,\(\beta\))=inf over A in \(Alg_{ad}(n)\) of W(A).

\(\mathbb R_{n}^{ob}\)(L,\(\beta\))=inf over A in \(Alg_{ob}(n)\) of W(A).

For every η\(\geq0, N_{1}^{*}\)(L,\(\beta\),η) is the least n with \(\mathbb R_{n}^{ad}\)(L,\(\beta)\leq\)η, with \(\infty\) if no such n exists.

The dimension d is a positive integer, and the search domain is {[}0,1{]}\(^{d}\) equipped with the \(\infty\) norm.

The objective f:{[}0,1{]}\(^{d}\) to the real numbers is L-Lipschitz for a known L\(>\)0.

The oracle returns y(x)=f(x)+b(x), where b:{[}0,1{]}\(^{d}\) to the real numbers is one fixed field throughout the run and satisfies \(\mid\) b(x)\(\mid \leq \beta\) at every point; no regularity of b, gradients, or stochastic samples are available.

The algorithm is deterministic, may choose each query adaptively from previous returned values, knows d, L, \(\beta\), and \(\varepsilon\), and may output any point x̂ in the cube whether or not it was queried.

For \(\varepsilon \geq0, N_{d}^{*}\)(L,\(\beta,\varepsilon\)) is the least worst-case evaluation cap guaranteeing f(x̂)-\(\min_{x}\) f(x)\(\leq \varepsilon\) for every admissible pair (f,b), and equals \(\infty\) when no finite cap has this property.

The principal positive-bias regime is 0\(< \beta <\) L/4; for the superthreshold asymptotic, \(\varepsilon=2 \beta+\delta\) with \(\delta >\)0 while d, L, and \(\beta\) remain fixed and \(\delta\) decreases to zero.

One-dimensional deterministic global minimization on {[}0,1{]} with one fixed arbitrary field satisfying \(\mid\) b(x)\(\mid \leq \beta\). Both algorithm classes may output any queried or unqueried point as a function of their observations. This is a finite-horizon structural comparison, not a restatement of the previously attempted leading-constant limit.

Fix d=1 and X={[}0,1{]} with its usual distance.

An oblivious algorithm fixes all query sites before seeing oracle values, although its output may depend on the complete observation vector.

For each integer n\(\geq\)1, both classes use at most n evaluations and are compared by worst-case simple regret.

Each algorithm is chosen before the adversarial pair (f,b) and must work uniformly over every f in \(F_{L}\) and b in \(B_{\beta}\).

In dimension d=1, determine whether deterministic adaptivity is completely powerless at every finite query horizon: for every L>0, every 0<\(\beta\)<L/4, and every integer n\(\geq\)1, is the minimax worst-case simple regret among adaptive n-query algorithms equal to that among algorithms whose n query sites are fixed before any oracle value is observed?
\end{problem}

\clearpage
\section{P037: Static topology cost in directed gradient tracking}
\label{problem:P037}

\begin{problem}
For I in \(D_{\mathrm{STATIC}}(η)\), z\(\star=(\sum_{i=1}^{3} h_{i} r_{i})/(\sum_{i=1}^{3} h_{i}\)).

For an anchor-defined trajectory, \(E_{k}^{\alpha}\)=(norm\((x_{k}-1 z\star)_{2}^{2}+\alpha^{2}\) norm\((y_{k})_{2}^{2})^{(1/2)}\).

\(\mathbb R_{k}\)(I,\(\alpha\)) is the supremum of \(E_{k}^{\alpha/E_{0}} \alpha\) over all \(x_{0}\) satisfying the tracker-consistent initialization \(y_{0}=\nabla\) F(\(x_{0}\)) and \(E_{0}^{\alpha} >\)0.

\[
\begin{aligned}
S_{\mathrm{STATIC}}(I)=\bigl\{\alpha>0:{}&
\sup_{k\geq0}\mathbb R_k(I,\alpha)<+\infty,\\
&\lim_{k\to+\infty}\mathbb R_k(I,\alpha)=0\bigr\}.
\end{aligned}
\]

\(T_{\varepsilon}\)(I,\(\alpha\))=inf\{K in \{0,1,2,\ldots\}: \(\sup_{k\geq K} \mathbb R_{k}\)(I,\(\alpha)\leq \varepsilon\)\}, with value +\(\infty\) when this set is empty.

\[
N_{\mathrm{STATIC}}(\varepsilon,\eta)
=\sup_{I\in D_{\mathrm{STATIC}}(\eta)}
\inf_{\alpha\in S_{\mathrm{STATIC}}(I)}T_\varepsilon(I,\alpha),
\]
with the inner infimum equal to \(+\infty\) when \(S_{\mathrm{STATIC}}(I)\) is empty.

There are n\(\geq\)3 agents with scalar objectives \(f_{i}(z)=h_{i} (z-r_{i})^{2/2}\), where 1\(\leq h_{i} \leq\)2. Their average has the unique minimizer \(z\star\), and \(\nabla\) F(x)=(\(h_{i}(x_{i}-r_{i}))_{i=1} ^{n}\).

For a constant \(\alpha >\)0, the algorithm is \(x_{k+1}=A_{k} x_{k}-\alpha y_{k}\) and \(y_{k+1}=B_{k} y_{k}+\nabla\) F(\(x_{k+1})-\nabla\) F(\(x_{k}\)), with tracker-consistent initialization \(y_{0}=\nabla\) F(\(x_{0}\)).

Each \(A_{k}\) is row-stochastic and each \(B_{k}\) is column-stochastic. Both have positive diagonal entries, and for i≠j, \((A_{k})_{ij} >\)0 if and only if \((B_{k})_{ij} >\)0 if and only if the directed edge j\(\to\)i is present at time k. Every positive entry of either matrix is at least η, where 0\(<\) η\(\leq\)1/2.

For a fixed integer H\(\geq\)1, the union of the directed graphs in every H consecutive times is strongly connected.

An admissible instance I consists of the objective data and matrix schedule satisfying A1-A4, but not an initialization. For each I and \(\alpha, E_{k}^{\alpha}, \mathbb R_{k}, T_{\varepsilon}\), S(I), and \(N_{\mathrm{GT}}\) are defined as in the claim shape, with \(\mathbb R_{k}\) taking the supremum over all tracker-consistent \(x_{0}\) having \(E_{0}^{\alpha} >\)0 and \(N_{\mathrm{GT}}\) taking the supremum over all admissible instances. An instance with S(I) empty makes \(N_{\mathrm{GT}}\) infinite. No finiteness or heterogeneity-control property of the chosen normalization is assumed.

All scalar-quadratic data allowed by A1 and all constant row-stochastic and column-stochastic matrix pairs satisfying A2-A5 with n=3, H=1, common strongly connected positive off-diagonal support, positive diagonals, and every positive entry at least η. Stepsizes are instance-aware and optimized over the explicitly defined tracker-reachable stability set \(S_{\mathrm{STATIC}}(I)\).

Fix n=3 and H=1 and require \(A_{k}\)=A and \(B_{k}\)=B for every nonnegative integer k.

Take the supremum over every admissible static common-support pair and every scalar-quadratic objective allowed by A1.

Optimize over every \(\alpha\) in \(S_{\mathrm{STATIC}}(I)\), where \(S_{\mathrm{STATIC}}(I)\) is defined solely through the tracker-consistent envelopes \(\mathbb R_{k}\)(I,\(\alpha\)).

Restrict A1-A5 to n=3, H=1, and constant matrices \(A_{k}\)=A and \(B_{k}\)=B. For each static admissible instance I, explicitly define \(S_{\mathrm{STATIC}}(I)\) as the set of \(\alpha\)>0 for which the tracker-consistent envelope is uniformly bounded and converges to zero. Define \(N_{\mathrm{STATIC}}(\varepsilon\),η) using the infimum over \(S_{\mathrm{STATIC}}(I)\). Determine whether \(N_{\mathrm{STATIC}}(\varepsilon\),η) is uniformly comparable to \(η^{(-2)}\) log(1/\(\varepsilon\)) for all sufficiently small η and \(\varepsilon\).
\end{problem}

\clearpage
\section{P038: Finite length of anchored variable-metric proximal orbits}
\label{problem:P038}

\begin{problem}
Define \(d_{k}=x_{k+1} - x_{k}\).

Define \(L_N=\sum_{k=0}^N\lVert d_k\rVert\).

The orbit has finite length when \(\sup_N L_N<\infty\), equivalently \(\sum_{k\geq0}\lVert d_k\rVert<\infty\).

H is a real Hilbert space; T:H⇉H is maximal monotone; and S=\{x\(\in\)H:0\(\in\)T(x)\} is nonempty.

Each \(M_{k}\) is bounded, self-adjoint, and strongly positive, with \(\alpha\)I\(\preccurlyeq M_{k}\preccurlyeq\beta\)I for fixed 0\(< \alpha \leq \beta < \infty\), and \((1+η_{k})^{-1} M_{k}\preccurlyeq M_{k+1}\preccurlyeq(1+η_{k})M_{k}\) for \(η_{k} \geq\)0 with \(\Sigma_{k} η_{k} < \infty\). These conditions make (\(M_{k}\)) operator-norm Cauchy; denote its strongly positive limit by \(M_{\infty}\).

The parameters satisfy 0\(< c_{min} \leq c_{k} \leq c_{max} < \infty\) and 0\(\leq \sigma_{k} \leq \bar{\sigma} <\)1.

Define
\[
\begin{aligned}
T^\varepsilon(y)=\{v:\;&
\langle y-z,v-w\rangle\geq-\varepsilon\\
&\text{for every }(z,w)\in\operatorname{gra}T\}.
\end{aligned}
\]
Given \(x_k\), choose \((y_k,v_k,\varepsilon_k)\) with
\(\varepsilon_k\geq0, v_k\in T^{\varepsilon_k}(y_k)\), and
\[
\delta_k=c_kM_kv_k+y_k-x_k,
\]
such that
\[
\begin{aligned}
\|\delta_k\|_{M_k^{-1}}^2+2c_k\varepsilon_k
\leq\sigma_k^2\bigl(&\|c_kM_kv_k\|_{M_k^{-1}}^2\\
&+\|y_k-x_k\|_{M_k^{-1}}^2\bigr).
\end{aligned}
\]

Set \(H_{k}\)=\{z\(\in\)H:⟨z\(- y_{k},v_{k}\)⟩\(\leq \varepsilon_{k}\)\}. Let \(W_{0}\)=H; for k\(\geq\)1 let \(W_{k}\)=\{z\(\in\)H:⟨z\(- x_{k},M_{k-1}^{-1}(x_{0} - x_{k}\))⟩\(\leq\)0\}. Define \(x_{k+1}=\operatorname*{arg\,min}_{z\in H_{k}\cap W_{k}}\) ½‖z\(- x_{0}\)‖\_\{\(M_{k}^{-1}\)\}\(^2\).

All real Hilbert spaces and exactly the recurrence, metric drift, parameter bounds, and enlargement-relative-error model in A1--A5. The claim quantifies over every admissible triple sequence and adds no compactness, regularity, finite-dimensionality, or error-summability assumption.

Quantify over every sequence of triples satisfying A4 and the iterates generated by A5.

Measure length by \(\sum_{k\geq0}\lVert x_{k+1}-x_k\rVert\) in the ambient Hilbert norm.

The target is finite total variation, not the already resolved norm convergence or square-summability of increments.

Under A1–A5, determine whether every admissible anchored orbit has finite total variation, namely whether \(\sum_{k\geq0}\lVert x_{k+1}-x_k\rVert\) converges.
\end{problem}

\clearpage
\section{P039: Last residuals under square-summable metric drift}
\label{problem:P039}

\begin{problem}
\(T^\varepsilon(y)=\{v:\langle v-w,y-z\rangle\geq-\varepsilon\ \text{for every }(z,w)\in\operatorname{gra}T\}\).

\(\lVert u\rVert_{M_k^{-1}}^2=\langle u,M_k^{-1}u\rangle\).

\(\delta_k=c_kM_kv_k+y_k-x_k\).

\(e_{k}=y_{k} - x_{k}\).

The space is \(\mathbb R^{n}\) for some finite n, T:\(\mathbb R^{n}\)⇒\(\mathbb R^{n}\) is maximally monotone, zer(T) is nonempty, and \(x_{0}\) is arbitrary.

Each \(M_{k}\) is symmetric positive definite, fixed constants 0\(<\) m\(\leq\)L\(< \infty\) satisfy mI\(\preccurlyeq M_{k}\preccurlyeq\)LI, and ∑\(_{k\geq0}\) ‖\(M_{k+1} - M_{k}\)‖\(_{op} ^2 < \infty\); neither ∑\(_{k\geq0}\) ‖\(M_{k+1} - M_{k}\)‖\(_{op} < \infty\) nor convergence or commutativity of the metrics is assumed.

The proximal parameters satisfy 0\(< c_{min} \leq c_{k} \leq c_{max} < \infty\).

For one fixed \(\sigma\in[0,1)\), each \(\varepsilon_k\geq0\) and \(v_k\in T^{\varepsilon_k}(y_k)\). With \(\delta_k=c_kM_kv_k+y_k-x_k\), require
\[
\lVert\delta_k\rVert_{M_k^{-1}}^2+2c_k\varepsilon_k
\leq\sigma^2\lVert y_k-x_k\rVert_{M_k^{-1}}^2.
\]

The accepted update is \(x_{k+1}=x_{k} - c_{k} M_{k} v_{k}\). No boundedness, residual convergence, error summability, strong monotonicity, metric subregularity, simultaneous diagonalization, or eventual metric stationarity is assumed.

Finite-dimensional maximal monotone inclusions with one fixed operator, bounded positive proximal parameters, the graph/enlargement-relative error condition of A4, and the direct update of A5. The conclusion concerns every computed last-evaluation residual, not an ergodic or best-iterate quantity.

The same maximally monotone operator T is used at every iteration.

Only the square-summability required in A2 is available; metric convergence, commutativity, and summable total variation cannot be invoked.

The target is the full-sequence limit of ∥\(v_{k}\)∥+\(\varepsilon_{k}\) for the actual evaluations appearing in A4.

Determine whether every trajectory satisfying A1–A5 necessarily obeys \(\lim_{k\to\infty}\)(∥\(v_{k}\)∥+\(\varepsilon_{k}\))=0 when the uniformly conditioned metrics have square-summable operator-norm increments but may have infinite total variation.
\end{problem}

\clearpage
\section{P040: A class-wide hyperbolic growth exponent}
\label{problem:P040}

\begin{problem}
\[
D_{\phi}(x,y)=\phi(x)-\phi(y)-< \nabla \phi(y),x-y>
\]

\[
P_{\phi}(y)=\operatorname*{arg\,\min}_{x \in N}\{‖x‖_{2}+\lambda^{-1} D_{\phi}(x,y)\},
\]
\[
e_{\phi}(y)=‖P_{\phi}(y)‖_{2}+\lambda^{-1} D_{\phi}(P_{\phi}(y),y), G_{\phi}(y)=\nabla e_{\phi}(y),
\]
and \(V_{\phi}(y)=\lambda^{-1}(\nabla \phi\)(y)-\(\nabla \phi(P_{\phi}\)(y))).

\[
T_{\phi}(s,z_{2})=(s,\partial_{2} \phi(s,z_{2})-\partial_{2} \phi(0)),
\]
\(\Omega_{\phi}=T_{\phi}(W)\), and \(zeta_{\phi}^{0}=T_{\phi}(0)\). Strict positivity of
\(\partial_{22} \phi\) makes \(T_{\phi}\) a diffeomorphism from W onto \(\Omega_{\phi}\).

For every simply connected proper plane domain \(\Omega, d_{\Omega}\)
denotes the curvature-minus-four Poincare distance, normalized on
the unit disk by \(d_{D}(0,r)\)=artanh(r).

For \(x=P_{\phi}(y)\) nonzero, \(Delta_{\phi}(y)=2 \max_{t \in [0,1]} d_{\Omega,\phi}(zeta_{\phi}^{0},T_{\phi}\)(x+t(y-x))).

Q(\(\mathbb R\))=max\{1,sup\{‖\(V_{\phi}(y)\)‖/‖\(G_{\phi}(y)\)‖:(\(\phi\),y) in \(E_{R}\)\}\}, interpreted in
{[}1,+\(\infty\){]} and with the supremum of an empty set omitted by the outer
maximum.

\[
L_{\mathrm{MA}}=limsup_{\mathbb R \to \infty} \log(Q(\mathbb R))/\mathbb R in [0,+\infty]
\]

The function f is proper and lower semicontinuous and is relatively prox-regular at a
stationary pair (\(\bar{x}\),0) with respect to a fixed reference Legendre kernel on a
neighborhood of \(\bar{x}\).

There are convex neighborhoods V contained in the interior of N, with compact closure of N
contained in a common open domain U, and a fixed \(\lambda >\)0 used for every kernel.

Each \(\phi_{k}\) is a C2 very strictly convex Legendre function on U; the kernels share their
domain but may repeat, reverse, recover, or have unbounded ambient condition numbers as k
varies.

For every k and y in V, the minimization of f(x)+\(\lambda^{-1} D_{\phi,k}(x,y)\) over x in N has a
unique minimizer \(P_{k}(y)\) in the interior of N, and the resulting localized envelope \(e_{k}\) is C1
on V.

All gradients, Hessians, and residual norms are measured in one fixed finite-dimensional
coordinate system; affine changes to \(\phi_{k}\) are identified because they do not change its
Bregman divergence.

A class-wide, two-dimensional local Monge-Ampere subclass of the anchored variable-kernel problem.
The objective, proximal parameter, localization neighborhoods, normalization point, determinant-one
neighborhood, and hyperbolic normalization are fixed. The extremum ranges directly over individual
current kernels and actual localized proximal chords, not over an ambiguously quantified kernel
family. Kernel order and accumulated drift do not enter Q. This functional hyperbolic exponent
problem is mathematically distinct from the established three-dimensional quadratic
longitudinal-cone cell with constant lower ratio 1/4. Literature overlap remains unresolved because
the supplied source pack is empty and no independent retrieval route is available to this role.

Set n=2, \(\bar{x}=0, \lambda\)=1/64, f(z)=‖z‖\(_{2}, U=\mathbb R^{2}\), V=B(0,1/16),
N=B(0,1/4), and W=B(0,1/2).

Each \(\phi\) in \(K_{\mathrm{MA}}\) is a global C2 very strictly convex Legendre kernel, is C3 on
W, satisfies \(nabla^{2} \phi(0)\)=I, obeys det(\(nabla^{2} \phi(z)\))=1 for all z in W, and
satisfies the per-kernel A4 localization and differentiability conditions on
the fixed V and N.

The extremal set consists directly of pairs (\(\phi\),y), with \(\phi\) in
\(K_{\mathrm{MA}}\) and \(x=P_{\phi}(y)\) nonzero; no kernel index or fixed-family
quantifier occurs in Q.

Every pair entering Q(\(\mathbb R\)) satisfies max\{‖\(P_{\phi}(y)\)‖,‖y‖\}\(\leq1/\mathbb R\).

The underlying Poincare metric has curvature minus four and
unit-disk normalization \(d_{D}(0,r)\)=artanh(r), while \(Delta_{\phi}(y)\) is
twice the maximum such distance along the transformed current
chord.

Specialize A1-A5 to dimension two with \(\bar{x}=0, \lambda\)=1/64, f(z)=‖z‖\(_{2}, U=\mathbb R^{2}\), V=B(0,1/16),
N=B(0,1/4), and W=B(0,1/2). Let \(K_{\mathrm{MA}}\) be the class of individual global C2 very strictly convex
Legendre kernels \(\phi\) that are C3 on W, satisfy \(nabla^{2} \phi(0)\)=I and det(\(nabla^{2} \phi\))=1 on W, and
satisfy the per-kernel A4 requirements for every y in V. For \(T_{\phi}\)(s,\(z_{2}\))=(s,\(\partial_{2} \phi\)(s,\(z_{2})-\partial_{2} \phi\)(0)), set \(\Omega_{\phi}=T_{\phi}(W)\). Normalize the Poincare distance \(d_{\Omega,\phi}\) to
curvature minus four, so \(d_{D}(0,r)\)=artanh(r) on the unit disk, and define the exponent-normalized
reach \(Delta_{\phi}(y)=2 \max_{t \in [0,1]} d_{\Omega,\phi}(T_{\phi}(0),T_{\phi}(P_{\phi}\)(y)+t(y-\(P_{\phi}(y)\)))). For
\(\mathbb R \geq\)64, define Q(\(\mathbb R\)) in [1,+\(\infty\)] as the class-wide supremum of ‖\(V_{\phi}(y)\)‖/‖\(G_{\phi}(y)\)‖ over
pairs (\(\phi\),y) with \(\phi\) in \(K_{\mathrm{MA}}, P_{\phi}(y)\) nonzero, max{‖\(P_{\phi}(y)\)‖,‖y‖}\(\leq1/\mathbb R\), and \(Delta_{\phi}(y)\leq \mathbb R\).
Determine whether the class-wide growth exponent reaches the universal barrier two.
\end{problem}

\clearpage
\section{P041: Full proximal decay at the critical perturbation order}
\label{problem:P041}

\begin{problem}
S(x)=T(x)+b(x).

p=1/q\(>\)1.

\[
K=\kappa^{(-1/q)},\qquad c_{gap}=K-L>0
\]

\(x_{k+1}=(I+\lambda S)^{(-1)}(x_{k}\)).

\[
d_{k}=‖x_{k}-\bar{x}‖
\]

\(a_{rate}\)=q/(1-q)=1/(p-1).

The ambient space is a finite-dimensional real Hilbert space, 0\(<\) q\(<\)1, T is maximal monotone, and \(\bar{x}\) is an isolated member of \(T^{(-1)}\)(0).

T is metrically q-subregular at (\(\bar{x}\),0) with exact finite positive modulus \(\kappa\) for the local inequality ‖x-\(\bar{x}\)‖\(\leq \kappa\) dist(0,T(x))\(^{q}\).

The perturbation b is single-valued on the ambient space and continuous near \(\bar{x}\), satisfies b(\(\bar{x}\))=0, and has critical-order coefficient \(L=limsup_{x\to\bar{x}, x≠\bar{x}}\) ‖b(x)‖/‖x-\(\bar{x}\)‖\(^{(1/q)}\) with 0\(\leq\)L\(\leq \kappa^{(-1/q)}\).

The actual pointwise sum x↦T(x)+b(x) is maximal monotone and contains 0 at \(\bar{x}\); no subregularity, isolation, or convergence property of T+b is assumed.

For the algorithmic classification, \(x_{k+1}\)=(I+\(\lambda(T+b))^{(-1)}(x_{k}\)) uses a fixed \(\lambda >\)0 and a sufficiently local initial point, and the rate observable is ‖\(x_{k}-\bar{x}\)‖; locality, convergence, and a rate remain conclusions to be classified rather than premises.

All finite-dimensional A1--A5 instances with 0\(\leq\)L\(< \kappa^{(-1/q)}\), without monotonicity of b, differentiability, cyclic monotonicity, homogeneity, or a sector assumption. The claim is orbitwise and local for each fixed \(\lambda >\)0.

The coefficient in A3 satisfies 0\(\leq\)L\(< \kappa^{(-1/q)}\).

Do not assume b is monotone; only the actual sum S=T+b is maximal monotone as required by A4.

For each instance and \(\lambda\), the localization radius and finite asymptotic constant may depend on the instance, \(\lambda\), and initial point.

Let S=T+b satisfy A1–A5 with strict critical-order gap 0\(\leq\)L<\(\kappa^{(-1/q)}\). Determine whether every sufficiently local fixed-parameter proximal orbit has distance decay \(d_{k}\)=O(\(k^{-q/(1-q)}\)), rather than only the slower exponent obtained from the norm-only Fejér estimate.
\end{problem}

\clearpage
\section{P042: Affine risk budgets in a two-step Gaussian controller}
\label{problem:P042}

\begin{problem}
For bounded Borel f on K, define \(Q_{u}\) f(x)=E{[}f\((X_{1})1_{X_{1} \in K}\) given \(X_{0}\)=x and \(u_{0}\)=u{]}.

Define \(V_{1}(x)=\max_{u \in U} Q_{u}\) 1(x).

For r=1-s and g=1-h, the propagated allocation inequality is equivalent to \(Q_{u}\) h(x)\(\geq\)s. One-step tail feasibility is equivalent to h(y)\(\leq V_{1}(y)\) for every y in K.

Define B(x)=max over u in U and h in H of \(Q_{u}\) h(x). A state-slack pair (x,s) is feasible exactly when s\(\leq\)B(x).

If a feasible selection chooses (\(u_{t},h_{t}\)), then on a safe transition \(s_{t+1}=h_{t}(X_{t+1}\)); the next pre-action feasibility test fails exactly when \(s_{t+1} >\) B(\(X_{t+1}\)).

Let \(\lambda_{affine,plus}\) be the supremum over Borel minimizing selections of limsup as T tends to \(\infty\) of P\((\tau> T and zeta> T)^{(1/T)}\), starting from (\(x_{0},s_{0})=(0,10^{(-4)}\)).

Let \(\lambda_{\infty}\) be the horizonwise unrestricted survival root rate from A6 for the same state dynamics and initial state.

Fix \(x_{0}\) in K={[}-1,1{]}. The observed state satisfies \(X_{t+1}=aX_{t}+u_{t}+\sigma W_{t+1}\), where the \(W_{t}\) are independent standard normal variables, \(\sigma >\)0, and U is a fixed nonempty finite subset of a bounded interval.

The exit time is \(\tau\)=inf\{t\(\geq1:X_{t}\) is not in K\}. The absorbing violation variable at time s is \(1_{\tau\leq s}\), and the only risk operator is conditional expectation given the observed history, so an N-step risk constraint at time t is exactly P(\(\tau \leq\)t+N given the current history)\(\leq r_{t}\).

Fix N\(\geq\)2, a bounded continuous running cost, and a bounded continuous terminal cost. At each integer time t, feasibility is tested only on \{\(\tau >\) t\} and before any time-t action is selected. Define zeta=inf\{t\(\geq0:\tau >\) t and no feasible N-step plan exists from (\(X_{t},r_{t}\))\}, with inf of the empty set equal to \(\infty\); if zeta=t the controller stops before acting, and no feasibility test is made after exit. At a feasible pair the controller minimizes the N-step expected cost, applies the first action, and reoptimizes. The reset protocol has \(r_{0}\)=r for a fixed r in (0,1) and resets \(r_{t}\)=r after every safe transition; the propagated protocol starts from a fixed \(r_{0}\) in (0,1). Recursive feasibility before exit means zeta\(\geq \tau\) almost surely, equivalently P(\(\tau >\) T and zeta\(\leq\)T)=0 for every integer T\(\geq\)0.

Fix knots -1=\(k_{0} < k_{1} <\)\ldots\(< k_{m}\)=1. A propagated successor budget is the continuous piecewise-affine interpolant \(g_{c}\) with coefficient vector c in {[}0,1{]}\(^{m+1}\); this compact Euclidean cube supplies its topology and Borel structure. At state-budget pair (x,r), a first action u and coefficient vector c are admissible only if P(\(X_{t+1}\) not in K given \(X_{t}\)=x,u)+\(\in t_{K} g_{c}(y)\)P(dy given x,u)\(\leq\)r and an explicitly length-(N-1) remaining tail is feasible from (y,\(g_{c}(y)\)) for almost every safe successor y. On a safe transition, \(r_{t+1}=g_{c}(X_{t+1}\)); the next time then tests a new length-N problem before selecting its action.

For protocol \(\mathbb R\) and its class \(S_{R}\) of Borel selections from the finite-horizon minimizing correspondence wherever that correspondence is nonempty, define, when \(S_{R}\) is nonempty, \(\lambda_{R}^{+}(x_{0},r_{0})=\sup_{s \in S_{R}} limsup_{T \to \infty} P_{s} (\tau> T and zeta> T)^{1/T}\) and \(\lambda_{R}^{-}(x_{0},r_{0})=\in f_{s \in S_{R}} liminf_{T \to \infty} P_{s} (\tau> T and zeta> T)^{1/T}\). Selector nonexistence is recorded separately and is not assigned numerical rate zero. Establishing nonempty compact feasible sections and a Borel graph from A1-A4 is part of the target, not an assumed selector certificate.

The infinite-horizon comparator ranges over all nonanticipative Borel U-valued policies, has no budget or infeasibility stopping rule, and is \(\lambda_{\infty}(x_{0})=limsup_{T \to \infty}(\sup_{pi} P_{pi}(\tau >\) T))\(^{1/T}\). The supremum is horizonwise and may use different policies for different T. An optional terminal continuation set C is a specified closed subset of K times {[}0,1{]}; any claimed continuation property must state that an admissible action and coefficient return almost every safe successor pair to C. Nonemptiness of C and its consequences are conclusions, not premises.

Fix K={[}-1,1{]}, \(x_{0}\)=0, a=1, \(\sigma\)=1/4, U=\{-1/2,1/2\}, N=2, zero running and terminal costs, no terminal continuation set, and propagated-budget knots \{-1,1\}. Write slack as s=1-r and fix \(s_{0}=10^{(-4)}\), so \(r_{0}\)=0.9999. Successor budgets are affine because the grid has only its two endpoints.

The dynamics are \(X_{t+1}=X_{t}+u_{t}+(1/4)W_{t+1}\) on K={[}-1,1{]}, with U=\{-1/2,1/2\} and \(x_{0}\)=0.

Use N=2 and the propagated protocol with knots \(k_{0}\)=-1 and \(k_{1}\)=1, so every successor budget and successor slack is globally affine.

Fix \(s_{0}=1-r_{0}=10^{(-4)}\).

Running and terminal costs vanish, so every feasible two-step plan is minimizing and the selector envelope ranges over all Borel feasible allocation choices.

For the fixed controlled Gaussian system below, prove or refute that two-step propagated budgets restricted to globally affine successor functions have a strictly smaller best joint survival-feasibility root rate than the unrestricted controller, even when the initial risk budget is 0.9999.
\end{problem}

\clearpage
\section{P043: A drift-free last-iterate modulus for variable-metric PDHG}
\label{problem:P043}

\begin{problem}
g\(\star\) denotes the Fenchel conjugate of g.

T(x,y)=(partial f(x)+K\emph{y,partial g}(y)-Kx), where K\(\star\) is the Hilbert adjoint of K.

Set \(H_{k}\)=((\(P_{k}\),-K\(\star\)),(-K,\(Q_{k}\))). Under A4, \(H_{k}\) is positive definite; write ‖u‖\_\((H_{k})^{2}\)=\(< H_{k}\) u,u\(>\).

Set \(d_{k}\)=(\(x_{k+1}\)-\(x_{k}\),\(y_{k+1}\)-\(y_{k}\)), \(e_{k+1}\)=(\(e^{x}\)\emph{(k+1),\(e^{y}\)}(k+1)), and \(w_{k+1}\)=-\(H_{k} d_{k}\)-\(e_{k+1}\).

For h\(\geq\)0 set \(A_{h}\)=(\(\sum_{j=h}^{\infty}\) ‖\(d_{j}\)‖\_\((H_{j})^{2}\))\(^{(1/2)}\).

For h\(\geq\)0 set \(B_{h}\)=\(\sup_{j\geq h}\) √(j+1)(‖\(e^{x}\)\emph{(j+1)‖+‖\(e^{y}\)}(j+1)‖).

For \(h_{k}\)=floor(k/2), set \(F_{k}\)=\(A_{h_{k}}\)+\(B_{h_{k}}\).

Set \(S_{k}\)=√(k+1)‖\(w_{k+1}\)‖/\(F_{k}\) when \(F_{k} >\)0, set \(S_{k}\)=0 when both numerator and denominator vanish, and set \(S_{k}\)=\(\infty\) when \(F_{k}\)=0 but the numerator is positive.

\(C_{nodrift}\)(m,M,\(\delta\),\(\Lambda\)) is the supremum of \(S_{k}\) over all admitted finite-dimensional A1-A6 data, all orbits with the fixed structural bounds and total drift at most \(\Lambda\), and all k\(\geq\)0.

X and Y are finite-dimensional real Hilbert spaces; f on X and g on Y are proper lower-semicontinuous convex functions, and K:X\(\to\)Y is linear.

The KKT operator T(x,y)=(\(\partial\)f(x)+K\emph{y, \(\partial\)g}(y)-Kx) has a nonempty zero set.

For every k, the iterates satisfy 0\(\in \partial\)f(\(x_{k+1}\))+K*\(y_{k}\)+\(P_{k}\)(\(x_{k+1}\)-\(x_{k}\))+\(e^{x} _{k+1}\) and 0\(\in \partial\)g*(\(y_{k+1}\))-K(2\(x_{k+1}\)-\(x_{k}\))+\(Q_{k}\)(\(y_{k+1}\)-\(y_{k}\))+\(e^{y} _{k+1}\), so \(e^{x}\) and \(e^{y}\) are the declared inexact-resolvent optimality residuals.

\(P_{k}\) and \(Q_{k}\) are self-adjoint positive-definite block metrics with mI\(\preccurlyeq\)\(P_{k}\),\(Q_{k}\)\(\preccurlyeq\)MI for fixed 0\(<\) m\(\leq\)M, and ‖\(Q_{k}^{(-1/2)}\) K \(P_{k}^{(-1/2)}\)‖\(\leq\)1-\(\delta\) for a fixed \(\delta >\)0.

There are nonnegative \(η_{k}\) with \(\sum_{k} η_{k} < \infty\) such that \((1+η_{k})^{(-1)} P_{k}\)\(\preccurlyeq\)\(P_{k+1}\)\(\preccurlyeq\)\((1+η_{k})P_{k}\) and the same two-sided relation holds for \(Q_{k}\).

The inexactness is norm-summable: \(\sum_{k}\) (‖\(e^{x}\)\emph{(k+1)‖+‖\(e^{y}\)}(k+1)‖)\(< \infty\).

All finite-dimensional A1-A6 variable-metric inexact PDHG orbits with fixed bounds 0\(<\) m\(\leq\)M, 0\(< \delta \leq\)1, and total relative metric drift at most \(\Lambda\). The observable is the graph-valid current KKT element. No averaging, best-iterate selection, restart, strong monotonicity, bounded variation of errors, or simultaneous diagonalization of the metrics is allowed.

For \(d_{k}\)=(\(x_{k+1}\)-\(x_{k}\),\(y_{k+1}\)-\(y_{k}\)), use \(w_{k+1}\)=(K\(\star\)(\(y_{k+1}\)-\(y_{k}\))-\(P_{k}\)(\(x_{k+1}\)-\(x_{k}\))-\(e^{x}\)\emph{(k+1),K(x}(k+1)-\(x_{k}\))-\(Q_{k}\)(\(y_{k+1}\)-\(y_{k}\))-\(e^{y} _{k+1}\)).

For \(h_{k}\)=floor(k/2), normalize only by \(A_{h_{k}}\)+\(B_{h_{k}}\), omitting every term involving the metric-drift tail \(D_{h_{k}}\).

Take the supremum over orbits with one fixed quadruple 0\(<\) m\(\leq\)M, 0\(< \delta \leq\)1, and \(\Lambda \geq\)0 satisfying \(\sum_{j} \eta_{j} \leq \Lambda\).

Allow every norm-summable optimality-error sequence authorized by A6 and every noncommuting metric path authorized by A4-A5.

Determine whether the established last-iterate tail modulus remains universally finite after deleting the term that multiplies the remaining metric drift by the total increment energy.
\end{problem}

\clearpage
\section{P044: Period-three tests under arbitrary delay switching}
\label{problem:P044}

\begin{problem}
x\(\star=(b_{1}+b_{2}+b_{3}\))/3.

For \(\delta=(\delta_{0},\delta_{1},\delta_{2}), \delta^{\infty}\) is the delay word satisfying \(d_{k}=\delta_{k mod 3}\).

At the start of round k, set \(u_{10}^{k}=(c_{k-1}^{x},c_{k-1}^{y}\)) if k\(\geq\)1 and \(d_{k-1}\)=0 and set it to zero otherwise; set \(u_{20}^{k}=(c_{k-2}^{x},c_{k-2}^{y}\)) if k\(\geq\)2 and \(d_{k-2}\)=1 and set it to zero otherwise; set \(u_{11}^{k}=(c_{k-1}^{x},c_{k-1}^{y}\)) if k\(\geq\)1 and \(d_{k-1}\)=1 and set it to zero otherwise. Define \(L_{d}^{k}=(x^{k}\)-x\(\star 1,y^{k},z^{k}\)-x\(\star 1,h^{k}\)-x\(\star 1,e_{x}^{k},e_{y}^{k},u_{10}^{k},u_{20}^{k},u_{11}^{k}\)) in this fixed order.

\(E_{d}(k)\) is the maximum absolute value of the 24 scalar coordinates of \(L_{d}^{k}\).

Frozen(q,\(\alpha\)) holds when for every \(\delta\) in \{0,1\}\(^{3}\) there exist \(C_{\delta} \geq\)1 and \(\rho_{\delta}\) in (0,1) such that \(E_{\delta^\infty}(k)\leq C_{\delta} \rho_{\delta}^{k} E_{\delta^\infty}(0)\) for every b,x0 and k.

Switched(q,\(\alpha\)) holds when there exist C\(\geq\)1 and \(\rho\) in (0,1) such that \(E_{d}(k)\leq\)C \(\rho^{k} E_{d}(0)\) for every b,x0,d and k.

Agents are indexed modulo three, with successor s(i)=i+1 and predecessor p(i)=i-1. The primal channel uses half self-weight and half of the stored predecessor reconstruction, while the tracking channel retains half of an active agent's post-delivery tracker mass and sends the other half to its successor.

The computing agent at round k is \(a_{k}\)=1+(k mod 3). Only \(a_{k}\) performs a primal and gradient-difference computation; packet deliveries do not count as activations.

At the end of round k, the active agent forms one paired packet \(P_{k}=(a_{k}\),s(\(a_{k}),c_{k}^{x},c_{k}^{y}\)) and chooses \(d_{k}\) in \{0,1\}; \(P_{k}\) is consumed at the start of round k+1+\(d_{k}\), before that round's computation. The delay word is an arbitrary element of \{0,1\} indexed by the nonnegative integers and need not be periodic. At the start of round k, \(Q^{k}\){[}0{]} and \(Q^{k}\){[}1{]} are multisets of packets scheduled for rounds k and k+1. Every packet in \(Q^{k}\){[}0{]} is delivered once and removed. After computation, the remaining queue is shifted and \(P_{k}\) is appended according to \(Q^{k+1}{[}0{]}=Q^{k}\){[}1{]} multiset-union \{\(P_{k}\) if \(d_{k}\)=0\} and \(Q^{k+1}\){[}1{]}=\{\(P_{k}\) if \(d_{k}\)=1\}. The paired components share \(d_{k}\), no packet is consumed in its formation round, and packets on each edge are delivered FIFO.

Let \(z_{r}^{k}\) be receiver r's stored reconstruction of its predecessor's primal state, \(h_{i}^{k}\) sender i's primal encoder reference, and \(e_{i}^{x,k}, e_{i}^{y,k}\) sender i's primal and tracking compression residuals. At the start of round k, for each receiver r set bar \(z_{r}^{k}=z_{r}^{k}\) plus the sum of \(c^{x}\) over packets in \(Q^{k}\){[}0{]} addressed to r, and bar \(y_{r}^{k}=y_{r}^{k}\) plus the corresponding sum of \(c^{y}\). After these packets are consumed, set \(z_{r}^{k+1}\)=bar \(z_{r}^{k}\) for every r, so every delivered primal increment persists in the stored reconstruction. For \(i=a_{k}\) set \(x_{i}^{k+1}=(x_{i}^{k}\)+bar \(z_{i}^{k})/2-\alpha\) bar \(y_{i}^{k}\) and \(y_{i}^{k+1}\)=bar \(y_{i}^{k}\)/2+{[}gradient \(f_{i}(x_{i}^{k+1}\))-gradient \(f_{i}(x_{i}^{k}\)){]}; for j not equal to i set \(x_{j}^{k+1}=x_{j}^{k}\) and \(y_{j}^{k+1}\)=bar \(y_{j}^{k}\). The active sender forms \(v_{i}^{x}=x_{i}^{k+1}-h_{i}^{k}+e_{i}^{x,k}, c_{k}^{x}=qv_{i}^{x}, e_{i}^{x,k+1}=v_{i}^{x}-c_{k}^{x}\), and \(h_{i}^{k+1}=h_{i}^{k}+c_{k}^{x}\), together with \(v_{i}^{y}\)=bar \(y_{i}^{k}/2+e_{i}^{y,k}, c_{k}^{y}=qv_{i}^{y}\), and \(e_{i}^{y,k+1}=v_{i}^{y}-c_{k}^{y}\). All inactive encoder references and residuals remain unchanged, and \(P_{k}\) is inserted into the shifted queue exactly as prescribed in A3.

Each objective is \(f_{i}(x)=(x-b_{i})^{2/2}\) on the real line, where the offsets \(b_{i}\) are arbitrary, and the target optimizer is x\(\star=(b_{1}+b_{2}+b_{3}\))/3.

The stepsize \(\alpha\) is constant and positive and q belongs to (0,1{]}. Initial primal states are arbitrary; \(y_{i}^{0}=x_{i}^{0}-b_{i}, z_{i}^{0}=x_{p(i)}^{0}, h_{i}^{0}=x_{i}^{0}\), both residuals are zero, and \(Q^{0}\){[}0{]} and \(Q^{0}\){[}1{]} are empty.

This generation studies a conditional frozen-to-switched stability principle, not the previous unconditional parameter rectangle. It asks whether the special reachable structure of the specified 24-coordinate recurrence prevents the joint-product instability that individual stable monodromies do not exclude in general.

The parameters satisfy 1/2 \(\leq\) q \(\leq\) 1 and 0 \(<{} \alpha <\){} 1/2.

The premise requires geometric exact convergence for all eight delay words obtained by periodically repeating a block \(\delta\) in \{0,1\}\(^{3}\).

The conclusion is uniform over every infinite delay word d in \{0,1\}\(^{N_{0}}\), with constants independent of d, the offsets, and the primal initialization.

Both frozen stability and switched failure are evaluated on trajectories generated from A6 initialization, and every failure must affect a declared lifted coordinate.

For the recurrence fixed by A1-A6 and every q in [1/2,1] and \(\alpha\) in (0,1/2), determine whether geometric exact convergence for each of the eight periodically repeated three-round delay blocks implies delay-word-uniform geometric convergence over the entire binary delay language.
\end{problem}

\clearpage
\section{P045: The positive lobe of a smoothed LCP map}
\label{problem:P045}

\begin{problem}
The feasible cells occur cyclically as empty, \{1\}, \{1,3\}, \{1,2,3\}, \{2,3\}, \{2\}. Writing \(r_{\eta}\)=2/(1-η) and \(d_{\eta}\)=4eta/((1-η)(1+2eta)), their gradients are respectively (0,0), (2,0), (0,-\(r_{\eta}), (-d_{\eta},-d_{\eta}), (-r_{\eta}\),0), and (0,2).

\(P_{\eta}=G_{\eta}(\mathbb R^{2}\)) intersected with \((0,\infty)^{2}\).

For 0\(<\) t\(<2, b_{\eta}(t)\)=sup(\{y\(>\)0: there exists z in \(\mathbb R^{2}\) with \(G_{\eta}(z)\)=(t,y)\}), with supremum zero when the set is empty.

\[
Fib_{\eta}(q)=\{z in \mathbb R^{2}:G_{\eta}(z)=q\}
\]

\(C_{\eta}^{+}\)=\{z in \(\mathbb R^{2}:G_{\eta}(z)\) is in \((0,\infty)^{2}\) and det(\(DG_{\eta}(z)\))=0\}.

A point z is a nondegenerate fold when \(DG_{\eta}(z)\) has rank one and the directional derivative of det(\(DG_{\eta}\)) along any nonzero vector in ker(\(DG_{\eta}(z)\)) is nonzero.

\(\Gamma_{\eta}\) is the closed polygonal loop through the cyclic gradients (0,0), (2,0), (0,-\(r_{\eta}), (-d_{\eta},-d_{\eta}), (-r_{\eta}\),0), (0,2), and back to (0,0).

\(\rho_{\eta}\)=(1-2eta)/(1+2eta) is the positive-p inner fan-ray slope and tends to zero as η increases to 1/2.

The leader variable is x in \(\mathbb R^{2}\). Let B be the 3 by 2 matrix with rows (1,0), (0,1), and (1,1). For a fixed η in (0,1/2), let \(M_{\eta}\) be the symmetric 3 by 3 matrix with diagonal entries 1 and off-diagonal entries η. The lower equilibrium is the unique \(u_{\eta}(x)\) satisfying u \(\geq 0, M_{\eta}\) u-Bx \(\geq\) 0, and \(u_{i} (M_{\eta} u-Bx)_{i}\)=0 for i=1,2,3.

The upper objective is \(F_{q}(x,u)\)=c dot u-q dot x with c=(2,2,-2) and q in \(\mathbb R^{2}\). Because the lower solution is unique, the exact optimistic marginal is \(\phi_{\eta}\),q(x)=c dot \(u_{\eta}(x)\)-q dot x.

Only feasible full-dimensional active-set cells of this LCP are used. On the relative interior of such a cell I, \(u_{I}=M_{\eta}\){[}I,I{]}\(^{-1} B_{I}\) x, inactive coordinates vanish, the defining complementarity inequalities are strict, and the unshifted cell gradient is \(g_{I}=B_{I}^{T} M_{\eta}\){[}I,I{]}\(^{-1} c_{I}\).

The solution map \(u_{\eta}\) and every \(\phi_{\eta}\),q are globally continuous, positively homogeneous, and piecewise linear with linear growth. Gaussian convolution is taken with a standard two-dimensional Gaussian, so \(G_{\eta}(z)\)=E{[}\(\nabla\)(c dot \(u_{\eta}\))(z+Z){]} is finite and fan faces are null sets.

The transfer question uses \(\varepsilon\) decreasing to zero and \(x=\varepsilon\) z with z in a fixed compact set. Membership q in \(G_{\eta}(\mathbb R^{2}\)), an omitted point q in relint(\(D_{\eta}\)), strict directional descent of \(\phi_{\eta}\),q at zero, and an exact zero of the finite-\(\varepsilon\) smoothed gradient are distinct predicates.

The single coupled LCP in A1-A5, every η in (0,1/2), finite shifts z in \(\mathbb R^{2}\), and outputs with both coordinates strictly positive. The target classifies the positive image, its complete finite fibers, and its critical boundary. It excludes the full-plane critical locus, product surrogates, metric projections, and the established diagonal-fold and central-fiber results.

Only q in \((0,\infty)^{2}\) and finite solutions of \(G_{\eta}(z)\)=q are classified.

All cells, gradients, recession limits, and Jacobian jumps must arise from the same \(M_{\eta}\).

The points (2,0) and (0,2) are limiting endpoints of the boundary graph and are not asserted to be attained at finite shifts.

Fiber multiplicity counts distinct finite preimages, and every claimed regular preimage must have nonzero Jacobian determinant.

For every η in (0,1/2), define \(P_{\eta}=G_{\eta}(\mathbb R^{2}\)) intersected with \((0,\infty)^{2}, b_{\eta}\)(t)=sup({y>0: there exists z in \(\mathbb R^{2}\) with \(G_{\eta}(z)\)=(t,y)}), and \(Fib_{\eta}(q)\)={z in \(\mathbb R^{2}:G_{\eta}(z)\)=q}. Prove or refute the following conjunction: \(b_{\eta}\) is positive, continuous, strictly decreasing, involutive, and has endpoint limits 2 and 0; \(P_{\eta}\) is exactly the subgraph 0<\(q_{2} \leq b_{\eta}(q_{1}\)) over 0<\(q_{1}\)<2; every value strictly below this graph has exactly two nonsingular preimages of opposite determinant signs; every value on the graph has exactly one preimage, which is a nondegenerate fold; and the complete positive critical set is one embedded curve mapped bijectively onto that graph.
\end{problem}

\clearpage
\section{P046: An explicit PL threshold for a scalar exact penalty}
\label{problem:P046}

\begin{problem}
q(x)=max\{p(x),0\} and \(Phi_{r}(x)\)=f(x)+r q(x).

B=η+f(xbar)-\(\min_{x \in U}\)f(x), \(L=\max_{x \in U}\)‖\(\nabla\) f(x)‖, and T=max\{\(r_{ex}(U)\),B/\(\tau\),L/\(\sigma\)\}.

At every positive-gap point, \(Q_{r}(x)\)=dist(0,partial \(Phi_{r}(x))^{2}/{[}2(Phi_{r}\)(x)-f(xbar)){]}.

The assertion \(r_{PL}(U,η)\leq\)T means that for every \(\varepsilon >\)0 some \(\mu >\)0 satisfies \(Q_{r}(x)\geq \mu\) for all r\(\geq\)T+\(\varepsilon\) and all positive-gap band points.

The ambient space is finite-dimensional Euclidean space. A fixed compact neighborhood U contains xbar in its interior. The maps f:\(\mathbb R^{n}\) to \(\mathbb R\) and P:\(\mathbb R^{n}\) to \(\mathbb R^{m}\) are \(C^{2}\) on a neighborhood of U, with fixed bounds on their first and second derivatives.

The feasible set is C=\{x:P(x) in \(\mathbb R _{-} ^{m}\)\}. The point xbar is a local constrained minimizer, and S=\{x in C intersect U:f(x)=f(xbar)\} is nonempty.

Feasible quadratic growth holds on C intersect U: f(x)\(\geq\)f(xbar)+(\(\alpha\)/2)dist\((x,S)^{2}\) for a fixed \(\alpha >\)0.

A fixed residual error bound holds throughout U: dist(x,C)\(\leq \kappa\) dist(P(x),\(\mathbb R _{-} ^{m}\)) for a fixed \(\kappa\).

A height η\(>\)0 is fixed before selecting r. Within U there is no feasible point x with 0\(<\) f(x)-f(xbar)\(\leq\)η satisfying 0 in \(\nabla\) f(x)+\(N_{C}(x)\), where \(N_{C}\) is the limiting normal cone.

For r\(>\)0 define \(Phi_{r}(x)\)=f(x)+r dist(P(x),\(\mathbb R _{-} ^{m}\)). Define \(r_{ex}(U)\) as the infimum of r such that \(Phi_{r}(x)\geq\)f(xbar) for every x in U. Define \(r_{PL}(U,η)\) as the infimum of r0\(\geq r_{ex}(U)\) for which some \(\mu >\)0 satisfies dist(0,partial \(Phi_{r}(x))^{2} \geq\)2mu(\(Phi_{r}(x)\)-f(xbar)) for every r\(\geq\)r0 and every x in U with 0\(< Phi_{r}(x)\)-f(xbar)\(\leq\)η, where partial is the limiting subdifferential.

One smooth scalar inequality p(x)\(\leq\)0 with q(x)=max\{p(x),0\}, a fixed compact neighborhood U, and a uniform residual-gradient lower bound \(\sigma\) on the exterior tube 0\(\leq\)p\(\leq \tau\). This seeks an explicit multiplier-geometric threshold and is strictly stronger than the established instancewise finiteness result.

The constraint dimension is m=1, P=p is scalar-valued, C=\{x:p(x)\(\leq\)0\}, and q(x)=dist(p(x),\(\mathbb R _{-}\))=max\{p(x),0\}.

There are fixed \(\tau >\)0 and \(\sigma >\)0 such that ‖\(\nabla\) p(x)‖\(\geq \sigma\) for every x in U satisfying 0\(\leq\)p(x)\(\leq \tau\).

Set B=η+f(xbar)-\(\min_{x \in U}\)f(x), \(L=\max_{x \in U}\)‖\(\nabla\) f(x)‖, and T=max\{\(r_{ex}(U)\),B/\(\tau\),L/\(\sigma\)\}.

The inequality \(r_{PL}(U,η)\leq\)T is interpreted through threshold infima: for every \(\varepsilon >\)0 the modulus may depend on the complete instance and \(\varepsilon\) but must be uniform for all r\(\geq\)T+\(\varepsilon\).

Prove or refute that for every scalar C2 inequality instance in \(K_{ST}\), the explicit threshold T=max{\(r_{ex}(U)\),B/\(\tau\),L/\(\sigma\)}, where B=η+f(xbar)-\(\min_{U}\) f and \(L=\max_{U}\) ‖\(\nabla\) f‖, satisfies \(r_{PL}(U,η)\leq\)T. Equivalently, for every \(\varepsilon\)>0 there is a positive modulus valid simultaneously for all r\(\geq\)T+\(\varepsilon\) on the fixed neighborhood and band.
\end{problem}

\subsection{Unresolved Problems}
\clearpage
\section{P047: The four-step minimax gain}
\label{problem:P047}

\begin{problem}
\(E_{n}(instance, h)\) = {[}f(\(x_{n}\))-f(\(x\star\)){]}/(L \(D^{2}\)), where \(x_{n}\) is the final point of trajectory.

\(W_{n}(h)\) = sup\{\(E_{n}(instance, h)\): instance is admissible in any positive finite dimension\}.

\[
V_{n} = \inf\{W_{n}(h):h in \mathbb R^{n}\}
\]

\[
H_{n} = \{h in \mathbb R^{n}:W_{n}(h) = V_{n}\}
\]

\(X_{n}(h)\) is the set of admissible instances for which \(E_{n}(instance, h) = W_{n}(h)\).

\(P_{4}\) is the proposition \(V_{4} < V_{3}\).

L\(>\)0 and D\(>\)0, and the ambient dimension d may be any positive finite integer.

f:\(\mathbb R^{d} \to \mathbb R\) is convex, differentiable, and L-smooth, attains its minimum at \(x\star\), and the initial point satisfies ‖\(x_{0}-x\star\)‖ \(\leq\) D for at least one minimizer.

For a horizon n, h = (\(h_{0}\), \ldots, \(h_{n-1}\)) is any real scalar schedule fixed before observing gradients, and \(x_{t+1} = x_{t}-(h_{t}\)/L) \(\nabla\) f(\(x_{t}\)). Zero and negative entries are not excluded a priori.

The method uses only the current gradient and the fixed scalar schedule: no momentum, memory variables, auxiliary sequences, line search, or adaptivity.

For each h in \(\mathbb R^{n}, W_{n}(h)\) is the supremum of {[}f(\(x_{n}\))-f(\(x\star\)){]}/(L \(D^{2}\)) over the class in A1-A4 and all allowed finite dimensions, and \(V_{n} = \in f_{h \in \mathbb R^{n}} W_{n}(h)\); neither the infimum nor any supremum is assumed to be attained.

Only horizons 3 and 4 for the unrestricted fixed real scalar schedules and finite-dimensional smooth-convex instances of A1-A5 are in scope; horizon 3 is used solely as the sharp comparator for the four-step cell.

Every strictness or equality certificate must compare \(V_{4}\) with the exact \(V_{3}\); the chapter-reported silver coefficient \(r_{2}\) may be used only as an upper bound for one three-step schedule.
Upper certificates and lower obstructions must cover all positive finite dimensions and all convex differentiable L-smooth objectives allowed by A2, not only quadratics or one-dimensional functions.
No minimizer of \(W_{4}\) and no maximizer defining \(W_{4}(h)\) may be assumed. Relevant suprema are \(W_{4}(h)\) for h in \(H_{4}\) and any \(W_{3}(g)\) used in a sharp \(V_{3}\) comparison.
A numerical optimizer or floating-point SDP gap is diagnostic only; the final comparison and characterization require exact inequalities, exact algebraic data, or rigorously bounded rational certificates.
The required classification is \(H_{4}\) and extremal data at its schedules; no arbitrary-horizon or asymptotic silver-optimality claim is requested.

Under A1-A5, decide whether the exact four-step minimax value satisfies \(V_{4}<V_{3}\). Separately determine the attainment branch for \(H_{4}\) = {h in \(\mathbb R^{4}:W_{4}(h) = V_{4}\)}: if \(H_{4}\) is nonempty, characterize \(H_{4}\) exactly and, for every h in \(H_{4}\), give an exact finite interpolation description of at least one admissible instance attaining \(W_{4}(h)\); if \(H_{4}\) is empty, characterize an explicit minimizing schedule sequence and corresponding asymptotically extremal instances. The comparator is the exact \(V_{3}\), not the reported silver-schedule coefficient.
\end{problem}

\clearpage
\section{P048: Cost variation within an exact schedule profile}
\label{problem:P048}

\begin{problem}
For \(r_{0}\)=0 and \(r_{t+1}=r_{t}+s_{t}\), set \(Abar_{t}=A_{r_{t}+s_{t}-1}\)\ldots \(A_{r_{t}}\) and
\[
Bbar_{t}=B_{r_{t}+s_{t}-1}\ldots B_{r_{t}}
\]

\(H_{t}\) is generated by \(\varepsilon\), g, the declared constants, the full
deterministic schedule S, the initial pair (\(x^{0},y^{0}\)), and all completed
records (\(s_{u},\alpha_{u},x^{u+1},y^{u+1}\)) for 0\(\leq\)u\(<\) t. The controls \(s_{t}\) and
\(\alpha_{t}\) are selected from \(H_{t}\) before macro-update t. A stopping decision at
T is made only after update T-1 and is measurable with respect to \(H_{T}\).

After the \(H_{t}\)-measurable controls are fixed, the macro-update is
\[
x^{t+1}=Abar_{t} x^{t}-\alpha_{t} y^{t},\qquad y^{t+1}=Bbar_{t} y^{t}+\nabla
\]
F(\(x^{t+1})-\nabla\) F(\(x^{t}\)).

\[
g_{\kappa}(\varepsilon)=1+ceil(\kappa \log(1/\varepsilon))
\]

\(C_{S}(\varepsilon\),g) is the infimum, over policies satisfying A5 and SC5 that halt
after T\(\geq\)1 with T+1\(\leq\)g and attain \(\max_{i} \mid x_{i}^{T}\)-x\(\star\mid \leq \varepsilon\) on every
normalized instance, of the supremum over those instances of
\(\sum_{t=0}^{T-1} s_{t}\). Set \(C_{S}(\varepsilon\),g)=\(\infty\) if no such policy exists.

Define the profile
\[
\begin{aligned}
E(S)=(&n,q,a,b,D,\phi_{\min},\psi_{\min},\\
&\gamma_A,\gamma_B,\omega_{\min}),
\end{aligned}
\]
with the periodic Perron sequences and projected Euclidean descriptors defined
exactly in A7.

The triple (S,\(S_{\prime},\varepsilon\)) is active when E(S)=E(\(S_{\prime}\))=E0 and both
\(C_{S}(\varepsilon,g_{\kappa}(\varepsilon\))) and \(C_{S,\prime}(\varepsilon,g_{\kappa}(\varepsilon\))) are
finite and strictly positive.

For an active triple,
\[
\mathbb R(S,S_{\prime},\varepsilon)=\max\{C_{S}(\varepsilon,g_{\kappa}(\varepsilon))/C_{S,\prime}(\varepsilon,g_{\kappa}(\varepsilon)),
\]
\(C_{S,\prime}(\varepsilon,g_{\kappa}(\varepsilon))/C_{S}(\varepsilon,g_{\kappa}(\varepsilon\)))\}.

There are n\(\geq\)3 agents using deterministic exact local gradients in dimension p\(\geq\)1.

Each local objective \(f_{i}\) is continuously differentiable, has L-Lipschitz gradient, and is
\(\mu\)-strongly convex; the average objective has unique optimizer \(x\star\), and the local
objectives need not share \(x\star\).

The schedule is indexed from the fixed phase k=0 and is periodic with a declared class
period q. At every round the positive support graph is strongly connected and contains every
self-loop.

At round k, \(A_{k}\) is row-stochastic and \(B_{k}\) is column-stochastic. Their positive supports
realize the declared round graph, and every positive entry is at least the fixed class
bounds a\(>\)0 and b\(>\)0, respectively. Under A3 these self-looped irreducible factors have
primitive q-step products.

Every macro-update follows exactly the recurrence in \(central_{object}\). One communication round
applies \(A_{k}\) and \(B_{k}\) simultaneously, and one gradient refresh evaluates all local gradients
at the new primal state. Each block length and stepsize is selected from the schedule and
the information available through the current macro-update. A policy halting after T
completed macro-updates is admissible for budget g only if T+1\(\leq\)g on every run; it may leave
budget unused, and it may halt only after a macro-update.

Every schedule is compared on the same normalized objective and initialization class: \(\max_{i}\)
norm(\(x_{i}^{0}\)-x\(\star)=1, \max_{i}\) norm(\(\nabla f_{i}\)(x\(\star))\leq\)H for fixed finite H, and \(y_{i}^{0}=\nabla f_{i}(x_{i}^{0}\)). The order is infimum over admissible policies followed by supremum over this
class; successful policies and finite costs are not assumed.

Relative to the fixed phase and declared q,a,b, define \(\phi\) and ψ as the unique normalized
q-periodic absolute-probability sequences satisfying \(\phi_{k+1}^{T} A_{k}=\phi_{k}^{T}, B_{k} psi_{k}=psi_{k+1}, \phi_{k}^{T} 1=1, 1^{T} psi_{k}=1, \phi_{k+q}=\phi_{k}\), and \(psi_{k+q}=psi_{k}\). Set
\[
\begin{aligned}
P_k^A&=I-\mathbf 1\phi_k^{\mathsf T},
&P_k^B&=I-\psi_k\mathbf 1^{\mathsf T},\\
\phi_{\min}&=\min_{0\leq k<q,\,i}\phi_{k,i},
&\psi_{\min}&=\min_{0\leq k<q,\,i}\psi_{k,i}.
\end{aligned}
\]
\[
\begin{aligned}
\gamma_A&=\max_{0\leq k<q}
\left\lVert P_{k+1}^A A_kP_k^A\right\rVert_2,\\
\gamma_B&=\max_{0\leq k<q}
\left\lVert P_{k+1}^B B_kP_k^B\right\rVert_2.
\end{aligned}
\]
\[
\omega_{\min}=
\min_{0\leq k<q}\phi_k^{\mathsf T}\psi_k.
\]
\(D=\max_{0\leq k<q}\)diameter(\(G_{k}\)). The canonical tuple is
E(S)=(n,q,a,b,D,\(\phi_{min},psi_{min},\gamma_{A},\gamma_{B},\omega_{min}\)), with exact coordinatewise
equality required. Existence of a distinct matched-tuple pair is a witness obligation, not a
premise.

The frontier retains the anchor recurrence and normalized minimax objective class, specializes to
n=3, p=1, and declared q=3, and restricts every schedule in both targets to the same explicitly
noncommuting domain. Ratios are formed only where both minimax costs are finite and positive. Static
q=1, fixed-block, fixed-frequency, single-trajectory, and finite-accuracy transient discrepancies
are excluded from the claimed contribution.

Fix n=3, p=1, and declared q=3. Every schedule considered must have
phases i,j in \{0,1,2\} for which \(A_{i} A_{j}\) is not equal to \(A_{j} A_{i}\) or
\(B_{i} B_{j}\) is not equal to \(B_{j} B_{i}\).

Every compared pair must satisfy exact coordinatewise equality E(S)=E(\(S_{\prime}\))
for the fixed phase and declared class values of q, a, and b.

A multiplicative ratio is admissible only at points where
0\(< C_{S}(\varepsilon,g_{\kappa}(\varepsilon))< \infty\) and
0\(< C_{S,\prime}(\varepsilon,g_{\kappa}(\varepsilon))< \infty\).

Every cost retains the infimum over deterministic
schedule-aware causal variable-block policies followed by the
supremum over the full normalized objective and initialization
class.

At the beginning of macro-update t, the policy selects an integer \(s_{t} \geq\)1
and a finite scalar \(\alpha_{t}\) in {[}0,\(\infty\)), both measurable with
respect to the pre-update history \(H_{t}\); neither choice may depend on
\(x^{t+1}, y^{t+1}\), or the gradient refresh performed during macro-update
t.

For three-agent, scalar, declared period-three directed AB/Push-Pull schedules, determine whether
the policy-optimized communication-cost profiles on every exact canonical-tuple fiber have uniformly
bounded multiplicative diameter on some terminal accuracy interval. Equivalently, decide whether
temporal order omitted by E(S) can produce arbitrarily large active cost ratios at accuracies
tending to zero while the entire tuple and all objective scales remain fixed.
\end{problem}

\clearpage
\section{P049: Rate distortion under tracker reordering}
\label{problem:P049}

\begin{problem}
Write \(P_{j}\)=(A,\(B_{j}\)), S=(\(P_{0},P_{1},P_{2}\)), and S'=(\(P_{0},P_{2},P_{1}\)), with phase indices
extended periodically.

For \(P_{j}\)=(A,\(B_{j}\)), define \(M_{j}(\alpha\),H)={[}{[}A,-\(\alpha I_{3}\){]},{[}H(A-\(I_{3}),B_{j}-\alpha\) H{]}{]},
so \(z_{t+1}=M_{j} (\alpha,H)z_{t}\).

\(Mono_{T,r}(\alpha\),H) is the product of the three consecutive one-step maps for
order T beginning at phase r, restricted to \(Z_{H}\).

\(Q_{3}(T)=(q_{1}(T),q_{2}(T),q_{3}(T)\)), where \(q_{l}\) is the phase-maximized
primal-or-tracker Dobrushin quantity from A5.

\(\beta(T,H)\)=inf over stable \(\alpha\) of \(\rho(Mono_{T,0}(\alpha\),H)\(\mid Z_{H}\)), and
\(\gamma(T,H)=\beta (T,H)^{(1/3)}\).

\(T_{slow}\) is an order attaining \(\gamma_{+}\)=max\{\(\gamma(S,H),\gamma(S',H)\)\}, and
\(T_{fast}\) is the other order with \(\gamma_{-}\)=min\{\(\gamma(S,H),\gamma(S',H)\)\}.

\(\mathbb R_{lead}\)=(-log \(\gamma_{-}\))/(-log \(\gamma_{+}\)). For positive stable rates, \(\mathbb R_{lead}\)
is the limit as \(\varepsilon\) tends to zero of
\[
C_{\varepsilon}(T_{slow},H)/C_{\varepsilon}(T_{fast},H)
\]

\(C_{\varepsilon}(T,H)\) is exactly the optimized phase-uniform Euclidean tail complexity
on \(Z_{H}\) defined in A6.

There are n\(\geq\)3 agents and a deterministic periodic schedule of period p.~Every graph has
self-loops, and every cyclic block of W consecutive graphs has a strongly connected union.

At time t, \(A_{t}\) is graph-compatible and row-stochastic, \(B_{t}\) is graph-compatible and
column-stochastic, and every positive entry of either matrix is at least η\(>\)0.

The recursion is \(x_{t+1}=A_{t} x_{t}-\alpha y_{t}\) and \(y_{t+1}=B_{t} y_{t}+\nabla\) F(\(x_{t+1})-\nabla\) F(\(x_{t}\)),
with \(y_{0}=\nabla\) F(\(x_{0}\)) and one communication round per iteration.

Local objectives are scalar quadratics \(f_{i}(x)=h_{i} (x-c_{i})^{2/2}\) with \(\mu \leq h_{i} \leq\)L. When two
schedules are compared, the diagonal Hessian H, centers c, and optimizer \(x\star\) are
identical; the positive branch permits a fixed heterogeneous H, while the negative branch
ranges over this Hessian class.

Compared schedules are noncyclic orderings of the same \(p\) mixing pairs and have identical
\(n,p,W,\eta\) and graph union. For row-stochastic \(P\), define
\[
\tau(P)=\frac12\max_{i,j}\lVert e_i^\top P-e_j^\top P\rVert_1
\]
and
\[
\begin{aligned}
q_l(S)=\max_{0\leq r<p}\max\bigl\{&
\tau(A_{r+l-1}\cdots A_r),\\
&\tau((B_{r+l-1}\cdots B_r)^\top)
\bigr\},
\end{aligned}
\]
with
indices modulo p.~The positive branch requires equality of the exact \(q_{W}\) values, not
equality of analytical upper bounds, and restricts candidates to genuinely nonsymmetric
directed pairs not related by a node permutation followed by a cyclic phase shift.

Let \(z=(u,v)=(x-\mathbf 1x_\star,y)\) and \(Z_H=\{(u,v):\mathbf 1^\top v=\mathbf 1^\top Hu\}\). This subspace is invariant under
exact initialization. A stepsize \(\alpha >\)0 is stable when every phase-p error map restricted
to \(Z_H\) has spectral radius below 1. With \(\Phi_{S,r}(m,0;\alpha,H)\) denoting \(m\)-round evolution
from phase \(r, C_\varepsilon(S,H)\) is the minimum over stable \(\alpha\) of the least \(k\) such that
\[
\max_{0\leq r<p}\sup_{\substack{z\in Z_H\\z\neq0}}
\sup_{m\geq k}
\frac{\lVert\Phi_{S,r}(m,0;\alpha,H)z\rVert_2}{\lVert z\rVert_2}
\leq\varepsilon;
\]
it
is \(\infty\) if no stable \(\alpha\) exists.

The positive branch of A1-A6 with n=p=W=3, η=1/8, \(\mu\)=1, L=4, H=diag(1,2,4), zero centers,
individual-round counting, and the Euclidean operator norm on \(Z_{H}\). The primal matrix is fixed across
phases and strictly dominates every tracker-product Dobrushin channel, so exact \(Q_{3}\) equality is
shielded from tracker reordering. Unlike the accepted equal-rate additive log-log problem, this
frontier asks for unbounded leading multiplicative distortion. Unlike the previous one-sided
finite-time endpoint, every schedule in the target has positive optimized rate. Literature novelty
remains unresolved because the supplied source pack is empty and no independent retrieval route is
available to this role.

Fix n=p=W=3, η=1/8, \(\mu\)=1, L=4, H=diag(1,2,4), c=0, \(x\star\)=0, and
A=(1/8){[}{[}5,2,1{]},{[}1,5,2{]},{[}2,1,5{]}{]} at every phase.

Each \(B_{j}\) is a column-stochastic 3-by-3 matrix with entries in {[}1/8,3/4{]} and
asymmetry margin \(\max_{i,k} \mid (B_{j})_{ik}-(B_{j})_{ki} \mid\){} at least 1/16. Distinct
trackers satisfy ‖\(B_{j}-B_{k}\)‖\(_{F} ^{2} \geq\)1/4096.

For S=(\(B_{0},B_{1},B_{2}\)) and S'=(\(B_{0},B_{2},B_{1}\)), every node-permutation matrix Pi
and cyclic shift s in \{0,1,2\} satisfy max\_\{t in \{0,1,2\}\}‖\(B_{t}^{S}-Pi^{T} B_{t+s}^{S'}\)Pi‖\(_{F} ^{2} \geq\)1/4096.

For T in \{S,S'\}, l in \{1,2,3\}, and every starting phase r,
\(\tau((B_{r+l-1}^{T}\ldots B_{r}^{T})^{T})\leq \tau(A^{l}\))-1/100.

Both orders have nonempty stable-stepsize sets and optimized
per-round restricted spectral rates satisfying 0\(< \gamma(T,H)<\)1.

No assertion about literature novelty is part of the mathematical target.

For three-agent period-three directed push-pull schedules with one fixed primal mixing matrix and
three reordered tracker matrices, determine whether equality of the complete exact profile \(Q_{3}\) can
coexist with arbitrarily large leading communication-complexity ratios while both globally optimized
restricted spectral rates remain strictly positive.
\end{problem}

\clearpage
\section{P050: A memory-dependent stability switch}
\label{problem:P050}

\begin{problem}
For each phase first form \(H_{x}=(1-q)h_{x}\)+qx and \(H_{y}=(1-q)h_{y}\)+qy. In phase zero use the preceding pair (\(h_{x},h_{y}\)) in the \(A_{0}\) and \(B_{0}\) mixing terms; in phase one use (\(H_{x},H_{y}\)). Apply A5 and store (\(H_{x},H_{y}\)), defining maps \(T_{0}\)(q,a,b,\(\alpha\)) and \(T_{1}\)(q,a,b,\(\alpha\)) on \(\mathbb R^{8}\).

Define \(M_{d}\)(q,a,b,\(\alpha)=T_{1} (q,a,b,\alpha)T_{0}\)(q,a,b,\(\alpha\)).

Let I=\{(x0,x0-r,\(0_{2},0_{2}\)): x0,r in \(\mathbb R^{2}\)\} and K(M)=span\{\(M^{jv}\): v in I and 0\(\leq\)j\(\leq\)7\}.

Let o=(\(1_{2},0_{2},1_{2},0_{2}\)) and O=span\{o\}; remove only K(M) intersect O from the reachable space.

For s=(x,y,\(h_{x},h_{y}\)), set \(J(s)=1^{T}(y-x)\). The monodromy M is tracker-reachably Schur stable when O is contained in K(M), Mo=o, J(Ms)=J(s) on K(M), and the induced map on K(M)/O has spectral radius strictly less than one.

Let S(q,a,b) be the positive \(\alpha\) for which \(M_{d}\)(q,a,b,\(\alpha\)) is tracker-reachably Schur stable. If S(q,a,b) has a connected component C(q,a,b) whose closure contains zero, define \(\alpha_{\star}(q,a,b)\)=sup C(q,a,b) when this supremum is finite and positive.

Define Max(q)=\{(a,b) in {[}1/4,3/4{]}\(^{2}: \alpha_{\star}(q,a,b)\)=max \(\alpha_{\star}\)(q,\(a_{\prime},b_{\prime}\)), where the maximum ranges over (\(a_{\prime},b_{\prime}\)) in {[}1/4,3/4{]}\(^{2}\)\}.

There are two agents with scalar objectives \(f_{i}(z)=(z-r_{i})^{2/2}\) for arbitrary real offsets \(r_{1}\) and \(r_{2}\); the aggregate minimizer is (\(r_{1}+r_{2}\))/2.

Communication alternates with period two. At even rounds use \(A_{0}\)={[}{[}1,0{]},{[}a,1-a{]}{]} and \(B_{0}\)={[}{[}1-b,0{]},{[}b,1{]}{]}; at odd rounds use \(A_{1}\)={[}{[}1-a,a{]},{[}0,1{]}{]} and \(B_{1}\)={[}{[}1,b{]},{[}0,1-b{]}{]}, where a and b belong to {[}η,1-η{]} for a fixed η in (0,1/2{]}.

The phase delays are \(d_{2m}=\delta_{0}\) and \(d_{2m+1}=\delta_{1}\) with \(\delta_{0},\delta_{1}\) in \{0,1\}; histories required at index -1 are initialized to zero. The principal frontier cell is (\(\delta_{0},\delta_{1}\))=(1,0).

For q in (0,1{]}, define the algorithmic memory-relaxation map \(Q_{q}(v)\)=qv componentwise. With \(h_{x}^{-1}=h_{y}^{-1}\)=0, update \(h_{x}^{k}=h_{x}^{k-1}+Q_{q}(x^{k}-h_{x}^{k-1}\)) and \(h_{y}^{k}=h_{y}^{k-1}+Q_{q}(y^{k}-h_{y}^{k-1}\)). No information-loss or bit-compression interpretation is assumed.

For a constant stepsize \(\alpha >\)0, first form the memories in A4 and then update \(x^{k+1}=x^{k}+(A_{k}-I)h_{x}^{k-d_{k}}-\alpha y^{k}\) and \(y^{k+1}=y^{k}+(B_{k}-I)h_{y}^{k-d_{k}}+(x^{k+1}\)-r)-(\(x^{k}\)-r), with \(y^{0}=x^{0}\)-r.

Every initial primal vector \(x^{0}\) in real two-space is allowed, and convergence or failure is evaluated only on states reachable from the initialization and memory conventions in A3-A5.

Use exactly A1-A6 with η=1/4, q in {[}1/4,1/2{]}, (a,b) in {[}1/4,3/4{]}\(^{2}\), and delay placement (\(\delta_{0},\delta_{1}\))=(1,0). Phase zero uses the preceding memories and phase one uses the newly formed memories. Stability is evaluated on the initialized reachable Krylov quotient after removing exactly the optimizer-family direction.

Allow q in {[}1/4,1/2{]} and (a,b) in {[}1/4,3/4{]}\(^{2}\).

Fix (\(\delta_{0},\delta_{1}\))=(1,0) with the memory-before-mixing order of A4-A5.

Use only modes in the Krylov span generated by A6 initializations, and quotient only the fixed optimizer direction.

Classify maximizers over the entire closed weight square, not merely among its four corners or on a sampled grid.

Prove or refute the following fully quantified proposition. For every (q,a,b) in [1/4,1/2] x [1/4,3/4]\(^{2}\), the connected-from-zero tracker-reachable stability component C(q,a,b) exists, its endpoint \(\alpha_{\star}(q,a,b)\) is finite and positive, and \(\alpha_{\star}\) is continuous on the closed prism. Moreover, there exists a unique \(q_{c}\) in (7/20,3/8) such that Max(q)={(3/4,1/4)} for 1/4\(\leq\)q<\(q_{c}\), Max(\(q_{c}\))={(3/4,1/4),(1/4,1/4)}, and Max(q)={(1/4,1/4)} for \(q_{c}\)<q\(\leq\)1/2.
\end{problem}

\clearpage
\section{P051: Exact four-query regret with bounded bias}
\label{problem:P051}

\begin{problem}
W(A;L,\(\beta\)) is the supremum over f in \(D_{OBJECTIVES}\) and b in \(D_{BIASES}\) of f(\(\widehat{x}_{A})-\min_{x \in [0,1]}\) f(x).

\[
\mathbb R_{4}(L,\beta)=\in f_{A \in D_{ALGORITHMS}} W(A;L,\beta)
\]

\(A_{grid}\) queries \(q_{i}\)=(2i-1)/8 for i=1,2,3,4 and returns a queried point having the smallest observed value.

The domain is {[}0,1{]} with its usual distance, L\(>\)0, and 0\(< \beta \leq\)L/16.

The unknown objective f:{[}0,1{]}\(\to \mathbb R\) is L-Lipschitz, with no convexity, smoothness, or derivative information assumed.

Each query x returns y(x)=f(x)+b(x), where b:{[}0,1{]}\(\to \mathbb R\) is one fixed unknown function satisfying \(\mid\) b(x)\(\mid \leq \beta\) everywhere and has no further regularity.

An algorithm is deterministic, makes at most four adaptively selected evaluations, and may return any transcript-dependent point of {[}0,1{]}, whether queried or unqueried.

\(\mathbb R_{4}\)(L,\(\beta\)) is the infimum over such algorithms of the supremum of f(x̂)-\(\min_{x \in [0,1]}\) f(x) over all pairs (f,b) satisfying A2 and A3.

One-dimensional L-Lipschitz global optimization on {[}0,1{]} with one fixed arbitrary bias field bounded by \(\beta\), at most four deterministic adaptive evaluations, and a recommendation that may be queried or unqueried. The task is the exact minimax equality in the stated small-bias cell; randomized algorithms, regular bias fields, higher query budgets, and asymptotic rates are excluded.

The evaluation budget is at most four and the parameter range is L\(>\)0 and 0\(< \beta \leq\)L/16.

Repeated evaluation of a point returns the same value f(x)+b(x), and b need not be continuous or regular.

The final recommendation may be any transcript-dependent point of {[}0,1{]}.

Determine whether \(\mathbb R_{4}\)(L,\(\beta\))=L/8+2 \(\beta\) for every L>0 and 0<\(\beta \leq\)L/16 under the deterministic, adaptive, unrestricted-recommendation oracle model of A1-A5.
\end{problem}

\clearpage
\section{P052: Adaptation to an unknown tail index}
\label{problem:P052}

\begin{problem}
\[
f(x)=n^{-1} \sum_{i=1}^{n} f_{i}(x),\qquad \\Phi(x)=f(x)+r(x)
\]

η=1/(2L) and \(P_{\eta}(x)=η^{-1}\){[}x-\(prox_{η r}\)(x-η \(\nabla\) f(x)){]}.

\(\mathbb R_{p}(T;n,L,Δ)\)=inf over A in \(Alg_{p}(T)\) of sup over d\(\geq\)1 and I in \(I_{p}(d;n,L,Δ)\) of \(E_{I}\){[}‖\(P_{\eta}(X_{T}^{A}\))‖{]}.

\[
\begin{aligned}
M_p(\varepsilon;n,L,\Delta)
=\inf\bigl\{T\in\mathbb Z_{\geq0}:{}&
\mathbb R_p(T;n,L,\Delta)\\
&\leq\varepsilon\bigr\}.
\end{aligned}
\]

\(T_{B}\)(p,\(\varepsilon\);n,L,Δ)=inf\{T in nonnegative integers: sup over d\(\geq\)1 and I in \(I_{p}(d;n,L,Δ)\) of \(E_{I}\){[}‖\(P_{\eta}(X_{T}^{B}\))‖{]}\(\leq \varepsilon\)\}.

\[
\Lambda=\log(e+n+L \Delta/\varepsilon^{2})
\]

Let d be arbitrary and finite, n\(\geq\)2, L\(>\)0, Δ\(>\)0, and p0 in (1,2). Each \(f_{i}:\mathbb R^{d} \to \mathbb R\) is differentiable and L-smooth but may be nonconvex; r is proper, closed, convex, and has a computable proximal map.

The objective \(\Phi(x)=n^{-1} \sum_{i} f_{i}(x)\)+r(x) is bounded below, and the supplied initial point x0 satisfies \(\Phi(x0)\)-inf \(\Phi \leq\)Δ.

For each p in {[}p0,2{]}, \(I_{p}\) consists of instances whose query (i,x) returns \(\nabla f_{i}(x)\)+ξ with E{[}ξ\(\mid\) history,i,x{]}=0 and E{[}‖ξ‖\(^{p} \mid\) history,i,x{]}\(\leq\)1. Noises are fresh and conditionally independent across calls, with no shared-noise component-difference access, cross-point coupling, or oracle-increment regularity.

An admissible T-query algorithm is a dimension-indexed randomized family that adaptively selects component indices and query points from its transcript and outputs \(X_{T}\). All algorithms know n,L,Δ,p0,r and x0; the p-aware families defining \(M_{p}\) additionally know p, while the single family B does not. The guarantee is required separately over every nested class \(I_{p}\) and does not posit a latent identifiable p.

Let η=1/(2L), \(f=n^{-1} \sum_{i} f_{i}\), and \(P_{\eta}(x)=η^{-1}\){[}x-\(prox_{η r}\)(x-η \(\nabla\) f(x)){]}. Accuracy is E{[}‖\(P_{\eta}(X_{T}\))‖{]}\(\leq \varepsilon\) for 0\(< \varepsilon <\)√(L\emph{Δ), \(\Lambda\) means log(e+n+L}Δ/\(\varepsilon^{2}\)), the permitted slack is exactly \(\Lambda^{2}\), and the investigated noise-dominated tuples satisfy 0\(< M_{p}(\varepsilon\);n,L,Δ)\(< \infty\) and \(M_{p}(\varepsilon\);n,L,Δ)\(\geq\)4n.

Centralized finite-sum composite optimization under A1-A5. The performance criterion is worst-case expected proximal-gradient norm. Oracle noises are fresh and pointwise, with no shared-noise component differences or cross-point regularity. The question concerns simultaneous nested-class risk rather than identification of a latent tail index.

Performance is measured by E{[}‖\(P_{\eta}(X_{T}\))‖{]}, never by an expected squared norm.

Queries provide only fresh pointwise noisy component gradients; paired component differences and cross-point noise coupling are unavailable.

The comparison is required only when 0 \(<{} M_{p}(\varepsilon\);n,L,Δ) \(<{} \infty\) and \(M_{p}(\varepsilon\);n,L,Δ) \(\geq\) 4n, with slack exactly \(\Lambda^{2}\).

For every p0 in (1,2), determine whether there exist one anytime admissible algorithm family B, not told p, and a finite constant C depending only on p0 such that \(T_{B}\)(p,\(\varepsilon) \leq\) C \(\Lambda^{2} M_{p}(\varepsilon\);n,L,Δ) simultaneously for every p in [p0,2] and every admissible tuple satisfying 0 < \(M_{p} < \infty\) and \(M_{p} \geq\) 4n.
\end{problem}

\clearpage
\section{P053: The adaptive query threshold on the square}
\label{problem:P053}

\begin{problem}
At query \(x_{t}\) after history \(h_{t-1}\), the report is \(y_{t}\)=f(\(x_{t}\))+E(\(x_{t},h_{t-1}\)).

The simple regret of x̂ is \(\mathbb R(f,x̂)\)=f(x̂)-\(\min_{x \in X}\)f(x).

\(N_{ad}\) is the least positive integer N for which some algorithm in \(A_{N}\) guarantees \(\mathbb R(f,x̂)\leq\)3L/8 for every f in \(F_{L}\) and every E in \(O_{L/8}\).

G consists of \(x_{ij}\)=((2i+1)/8,(2j+1)/8) for i,j in \{0,1,2,3\}; the index distance between \(x_{ij}\) and \(x_{kl}\) is max\{\(\mid\) i-k\(\mid,\mid\) j-l\(\mid\)\}.

For q in S subset G, \(W_{14}(S,q)\) holds at \(\mid\) S\(\mid\)=14 exactly when every point of G outside S has index distance at most one from q. For \(\mid\) S\(\mid <\)14 it holds exactly when some x in G outside S satisfies both \(W_{14}(S union \{x\},q)\) and \(W_{14}(S union \{x\},x)\).

Policy \(P_{14}\) first queries \(x_{00}\). In state (S,q), where q is the earliest queried point having minimum report, it queries the lexicographically first x witnessing the recursion for \(W_{14}(S,q)\). It replaces q by x exactly when the new report is strictly smaller, stops after fourteen queries, and recommends q.

The search space is X={[}0,1{]}\(^{d}\) for a fixed positive integer d, equipped with the \(\infty\) norm.

The unknown objective f:X to \(\mathbb R\) is globally L-Lipschitz for a known constant L\(>\)0.

At query t the oracle returns \(y_{t}\)=f(\(x_{t})+e_{t}\), where the errors are supplied by one globally defined deterministic rule of the current query and preceding transcript, defined for every counterfactual history and repeated query, and satisfy \(\mid e_{t} \mid \leq \beta\) everywhere.

The bias magnitude satisfies 0\(< \beta <\) L/4.

For a positive integer N, the algorithm is deterministic, may choose at most N queries adaptively from past observations, and may recommend any point of X, queried or unqueried.

Performance is worst-case simple regret f(x̂)-\(\min_{X}\) f with the non-strict success requirement f(x̂)-\(\min_{X}\) f\(\leq \varepsilon\), where 0\(< \varepsilon <\) L/2.

The domain is {[}0,1{]}\(^{2}\) with the \(\infty\) norm. The objective is any real-valued L-Lipschitz function for arbitrary L\(>\)0. A deterministic algorithm may choose each query from previous reports, stop after at most N queries, and recommend any queried or unqueried point. One deterministic error rule defined on every query-history pair supplies errors bounded by L/8. Success is non-strict simple regret at most 3L/8. Fixed-only designs, randomized algorithms, stochastic noise, other parameter ratios, and other dimensions are excluded.

Specialize to d=2, \(\beta\)=L/8, and \(\varepsilon\)=3L/8 for arbitrary L\(>\)0.

Every query and the stopping decision may depend deterministically on the complete preceding report transcript.

The bounded error rule must be defined on every counterfactual query-history pair, including repeated queries.

The final recommendation may be queried or unqueried.

Regret equal to 3L/8 is successful.

Fix d=2, \(\beta\)=L/8, and \(\varepsilon\)=3L/8. Determine whether the deterministic adaptive minimax evaluation complexity is exactly fourteen under globally defined transcript-dependent bounded corruption and unrestricted recommendations.
\end{problem}

\clearpage
\section{P054: The fixed-design query threshold on the square}
\label{problem:P054}

\begin{problem}
At the t-th fixed query \(q_{t}\) after history \(h_{t-1}\), the report is \(y_{t}\)=f(\(q_{t}\))+E(\(q_{t},h_{t-1}\)).

The simple regret of x̂ is \(\mathbb R(f,x̂)\)=f(x̂)-\(\min_{x \in X}\) f(x).

\(N_{fix}\) is the least positive integer N for which some design in \(D_{N}\) guarantees \(\mathbb R(f,x̂)\leq\)3L/8 for every f in \(F_{L}\) and every E in \(O_{L/8}\).

G=\{1/8,3/8,5/8,7/8\}\(^{2}\) is the sixteen-point tensor grid in X.

The search space is X={[}0,1{]}\(^{d}\) for a fixed positive integer d, equipped with the \(\infty\) norm.

The unknown objective f:X to \(\mathbb R\) is globally L-Lipschitz for a known constant L\(>\)0.

At query t the oracle returns \(y_{t}\)=f(\(x_{t})+e_{t}\), where the errors are supplied by one globally defined deterministic rule of the current query and preceding transcript, defined for every counterfactual history and repeated query, and satisfy \(\mid e_{t} \mid \leq \beta\) everywhere.

The bias magnitude satisfies 0\(< \beta <\) L/4.

For a positive integer N, the algorithm is deterministic, may choose at most N queries adaptively from past observations, and may recommend any point of X, queried or unqueried.

Performance is worst-case simple regret f(x̂)-\(\min_{X}\) f with the non-strict success requirement f(x̂)-\(\min_{X}\) f\(\leq \varepsilon\), where 0\(< \varepsilon <\) L/2.

The domain is X={[}0,1{]}\(^{2}\) with the \(\infty\) norm. The objective is any real-valued L-Lipschitz function, where L\(>\)0. Query locations must be fixed before any reports are observed, although they are queried sequentially under one globally defined history-dependent error rule bounded by L/8. The deterministic recommendation may be any report-dependent point of X. The target is the exact fixed-design query count for non-strict simple regret at most 3L/8; adaptive placement, randomized algorithms, stochastic noise, other dimensions, and fixed spatial-bias restrictions are excluded.

Specialize to d=2, \(\beta\)=L/8, and \(\varepsilon\)=3L/8 for arbitrary L\(>\)0.

All query locations are chosen before reports are observed, but the recommendation is an arbitrary deterministic function of the complete report vector.

The error rule is defined and bounded on every query-history pair, including counterfactual histories and repeated locations.

Success means simple regret at most 3L/8, including equality.

Fix d=2, \(\beta\)=L/8, and \(\varepsilon\)=3L/8. Determine whether the deterministic fixed-design minimax evaluation complexity is exactly sixteen when reports have globally defined transcript-dependent bounded errors and the recommendation may depend on all reports and be unqueried.
\end{problem}

\clearpage
\section{P055: A sharp residual threshold for moving HJB selectors}
\label{problem:P055}

\begin{problem}
\[
\mathbb R_{n}=\delta_{n}^{(1/6)}, a_{n}=\mathbb R_{n}^{2},,\qquad c_{n}=\gamma \mathbb R_{n}/a_{n}
\]

For the fixed canonical model, V=0, \(H_{v}\){[}V{]}=0, \(H_{r}\){[}V{]}=\(x^{2}\), and \(G_{V}=x^{2}\).

For a closed-loop solution X, \(\tau_{0}(X)\)=inf\{t\(\geq0:X_{t}\)=0\}, with \(inf(empty set)=\infty\).

For a specified candidate sequence w, \(P_{n}^{w}(\lambda\)) is the supremum of P(\(\tau_{0}(X)\leq\)T) over every A6-authorized feedback and every corresponding weak solution with \(X_{0}=\lambda \mathbb R_{n}\).

\(F_{n}(t,x)\)=integral from 0 to x of \((x-y)y^{2}\) k((y-\(c_{n}\) t)/\(a_{n}\))dy.

Set \(w_{n}=F_{n}\) for x\(\leq 2R_{n}\). For z=(x-\(2R_{n})/a_{n}\) in {[}0,\(L_{cut}\){]}, define \(w_{n}\)(t,\(2R_{n}+a_{nz}\)) as the unique polynomial of degree at most five whose value and first two z-derivatives at z=0 are \(F_{n}, a_{n} (F_{n})_{x}\), and \(a_{n}^{2} (F_{n})_{xx}\), and whose value and first two z-derivatives at \(z=L_{cut}\) are zero. Set \(w_{n}\)=0 for x\(\geq 2R_{n}+L_{cut} a_{n}\).

A critical template includes one specified Borel feedback which selects r on raw-region phase plateaus where k((x-\(c_{n}\) t)/\(a_{n}\))=-2 and selects an \(\varepsilon_{n}\)-minimizing action elsewhere, including throughout the Hermite transition.

A critical template is valid if its profile and Hermite data satisfy \(SC_{cutoff,bounds}\), its candidate satisfies A4-A5 with \(\delta_{n}=\mathbb R_{n}^{6}\) and \(\varepsilon_{n} \leq K_{crit} \mathbb R_{n}^{3}\) for one fixed \(K_{crit}\), and its specified feedback satisfies A6.

For a valid critical template q and C\(>0, Q_{n}(q,C)\) is the infimum of P(\(\tau_{0}(X)\leq\)min\{\(CR_{n}\),T\}) over all weak solutions corresponding to q's specified feedback with \(X_{0}=\mathbb R_{n}\). This set is nonempty by A1.

Fix T\(>\)0, D=(0,1), and U=\{v,r\}. Under action a, \(dX_{t}=b_{a}\)(t,\(X_{t}\))dt+\(\sigma_{a} (t,X_{t})dW_{t}\). The coefficients are bounded and continuous on {[}0,T{]} times {[}0,1{]}, locally Lipschitz in x on D, and every tested Borel feedback admits at least one weak solution up to the exit time \(\tau\) from D.

There are \(r_{0},c_{0},C_{0},s_{0} >\)0 such that for x\(< r_{0}, b_{v}(t,x)\geq c_{0}\) and \(\mid \sigma_{v}(t,x)\mid \leq C_{0}\) x, while \(\mid \sigma_{r}(t,x)\mid \geq s_{0}\). For x\(>1-r_{0}\) and both actions, \(b_{a}(t,x)\leq-c_{0}\) and \(\mid \sigma_{a}(t,x)\mid \leq C_{0}(1-x)\). Thus the upper boundary has an explicit action-uniform invariance condition and only the lower boundary distinguishes v from r.

Running costs \(l_{a}\) and terminal cost g are bounded and continuous. The finite-horizon state-constrained value V extends spatially as a \(C^{2,\alpha}\) function on {[}0,\(r_{0}\){]} for a fixed \(\alpha\) in (0,1{]}, solves \(V_{t}+\min_{a} H_{a}\){[}V{]}=0 in the lower collar, and has exact gap \(G_{V}(t,x)=H_{r}\)\[V\](t,x)-\(H_{v}\)\[V\](t,x) satisfying \(c_{-} x^{p} \leq G_{V}(t,x)\leq c_{+} x^{p}\) for fixed p,\(c_{-},c_{+} >\)0. Here \(H_{a}\){[}z{]}=\(l_{a}+b_{a} z_{x}+(1/2)\sigma_{a}^{2} z_{xx}\).

Candidates \(w_{n}\) are bounded \(C^{1,2}\) functions with \(\varepsilon_{n},\delta_{n}\) tending to zero and satisfy \(\sup_{t<T,x \in D} \mid (w_{n})_{t}+\min_{a} H_{a}{[}w_{n}{]}\mid \leq \varepsilon_{n}, \sup_{t,x} \mid w_{n}\)-V\(\mid \leq \delta_{n}\), and \(\sup_{x} \mid w_{n}(T,x)\)-g(x)\(\mid \leq \varepsilon_{n}\).

For one fixed M\(< \infty, \sup_{n} \sup_{t}\) ‖\(w_{n}(t,.)\)‖\emph{\{\(C^{2,\alpha}({[}0,r_{0}\){]})\}+\(\sup_{t}\) ‖V(t,.)‖}\{\(C^{2,\alpha}({[}0,r_{0}\){]})\}\(\leq\)M. The stopped Ito families required for later cost comparison are uniformly integrable.

For every \(n, u_n\) is any Borel feedback satisfying
\[
H_{u_n(t,x)}[w_n](t,x)
\leq \min_a H_a[w_n](t,x)+\varepsilon_n.
\]
All viability and exit conclusions quantify over every weak closed-loop solution under \(u_n\); the unstopped cost is considered only after lower- and upper-boundary viability have been established.

One fixed finite-horizon one-dimensional model satisfying A1-A6, with \(\alpha\)=1 and p=2. The upper assertion ranges over all A4 candidate sequences, all A6-authorized Borel feedbacks, and all corresponding weak solutions. A critical template includes a specified authorized feedback, and its exit probability is the infimum over all corresponding weak solutions. The moving wave has a precisely defined spatial cutoff and no terminal cutoff.

Fix a smooth function θ on {[}0,1{]} with θ=0 on {[}0,1/4{]} and θ=1 on {[}3/4,1{]}. Set \(b_{v}\)=1-2theta, \(b_{r}\)=-θ, \(\sigma_{v}\)=(1-θ)x+θ(1-x), \(\sigma_{r}\)=(1-θ)+θ(1-x), \(l_{v}=0, l_{r}=x^{2}\), and g=0.

Let \(\delta_{n} >\)0 tend to zero, \(\mathbb R_{n}=\delta_{n}^{(1/6)}\), and \(a_{n}=\mathbb R_{n}^{2}\).

The universal assertion assumes \(\varepsilon_{n}/\mathbb R_{n}^{3}\) tends to zero.

The upper assertion starts at \(X_{n}(0)=\lambda \mathbb R_{n}\) for fixed \(\lambda >\)0, while the critical construction starts at \(X_{n}(0)=\mathbb R_{n}\).

A critical template uses a smooth one-periodic mean-zero function k with k\(\geq\)-2 and k=-2 on an interval of positive length, speed \(c_{n}=\gamma \mathbb R_{n}/a_{n}\), and raw wave \(F_{n}(t,x)\)=integral from 0 to x of \((x-y)y^{2}\) k((y-\(c_{n}\) t)/\(a_{n}\))dy for x\(\leq 2R_{n}\). Across {[}\(2R_{n},2R_{n}+L_{cut} a_{n}{]}, w_{n}\) is the unique degree-at-most-five Hermite polynomial matching \(F_{n}, a_{n} (F_{n})_{x}\), and \(a_{n}^{2} (F_{n})_{xx}\) at the left endpoint and matching a zero two-jet at the right endpoint. Set \(w_{n}\)=0 beyond that interval.

A valid critical template has constants \(C_{0}\),\ldots,\(C_{4}\) and \(C_{H}\) independent of n such that in the transition \(\mid w_{n} \mid \leq C_{0R,n}^{6}, \mid (w_{n})_{x} \mid \leq C_{1R,n}^{4}, \mid (w_{n})_{xx} \mid \leq C_{2R,n}^{2}\), the spatial Lipschitz seminorm of \((w_{n})_{xx}\) is at most \(C_{3}, \mid (w_{n})_{t} \mid \leq C_{4R,n}^{3}\), and \((w_{n})_{xx} \geq-2x^{2}-C_{HR,n}^{3}\).

The moving wave is not multiplied by a terminal-time cutoff.

In the fixed canonical two-action diffusion with exact value V=0, \(\delta_{n}=\mathbb R_{n}^{6}\), and interpolation width \(a_{n}=\mathbb R_{n}^{2}\), prove or refute the conjunction of a universal subcritical exit bound and an existential critical moving-lattice construction, with all probabilities explicitly extremized over corresponding weak solutions.
\end{problem}

\clearpage
\section{P056: Pointwise errors in variable-metric PDHG}
\label{problem:P056}

\begin{problem}
\(g\star\) denotes the Fenchel conjugate of g.

T(x,y)=(partial f(x)+K\emph{y, partial g}(y)-Kx), where \(K\star\) is the Hilbert adjoint of K.

Set \(e_{k+1}=(e^{x}\)\emph{(k+1),\(e^{y}\)}(k+1)).

With Δ \(x_{k}=x_{k+1}-x_{k}\) and Δ \(y_{k}=y_{k+1}-y_{k}\), define \(w_{k+1}=(K\star\)Δ \(y_{k}-P_{k}\) Δ \(x_{k}-e^{x}\)\emph{(k+1), K Δ \(x_{k}-Q_{k}\) Δ \(y_{k}-e^{y}\)}(k+1)); A3 gives \(w_{k+1}\) in T(\(x_{k+1},y_{k+1}\)).

The error sequence is pointwise root-small when √(k+1)(‖\(e^{x}\)\emph{(k+1)‖+‖\(e^{y}\)}(k+1)‖)\(\to\)0.

X and Y are finite-dimensional real Hilbert spaces; f on X and g on Y are proper lower-semicontinuous convex functions, and K:X\(\to\)Y is linear.

The KKT operator T(x,y)=(\(\partial\)f(x)+K\emph{y, \(\partial\)g}(y)-Kx) has a nonempty zero set.

For every k, the iterates satisfy 0\(\in \partial\)f(\(x_{k+1}\))+K*\(y_{k}+P_{k}(x_{k+1}-x_{k})+e^{x} _{k+1}\) and 0\(\in \partial\)g*(\(y_{k+1}\))-K(2\(x_{k+1}-x_{k})+Q_{k}(y_{k+1}-y_{k})+e^{y} _{k+1}\), so \(e^{x}\) and \(e^{y}\) are the declared inexact-resolvent optimality residuals.

\(P_{k}\) and \(Q_{k}\) are self-adjoint positive-definite block metrics with mI\(\preccurlyeq P_{k},Q_{k}\preccurlyeq\)MI for fixed 0\(<\) m\(\leq\)M, and ‖\(Q_{k}^{(-1/2)}\) K \(P_{k}^{(-1/2)}\)‖\(\leq1-\delta\) for a fixed \(\delta >\)0.

There are nonnegative \(η_{k}\) with \(\sum_{k} η_{k} < \infty\) such that \((1+η_{k})^{(-1)} P_{k}\preccurlyeq P_{k+1}\preccurlyeq(1+η_{k})P_{k}\) and the same two-sided relation holds for \(Q_{k}\).

The inexactness is norm-summable: \(\sum_{k}\) (‖\(e^{x}\)\emph{(k+1)‖+‖\(e^{y}\)}(k+1)‖)\(< \infty\).

All finite-dimensional A1-A6 variable-metric PDHG data and their original last-iterate sequences. The observable is the selected graph-valid KKT element. No averaging, best-iterate selection, strong monotonicity, bounded-variation assumption, or restriction to quadratic objectives is permitted.

At \(z_{k+1}=(x_{k+1},y_{k+1}\)), use \(w_{k+1}\)=(K\(\star(y_{k+1}-y_{k})-P_{k}(x_{k+1}-x_{k})-e^{x}\)\emph{(k+1), K(x}(k+1)-\(x_{k})-Q_{k}(y_{k+1}-y_{k})-e^{y} _{k+1}\)).

The tested stratum satisfies √(k+1)(‖\(e^{x}\)\emph{(k+1)‖+‖\(e^{y}\)}(k+1)‖)\(\to\)0 in addition to the norm summability already required by A6.

The metrics may vary noncommutatively subject only to A4-A5; they need not become constant after finitely many iterations.

The limit is taken along every original iterate k and not along a subsequence, average, or running minimum.

Determine whether norm-summable optimality errors satisfying the pointwise condition √(k+1)(‖\(e^{x} _{k+1}\)‖+‖\(e^{y} _{k+1}\)‖)\(\to\)0 necessarily preserve the exact-method last-iterate law √(k+1)‖\(w_{k+1}\)‖\(\to\)0 under all metric sequences allowed by A4-A5, without any bounded-variation or dyadic-tail hypothesis.
\end{problem}

\clearpage
\section{P057: Heavy-tail minimax equivalence}
\label{problem:P057}

\begin{problem}
\(N_{C}(\rho)=N_{C}(1/8;1,3/2,1,1,1,1,\rho\)) for C in \{U,G\}, with the argument order fixed by A8.

\[
d_{ratio}=\{\rho >0:0< N_{G}(\rho)< \infty\}
\]

Leading-order equivalence holds when some \(r_{threshold}\) satisfies {[}\(r_{threshold},\infty\)) subset of \(d_{ratio}\) and, for every \(\delta >\)0, some \(\mathbb R >\)0 satisfies 1-\(\delta < N_{U}(\rho)/N_{G}(\rho)\leq\)1 for every \(\rho \geq \mathbb R\).

Fix an integer \(d\geq 1, 1<p<2, 0<\mu\leq L, R_0>0, \sigma_\star>0\), and \(0<\varepsilon<R_0/4\). The decision space is \(\mathbb R^d, \lVert x_0-x_\star\rVert\leq R_0\), and no bounded-domain assumption is imposed.

Each instance has g=f+\(\phi\), where f is differentiable, L-smooth, and \(\mu\)-strongly convex, while \(\phi\) is proper, closed, and convex; consequently g has a unique minimizer \(x\star\).

For every declared positive step size \(\eta\), an exact call to \(\operatorname{prox}_{\eta\phi}\) is available and counts as one proximal call.

For every adaptively selected query \(Q_t\) and the \(\sigma\)-field \(\mathcal F_{t-1}\) generated by all earlier states, queries, oracle values, and internal randomness, the oracle sample \(G_t\) satisfies
\[
\mathbb E[G_t\mid\mathcal F_{t-1},Q_t]=\nabla f(Q_t)
\]
and
\[
\left(\mathbb E\left[\lVert G_t-\nabla f(Q_t)\rVert^p
\mid\mathcal F_{t-1},Q_t\right]\right)^{1/p}
\leq \sigma_\star+\sigma_L\lVert Q_t-x_\star\rVert.
\]
Samples drawn after a fixed query are conditionally independent given that query and the past.

The parameters d, p, L, \(\mu\), R0, \(\sigma_{*}\), and \(\sigma_{L}\) are declared to algorithms, but \(x\star\), the instance, and the oracle distribution are not. The occurrence of \(x\star\) in A1 and A4 defines the instance class and supplies no algorithmic side information.

Class U consists of measurable state machines with arbitrary standard-Borel persistent state. They may retain past oracle information, choose current or off-current queries adaptively, use robust estimation or clipping, and make exact proximal calls. Every oracle sample and proximal call consumes the common budgets.

Class G is the precommitted geometric-median proximal template family. Before observing oracle values, an algorithm fixes a finite horizon \(T\), positive step sizes \(\eta_t\), positive batch sizes \(b_t\), and odd block counts \(k_t\) dividing \(b_t\), as functions only of the declared parameters, with \(T\leq M(\varepsilon)\). At round \(t\), it draws \(b_t\) samples at \(X_t\), partitions them into \(k_t\) equal blocks, averages within blocks, selects a measurable Euclidean geometric median \(m_t\) of the block averages, and sets
\[
X_{t+1}=\operatorname{prox}_{\eta_t\phi}(X_t-\eta_t m_t).
\]
It returns \(X_T\), makes no off-current query, retains no auxiliary oracle summary, performs no clipping, and adds no update term outside this prescribed template. Its oracle count is \(\sum_t b_t\), and \(N_G\) optimizes over every schedule allowed here.

Let \(\kappa=L/\mu\) and
\[
M(\varepsilon)=\left\lceil 64\kappa\log(R_0/\varepsilon)\right\rceil.
\]
For \(C\in\{U,G\}\), define \(N_C\) as the infimum over algorithms \(A\) in class \(C\) using at most \(M(\varepsilon)\) proximal calls of the supremum over admissible instances \(P\) of \(A\)'s deterministic maximum oracle-sample count, subject to
\[
\sup_P\left(\mathbb E_P\lVert X_A-x_P^\star\rVert^p\right)^{1/p}
\leq\varepsilon.
\]
Set \(N_C=\infty\) if no such algorithm exists. Study \(\rho=\sigma_L/\mu\) tending to infinity with \(d,p,L,\mu,R_0,\sigma_\star,\varepsilon\), and 64 fixed, and form \(N_U/N_G\) only where \(0<N_G<\infty\).

Use the full instance and algorithm classes of A1-A8, specialized only to d=1, p=3/2, \(L=\mu=R0=\sigma_{*}=1, \varepsilon\)=1/8, and \(\sigma_{L}=\rho\). The comparison remains worst-case over every admissible composite objective and conditional oracle; neither class may receive the minimizer or oracle distribution.

Set d=1, p=3/2, \(L=\mu=R0=\sigma_{*}=1, \varepsilon\)=1/8, and \(\sigma_{L}=\rho\) with \(\rho >\)0.

Compute \(N_{U}\) and \(N_{G}\) using the complete classes and worst-case instance supremum in A6-A8, without restricting the oracle to a selected toy family.

Compare the unnormalized complexities through \(N_{U}(\rho)/N_{G}(\rho\)); equality of polynomial exponents alone is insufficient.

On the fixed ray d=1, p=3/2, \(L=\mu=R0=\sigma_{*}=1, \varepsilon\)=1/8, and \(\sigma_{L}=\rho\), determine whether the unrestricted class U and the precommitted geometric-median proximal class G have asymptotically equal minimax oracle complexities: the admissible ratio domain contains a tail and \(N_{U}(\rho)/N_{G}(\rho\)) tends to 1 as \(\rho\) tends to \(\infty\).
\end{problem}

\clearpage
\section{P058: Intermediate-bias minimax regret}
\label{problem:P058}

\begin{problem}
For (f,b)\(\in F_{L} \times B_{\delta}\), define the fixed oracle \(y_{f,b}(x)\)=f(x)+b(x).

If \(\widehat{x}_{A}(f,b)\) is A's recommendation, define W(A;L,\(\delta)=\sup_{f\in F_{L},b\in B_{\delta}}\){[}f(\(\widehat{x}_{A}(f,b))-\min_{x\in[0,1]}\)f(x){]}.

\[
\mathbb R_{n}(1,L,\delta)=\in f_{A\in Alg_{n}}W(A;L,\delta)
\]

\(V_{n}\)(L,\(\delta\))=min\{L/2,L/(2n)+2\(\delta\)\}.

\(N_{L,\delta}(\varepsilon\))=min\{n\(\geq5:\mathbb R_{n}\)(1,L,\(\delta)\leq2\delta+\varepsilon\)\}, when this set is nonempty.

The domain is {[}0,1{]}\(^{d}\) with the \(\ell\infty\) metric, where d is a positive integer.

The unknown objective f is real-valued and L-Lipschitz on the domain for a known constant L\(>\)0; no convexity, differentiability, or additional regularity is assumed.

The oracle is the fixed map y(x)=f(x)+b(x), where b is chosen adversarially before the interaction and satisfies \(\mid\) b(x)\(\mid \leq \delta\) for every x, for a known \(\delta \geq\)0; no regularity of b is assumed.

The query budget n is a positive integer. An algorithm is deterministic, may choose each query from earlier returned values, uses at most n queries, and then recommends any point of the domain, whether queried or unqueried. Because y is one fixed map, repeated queries at the same point return the same value.

For an algorithm A with recommendation x̂\(_{A}\), its risk is \(\sup_{f,b}\){[}f(x̂\(_{A})- \min_{x}\) f(x){]}, with the supremum taken jointly over all pairs satisfying A2 and A3; \(\mathbb R_{n}\)(d,L,\(\delta\)) is the infimum of this risk over all algorithms satisfying A4.

No gradients, randomized observations, distributional noise assumptions, or side information beyond d, L, \(\delta\), and returned biased values may be used.

One-dimensional deterministic global minimization on {[}0,1{]} under A1-A6. This frontier covers the intermediate-bias region omitted by the established low-bias identity and excludes independently varying errors, queried-point-only recommendations, higher dimensions, and \(\delta \geq\)L/4.

Fix d=1 with metric \(\mid\) x-x'\(\mid\){} on {[}0,1{]}.

The budget n is an arbitrary integer n\(\geq\)5.

The bias radius satisfies L/(4n)\(< \delta <\) L/4.

The oracle is one fixed spatial map y=f+b, so repeated queries receive identical reports.

The final recommendation may be queried or unqueried.

Determine whether, for every L>0, integer n\(\geq\)5, and L/(4n)<\(\delta\)<L/4, the exact identity \(\mathbb R_{n}\)(1,L,\(\delta\))=min{L/2,L/(2n)+2\(\delta\)} holds under one fixed arbitrary bias field and an unrestricted final recommendation.
\end{problem}

\clearpage
\section{P059: Low-bias minimax regret}
\label{problem:P059}

\begin{problem}
For an admissible pair (f,b), the oracle is the fixed map \(y_{f,b}(x)\)=f(x)+b(x).

If \(\widehat{x}_{A}(f,b)\) is A's recommendation after interacting with \(y_{f,b}\), define W(A;L,\(\delta)=\sup_{f\in F_{L},b\in B_{\delta}}\){[}f(\(\widehat{x}_{A}(f,b))-\min_{x\in[0,1]}\)f(x){]}.

\[
\mathbb R_{n}(1,L,\delta)=\in f_{A\in Alg_{n}}W(A;L,\delta)
\]

The midpoint-grid algorithm queries \(q_{i}\)=(2i-1)/(2n), i=1,\ldots,n, and recommends a queried point with minimum reported value, using a fixed deterministic tie rule.

The domain is {[}0,1{]}\(^{d}\) with the \(\ell\infty\) metric, where d is a positive integer.

The unknown objective f is real-valued and L-Lipschitz on the domain for a known constant L\(>\)0; no convexity, differentiability, or additional regularity is assumed.

The oracle is the fixed map y(x)=f(x)+b(x), where b is chosen adversarially before the interaction and satisfies \(\mid\) b(x)\(\mid \leq \delta\) for every x, for a known \(\delta \geq\)0; no regularity of b is assumed.

The query budget n is a positive integer. An algorithm is deterministic, may choose each query from earlier returned values, uses at most n queries, and then recommends any point of the domain, whether queried or unqueried. Because y is one fixed map, repeated queries at the same point return the same value.

For an algorithm A with recommendation x̂\(_{A}\), its risk is \(\sup_{f,b}\){[}f(x̂\(_{A})- \min_{x}\) f(x){]}, with the supremum taken jointly over all pairs satisfying A2 and A3; \(\mathbb R_{n}\)(d,L,\(\delta\)) is the infimum of this risk over all algorithms satisfying A4.

No gradients, randomized observations, distributional noise assumptions, or side information beyond d, L, \(\delta\), and returned biased values may be used.

One-dimensional deterministic global minimization on {[}0,1{]} under A1-A6. The frontier concerns the entire general-budget phase n\(\geq\)5 with 0\(\leq \delta \leq\)L/(4n), not an isolated low-budget cell, independently varying errors, queried-point-only recommendations, higher dimensions, or the saturated regime.

Fix d=1 and use the absolute-value metric on {[}0,1{]}.

The query budget n is an arbitrary finite integer satisfying n\(\geq\)5.

The bias radius satisfies 0\(\leq \delta \leq\)L/(4n).

For each instance, b is one globally defined field on {[}0,1{]}, and repeated queries at one point return the same value.

The recommendation may be any point of {[}0,1{]}, whether queried or unqueried.

Determine whether the exact identity \(\mathbb R_{n}\)(1,L,\(\delta\))=L/(2n)+2\(\delta\) holds simultaneously for every finite integer n\(\geq\)5 and every 0\(\leq \delta \leq\)L/(4n) under one fixed arbitrary bias field and an unrestricted final recommendation.
\end{problem}

\clearpage
\section{P060: A topology exponent for directed gradient tracking}
\label{problem:P060}

\begin{problem}
For I in \(D_{\mathrm{FULL}}(η)\), z\(\star=(\sum_{j=1}^{3} h_{j} r_{j})/(\sum_{j=1}^{3} h_{j}\)).

\[
E_{k}^{\alpha}=(norm(x_{k}-1 z\star)_{2}^{2}+\alpha^{2} norm(y_{k})_{2}^{2})^{(1/2)}
\]

\(\mathbb R_{k}\)(I,\(\alpha\))=sup \(E_{k}^{\alpha/E_{0}} \alpha\) over all tracker-consistent \(x_{0}\) with \(E_{0}^{\alpha} >\)0.

S(I)=\{\(\alpha >0: \sup_{k\geq0} \mathbb R_{k}\)(I,\(\alpha)<+\infty\) and \(\lim_{k \to +\infty} \mathbb R_{k}\)(I,\(\alpha\))=0\}.

\(T_{\varepsilon}\)(I,\(\alpha\))=inf\{K in \{0,1,2,\ldots\}: \(\sup_{k\geq K} \mathbb R_{k}\)(I,\(\alpha)\leq \varepsilon\)\}, with value +\(\infty\) when the set is empty.

\(N_{\mathrm{GT}}(\varepsilon\);3,2,η)=\(\sup_{I \in D_{\mathrm{\mathrm{FULL}}}(η)} \in f_{\alpha \in S(I)} T_{\varepsilon}\)(I,\(\alpha\)), with an empty inner infimum equal to +\(\infty\).

There are n\(\geq\)3 agents with scalar objectives \(f_{i}(z)=h_{i} (z-r_{i})^{2/2}\), where 1\(\leq h_{i} \leq\)2. Their average has the unique minimizer \(z\star\), and \(\nabla\) F(x)=(\(h_{i}(x_{i}-r_{i}))_{i=1} ^{n}\).

For a constant \(\alpha >\)0, the algorithm is \(x_{k+1}=A_{k} x_{k}-\alpha y_{k}\) and \(y_{k+1}=B_{k} y_{k}+\nabla\) F(\(x_{k+1})-\nabla\) F(\(x_{k}\)), with tracker-consistent initialization \(y_{0}=\nabla\) F(\(x_{0}\)).

Each \(A_{k}\) is row-stochastic and each \(B_{k}\) is column-stochastic. Both have positive diagonal entries, and for i≠j, \((A_{k})_{ij} >\)0 if and only if \((B_{k})_{ij} >\)0 if and only if the directed edge j\(\to\)i is present at time k. Every positive entry of either matrix is at least η, where 0\(<\) η\(\leq\)1/2.

For a fixed integer H\(\geq\)1, the union of the directed graphs in every H consecutive times is strongly connected.

An admissible instance I consists of the objective data and matrix schedule satisfying A1-A4, but not an initialization. For each I and \(\alpha, E_{k}^{\alpha}, \mathbb R_{k}, T_{\varepsilon}\), S(I), and \(N_{\mathrm{GT}}\) are defined as in the claim shape, with \(\mathbb R_{k}\) taking the supremum over all tracker-consistent \(x_{0}\) having \(E_{0}^{\alpha} >\)0 and \(N_{\mathrm{GT}}\) taking the supremum over all admissible instances. An instance with S(I) empty makes \(N_{\mathrm{GT}}\) infinite. No finiteness or heterogeneity-control property of the chosen normalization is assumed.

All scalar-quadratic objectives and all infinite matrix schedules satisfying A1-A5 with n=3, H=2, and 0\(<\) η\(\leq\)1/3. Stepsizes are instance-aware, constant, and optimized over the tracker-reachable stability set S(I); no periodicity, common Perron vector, or balancedness is imposed.

Fix n=3 and H=2 but retain every admissible objective, support sequence, and communication weight sequence.

For each instance, optimize over every \(\alpha\) in S(I), with stability and settling evaluated only through tracker-consistent envelopes.

The same finite exponent and comparison constants must work for every sufficiently small η and \(\varepsilon\).

For the anchor-defined push-pull recursion with n=3 and H=2, determine whether the full admissible-class complexity \(N_{\mathrm{GT}}(\varepsilon\);3,2,η) is uniformly comparable to \(η^{(-\beta)}\) log(1/\(\varepsilon\)) for some finite \(\beta \geq\)0 as η and \(\varepsilon\) tend to zero.
\end{problem}

\clearpage
\section{P061: Strong-bias asymptotics for a single jump}
\label{problem:P061}

\begin{problem}
A query at x returns y(x)=f(x)+b(x), with the same f and fixed step b used for the entire transcript.

\(\mathbb R_{N}^{step}\)(L,\(\delta\)) is the infimum over deterministic N-query algorithms of the supremum of f(x̂)-\(\min_{x \in [0,1]}\) f(x) over all instances authorized by A1-A4.

\[
S_{N}(L)=\sup_{\delta\geq0} \mathbb R_{N}^{step}(L,\delta)
\]

The objective f:{[}0,1{]} to the real numbers is L-Lipschitz for a known L\(>\)0.

The bias has the fixed form \(b(x)=c_{-}\) for x\(< \tau\) and \(b(x)=c_{+}\) for x\(\geq \tau\), where \(\tau\) is in (0,1) and \(\mid c_{-} \mid,\mid c_{+} \mid\){} are at most \(\delta\).

A deterministic algorithm chooses N query points adaptively, observes y(x)=f(x)+b(x), and repeated queries receive the same value.

After N queries the algorithm may output any point of {[}0,1{]}, and performance is f(x̂)-\(\min_{x \in [0,1]}\) f(x) under the quantifier order infimum over algorithms followed by supremum over admissible f and b.

One-dimensional L-Lipschitz objectives, one fixed two-level bias with a strictly interior jump, deterministic adaptive algorithms using N positive function-value queries, arbitrary output points, and the worst case taken uniformly over every finite bias bound \(\delta\). This is the strong-bias, uniform-in-\(\delta\) asymptotic and is mathematically distinct from existence of the already established finite-q crossover.

For each N and L, \(S_{N}(L)\) is the supremum of \(\mathbb R_{N}^{step}\)(L,\(\delta\)) over all finite \(\delta \geq0; \delta\) may depend on N inside this supremum.

The asymptotic is over positive integers N tending to \(\infty\), with L\(>\)0 fixed.

Every adversarial oracle uses one \(\tau\) in (0,1), \(b(x)=c_{minus}\) for x\(< \tau\) and \(b(x)=c_{plus}\) for x\(\geq \tau\), and the same \(c_{minus},c_{plus}\) throughout its transcript.

Let \(S_{N}(L)=\sup_{\delta\geq0} \mathbb R_{N}^{step}\)(L,\(\delta\)). Determine whether, for every L>0, the limit of N \(S_{N}(L)\)/L as N tends to \(\infty\) exists and equals 5/2.
\end{problem}

\clearpage
\section{P062: A sharp joint resolution constant}
\label{problem:P062}

\begin{problem}
L=1, \(\sigma=1, \beta\)=r \(\gamma, \varepsilon\)=(2r+1) \(\gamma\), and \(\alpha=\gamma^{\lambda}\).

\(a_{d}\)=d/(d+2).

\(C_{d}=2^{(1-d)} a_{d}^{(-d)} (1-a_{d})^{(-2)}\).

\[
\mathbb R_{d,r,\lambda}(\gamma)=\gamma^{(d+2)} N\star(d,1,r \gamma,1,(2r+1) \gamma,\gamma^{\lambda)/\log(1/\gamma} \lambda)
\]

The dimension d is a positive integer, the domain is {[}0,1{]}\(^{d}\) with the \(\infty\) norm, and f is an unknown real-valued L-Lipschitz function with known L\(>\)0.

At query \(x_{t}\) the oracle returns \(Y_{t}\)=f(\(x_{t}\))+b(\(x_{t})+xi_{t}\), where b is one fixed unknown function throughout the transcript, \(\sup_{x} \mid\) b(x)\(\mid \leq \beta, \beta \geq\)0 is known, and b has no assumed regularity.

For every admissible oracle law, the variables \(xi_{t}\) are independent and centered and satisfy E{[}exp(\(\lambda xi_{t}){]}\leq\)exp(\(\sigma^{2} \lambda^{2/2}\)) for every real \(\lambda\), with known \(\sigma \geq\)0; the guarantee is uniform over all laws satisfying this condition.

The algorithm is deterministic and adaptive, may repeat query points, uses a budget N in \{0,1,2,\ldots\}, and outputs a point in {[}0,1{]}\(^{d}\); when N=0 its output is fixed before observing the oracle.

The budget \(N\star\) is the smallest nonnegative N for which one algorithm, uniformly over every admissible f, fixed b, and noise law, outputs x̂ satisfying f(x̂)-\(\min_{x}\) f(x)\(\leq \varepsilon\) with probability at least 1-\(\alpha\).

The active strip has 0\(< \alpha \leq1/4, 0\leq \beta <\) L/4, \(\gamma=\varepsilon-2 \beta >\)0, and \(\varepsilon=2 \beta+\gamma <\) L/2. The regimes \(\varepsilon \geq\)L/2, \(\varepsilon <\) min\{2 \(\beta\),L/2\}, and all equality endpoints are classified separately and are not included in the active-rate claim.

Deterministic adaptive value-query algorithms under A1-A6 on {[}0,1{]}\(^{d}\) with the \(\infty\) norm. The parameters d, r, and \(\lambda\) are fixed before \(\gamma\) tends to zero. Only \(\gamma\) small enough that \(\varepsilon <\)1/4 and \(\alpha \leq\)1/4 is included. The guarantee remains uniform over every fixed bias field and every independent centered 1-sub-Gaussian noise law.

Set L=1, \(\sigma=1, \beta\)=r \(\gamma, \varepsilon\)=(2r+1) \(\gamma\), and \(\alpha=\gamma^{\lambda}\) for fixed d in \{1,2,\ldots\}, r\(\geq\)0, and \(\lambda >\)0.

Restrict \(\gamma\) to 0\(< \gamma <\) min\{1/{[}4(2r+1){]},\(4^{(-1/\lambda)}\)\}.

For every oracle instance, b is one fixed function throughout the transcript and satisfies \(\sup_{x} \mid\) b(x)\(\mid \leq\)r \(\gamma\).

Fix a positive integer d, a bias ratio r\(\geq\)0, and a confidence-resolution exponent \(\lambda\)>0. On the canonical ray \(L=\sigma=1, \beta\)=r \(\gamma, \varepsilon\)=(2r+1) \(\gamma\), and \(\alpha=\gamma^{\lambda}\), determine whether the exact leading constant of the minimax budget is \(C_{d}\) in the limit \(\gamma\) down to zero, where \(a_{d}\)=d/(d+2) and \(C_{d}=2^{(1-d)} a_{d}^{(-d)} (1-a_{d})^{(-2)}\).
\end{problem}

\clearpage
\section{P063: Weak-only convergence in a variable metric}
\label{problem:P063}

\begin{problem}
\(J_{cMA}:=(I+cMA)^{-1}\).

\(x_{k}\)⇀0 means that ⟨\(x_{k},test_{vector}\)⟩\(\to\)0 for every \(test_{vector} \in\)H.

Failure of strong convergence to 0 means \(limsup_{k\to\infty}\)∥\(x_{k}\)∥\(>\)0, equivalently there are \(\varepsilon >\)0 and infinitely many k with ∥\(x_{k}\)∥\(\geq \varepsilon\).

A frozen replay associated with M and c is \(y_{j+1}=J_{cMA} y_{j}\) with M and c independent of j.

H is a separable infinite-dimensional real Hilbert space.

A:H\(\to\)H is bounded, self-adjoint, positive semidefinite, and maximally monotone, with nonempty infinite-dimensional kernel and nonclosed range.

The iterates are exact and satisfy \(x_{k+1}=J_{c_{k} M_{k} A} x_{k}\) with 0\(< c_{min} \leq c_{k} \leq c_{max} < \infty\).

Every \(M_{k}\) is bounded, self-adjoint, and uniformly positive definite, with mI\(\preccurlyeq M_{k}\preccurlyeq\)MI for fixed 0\(<\) m\(\leq\)M\(< \infty\).

There are \(η_{k} \geq\)0 with ∑\(_{k} η_{k} < \infty\) such that \((1+η_{k})^{-1} M_{k}\preccurlyeq M_{k+1}\) for every k.

There are no errors, relaxation, inertia, anchoring projections, or iteration-dependent operators.

Separable infinite-dimensional real Hilbert spaces; bounded self-adjoint positive semidefinite linear A with infinite-dimensional kernel and nonclosed range; exact unrelaxed proximal iterations under uniformly conditioned metrics and one-sided summable Loewner variation. Closed-range and two-sided-variation recovery questions are excluded from this candidate.

The preconditioning convention is \(J_{cMA}=(I+cMA)^{-1}\); it may not be replaced by J\_\{\(cM^{-1}\)A\}.

The positive witness must satisfy the Loewner inequality for every k with one summable sequence (\(η_{k}\)), not merely on isolated finite blocks.

For every fixed metric M occurring in the schedule, every fixed c in {[}\(c_{min},c_{max}\){]}, and every \(y_{0}\) in H, the replay \(y_{j+1}=J_{cMA} y_{j}\) must be shown to converge strongly to a point of ker(A).

Determine whether there exists an admissible datum satisfying A1–A6 and an initial point \(x_{0}\) such that the exact orbit \(x_{k+1}=J_{c_{kM,kA}} x_{k}\) converges weakly to 0 but not strongly to 0. Any positive witness must verify the global one-sided Loewner condition for the entire infinite schedule and must derive, rather than assume, strong convergence of every associated frozen-metric replay.
\end{problem}

\clearpage
\section{P064: Reciprocal instability of perturbed OMWU}
\label{problem:P064}

\begin{problem}
J0={[}{[}0,1,0,1{]},{[}-1,0,1,0{]},{[}0,-1,0,1{]},{[}-1,0,-1,0{]}{]}, \((E12)_{12}\)=1 with all other entries zero, and \(J_{e}\)=J0-eE12.

\[
\begin{aligned}
q^{(t+1)}&=q^t+\eta J_e
\bigl(2z^t-z^{(t-1)}\bigr),\\
z_i^t&=\tanh(q_i^t/2),
&q^{(-1)}&=q^0,\\
\eta&=\frac{1}{20\sqrt 2}.
\end{aligned}
\]

\(Bad_{e}(q0)\) holds when the corresponding orbit satisfies limsup as t tends to \(\infty\) of ‖\(x^{t}-(1/2,1/2)^{4}\)‖\(_{2} >\)0.

\(\beta(e)\)=inf\{‖q0‖\_\(\infty\): q0 is finite and \(Bad_{e}(q0)\) holds\}, with the infimum of the empty set equal to +\(\infty\).

The interaction graph is finite, connected, and undirected; every player has two pure actions, total payoff is the sum of incident-edge payoffs, and every entry of every perturbed edge-payoff matrix has absolute value at most one.

Write \(z_{i}=x_{i,1}-x_{i,2}\) and let \(g_{i}(z)\) be player i's payoff difference between actions 1 and 2. Euclidean norm is used on z-space and its induced spectral norm on matrices. The zero-sum baseline has \(g_{0}(z)=J_{0z}\) with \(J_{0}\) skew-symmetric, smallest singular value at least a fixed \(\sigma\) greater than zero, and a unique fully mixed Nash equilibrium.

For the perturbed payoff-difference Jacobian J, the defect is \(\delta\)=‖J-\(J_{0}\)‖\(_{2}\). With cost pseudogradient F=-g, the symmetric part of DF has a negative eigenvalue of magnitude at most \(\delta\). Moreover, no positive weights \(w_{i}\) satisfy \(w_{i} J_{ij}=-w_{j} J_{ji}\) on every interaction edge, so the perturbation cannot be removed by positive player-wise payoff rescaling.

The perturbed game has a unique Nash equilibrium \(x\star\), it is fully mixed, and every action probability at \(x\star\) is at least a fixed \(\kappa\) greater than zero.

Players maximize payoff and use \(x_{i,a}^{t+1}=x_{i,a}^{t}\) exp(η(\(2u_{i,a}(x_{-i}^{t})-u_{i,a}(x_{-i}^{t-1})))/\sum_{b} x_{i,b}^{t}\) exp(η(\(2u_{i,b}(x_{-i}^{t})-u_{i,b}(x_{-i}^{t-1}\)))). The history is initialized by \(x^{-1}=x^{0}\) with \(x^{0}\) strictly interior, and 0\(<\) η L\(\leq\)1/10, where L=‖J‖\(_{2}\).

The frontier quantifies how far an initialization must retreat toward the simplex boundary before the explicit one-edge imbalance can defeat last-iterate convergence. It neither repeats the established positive radius for a different fixed orientation nor asks again whether a bad vanishing-defect orientation exists.

Fix J0={[}{[}0,1,0,1{]},{[}-1,0,1,0{]},{[}0,-1,0,1{]},{[}-1,0,-1,0{]}{]} and \(J_{e}\)=J0-eE12 for 0\(<\) e\(<\)1/4, where \((E12)_{12}\)=1 and all other entries vanish.

Set η=1/(20 √(2)) and use \(q^{(t+1)}=q^{t}\)+η \(J_{e}(2z^{t}-z^{(t-1)}), z_{i}^{t}\)=tanh(\(q_{i}^{t}\)/2), and \(q^{(-1)}=q^{0}\).

An orbit is bad when limsup as t tends to \(\infty\) of ‖\(x^{t}-(1/2,1/2)^{4}\)‖\(_{2}\) is positive.

The target is a lower bound on the initialization scale of bad orbits; it does not assert an upper bound or a complete basin classification.

For 0<e<1/4, let \(J_{e}\)=J0-eE12, where E12 has its sole nonzero entry equal to one in row 1, column 2. Let \(\beta(e)\) be the infimum of ‖q0‖\_\(\infty\) over finite initial log-odds vectors whose initialized OMWU orbit is bad, with \(\beta(e)=+\infty\) if no bad orbit exists. Prove or refute that liminf as e decreases to zero of e \(\beta(e)\) is positive.
\end{problem}

\clearpage
\section{P065: Heavy-tail objective rates for clipped proximal SAGA}
\label{problem:P065}

\begin{problem}
\(\Phi(x)=n^{(-1)} \sum_{i} f_{i}\)(x)+r(x), and \(x\star\) is its unique minimizer.

Initialize \(h_{i,0}\)=0; form \(\bar{h}_{t}\), sample \(I_{t}\) from \(p_{t}\), query \(Z_{t}\), set \(Y_{t}=\bar{h}_{t}+{[}Z_{t}-h_{I_{t},t}\){]}/{[}n \(p_{I_{t},t}{]}, v_{t}=clip_{\tau_{t}}(Y_{t}), x_{t+1}=prox_{\eta_{t} r}(x_{t}-\eta_{t} v_{t}\)), and refresh only \(h_{I_{t},t}\) with \(clip_{\tau_{t}}(Z_{t}\)).

\[
\begin{aligned}
\eta_t&=\frac{2}{\mu(t+t_0)},
&t_0&\geq\frac{8L_{\max}}{\mu},\\
\tau_t&=C(q,n,p_{\min},\mu,L_{\max},(\sigma_i))\\
&\quad\cdot(t+t_0)^{1/q}.
\end{aligned}
\]

\[
\bar{x}_{T}=2[T(T+1)]^{(-1)} \sum_{t=1}^{T} t x_{t}
\]

\[
\begin{aligned}
\mathbb R_C(q,\mathrm{instance})
={}&\sup_p\sup_{T\geq1}T^{2(q-1)/q}\\
&\quad\cdot\mathbb E\bigl[
\Phi(\bar x_T)-\Phi(x_\star)\bigr].
\end{aligned}
\]

With \(H_{t}=n^{(-1)} \sum_{i}\) ‖\(h_{i,t}-\nabla f_{i}(x_{t}\))‖\(^{q}\), define \(Q_{t}=2^{(q-1)}\)(‖\(\nabla\) F(\(x_{t}\))‖\(^{q}+2^{(2q-1)} (n p_{min})^{(1-q)}(H_{t}+n^{(-1)} \sum_{i} \sigma_{i}^{q}\))).

The objective is \(\Phi(x)=n^{-1} \sum_{i=1}^{n} f_{i}(x)\)+r(x) on a finite-dimensional Euclidean space, and \(x_{0}\) belongs to dom(r).

Each \(f_{i}\) is convex with \(L_{i}\)-Lipschitz gradient, their average is \(\mu\)-strongly convex for \(\mu >\)0, and r is proper, closed, convex, and has an exactly evaluable proximal map.

A query to component i at x returns \(G_{i}(x)=\nabla f_{i}(x)+xi_{i}\). For one fixed q in (1,2), E{[}\(xi_{i}\) given the pre-query history and \(I_{t}\)=i{]}=0 and E{[}‖\(xi_{i}\)‖\(^{q}\) given that information{]}\(\leq \sigma_{i}^{q}\); no second moment is assumed.

Before query t, \(p_{t}\) is measurable with respect to the pre-query history, 0\(< p_{min} \leq\)1/n, \(p_{i,t} \geq p_{min}\) for every i, and \(I_{t}\) has conditional distribution \(p_{t}\). The target is uniform over all such predictable rules, including \(p_{i,t}\)=1/n.

Fresh oracle noise obeys A3 conditional on the past and selected component at every queried point.

The algorithm is exactly the recursion in \(central_{object}\). Importance weighting occurs before clipping, clipping occurs only in the displayed estimator and table updates, and the same noisy query \(Z_{t}\) is shared by both uses.

The stepsizes are \(\eta_{t}\)=min\{1/(\(4L_{max}),2/{[}\mu\)(t+\(t_{0}\)){]}\} with \(t_{0} \geq 8L_{max}/\mu\). An admissible threshold rule C maps only (q,n,\(p_{min},\mu,L_{max},(\sigma_{i})_{i=1}^{n}\)) to a finite \(c_{\tau}\) in (0,\(\infty\)), before any oracle draws, and \(\tau_{t}=c_{\tau} (t+t_{0})^{1/q}\). The same selected \(c_{\tau}\) is used when comparing adaptive and uniform sampling. Complexity counts every noisy component query.

Use exactly the A1-A7 objective, oracle, recursion, schedules, and predictable sampling class. This frontier concerns the weighted composite objective gap, not the already established pointwise squared-distance and table-energy stability capability.

For T\(\geq\)1, use \(\bar{x}_{T}\)=2{[}T(T+1){]}\(^{(-1)} \sum_{t=1}^{T}\) t \(x_{t}\) and no alternative averaging scheme.

The threshold rule may depend only on (q,n,\(p_{min},\mu,L_{max},(\sigma_{i}\))) and must return a finite \(c_{\tau} >\)0 independently of \(t_{0}\), the objective offsets, and the sampling history.

The positive claim is uniform over every A4-predictable probability process, including uniform sampling.

Uniform failure may use a different predictable rule and horizon for each requested bound M, but must use one fixed admissible instance for each threshold rule C.

Determine whether the exact A1-A7 recursion admits a strictly positive parameter-only threshold rule whose linearly weighted average achieves the heavy-tail expected composite-objective rate uniformly over all admissible predictable sampling rules, or whether recurrent clipping and recycled-table bias create the exact quantifier-level obstruction.
\end{problem}

\clearpage
\section{P066: Sharpness of the realized chord bound}
\label{problem:P066}

\begin{problem}
\[
D_{\phi_{k}}(x,y)=\phi_{k}(x)-\phi_{k}(y)-< \nabla \phi_{k}(y),x-y>
\]

\[
P_{k}(y)=\operatorname*{arg\,\min}_{x \in N}\{f(x)+D_{\phi_{k}}(x,y)\},
\]
\[
\begin{aligned}
e_k(y)&=f(P_k(y))+D_{\phi_k}(P_k(y),y),\\
G_k(y)&=\nabla e_k(y),\\
V_k(y)&=\nabla\phi_k(y)-\nabla\phi_k(P_k(y)).
\end{aligned}
\]

For (k,y) in \(\mathbb R_{c}\), put \(x=P_{k}(y)\), d=y-x, q=d/‖d‖, and \(A_{c,k,y}(t)=nabla^{2} \phi_{k}(x+td)\)q.

Define
\[
\begin{aligned}
K_c=\sup_{\substack{(k,y)\in\mathbb R_c\\t\in[0,1]}}
\max\biggl\{&
\frac{\|A_{c,k,y}(t)\|}{\|A_{c,k,y}(1)\|},\\
&\frac{\|A_{c,k,y}(1)\|}{\|A_{c,k,y}(t)\|}\biggr\},
\end{aligned}
\]
and
\[
\beta_c=\inf_{\substack{(k,y)\in\mathbb R_c\\t\in[0,1]}}
\frac{\langle q,A_{c,k,y}(t)\rangle}{\|A_{c,k,y}(t)\|}.
\]
For an empty \(\mathbb R_c\), set \(K_c=\beta_c=1\).

\(\rho_{c}(k,y)\)=‖\(V_{k}(y)\)‖/‖\(G_{k}(y)\)‖=‖\(\in t_{0}^{1} A_{c,k,y}(t)\)dt‖/‖\(A_{c,k,y}(1)\)‖ for every (k,y) in \(\mathbb R_{c}\).

A pair s=((\(k_{j}),(y_{j}\))) belongs to S(c) exactly when the \(k_{j}\) are
pairwise distinct, \(y_{j}\) tends to 0, \(P_{k_{j}}(y_{j}\)) tends to 0, and there
exists j0 such that (\(k_{j},y_{j}\)) belongs to \(\mathbb R_{c}\) for every j\(\geq\)j0.

\[
E_{cell}=\inf\{liminf_{j \to \infty} \rho_{c}(k_{j},y_{j}): c in C_{cell} and
\]
s=((\(k_{j}),(y_{j}\))) in S(c)\}.

The function f is proper and lower semicontinuous and is relatively prox-regular at a
stationary pair (\(\bar{x}\),0) with respect to a fixed reference Legendre kernel on a
neighborhood of \(\bar{x}\).

There are convex neighborhoods V contained in the interior of N, with compact closure of N
contained in a common open domain U, and a fixed \(\lambda >\)0 used for every kernel.

Each \(\phi_{k}\) is a C2 very strictly convex Legendre function on U; the kernels share their
domain but may repeat, reverse, recover, or have unbounded ambient condition numbers as k
varies.

For every k and y in V, the minimization of f(x)+\(\lambda^{-1} D_{\phi,k}(x,y)\) over x in N has a
unique minimizer \(P_{k}(y)\) in the interior of N, and the resulting localized envelope \(e_{k}\) is C1
on V.

All gradients, Hessians, and residual norms are measured in one fixed finite-dimensional
coordinate system; affine changes to \(\phi_{k}\) are identified because they do not change its
Bregman divergence.

A fixed finite-dimensional sharpness cell inside the target anchor. Each admissible sequence uses
one normalized quadratic anchored instance and one fixed witnessing pair (r,L), although different
\(\varepsilon\)-level near-extremizers may use different tuples. Only actual localized proximal chords are
used. The question is whether Hessian integrability, normalization at the stationary point, and
simultaneous control of all realized chords in the tuple's fixed local window force a uniform strict
improvement over 1/4.

For each cell tuple c=(I,r,L), all active constants and admissible sequences
use the particular bundle \(\mathbb R_{c}\) consisting of pairs (k,y) with \(x=P_{k}(y)\),
‖x‖\(\leq\)r, ‖y‖\(\leq\)r, 0\(<\)‖y-x‖, and ‖\(V_{k}(y)\)‖\(\leq\)L.

The sharpness cell is \(K_{c} \leq\)2 and \(\beta_{c} \geq\)1/2, where \(K_{c}\) controls radial action
distortion and \(\beta_{c}\) controls the longitudinal fraction \(<\) q,A(t)\(>\)/‖A(t)‖ on
\(\mathbb R_{c}\).

Every instance component I must use n=3, \(\bar{x}=0, \lambda\)=1,
f(x)=‖x‖\(^{2/2}, U=\mathbb R^{3}\), and \(nabla^{2} \phi_{k}(0)\)=I for every k.

For each \(\varepsilon\)-level witness, the entire shrinking sequence must lie in one
tuple \(c_{\varepsilon}=(I_{\varepsilon},r_{\varepsilon},L_{\varepsilon}\)), so its objective, dimension,
localization, radius, and subgradient window do not change with the sequence
index. Different positive \(\varepsilon\) values may use different tuples.

Restrict to normalized three-dimensional quadratic instances inside A1-A5: \(\bar{x}=0, \lambda\)=1,
f(x)=‖x‖\(^{2/2}, U=\mathbb R^{3}\), and \(nabla^{2} \phi_{k}(0)\)=I for every k. Bind each cell as a tuple c=(I,r,L),
where I is one fixed anchored instance and r,L are fixed witnesses for the realized-bundle bounds
\(K_{c} \leq\)2 and \(\beta_{c} \geq\)1/2. The elementary projection bound gives \(\rho_{k}(y)\)=‖\(V_{k}(y)\)‖/‖\(G_{k}(y)\)‖\(\geq\)1/4 on
\(\mathbb R_{c}\). Determine whether 1/4 is the class infimum over tuple-bound admissible shrinking sequences: for
every \(\varepsilon\)>0, may one choose a single tuple \(c_{\varepsilon}\) and a sequence eventually contained in its
fixed \(\mathbb R_{c,\varepsilon}\) whose liminf calibration ratio is at most 1/4+\(\varepsilon\)?
\end{problem}

\clearpage
\section{P067: Tail-adaptive finite-sum optimization}
\label{problem:P067}

\begin{problem}
\[
\begin{aligned}
\Phi(x)&=\frac1n\sum_i f_i(x)+g(x),\\
\Phi_\star&=\min_x\Phi(x),&
\Delta_0&=\Phi(x_0)-\Phi_\star,
\end{aligned}
\]
and \(\kappa=L/\mu\).

For p in P, s(p) is the common value \(\sigma_{1}(p)\)=\ldots=\(\sigma_{n}(p)\).

\[
\begin{aligned}
B_{\mathrm{bal}}(I,T,p,C_0)
={}&C_0\exp\left(-\frac{T}{C_0(n+\kappa)}\right)\Delta_0\\
&+\frac{C_0}{\mu}s(p)^2
\frac{\log^2(2+T)}{T^{2(p-1)/p}}.
\end{aligned}
\]

In Euclidean space, \(\Phi(x)=n^{(-1)} \sum_{i} f_{i}\)(x)+g(x); every \(f_{i}\) is convex with L-Lipschitz gradient, the smooth average is \(\mu\)-strongly convex with \(\mu >\)0, g is proper closed convex with an exact Euclidean proximal map, and \(\Phi\) has a minimizer \(x\star\).

A query of component i at any history-measurable point x returns a fresh vector \(G_{i}(x)\) with conditional mean \(\nabla f_{i}(x)\). For p in (1,2{]}, \(\sigma_{i}(p)\) is the smallest uniform conditional Lp radius over all query points and all finite admissible prior query histories, independent of any candidate algorithm. Let P be the nonempty set of p for which every \(\sigma_{i}(p)\) is finite, and permit \(\sigma_{i}(p)\)=0. Oracle innovations are conditionally independent across calls given their query histories.

An admissible rule chooses before each call a strictly positive predictable distribution on the n components, samples one component, may retain finite state, and may apply exact proximal operations. It may use n, L, and \(\mu\) but is not given p, P, any \(\sigma_{i}(p)\), the final horizon, or a scale-dependent probability floor.

Every fresh component response is charged, including responses used for allocation statistics or stored state. For each T\(\geq1, \widehat{x}_{T}\) is measurable with respect to the first T charged responses and the rule's internal randomness.

The initial point \(x_{0}\) has finite \(Delta_{0}=\Phi(x_{0})-Phi_{*}\). Expectations are over oracle, component-sampling, and internal randomness.

One-dimensional composite finite-sum instances satisfying exactly A1-A5, with arbitrary finite n\(\geq\)2 and equal canonical component radii at every supported moment order. Components may have different objectives and oracle laws. The rule may use n, L, \(\mu\), and exact proximal operations but is not given p, P, the common scales, or T. The displayed bound is the sole target.

For every p in P there is a finite s(p)\(\geq\)0 such that \(\sigma_{i}(p)\)=s(p) for every component i.

The number of components satisfies n\(\geq\)2, and no equality or common-minimizer condition is imposed on the component objectives.

The same rule \(\mathbb R\) and constant C0 must work for all scoped instances, supported orders, and charged horizons without a scale-dependent probability floor.

The stochastic comparator remains the raw-seed expression; under scale balance it equals s\((p)^{2} log^{2}\)(2+T)/(\(\mu T^{2(p-1)/p}\)).

Decide whether there exist one centralized anytime stateful proximal finite-sum rule \(\mathbb R\) and one universal numerical constant C0 \(\geq\) 1 such that, for every scoped one-dimensional instance satisfying A1-A5 with n \(\geq\) 2 and \(\sigma_{1}(p)=...=\sigma_{n}(p)\) for every p in P, every integer T \(\geq\) 1, and every p in P, its output \(\widehat{x}_{T}\) satisfies E[\(\Phi(\widehat{x}_{T})-Phi_{*}] \leq\) C0 exp(-T/(C0(n+\(\kappa))) Delta_{0}\) + (C0/\(\mu)((\sum_{i} \sigma_{i}\)(p))/n)\(^{2} log^{2}(2+T)/T^{2(p-1)/p}\).
\end{problem}

\clearpage
\section{P068: Exact cross-kernel comparison constants}
\label{problem:P068}

\begin{problem}
\(P_{\phi}^{f}(y)\) is the unique interior localized branch from A4 and
\[
g_{\phi}^{f}(y)=\lambda^{-1} \phi'\,'(y)(y-P_{\phi}^{f}(y))
\]

For \(h=\phi\)-ψ, \(\chi_{\mathrm{UW}}(h)\)=sup\{\(\mid\) h'(x)-h'(z)\(\mid\): x in U, z in W\} and
\(\eta_{\mathrm{UW}}(h)\)=sup\{\(\mid\) h'\,'(x)(x-z)\(\mid\): x in U, z in W\}.

\(C_{P}\) is the supremum of \(\mid P_{\phi}^{f}(y)-P_{psi}^{f}(y)\mid/\chi_{\mathrm{UW}}(\phi\)-ψ), \(C_{g}\) is the
supremum of \(\mid g_{\phi}^{f}(y)-g_{psi}^{f}(y)\mid/{[}\chi_{\mathrm{UW}}(\phi\)-ψ)+\(\eta_{\mathrm{UW}}(\phi\)-ψ){]}, and
\(C_{g}^{\chi}\) uses \(\chi_{\mathrm{UW}}(\phi\)-ψ) alone; every supremum ranges over ordered
\(\phi\),ψ in K, f in F, and y in U'.

For \(\rho\) in K define \(A_{\rho}(u,v)={[}\rho\)'(u)-\(\rho\)'(v){]}/(u-v) when u≠v and
\(A_{\rho}(v,v)=\rho\)'\,'(v). Set a\(\star(u,v)=\min_{\rho \in K} A_{\rho}(u,v)\).

For an ordered pair \(\phi\),ψ, let \(h=\phi\)-ψ, d=u-v, c=h'(y)-h'(v), and
\(B_{\phi}(u,v)=A_{\phi}(u,v)-\lambda\) r a\(\star(u,v)\). The objective-free feasible
set \(\mathbb R\) consists of tuples (\(\phi\),ψ,y,u,v) in K x K x U' x W x W
satisfying c d\(\geq B_{\phi} (u,v)d^{2}\).

\(\mathbb R_{P}\) is the supremum over \(\mathbb R\) of \(\mid\) u-v\(\mid/\chi_{\mathrm{UW}}(h). \mathbb R_{g}\) is the supremum over \(\mathbb R\)
of \(\mid-\phi\)'`(y)(u-v)+h''(y)(y-v)\(\mid/{[}\lambda(\chi_{\mathrm{UW}}\)(h)+\(\eta_{\mathrm{UW}}(h)\)){]}, with the
fixed quotient convention.

Fix finite r and \(\mathbb R\), constants 0\(<\) a0\(\leq\)a1 and 0\(<\) b0\(\leq\)b1, and the geometric data in A2-A4. The
class F is defined extensionally as all proper lower-semicontinuous objectives equipped with
a reference limiting-subgradient pair that satisfy A3-A4; the minimax suprema range over
this entire class.

The admissible kernels form a fixed nonempty family, chosen before F, of C2 Legendre
functions on a common open neighborhood V. Modulo affine functions, the family is compact in
the local C2 topology of uniform gradient and Hessian convergence on compact subsets of V,
satisfies m I\(\leq\)Hess \(\phi \leq\)M I, and has a uniform local Lipschitz bound on its Hessians.

Every f in F is relatively prox-regular at its reference pair with respect to every
admissible kernel on common primal, dual, and function-value-attentive neighborhoods, with
modulus at most r. The proximal parameter \(\lambda\) is fixed and \(\kappa=1-\lambda\) r\(>\)0; no
asymptotic claim as \(\kappa\) tends to zero is authorized.

There are center neighborhoods U' and U with closure(U') contained in U and closure(U)
contained in V, and a fixed compact W contained in V, such that for every authorized f, y in
U, and kernel \(\phi\) there is a unique interior localized proximal selection \(P_{\phi}^{f}(y)\) in W
satisfying the unconstrained optimality inclusion. Every corresponding primal-dual graph
pair lies in all common neighborhoods required by A3, and sup ‖y-\(P_{\phi}^{f}(y)\)‖ over the
authorized indices is at most \(\mathbb R\). No global-minimizer assertion is included.

Kernel perturbations are identified modulo affine functions. For \(h=\phi\)-ψ define
\[
\begin{aligned}
\chi_{\mathrm{UW}}(h)
&=\sup_{(y,u)\in U\times W}
\lVert\nabla h(y)-\nabla h(u)\rVert,\\
\eta_{\mathrm{UW}}(h)
&=\sup_{(y,u)\in U\times W}
\lVert\nabla^2h(y)(y-u)\rVert.
\end{aligned}
\]
These discrepancies are fixed independently of f and its proximal
branch.

The fixed geometry is nonvacuously compatible with the probe family: F contains every
kernel-independent affine-quadratic probe \(f_{a,b}(x)=<\) a,x\(>\)+(b/2)‖x‖\(^{2}\) with a0\(\leq\)‖a‖\(\leq\)a1
and b in \{0\} union {[}b0,b1{]}. For every such probe, center, and admissible kernel, its actual
localized branch lies in one fixed compact W0 contained in the interior of W and its graph
pair lies in the common neighborhoods from A3. Singleton indicators may be used as boundary
tests when they satisfy A3-A4, but they are not the only admissible objectives.

One-dimensional fixed-margin classification for one explicit compact two-parameter kernel family,
fixed nested center intervals, one compact localization interval, and the full extensional objective
class authorized by A1-A6. The problem concerns exactly two finite sharp constants and one
chord-only divergence statement; it makes no pointwise-necessity, weakest-topology, or
vanishing-margin claim.

Set \(\lambda\)=1/2, r=1/2, \(\kappa\)=3/4, V=(-2,2), U=(-1/4,1/4), U'=(-1/8,1/8), W={[}-1,1{]},
W0={[}-3/4,3/4{]}, \(\mathbb R\)=1, a0=1/4, a1=1/2, b0=1/4, and b1=1/2.

The admissible family is K=\{\(\phi_{q,e}\): q in {[}1,3/2{]}, e in {[}0,1/4{]}\}, where \(\phi_{q,0}(x)\)=q
\(x^{2/2}\) and \(\phi_{q,e}(x)\)=q \(x^{2/2}+(e^{3/8}\))sin(x/e) for e\(>\)0.

The objective class remains the entire extensional class F specified by A1; it is not
replaced by the affine-quadratic probes or by a hand-picked witness subclass.

All normalized quotients use the A5 convention: 0/0 is zero and a positive numerator over a
zero denominator is positive \(\infty\).

Specialize A1-A6 to the one-dimensional compact kernel family \(\phi_{q,e}(x)\)=q \(x^{2/2}+(e^{3/8}\))sin(x/e),
with the perturbation defined as zero at e=0, q in [1,3/2], and e in [0,1/4]. Determine whether the
exact proximal and combined ordinary-gradient minimax constants equal the objective-free two-point
relaxations \(\mathbb R_{P}\) and \(\mathbb R_{g}\) forced by simultaneous relative hypomonotonicity. The same specialization
must retain the contrasting classification \(C_{g}^{\chi}=+\infty\) for the chord-only ordinary-gradient
denominator.
\end{problem}

\clearpage
\section{P069: The small-observation-noise horizon}
\label{problem:P069}

\begin{problem}
For a reached history \(h_{k}\) under the past receding-horizon law, define \(B_{k}\)(\(h_{k}\))=x0 plus the integral from 0 to \(t_{k}\) of the realized past control. The two mode-conditioned current states associated with that same history are \(x_{k}^{\sigma}\)(\(h_{k}\))=\(B_{k}\)(\(h_{k}\))+\(\sigma\) a \(t_{k}\).

Let \(pi_{k}^{\sigma}\)(\(h_{k}\)) be the fixed regular conditional probability of Θ=\(\sigma\) a given \(F_{k}\)=\(h_{k}\), so \(pi_{k}^{-}\) plus \(pi_{k}^{+}\) equals one.

For an admissible H-window observation-feedback continuation η, let \(P_{k,h_{k}}^{\sigma,η}\) be the law of its unique strong continuation on {[}\(t_{k}\),\(t_{k}\)+H{]}, initialized at \(x_{k}^{\sigma}\)(\(h_{k}\)), under Θ=\(\sigma\) a and the observation equation with noise intensity \(\nu\).

Define \(G_{k}\)(\(h_{k}\),η) as the sum over \(\sigma\) in \{-1,+1\} of \(pi_{k}^{\sigma}\)(\(h_{k}\)) times \(P_{k,h_{k}}^{\sigma,η}\)(\(X_{s}\) belongs to K for every s in {[}\(t_{k}\),\(t_{k}\)+H{]}).

Define \(C_{k}\)(\(h_{k}\))=\{η: η is an admissible H-window continuation and \(G_{k}\)(\(h_{k}\),η)\(\geq\)1-\(\varepsilon\)\}. Feasibility means \(C_{k}\)(\(h_{k}\)) is nonempty; no supremum over η is assumed to be attained.

An admissible receding-horizon law measurably selects \(\eta_{k}\)(\(h_{k}\)) in \(C_{k}\)(\(h_{k}\)) whenever \(C_{k}\)(\(h_{k}\)) is nonempty, applies \(\eta_{k}\) on {[}\(t_{k}\),\(t_{k+1}\)), and reoptimizes at \(t_{k+1}\); after the first empty \(C_{k}\) it uses the fixed fallback from A5.

For a receding-horizon law, \(\tau_{K}\) is the first time X leaves K and \(\tau_{F}\)=inf\{\(t_{k}\):\(C_{k}\)(\(h_{k}\)) is empty\}, with the conventions in A4-A5.

For \(\nu >\)0, let \(\mathbb R_{\nu}\) be the set of T\(\geq\)0 for which there exists an admissible receding-horizon law satisfying P(min(\(\tau_{K}\),\(\tau_{F}\))\(>\) T)\(\geq\)1-\(\alpha\). Define \(T_{\nu}\)=supremum(\(\mathbb R_{\nu}\) union \{0\}); apply the zero convention of A5 when no law can start.

For \(\sigma\) in \{-1,+1\}, define \(\tau_{\sigma}\)=(1-\(\sigma\) x0)/(a-U). This is the latest state-exit time attainable when \(\sigma\) is known initially, achieved by constant control -\(\sigma\) U.

Define \(ell_{\sigma}\)=Δ times (floor((\(\tau_{\sigma}\)-H)/Δ)+1). Under grid nonresonance, \(ell_{\sigma}\) is the first sampling time \(t_{k}\) at which \(t_{k}\)+H\(> \tau_{\sigma}\).

Define \(T_{FI}\) as the supremum of T\(\geq\)0 such that p0 times 1\{ell\_\(\to\)T\} plus (1-p0) times 1\{\(ell_{+} >\) T\} is at least 1-\(\alpha\).

For one common observation-feedback functional evaluated on the same observation path, the mode-conditioned observation-drift difference is 2at. Its deterministic likelihood-information clock is \(I_{\nu}\)(t)=\(4a^{2} t^{3}\)/(\(3nu^{2}\)), independently of the common controlled midpoint.

The hidden drift Θ takes values -a and a, where a\(>\)0, with prior probabilities p0 and 1-p0 for some p0 in (0,1).

For a fixed known x0 in (-1,1), the state satisfies \(dX_{t}\)=(Θ+\(u_{t}\))dt and \(\mid u_{t} \mid \leq\)U for a fixed U\(>\)0; x0 is an argument of the maximal-horizon value.

The observation satisfies \(dY_{t}\)=\(X_{t}\) dt+\(\nu dV_{t}\). Admissible controls are bounded nonanticipative Borel functionals of the observed path for which the coupled state-observation equations have a unique strong closed-loop solution under each drift mode. The regimes \(\nu >\)0 and \(\nu\)=0 are treated as distinct information structures.

The hard hidden-state constraint is K={[}-1,1{]}, and \(\tau_{K}\)=inf\{t\(\geq\)0:\(X_{t}\) is not in K\}, with inf of the empty set equal to \(\infty\).

Fix H\(>\) Δ\(>\)0, \(\varepsilon\) in (0,1/2), and \(\alpha\) in (0,1). At \(t_{k}\), \(C_{k}\) consists of admissible future observation-feedback controls satisfying the conditional mixture path-safety constraint on {[}\(t_{k}\),\(t_{k}\)+H{]}. A receding-horizon law is a measurable history-dependent selection from \(C_{k}\) whenever it is nonempty, applies the selected control for Δ time, and then reoptimizes; after \(C_{k}\) first becomes empty it follows a prescribed admissible fallback so that the closed loop and both stopping times remain defined. Set \(\tau_{F}\)=inf\{\(t_{k}\):\(C_{k}\) is empty\}, and set T\(\star\)=0 when \(C_{0}\) is empty or no admissible selection starts at time zero.

Assumptions A1-A5 with fixed a\(>\)0, 0\(<\) U\(<\) a, x0 in (-1,1), p0 in (0,1), H\(>\) Δ\(>\)0, \(\varepsilon\) in (0,1/2), and \(\alpha\) in (0,1). For each \(\nu >\)0, conditional probabilities are fixed regular conditional versions on canonical observation-path space. The safety event includes both endpoints of each prediction window. Every \(T_{\nu}\) uses the same admissibility, measurable-selection, fallback, and strict stopping-time conventions. \(T_{FI}\) is a counterfactual benchmark with the mode known initially; it does not identify the distinct \(\nu\)=0 information structure with a positive-noise model.

Require 0\(<\) U\(<\) a.

For every \(\sigma\) in \{-1,+1\}, \(\tau_{\sigma}\)=(1-\(\sigma\) x0)/(a-U) is strictly greater than H.

For every \(\sigma\) in \{-1,+1\}, (\(\tau_{\sigma}\)-H)/Δ is not an integer.

Α differs from p0 and from 1-p0.

Use fixed regular conditional posterior and continuation laws on canonical observation-path space, evaluated on reached sampled histories outside one common null set.

Fix all parameters except the positive observation-noise intensity \(\nu\) in the binary-drift model, assume 0<U<a, and define the sampled feasible sets \(C_{k}\) by the explicit conditional mixture path-safety predicate below. Let \(T_{\nu}\) be the resulting maximal reliable horizon and let \(T_{FI}\) be the full-information grid benchmark determined by the two modewise first prediction-infeasibility times. Prove or refute that \(T_{\nu}\) converges to \(T_{FI}\) as \(\nu\) decreases to zero.
\end{problem}

\clearpage
\section{P070: Weak frustration and optimistic recurrence}
\label{problem:P070}

\begin{problem}
For \(\varepsilon >\)0 and z in {[}-1,1{]}\(^{3}\), set \(d_{\varepsilon}\)(z)=\(J_{\varepsilon}\) z, where \(J_{\varepsilon}\)={[}{[}0,1,-(1+\(\varepsilon\)){]},{[}-1,0,1{]},{[}1,-1,0{]}{]}.

Set \(\eta_{t}\)=1/(64 √(t+1)), \(z_{m}^{t}\)=tanh(\(q_{m}^{t}\)/2) coordinatewise, and \(q_{m}^{(t+1)}\)=\(q_{m}^{t}\)+\(\eta_{t}\)(\(2d_{\varepsilon,m}\)(\(z_{m}^{t}\))-\(d_{\varepsilon,m}\)(\(z_{m}^{(t-1)}\))).

\(N_{\varepsilon}\) consists of z in {[}-1,1{]}\(^{3}\) such that \(d_{i}\)=0 when \(\mid z_{i} \mid <\)1, \(d_{i} \geq\)0 when \(z_{i}\)=1, and \(d_{i} \leq\)0 when \(z_{i}\)=-1. For every \(\varepsilon >\)0 in scope, these conditions imply \(N_{\varepsilon}\)=\{0\}.

Acc(\(z_{m}\)) is the set of limits of subsequences \(z_{m}^{(t_{k})}\) with \(t_{k}\) tending to \(\infty\).

Writing \(d_{i}^{t}\)=\((J_{\varepsilon,m} z_{m}^{t})_{i}\), define \(D_{i}\),+(T)=sum from t=0 to T-1 of \((1-z_{i}^{t})d_{i}^{t}\)/2 and \(D_{i}\),-(T)=sum from t=0 to T-1 of -\((1+z_{i}^{t})d_{i}^{t}\)/2. Set \(\mathbb R_{i}\)(T)=max(0,\(D_{i}\),+(T),\(D_{i}\),-(T)).

There are three players on the triangle graph, each with two pure actions represented by a centered mixed coordinate \(z_{i}\) in (-1,1).

Player i's payoff is the sum of its incident bilinear edge payoffs, with centered payoff-difference operator \(J_{\varepsilon}\)={[}{[}0,1,-(1+\(\varepsilon\)){]},{[}-1,0,1{]},{[}1,-1,0{]}{]}, under a common normalization keeping all pure payoffs bounded.

The admissible parameter range is 0\(\leq \varepsilon \leq\)1/100. The primary target quantifies existentially over a positive sequence \(\varepsilon_{m}\) decreasing to zero; \(\varepsilon\)=0 is used only as a balanced comparison cell.

All players use the displayed deterministic full-information optimistic multiplicative-weights log-odds recurrence with the fixed schedule \(\eta_{t}\)=1/(64 √(t+1)) for every integer t\(\geq\)0.

For each tested positive \(\varepsilon\), the two initial log-odds vectors \(q^{(-1)}\) and \(q^{0}\) are finite and may depend on \(\varepsilon\), so both initial strategy profiles are strictly interior; no single history is required to work for all \(\varepsilon\).

External regret compares each player's realized expected payoff with that of each fixed pure action against the same sequence of opponents' mixed profiles, and vanishing regret means the positive part of the larger of these two cumulative payoff differences is o(T).

For 0\(< \varepsilon \leq\)1/100, use exactly \(J_{\varepsilon}\)={[}{[}0,1,-(1+\(\varepsilon\)){]},{[}-1,0,1{]},{[}1,-1,0{]}{]}, \(\eta_{t}\)=1/(64 √(t+1)), \(z_{i}^{t}\)=tanh(\(q_{i}^{t}\)/2), and \(q_{i}^{(t+1)}\)=\(q_{i}^{t}\)+\(\eta_{t}\)(2\((J_{\varepsilon} z^{t})_{i}\)-(\(J_{\varepsilon} z^{(t-1)}\))\(_{i}\)). Histories \(q^{(-1)}\),\(q^{0}\) are finite and may depend on \(\varepsilon\). The balanced endpoint and deleted-edge path are diagnostic cells only.

The primary target requires a strictly positive sequence \(\varepsilon_{m}\) decreasing to zero, with 0\(< \varepsilon_{m} \leq\)1/100.

Each \(\varepsilon_{m}\) may have its own finite pair \(q_{m}^{(-1)}\),\(q_{m}^{0}\) in \(\mathbb R^{3}\).

For every positive \(\varepsilon\) in scope, distance is measured from the complete Nash set, which equals the singleton \{0\} in centered coordinates.

Two accumulation points must be witnessed by two separated infinite subsequences of exact iterates.

Unweighted regret must be calculated against both pure actions for the exact trajectory.

Determine whether arbitrarily small positive frustration in the specified three-player binary-action triangle admits parameter-dependent finite interior histories whose exact diminishing-step optimistic trajectories remain recurrent away from Nash equilibrium while every player has vanishing unweighted external regret.
\end{problem}

\clearpage
\section{P071: Attainment of the viable drift energy}
\label{problem:P071}

\begin{problem}
For the solution driven by v and started at x0, \(\tau_{v}\) is the infimum of times t\(\geq\)0 for which \(X_{t}\) is not in (0,1).

J(T,x0;v) is the expectation of the integral of v\((X_{t})^{2}\) from 0 to T wedge \(\tau_{v}\) when the probability of \(\tau_{v} >\) T is one, and is \(\infty\) otherwise.

\(A_{T}(x0)\) is the set of v in V for which J(T,x0;v) is finite.

M\(\star(T,x0)\) is the infimum of J(T,x0;v) over v in \(A_{T}(x0)\), with value \(\infty\) only when \(A_{T}(x0)\) is empty.

The state space before exit is (0,1), T is fixed with 0\(<\) T\(< \infty\), and x0 lies in (0,1).

The state satisfies \(dX_{t}\)=v(\(X_{t}\))dt+\(dW_{t}\) with a standard one-dimensional Brownian motion and constant diffusion coefficient one.

The ambient class V consists of stationary Markov drifts v:(0,1)\(\to \mathbb R\) that are Borel measurable and locally Lipschitz on (0,1); endpoint divergence is allowed, and endpoint asymptotics do not determine the drift in the interior.

For every v in V, the stochastic differential equation started at x0 has a unique law up to \(\tau\)=inf\{t\(\geq0:X_{t}\) is not in (0,1)\}.

Define J(T,x0;v)=E{[}integral from 0 to T wedge \(\tau\) of v\((X_{t})^{2}\) dt{]} when P(\(\tau >\) T)=1, and \(J(T,x0;v)=\infty\) otherwise. Define \(A_{T}(x0)\)=\{v in V:J(T,x0;v)\(< \infty\)\} and M\(\star(T,x0)\)=inf\{J(T,x0;v):v in \(A_{T}(x0)\)\}, with the convention that the infimum is \(\infty\) if \(A_{T}(x0)\) is empty.

No reflecting local time, impulse, projection, absorption convention, boundary reset, or volatility control may enforce survival.

Critical logarithmic profiles may be tested near zero in the form v(x)=1/(2x)+\(c_{0}\)/(x log(1/x)) plus an explicitly controlled lower-order remainder, with the reflected inward form near one. These profiles are diagnostic subclasses and are not restrictions on \(A_{T}(x0)\).

Convergence of feedbacks is compact-open convergence: \(v_{n}\) converges to v locally uniformly on every compact subset of (0,1). Any candidate limit must be rechecked for well-posedness, survival through T, and finite energy; shared endpoint profiles do not imply compact-open precompactness.

One-dimensional unit diffusion on (0,1), with arbitrary fixed finite T\(>\)0 and x0 in (0,1), stationary locally Lipschitz drift-only feedback, almost-sure survival through T, extended quadratic energy, and compact-open convergence used only to classify defects. No boundary actuator, time-dependent feedback, endpoint-envelope restriction, or relaxation of the feedback class is allowed.

The assertion is pointwise in the displayed pair (T,x0), and the positive target quantifies over every T\(>\)0 and x0 in (0,1); parameter dependence may not be suppressed.

The optimization is over the entire class V authorized by the anchor, not an endpoint-profile subclass, a uniformly Lipschitz subclass, or a prescribed singularity envelope.

Nonattainment requires both a sequence \(v_{n}\) in \(A_{T}(x0)\) with J(T,x0;\(v_{n}\)) converging to M\(\star(T,x0)\) and a proof that every v in \(A_{T}(x0)\) has strictly larger cost.

Any proposed minimizing sequence must be audited separately for interior concentration or oscillation, loss of endpoint viability, and failure of energy lower semicontinuity.

Determine whether, for every fixed \(T\in(0,\infty)\) and \(x_0\in(0,1)\), the finite infimum \(M_\star(T,x_0)\) over stationary locally Lipschitz drifts that keep \(dX_t=v(X_t)\,dt+dW_t\) inside \((0,1)\) through time \(T\) is attained.
\end{problem}

\clearpage
\section{P072: The sharp one-update contraction coefficient}
\label{problem:P072}

\begin{problem}
\(\eta_{0}=1/(\mu t_{0}\)).

\(V_{q}(p)=n^{(-q)} \sum_{i=1}^{n} \sigma_{i}^{q} p_{i}^{(1-q)}\).

\(P_{q}(\beta\)) holds if there exists finite \(K_{q}\) such that every authorized one-update datum satisfies E{[}‖\(X_{1}-x\star\)‖\(^{q}{]} \leq (1-\beta/t_{0}\))‖\(X_{0}-x\star\)‖\(^{q}+K_{q} \mu^{(-q)} V_{q}(p_{0})/t_{0}^{q}\).

\[
\beta_{q}^{*}=\sup\{\beta in [0,q]: P_{q}(\beta) holds\}
\]

\(\lambda_{q}\)=2q{[}1-(1+1/(2q))\(^{(-q)}\){]}.

\(F(x)=n^{(-1)} \sum_{i} f_{i}\)(x)+r(x) on a finite-dimensional Euclidean space, where n\(\geq\)2, each \(f_{i}\) is convex with \(L_{i}\)-Lipschitz gradient, r is proper, closed, convex, and proximable, and F is \(\mu\)-strongly convex. Its unique minimizer \(x\star\) satisfies \(\nabla f_{i}(x\star\))=0 for every i and 0 belongs to partial r(\(x\star\)).

A fresh query to component i at x returns \(G_{s,i}(x)=\nabla f_{i}(x)+xi_{s,i}\). Conditional on the pre-query history, \(xi_{s,i}\) is fresh, centered, independent of earlier oracle noise, independent of x, and satisfies E{[}‖\(xi_{s,i}\)‖\(^{q}{]}=\sigma_{i}^{q}\) for fixed unknown \(\sigma_{i}\) in (0,\(\infty\)) and q in (1,2). No second moment, clipping, free batch, or uncharged oracle call is available.

Charged rounds are indexed by s=0,1,\ldots,B-1. Before any oracle call, the complete indicator schedule (\(a_{s}\)) is fixed, or sampled using external randomness independent of every oracle output and then fixed. If \(a_{s}\)=0, the round makes one calibration query at a declared reference point and sets \(X_{s+1}=X_{s}\). If \(a_{s}\)=1, it draws \(I_{s}\) from positive predictable probabilities \(p_{s}\), queries \(G_{s,I_{s}}(X_{s}\)), forms \(v_{s}=G_{s,I_{s}}(X_{s}\))/(n \(p_{s,I_{s}}\)), and sets \(X_{s+1}=prox_{\eta_{s} r}(X_{s}-\eta_{s} v_{s}\)). Query components and update probabilities may depend on the observable past, but \(a_{s}\) and all future indicators may not depend on any optimization or calibration oracle output. Every call advances s, and the output is \(X_{B}\).

Fix \(\varepsilon\) in (0,\(2^{(q-1)}\)-1). Before observing oracle noise, the policy chooses \(t_{0} \geq\)max\{2q,\(8L_{max}/(\mu \varepsilon\))\}, where \(L_{max}=\max_{i} L_{i}\), and uses \(\eta_{s}=1/(\mu\)(s+\(t_{0}\))) on update rounds. At every update round, \(p_{s,i} \geq 4L_{i}\)/(n \(\mu\)(s+\(t_{0}\))) for all i. Thus calibration advances the floor-release and stepsize clocks, and \(t_{0}\) remains visible in the deterministic transient. Constants in the target inequality may depend only on q and the displayed numerical floor coefficient, not on n, B, \(\mu\), the \(L_{i}, t_{0}, \varepsilon\), or the unknown scales.

The frontier concerns exactly one externally scheduled update with B=1 and \(a_{0}\)=1. It ranges over every finite-dimensional A1-A2 datum, deterministic \(X_{0}\), admissible \(\varepsilon\) and \(t_{0}\), and pre-query probability vector satisfying A5. It excludes A3 and makes no charged-horizon composition claim.

Set B=1 and \(a_{0}\)=1, with \(X_{0}\) and \(p_{0}\) fixed before the fresh component and oracle noise are drawn.

The finite remainder constant \(K_{q}\) may depend only on q and the fixed numerical floor coefficient in A5.

The sharp coefficient is the supremum of all \(\beta\) in {[}0,q{]} for which the displayed inequality holds uniformly over the complete one-update class.

The bounded common-shape premise A3 is not used.

For q in (1,2), let \(\lambda_{q}\)=2q[1-(1+1/(2q))\(^{(-q)}\)]. Classify the largest coefficient \(\beta\) for which a uniform one-update bound E[‖\(X_{1}-x\star\)‖\(^{q}] \leq (1-\beta/t_{0}\))‖\(X_{0}-x\star\)‖\(^{q}+K_{q} \mu^{(-q)} V_{q}(p_{0})/t_{0}^{q}\) holds over every B=1, \(a_{0}\)=1 instance authorized by A1, A2, A4, and A5, with finite \(K_{q}\) depending only on q and the fixed floor coefficient. Determine whether this sharp coefficient equals \(\lambda_{q}\).
\end{problem}

\clearpage
\section{P073: The \(3/5\) transition for an exact rank-two kernel}
\label{problem:P073}

\begin{problem}
For an entry set E, \(P_{E}(D)\) agrees with D on E and is zero outside E.

Kern(\(\Omega\)) is the event that there exist distinct M,\(M_{\prime}\) in \(K_{n}\) and S,T separately in \(L_{1/32}(\Omega\)) such that \(P_{\Omega minus (S union T)}\)(M-\(M_{\prime}\))=0.

n is even and sufficiently large. The admissible signal class \(K_{n}\) consists of real n by n matrices with exact rank two, Frobenius norm one, condition number at most two, and every row-space and column-space leverage score at most 4/n.

\(\mathbb R\) and C are deterministic subsets of {[}n{]} of size n/2, fixed before sampling. Entries are sampled independently with probability q on \(\mathbb R\) x C and probability p elsewhere, where 0\(<\) q\(\leq\)p\(\leq\)1. The modulus is simultaneous over \(K_{n}\) after \(\Omega\) is realized, so a competing pair may depend on \(\Omega\). High probability means probability at least 1-\(n^{(-10)}\).

The corruption fraction is fixed at \(\rho\)=1/32. The feasible propensity region requires every expected row and column degree to be at least 64 log n, where log is natural; this degree condition is only a baseline and not the sought rare-rectangle criterion.

Corruption values are arbitrary and their support may depend on M, \(\Omega\), and the realized noise, but its row and column counts are capped by floor(\(\rho d_{i}(\Omega\))) and floor(\(\rho e_{j}(\Omega\))), respectively.

When two explanations are compared, their deletion or corruption supports S and T must each satisfy the A5 caps separately; they may not be replaced by one support obeying doubled caps.

Only the noiseless p=1 two-level design, exact finite endpoint pairs in \(K_{n}\), corruption fraction 1/32, and attained zero-residual secants are considered. Positive lower bounds for the global modulus, tangent-modulus bounds, noisy risk, and general propensity matrices are excluded. Unlike the established full-observation results and the earlier tangent frontier, this asks whether finite global ambiguity persists under nonuniform sampling.

Fix p=1 and allow q in (0,1{]}, with deterministic half-size sets \(r_{set}\) and \(c_{set}\) fixed before sampling.

Every kernel witness must use distinct finite matrices M and \(M_{\prime}\) in \(K_{n}\); a limiting tangent direction or vanishing residual sequence is insufficient.

The two supports S and T are subsets of the realized sample and must separately satisfy every floor-valued row and column cap at fraction 1/32.

The endpoint pair and both supports may depend on the realized sample \(\Omega\).

Set p=1. Determine whether q=3/5 is the sharp high-probability transition, outside a fixed constant window, between existence and nonexistence of an attained exact kernel over the full anchored rank-two class.
\end{problem}

\clearpage
\section{P074: The selected-limit envelope for gradient descent}
\label{problem:P074}

\begin{problem}
Z(f)=\{x\(\in\)Rⁿ:f(x)=0\}.

\(x_{k+1}=x_{k}\)-η\(\nabla\)f(\(x_{k}\)), where η=t/L.

q(χ,t)=1-t(2-t)/χ.

For an admissible instance with 0\(<\) e\(<\)1, set \(T=\infty\) if its trajectory does not converge to some x\(\infty \in\)Z(f); otherwise set \(T=\sup_{k\in K(q)}\) ‖\(x_{k}\)-x\(\infty\)‖/(r √(e) \(q^{k/2}\)). Set T=0 when e=0. For q=0, convergence additionally means x₁=x\(\infty\).

C★(χ,t) is the supremum of T over all safe admissible instances having the fixed certified parameters χ and t.

The dimension n is arbitrary and f:\(\mathbb R\)ⁿ\(\to{[}0,\infty\)) is continuously differentiable with nonempty zero set.

The gradient is globally L-Lipschitz for a certified L\(>\)0, and on B(x₀,r) it satisfies ‖\(\nabla\)f(x)‖\(^2 \geq2\mu\)f(x) for a certified \(\mu \in\)(0,L{]}.

The radius r\(>\)0 and initial point x₀ are fixed, with normalized energy e=2f(x₀)/(\(\mu\)r\(^2\)) and condition ratio χ=L/\(\mu\).

The iterates are exact gradient descent \(x_{k+1}=x_{k} -\)η\(\nabla\)f(\(x_{k}\)) with normalized step t=ηL\(\in\)(0,1{]}.

The frontier is classwise: Θ(χ,t) is defined using all dimensions and all functions satisfying A1--A4 with the stated certified constants, and closed-ball invariance is distinguished from strict interior invariance.

All dimensions and all objectives authorized by A1--A5, with L and \(\mu\) treated as certified constants. The supremum defining C★ ranges over 0\(\leq\)e\(<\)1 and every finite iterate. Failure to converge to a zero is assigned infinite cost. When q(χ,t)=0, only k=0 enters the normalized ratio and finite termination is checked separately.

L and \(\mu\) are the certified constants from A2, whether or not they are optimal for the objective.

Only instances with normalized initial energy 0\(\leq\)e\(<\)1 enter C★.

An instance with e=0 has normalized tail cost zero.

If q=0, the ratio is evaluated only at k=0 and the target additionally requires x₁=x\(\infty \in\)Z(f).

For exact gradient descent under A1–A5, determine whether the worst normalized nonasymptotic distance to the selected limit has the exact envelope C★(χ,t)=1 throughout the strict safe class e<1.
\end{problem}

\clearpage
\section{P075: Maxwell span under log-Lipschitz curvature}
\label{problem:P075}

\begin{problem}
H(s)=s log s-s+1 for s\(>\)0.

Extend A by one outside {[}1/4,2{]} and let \(G_{A}\) be the unique C2 function satisfying \(G_{A}(1)=G_{A}\)`(1)=0 and \(G_{A}\)''(s)=A(s)/s.

Set \(F_{A}(s)=G_{A}(s)\)-H(s)/2 and \(Phi_{A},q(s)=F_{A}(s)\)-q(s-1). Then \(F_{A}\)'\,'(s)=(A(s)-1/2)/s.

A pair (u,v), 0\(<\) u\(<\) v, is a Maxwell pair if some real q satisfies \(Phi_{A},q(u)=Phi_{A},q(v)=\min_{s>0} Phi_{A}\),q(s).

For J={[}ell,rgt{]}, let Δ(A) be the supremum of log(v/u) over all Maxwell pairs and set Sigma(A,J)=Δ(A)/log(rgt/ell).

\(Lip_{log}(A)\) is the least M in {[}0,\(\infty\){]} such that \(\mid\) log A(s)-log A(t)\(\mid \leq\)M\(\mid\) log s-log t\(\mid\){} for all s,t in {[}1/4,2{]}.

Let \(\sigma(M)\) be the infimum of Sigma(A,J) over all admissible pairs with \(Lip_{log}(A)\leq\)M, and set \(\sigma(M)=\infty\) if this class is empty.

For each admissible A and positive integer k, prescribe \(h_{k}\)'`(\(\varepsilon_{k}\) s)=A(s)/(\(\varepsilon_{k}\) s) on {[}\(\varepsilon_{k}/4,2epsilon_{k}\){]}, set \(h_{k}\)'`=h'' outside, and match \(h_{k}\) to h on the left exterior component. The two support moments imply equality on the right exterior component.

The ambient space is finite-dimensional, and h and every \(h_{k}\) are Legendre functions with the same nonempty open convex interior domain \(\Omega\).

Each kernel is twice continuously differentiable and has positive-definite Hessian on \(\Omega\), while \(h_{k}\) converges to h in C1 on every fixed compact subset of \(\Omega\); no uniform Hessian or divergence comparison is assumed near the boundary.

The objective f is fixed, proper, lower-semicontinuous, and prox-bounded relative to the reference kernel h.

There are attentive graph points (\(x_{k},v_{k}\)) with \(v_{k}\) in the limiting subdifferential of f at \(x_{k}, x_{k}\) approaching the boundary of \(\Omega\), shrinking neighborhoods \(U_{k}\), and a constant r such that f(z) is at least f(x)+\(<\) v,z-x\(>-rD_{h}(z,x)\) for every sufficiently large k and every authorized attentive graph point (x,v) and z in \(U_{k}\).

The proximal parameter \(\lambda\) is fixed and positive with \(\lambda\) r\(<\)1, and all localized subproblems under consideration remain inside \(\Omega\).

Use A1-A5 in dimension one with the fixed entropy landscape h(t)=t log t, f(t)=-(t log t)/2, \(\lambda\)=1, and shrinking anchors \(x_{k}=2^{(-k)}\). For each M, optimize the global Maxwell-span ratio over continuous exact-support profiles on {[}1/4,2{]} having a logarithmically wide subthreshold interval and intrinsic log-Lipschitz distortion at most M. The target is a two-sided asymptotic rate for the excess over seven sixths, not a positive M-independent gap.

For t\(>\)0 set h(t)=t log t and f(t)=-(t log t)/2; set h(0)=f(0)=0 and both functions to +\(\infty\) for t\(<\)0. Let \(\Omega=(0,\infty), \lambda=1, \varepsilon_{k}=2^{(-k)}, x_{k}=\varepsilon_{k}\), and \(v_{k}\)=f'(\(x_{k}\)).

A is continuous on {[}1/4,2{]}, equals one near both endpoints, satisfies 1/4\(\leq\)A(s)\(\leq\)5/4, and obeys integral\_\[1/4,2\](A(s)-1)ds=0 and integral\_\[1/4,2\](A(s)-1)ds/s=0.

J={[}ell,rgt{]} is contained in {[}1/2,1{]}, log(rgt/ell)\(\geq\)1/8, and A(s)\(\leq\)3/8 for every s in J.

A Maxwell pair consists of distinct global minimizers over (0,\(\infty\)) for one common tilt, and Δ(A) is the supremum of their logarithmic separations.

For M\(>\)0 require \(\mid\) log A(s)-log A(t)\(\mid \leq\)M\(\mid\) log s-log t\(\mid\){} for all s,t in {[}1/4,2{]}.

Determine whether imposing an intrinsic log-Lipschitz bound M on exact-support curvature profiles makes the optimal excess Maxwell span above seven sixths decay at the sharp order 1/M.
\end{problem}

\clearpage
\section{P076: Square-root gluing of clique moments}
\label{problem:P076}

\begin{problem}
For q=(x,w), \(b_{1}\)(q)=(1,x,w), \(b_{2}\)(q)=(1,x,w,\(x^{2}\),xw,\(w^{2}\)), \(M_{1}\)(y)=\(L_{y}\)(\(b_{1b,1}^{T}\)), and \(M_{2}\)(y)=\(L_{y}\)(\(b_{2b,2}^{T}\)). The box localizers are \(L_{y}\)(\((1-x^{2})b_{1b,1}^{T}\)) and \(L_{y}\)(\((1-w^{2})b_{1b,1}^{T}\)).

\(F_{\kappa,\delta}\) is the set of normalized pairs (\(y_{L}\),\(y_{R}\)) satisfying the moment and box-localizing positive-semidefinite constraints, \(y_{L}\)(\(x^{r}\))=\(y_{R}\)(\(x^{r}\)) for 0\(\leq\)r\(\leq\)4, \(\lambda_{2}\)(\(M_{1}\)(\(y_{L}\))),\(\lambda_{2}\)(\(M_{1}\)(\(y_{R}\)))\(\geq \kappa\), and \(\lambda_{3}\)(\(M_{2}\)(\(y_{L}\))),\(\lambda_{3}\)(\(M_{2}\)(\(y_{R}\)))\(\leq \delta\).

G is the set of pairs of degree-four clique marginals induced by Borel probability measures on {[}-1,1{]}\(^{3}\).

Δ(\(y_{L}\),\(y_{R}\))=max\{\(\lambda_{3}\)(\(M_{2}\)(\(y_{L}\))),\(\lambda_{3}\)(\(M_{2}\)(\(y_{R}\)))\}.

H\_\(\kappa\)(\(\delta\))=sup\{inf\{‖y-g‖\(_{2}\):g in G\}:y in \(F_{\kappa,\delta}\)\}.

\(y_{L}\) and \(y_{R}\) are real degree-four truncated moment sequences on variables (x,u) and (x,v), respectively, normalized by \(y_{L}\)(1)=\(y_{R}\)(1)=1.

For each clique, the order-two moment matrix and the order-one localizing matrices for its applicable box polynomials 1-\(x^{2}\), 1-\(u^{2}\), and 1-\(v^{2}\) are positive semidefinite.

The separator moments agree exactly: \(y_{L}\)(\(x^{j}\))=\(y_{R}\)(\(x^{j}\)) for every integer 0\(\leq\)j\(\leq\)4.

For each clique, the second-largest eigenvalue of the degree-one moment block is at least the fixed \(\kappa\), and the third-largest eigenvalue of the order-two moment matrix is at most \(\delta\); eigenvalues are ordered nonincreasingly.

The ambient vector contains every recorded degree-at-most-four moment from both cliques, distance is its Euclidean norm, and objective coefficients are restricted to the corresponding Euclidean unit ball.

The conditioning level \(\kappa\) is fixed with 0\(< \kappa <\)1, \(\delta\) is nonnegative, and G consists of the degree-four clique marginals of all probability measures on {[}-1,1{]}\(^{3}\) without imposing an extraction algorithm or an auxiliary low-rank approximation.

Degree-four truncated moments on the two cliques (x,u) and (x,v), order-two moment matrices, order-one box localizers, exact separator agreement through degree four, fixed Euclidean moment coordinates, and no assumptions beyond A1-A6.

For q=(x,w), use \(b_{1}\)(q)=(1,x,w) and \(b_{2}\)(q)=(1,x,w,\(x^{2}\),xw,\(w^{2}\)); \(M_{1}\)(y)=\(L_{y}\)(\(b_{1b,1}^{T}\)) and \(M_{2}\)(y)=\(L_{y}\)(\(b_{2b,2}^{T}\)).

The actual defect is
\[
\Delta(y_L,y_R)
=\max\!\left\{
\lambda_3\!\left(M_2(y_L)\right),
\lambda_3\!\left(M_2(y_R)\right)
\right\}.
\]

Distance is taken in the complete, nonduplicated pair of clique moment coordinates specified by A5.

Every sharpness witness must lie outside G, retain one fixed \(\kappa\), and be separated from every point of G.

For each \(\kappa\), all upper-bound constants and all sharpness witnesses are chosen before the universally quantified upper-bound variable \(\delta\) and sequence index j.

For every fixed \(\kappa\) in (0,1), the following conjunction is asserted. First, there exist C(\(\kappa\))>0 and \(\delta_{0}\)(\(\kappa\))>0 such that, for every \(\delta\) in [0,\(\delta_{0}\)(\(\kappa\))], H\_\(\kappa\)(\(\delta\))\(\leq\)C(\(\kappa\))√\(\delta\). Second, there exist c(\(\kappa\))>0 and sequences \(\delta_{j}\)>0 and \(y_{j}\) such that \(\delta_{j}\) decreases to zero, \(y_{j}\) belongs to \(F_{\kappa,\delta_{j}}\) minus G, Δ(\(y_{j}\))=\(\delta_{j}\), and \(\operatorname{dist}_{2}\)(\(y_{j}\),G)\(\geq\)c(\(\kappa\))√\(\delta_{j}\) for every positive integer j.
\end{problem}

\clearpage
\section{P077: Zero-memory delayed gradient tracking}
\label{problem:P077}

\begin{problem}
\(m_{\star,max}\) denotes m\(\star(2,2,1;1,1,1/2)\) under \(SC_{UNION,ACTIVE,DELIVERY}\) and the maximal local update interpretation.

GoodMax(\(pi_{0},\alpha\),C,\(\rho\),schedule,r,\(x_{initial}\)) means that the A5 execution has an exact two-periodic limiting orbit with both limiting primal coordinates equal to (\(r_{1}+r_{2}\))/2, both limiting trackers zero, and the A5 relative Euclidean estimate with constants C and \(\rho\) for every k\(\geq\)0.

Exact0Max(\(pi_{0}\)) means that there exists \(\bar{\alpha} >\)0 such that for every \(\alpha\) in (0,\(\bar{\alpha}\)) there exist finite \(C_{\alpha}\) and \(\rho_{\alpha}\) in (0,1) for which GoodMax holds for every schedule in \(D_{UNION,SCHEDULES}\) and every r,\(x_{initial}\) in \(\mathbb R^{2}\).

Fix n\(\geq\)2, P\(\geq\)2, D\(\geq1, 0< \mu \leq\)L, and 0\(<\) η\(\leq\)1/2. An admissible schedule consists of P-periodic matrices A(k) and B(k), where periodicity does not require fundamental period P. Each A(k) is row-stochastic, each B(k) is column-stochastic, all diagonal entries are positive, every positive entry is at least η, and the period-union supports of A and B are separately strongly connected. For every ordered nonself edge active at round k, exactly one packet of age \(d_{ji}(k)\) in \{0,\ldots,D\} is delivered, with P-periodic ages; distinct-edge deliveries commute and an edge receives at most one packet per round. This class is nonempty because every phase may use self-weight 1/2 and weight 1/2 on a directed cycle, with any allowed periodic delay pattern.

Each node has \(f_{i}(x)=h_{i} (x-r_{i})^{2/2}\) on the real line, where \(\mu \leq h_{i} \leq\)L and \(r_{i}\) is arbitrary. The target is x\(\star=(\sum_{i} h_{i} r_{i})/(\sum_{i} h_{i}\)). Node i knows \(h_{i}\), while other curvatures and centers are not globally available.

At the beginning of round k, node j forms exactly the packet \(p_{j}(k)=(x_{j}(k),y_{j}(k)\)). Node i has exactly two persistent node coordinates \(x_{i}(k), y_{i}(k)\), and for every possible ordered nonself edge (j,i) it has one baseline overwrite register \(b_{ji}(k)\) in \(\mathbb R^{2}\) and one auxiliary register \(u_{ji}(k)\) in \(\mathbb R^{m}\). No other persistent node-local, edge-local, global, counter, clock, or controller state is permitted. Scheduled packets are delivered before the state update and overwrite their corresponding \(b_{ji}\) registers; inactive registers retain their values. Packets carry no timestamp, generation index, delay label, counter, or other metadata. Phase, current incident weights, and edge activity are transiently observable, but realized delays are not.

For each m, one protocol \(Pi_{m}\) is chosen after n,P,D,\(\mu\),L,η are fixed but before the schedule, objectives, centers, initialization, and stepsize are chosen. It consists of deterministic local coefficient functions defining phase-indexed linear maps: \(x_{i}(k+1)\) is linear in \(x_{i}(k), y_{i}(k)\), and current incoming \(b_{ji}(k), u_{ji}(k)\); after forming the local gradient difference \(h_{i}(x_{i}\)(k+1)-\(x_{i}(k)\)), a second linear map produces \(y_{i}(k+1)\) and \(u_{ji}(k+1)\). Coefficients may depend only on \(\alpha\), phase k modulo P, \(h_{i}\), and current incident entries of A(k), B(k). They may not depend on centers, past arrivals except through declared state, realized delay labels, future weights or delays, or the complete schedule. On zero-delay schedules, admissible laws must preserve \(\sum_{i} y_{i}(k)=\sum_{i}\) gradient \(f_{i}(x_{i}\)(k)) under tracker-consistent initialization and preserve every common-optimizer fixed trajectory; these are baseline restrictions, not assumptions of delayed exactness.

For k=-D,\ldots,0 set \(x_{i}(k)=x_{i}^{0}, y_{i}(k)=h_{i}(x_{i}^{0}-r_{i}), p_{i}\)(k)=(\(x_{i}^{0},y_{i}(k)), b_{ji}(0)=p_{j}\)(0), and \(u_{ji}(0)\)=0. Let z(k) be the Euclidean vector of all declared persistent coordinates immediately after round-k deliveries and before the round-k update. For a fixed protocol, uniform exact linear convergence means that there exist \(\bar{\alpha} >\)0 and, for every \(\alpha\) in (0,\(\bar{\alpha}\)), constants \(C_{\alpha} <+\infty\) and \(\rho_{\alpha}\) in (0,1), independent of the admissible schedule, centers, allowed curvatures, and \(x^{0}\), such that each execution has an exact P-periodic limiting orbit \(z_{\infty}\) with all limiting primal coordinates equal to \(x\star\) and all limiting tracker coordinates zero, and norm(z(k)-\(z_{\infty}(k modulo P))_{2} \leq C_{\alpha} \rho_{\alpha}^{k}\) norm(z(0)-\(z_{\infty}(0))_{2}\) for every k\(\geq\)0. Define \(m\star\)(n,P,D;\(\mu\),L,η)=inf\{m in the nonnegative integers: one such \(Pi_{m}\) exists\}, with infimum of the empty set equal to +\(\infty\).

Use A1-A5 with n=2, P=2, D=1, \(\mu\)=L=1, η=1/2, and m=0. An ordered nonself edge (j,i) is active exactly when \(A_{ij}(k)>\)0 or \(B_{ij}(k)>\)0. Its single delivered packet \(p_{j}\)(k-\(d_{ji}(k)\)) overwrites \(b_{ji}\) before the update. The second local map may use every currently available local quantity listed in \(SC_{MAXIMAL,SECOND,MAP,SIGNATURE}\).

At round k, E(k) consists of the ordered nonself edges (j,i) for which \(A_{ij}(k)>\)0 or \(B_{ij}(k)>\)0. Each edge in E(k) delivers exactly \(p_{j}\)(k-\(d_{ji}(k)\)) before the update, using one common age if both entries are positive; an edge outside E(k) makes no delivery and retains \(b_{ji}\).

For m=0, after \(x_{i}(k+1)\) and \(\delta_{g,i}(k)=h_{i}(x_{i}\)(k+1)-\(x_{i}(k)\)) are formed, \(y_{i}(k+1)\) may be any linear function of \(x_{i}(k), y_{i}(k), x_{i}(k+1), \delta_{g,i}(k)\), and all current incoming overwrite registers \(b_{ji}(k)\).

Coefficients may be node-indexed and incoming-edge-indexed, subject to A4's restrictions on their observable arguments; they remain independent of centers, hidden ages, past arrivals outside declared state, future data, and the complete schedule.

In the fixed two-agent, two-phase, one-delay cell, with union-active packet delivery and the maximal local linear input signature specified below, determine whether every zero-additional-memory protocol fails uniform exact linear convergence.
\end{problem}

\clearpage
\section{P078: From tangent injectivity to a global rank-two gap}
\label{problem:P078}

\begin{problem}
\(P_{\Omega}\) retains entries indexed by \(\Omega\) and sets all other entries to zero.

\(\alpha_{\Omega}\)(X)=inf over nonzero H in \(T_{X}\) of ‖\(P_{\Omega}\) H‖\(_{F}\)/(√(p)‖H‖\(_{F}\)).

\(\beta_{\Omega}\)(X)=inf over Y in K(n,\(\mu\),\(\kappa\)), Y distinct from X, of ‖\(P_{\Omega}\)(Y-X)‖\(_{F}\)/(√(p)‖Y-X‖\(_{F}\)).

\(E_{T}\)(X) is the event \(\alpha_{\Omega}\)(X)\(\geq\)a.

For universally quantified L in (0,\(\infty\)), m=ceil(L \(\mu\) n log n) and p=m/\(n^{2}\).

n is at least 4, \(\Omega\) is sampled uniformly without replacement from {[}n{]}\(\times\){[}n{]} with \(\mid \Omega \mid\)=m, and p=m/n\(^2\).

\(\mu\) and \(\kappa\) are fixed parameters with \(\mu\) at least 1 and \(\kappa\) finite; K(n,\(\mu\),\(\kappa\)) consists of real n by n matrices of exact rank two, Frobenius norm one, condition number at most \(\kappa\), and left and right rank-two leverage scores at most 2\(\mu\)/n.

The target X belongs to K(n,\(\mu\),\(\kappa\)), and \(T_{X}\) is the tangent space of the exact-rank-two manifold at X.

For a fixed constant a\(>\)0, every H in \(T_{X}\) satisfies ‖P\_\(\Omega\)H‖\(_{F}\) at least a √(p)‖H‖\(_{F}\).

Candidate competing completions Y are finite matrices in the same class K(n,\(\mu\),\(\kappa\)); limiting tangent vectors, rank-deficient endpoints, and matrices whose condition number diverges are not admissible global witnesses.

The target is pointwise in a deterministic X chosen before sampling. It concerns the conditional distribution of uniformly sampled fixed-size observation sets at the order-\(\mu\) n log n scale. Competitors are all distinct finite matrices in the same normalized exact-rank-two, incoherent, bounded-condition class. Rank-deficient endpoints, divergent condition numbers, tangent vectors, and merely local competitors are excluded.

For each n, X is deterministic and fixed before \(\Omega\) is sampled; the theorem is not a simultaneous uniform assertion over sample-adaptive targets.

The probability statement is conditional on \(\alpha_{\Omega}\)(X)\(\geq\)a and is required only when this event has positive probability.

The global modulus takes the infimum over every Y in K(n,\(\mu\),\(\kappa\)) distinct from X, including remote isolated completions.

The lower-gap constant may depend on \(\mu\), \(\kappa\), a, and L but not on n, X, or the realized \(\Omega\).

The symbol L is a universally quantified real sampling coefficient in (0,\(\infty\)), and the schedule is defined by m=ceil(L \(\mu\) n log n) and p=m/\(n^{2}\).

Fix \(\mu \geq\)1, \(\kappa \geq\)1, a>0, and a sampling coefficient L>0. For each sufficiently large n set m=ceil(L \(\mu\) n log n), p=m/\(n^{2}\), and sample \(\Omega\) uniformly without replacement. Determine whether, for every deterministic X in K(n,\(\mu\),\(\kappa\)), quantitative injectivity on \(T_{X}\) conditionally forces a dimension-uniform global secant gap with polynomially high conditional probability.
\end{problem}

\clearpage
\section{P079: Pointwise stability of clipped proximal SAGA}
\label{problem:P079}

\begin{problem}
Let
\[
\Phi(x)=\frac1n\sum_{i=1}^n f_i(x)+r(x),
\]
and let \(x_\star\) be its unique minimizer.

Set
\[
\eta_t=\min\left\{\frac{1}{4L_{\max}},
\frac{2}{\mu(t+t_0)}\right\},
\qquad
\tau_t=c_\tau(t+t_0)^{1/q},
\]
where \(c_\tau=C(q,n,p_{\min},\mu,L_{\max},(\sigma_i)_{i=1}^n)\).

Initialize \(h_{i,0}=0\).  At time \(t\), draw \(I_t\) from \(p_t\) and set
\[
\begin{aligned}
\bar h_t&=\frac1n\sum_i h_{i,t},
&Z_t&=G_{I_t}(x_t),\\
Y_t&=\bar h_t+\frac{Z_t-h_{I_t,t}}{np_{I_t,t}},
&v_t&=\operatorname{clip}_{\tau_t}(Y_t),\\
x_{t+1}&=\operatorname{prox}_{\eta_t r}(x_t-\eta_t v_t).
\end{aligned}
\]
Only \(h_{I_t,t}\) is replaced by \(\operatorname{clip}_{\tau_t}(Z_t)\).

Finally, define
\[
H_t=\frac1n\sum_i\lVert h_{i,t}-\nabla f_i(x_t)\rVert^q.
\]

\[
\begin{aligned}
S_C(q,\mathrm{instance})
={}&\sup_p\sup_{t\geq0}\max\bigl\{
\mathbb E[H_t],\\
&\quad (t+t_0)^{2(q-1)/q}
\mathbb E\lVert x_t-x_\star\rVert^2
\bigr\},
\end{aligned}
\]
where \(p\) ranges over all admissible predictable sampling rules.

The objective is \(\Phi(x)=n^{-1} \sum_{i=1}^{n} f_{i}(x)\)+r(x) on a finite-dimensional Euclidean space, and \(x_{0}\) belongs to dom(r).

Each \(f_{i}\) is convex with \(L_{i}\)-Lipschitz gradient, their average is \(\mu\)-strongly convex for \(\mu >\)0, and r is proper, closed, convex, and has an exactly evaluable proximal map.

A query to component i at x returns \(G_{i}(x)=\nabla f_{i}(x)+xi_{i}\). For one fixed q in (1,2), E{[}\(xi_{i}\) given the pre-query history and \(I_{t}\)=i{]}=0 and E{[}‖\(xi_{i}\)‖\(^{q}\) given that information{]}\(\leq \sigma_{i}^{q}\); no second moment is assumed.

Before query t, \(p_{t}\) is measurable with respect to the pre-query history, 0\(< p_{min} \leq\)1/n, \(p_{i,t} \geq p_{min}\) for every i, and \(I_{t}\) has conditional distribution \(p_{t}\). The target is uniform over all such predictable rules, including \(p_{i,t}\)=1/n.

Fresh oracle noise obeys A3 conditional on the past and selected component at every queried point.

The algorithm is exactly the recursion in \(central_{object}\). Importance weighting occurs before clipping, clipping occurs only in the displayed estimator and table updates, and the same noisy query \(Z_{t}\) is shared by both uses.

The stepsizes are \(\eta_{t}\)=min\{1/(\(4L_{max}),2/{[}\mu\)(t+\(t_{0}\)){]}\} with \(t_{0} \geq 8L_{max}/\mu\). An admissible threshold rule C maps only (q,n,\(p_{min},\mu,L_{max},(\sigma_{i})_{i=1}^{n}\)) to a finite \(c_{\tau}\) in (0,\(\infty\)), before any oracle draws, and \(\tau_{t}=c_{\tau} (t+t_{0})^{1/q}\). The same selected \(c_{\tau}\) is used when comparing adaptive and uniform sampling. Complexity counts every noisy component query.

The exact A1-A7 shared-query clipped proximal SAGA recursion. Unlike the previous averaged expected-objective frontier, this generation concerns pointwise squared distance and table q-energy; it neither assumes nor directly asserts an objective-gap bound for the nonsmooth composite objective.

Use exactly the estimator, clipping, proximal update, and shared-query table refresh specified by A6.

C maps only (q,n,\(p_{min},\mu,L_{max},(\sigma_{i}\))) to a finite \(c_{\tau} >\)0 and cannot depend on \(t_{0}\), the initialization, component affine offsets, or realized history.

The finite stability constant may depend on the fixed admissible instance, including \(x_{0}, t_{0}\), and the component gradients at the minimizer, but not on the predictable sampling rule or time.

The target is the joint pointwise distance and table-energy observable, not the previously attempted weighted-average expected-objective observable.

Determine whether one admissible strictly positive parameter-only threshold rule C, independent of the fixed schedule offset \(t_{0}\), simultaneously gives the critical pointwise mean-square decay and a uniformly bounded recycled-table q-energy for every A1-A7 instance and every admissible predictable sampling rule.
\end{problem}

\clearpage
\section{P080: A quadratic-tail bound for binary OMWU}
\label{problem:P080}

\begin{problem}
Set \(\alpha_{1}\)=A11-A21, \(\alpha_{0}\)=A12-A22, \(\beta_{1}\)=B11-B12, and \(\beta_{0}\)=B21-B22. Define \(\alpha(q)=\alpha_{1}\) q+\(\alpha_{0}(1-q)\) and \(\beta(p)=\beta_{1}\) p+\(\beta_{0}(1-p)\).

For \(p_{t}=x_{1}^{t}, q_{t}=y_{1}^{t}, z_{t}\)=log(\(p_{t}/(1-p_{t}\))), and \(w_{t}\)=log(\(q_{t}/(1-q_{t}\))), define \(z_{t+1}=z_{t}\)+η(2alpha(\(q_{t})-\alpha(q_{t-1}\))) and \(w_{t+1}=w_{t}\)+η(2beta(\(p_{t})-\beta(p_{t-1}\))).

Define q\(\star=-\alpha_{0}/(\alpha_{1}-\alpha_{0}\)) and p\(\star=-\beta_{0}/(\beta_{1}-\beta_{0}\)). Under \(SC_{CYCLIC,SIGN,CELL}, (p\star,q\star\)) is the unique completely mixed Nash equilibrium.

Define \(\gamma(A,B)\)=min\{\(\mid \alpha_{0} \mid,\mid \alpha_{1} \mid,\mid \beta_{0} \mid,\mid \beta_{1} \mid\)\} and \(\rho(A,B)\)=min\{\(p\star,1-p\star,q\star,1-q\star\)\}.

Define \(H_{init}\)=max\{\(\mid z_{-1} \mid,\mid z_{0} \mid,\mid w_{-1} \mid,\mid w_{0} \mid\)\}.

Let \(d_{N}(t)\) be the Euclidean distance of (\(x^{t},y^{t}\)) from the unique mixed Nash equilibrium. Define \(T_{\varepsilon}\) as the least nonnegative integer T such that \(d_{N}(t)\leq \varepsilon\) for every integer t\(\geq\)T, with value \(\infty\) if no such T exists.

Define \(\mathbb R_{quad}=η^{(-2)} \gamma (A,B)^{(-2)} \rho (A,B)^{(-2)}((1+H_{init})^{2}\)+log(1/\(\varepsilon\))).

There are exactly two players and two pure actions per player, with row-player payoff matrix A and column-player payoff matrix B, and every payoff entry lies in {[}-1,1{]}.

For every integer t\(\geq\)0, player i uses the exact entropic optimistic multiplicative-weights update \(x_{i}^{(t+1)}\)(a) proportional to \(x_{i}^{t}(a)\) exp(η times (2 \(u_{i}\)(a,\(x_{-i}^{t})-u_{i}\)(a,\(x_{-i}^{(t-1)}\)))), with one common constant stepsize η and no projection, truncation, or noise.

The stepsize satisfies 0\(<\) η\(\leq\)1/16 under the payoff normalization in A1.

Both prescribed strategy profiles at times -1 and 0 are strictly interior.

The game is standard nondegenerate in the two-action sense: against each pure action of the opponent, a player's two pure actions have unequal payoffs. Equivalently here, \(\gamma(A,B)>\)0.

Games satisfy A1-A5 and have opposite-sign endpoint gaps for both players with opposite gap slopes, so they possess a unique completely mixed Nash equilibrium. All finite two-profile histories and all η in (0,1/16{]} are allowed. The target is quantitative: it strengthens established qualitative Nash-set convergence and repairs the false linear-radius rate by using the quadratic radius forced by the matching-pennies boundary excursion.

Writing \(\alpha_{1}\)=A11-A21, \(\alpha_{0}\)=A12-A22, \(\beta_{1}\)=B11-B12, and \(\beta_{0}\)=B21-B22, require \(\alpha_{0} \alpha_{1} <0, \beta_{0} \beta_{1} <\)0, and (\(\alpha_{1}-\alpha_{0})(\beta_{1}-\beta_{0})<\)0.

The bound uses the actual quantities \(\gamma(A,B)\) and \(\rho(A,B)\), not auxiliary lower bounds.

\(H_{init}\) is the maximum absolute value of all four log odds prescribed at times -1 and 0.

The initialization contribution is \((1+H_{init})^{2}\) rather than 1+\(H_{init}\).

\(T_{\varepsilon}\) is the least time after which every later iterate is \(\varepsilon\)-close to the unique Nash equilibrium.

Determine whether every cyclic-cell exact binary bimatrix OMWU orbit admits a universal eventual Nash-tail bound that is quadratic in the complete initial log-odds radius and explicitly controlled by the stepsize, payoff-gap margin, equilibrium-interiority margin, and accuracy.
\end{problem}

\clearpage
\section{P081: The scaling crossover for a single bias jump}
\label{problem:P081}

\begin{problem}
For an admissible objective and bias, a query at x returns y(x)=f(x)+b(x), and repeated queries return the same value.

The simple regret of an output x̂ is f(x̂)-\(\min_{x \in [0,1]}\) f(x).

\(\mathbb R_{N}^{step}\)(L,\(\delta\)) is the infimum over deterministic N-query algorithms of the supremum, over all f, \(\tau, c_{minus}\), and \(c_{plus}\) authorized by A1-A4, of the simple regret.

A function \(\Phi\) is a universal scaling limit at q when, for every L\(>\)0 and every pair of sequences \(N_{j}\) tending to \(\infty\) and \(\delta_{j} \geq\)0 satisfying \(N_{j} \delta_{j}\)/L tending to q, the values \(N_{j} \mathbb R_{N_{j}}^{step}\)(L,\(\delta_{j}\))/L tend to \(\Phi(q)\).

The objective f:{[}0,1{]} to the real numbers is L-Lipschitz for a known L\(>\)0.

The bias has the fixed form \(b(x)=c_{-}\) for x\(< \tau\) and \(b(x)=c_{+}\) for x\(\geq \tau\), where \(\tau\) is in (0,1) and \(\mid c_{-} \mid,\mid c_{+} \mid\){} are at most \(\delta\).

A deterministic algorithm chooses N query points adaptively, observes y(x)=f(x)+b(x), and repeated queries receive the same value.

After N queries the algorithm may output any point of {[}0,1{]}, and performance is f(x̂)-\(\min_{x \in [0,1]}\) f(x) under the quantifier order infimum over algorithms followed by supremum over admissible f and b.

One-dimensional L-Lipschitz objectives, one fixed two-level bias with a strictly interior jump, positive integer query budgets, deterministic adaptive queries, and arbitrary output points. The problem concerns finite-q scaling rather than merely the order L/N bound.

N is a positive integer, L\(>\)0, and \(\delta \geq\)0.

The jump satisfies \(\tau\) in (0,1), with b(\(\tau)=c_{plus}\) as prescribed by A2.

The same function \(\Phi\) must govern every positive-integer sequence \(N_{j}\) tending to \(\infty\) and every \(\delta_{j} \geq\)0 satisfying \(N_{j} \delta_{j}\)/L tending to the same q.

Determine whether there exist one function \(\Phi:[0,\infty\)) to [1/2,5/2] and one finite \(q\star\)>0 such that, for every L>0, every q\(\geq\)0, every sequence of positive integers \(N_{j}\) tending to \(\infty\), and every sequence \(\delta_{j} \geq\)0 satisfying \(N_{j} \delta_{j}\)/L tending to q, the normalized radius \(N_{j} \mathbb R_{N_{j}}^{step}\)(L,\(\delta_{j}\))/L tends to \(\Phi(q)\), with \(\Phi(0)\)=1/2 and \(\Phi(q\star\))>1/2.
\end{problem}

\clearpage
\section{P082: The fixed-confidence resolution constant}
\label{problem:P082}

\begin{problem}
L=1, \(\sigma=1, \beta\)=r \(\gamma, \varepsilon\)=(2r+1)\(\gamma\), and \(\alpha=\alpha_{0}\).

\[
a_{d}=d/(d+2),\qquad C_{d}=2^{(1-d)} a_{d}^{(-d)} (1-a_{d})^{(-2)}
\]

\[
S_{d,r,\alpha_{0}}(\gamma)=\gamma^{(d+2)} N\star(d,1,r \gamma,1,(2r+1)\gamma,\alpha_{0})
\]

The dimension d is a positive integer, the domain is {[}0,1{]}\(^{d}\) with the \(\infty\) norm, and f is an unknown real-valued L-Lipschitz function with known L\(>\)0.

At query \(x_{t}\) the oracle returns \(Y_{t}\)=f(\(x_{t}\))+b(\(x_{t})+xi_{t}\), where b is one fixed unknown function throughout the transcript, \(\sup_{x} \mid\) b(x)\(\mid \leq \beta, \beta \geq\)0 is known, and b has no assumed regularity.

For every admissible oracle law, the variables \(xi_{t}\) are independent and centered and satisfy E{[}exp(\(\lambda xi_{t}){]}\leq\)exp(\(\sigma^{2} \lambda^{2/2}\)) for every real \(\lambda\), with known \(\sigma \geq\)0; the guarantee is uniform over all laws satisfying this condition.

The algorithm is deterministic and adaptive, may repeat query points, uses a budget N in \{0,1,2,\ldots\}, and outputs a point in {[}0,1{]}\(^{d}\); when N=0 its output is fixed before observing the oracle.

The budget \(N\star\) is the smallest nonnegative N for which one algorithm, uniformly over every admissible f, fixed b, and noise law, outputs x̂ satisfying f(x̂)-\(\min_{x}\) f(x)\(\leq \varepsilon\) with probability at least 1-\(\alpha\).

The active strip has 0\(< \alpha \leq1/4, 0\leq \beta <\) L/4, \(\gamma=\varepsilon-2 \beta >\)0, and \(\varepsilon=2 \beta+\gamma <\) L/2. The regimes \(\varepsilon \geq\)L/2, \(\varepsilon <\) min\{2 \(\beta\),L/2\}, and all equality endpoints are classified separately and are not included in the active-rate claim.

Deterministic adaptive value-query algorithms under A1-A6 on the \(\infty\)-norm unit cube. The parameters d, r, and \(\alpha_{0}\) are fixed while \(\gamma\) decreases to zero through 0\(< \gamma <\)1/{[}2(2r+1){]}. The guarantee is uniform over every fixed bias field and every independent centered 1-sub-Gaussian noise law. This fixed-confidence regime is not the established polynomial-confidence ray \(\alpha=\gamma^{\lambda}\).

Set L=1, \(\sigma=1, \beta\)=r \(\gamma, \varepsilon\)=(2r+1)\(\gamma\), and \(\alpha=\alpha_{0}\) for fixed d in \{1,2,\ldots\}, r\(\geq\)0, and \(\alpha_{0}\) in (0,1/4{]}.

Restrict \(\gamma\) to 0\(< \gamma <\)1/{[}2(2r+1){]}.

The same deterministic capped algorithm must work for every fixed field b with \(\sup_{x} \mid\) b(x)\(\mid \leq\)r \(\gamma\) and every noise law authorized by A3.

Fix a positive integer d, a bias ratio r\(\geq\)0, and a constant failure probability \(\alpha_{0}\) in (0,1/4]. With \(L=\sigma=1, \beta\)=r \(\gamma\), and \(\varepsilon\)=(2r+1)\(\gamma\), determine whether \(\gamma^{(d+2)} N\star\)(d,1,r \(\gamma\),1,(2r+1)\(\gamma,\alpha_{0}\)) converges as \(\gamma\) decreases to zero to \(C_{d}\) log(1/\(\alpha_{0}\)), where \(a_{d}\)=d/(d+2) and \(C_{d}=2^{(1-d)} a_{d}^{(-d)} (1-a_{d})^{(-2)}\).
\end{problem}

\clearpage
\section{P083: Phase-dependent feedback from a stopped HJB equation}
\label{problem:P083}

\begin{problem}
L(r)=log(e/r).

\(z_{0}\)(r)=r and \(z_{1}\)(r)=1-r.

\(D_{i}\)(r)=L(r)(1/2-\(a_{h}\)(0,\(z_{i}\)(r))).

The state domain is (0,1), \(\sigma >\)0 is constant, the action set is {[}0,A{]} with A\(> \sigma ^2\)/2, \(\tau\) is the first hitting time of 0 or 1, and the process is stopped at \(\tau\) without reflection or boundary intervention.

Before \(\tau\), \(dX_{t}\)=\(a_{t}\)(1-\(2X_{t}\))/{[}\(X_{t}\)(1-\(X_{t}\)){]}dt+\(\sigma dW_{t}\). Admissible controls are progressively measurable {[}0,A{]}-valued processes for which a weak solution exists up to \(\tau\); admissibility does not require invariance.

For starting phase s and state x, the horizon-H cost is the expected running cost ∫\(_{s} ^{H∧\tau}\){[}q(\(X_{r}\))+\(\lambda a_{r} ^2\){]}dr, plus g(\(X_{H}\)) on \{\(\tau >\) H\}, plus \(h_{0}\)(\(\tau\)) or \(h_{1}\)(\(\tau\)) upon exit at 0 or 1. Here \(\lambda >\)0; q and g are symmetric, lower-bounded, continuous on {[}0,1{]}, and sufficiently regular in the interior; \(h_{0}\) and \(h_{1}\) are continuous; and \(h_{i}\)(H)=g(i).

For the data under study, the value V is assumed to be the unique continuous viscosity solution, with interior \(C^{1,2}\) regularity, of \(V_{s}\)+\((\sigma^2/2)V_{xx}\)+q(x)+\(\min_{0\leq a\leq A}\)\{\(\lambda\)a\(^2\)+a\((1-2x)V_{x}\)/{[}x(1-x){]}\}=0, with V(H,x)=g(x) and V(s,i)=\(h_{i}\)(s) for i\(\in\)\{0,1\}. The canonical feedback \(a_{H}\)(s,x) is the unique minimizer, namely the projection onto {[}0,A{]} of -\((1-2x)V_{x}\)/{[}2\(\lambda\)x(1-x){]}; no viability or boundary expansion is assumed.

Fix 0\(<\) Δ\(\leq\)H. On every interval {[}kΔ,(k+1)Δ), the controller applies \(a_{H}\)(t-kΔ,\(X_{t}\)), so every phase s\(\in\){[}0,Δ) recurs indefinitely.

The SDE driven by the periodically reset canonical feedback has a weak solution unique in law up to \(\tau\).

Specialize A1-A6 to \(\sigma\)=\(\lambda\)=A=H=Δ=1 and equal exit data \(h_{0}\)=\(h_{1}\). The problem is an existence question for autonomous HJB data, not a conditional accessibility theorem for a prescribed feedback. The data must satisfy A4 uniqueness and A6 weak uniqueness; the reset profile and later zero-control slab must both arise from the canonical projected minimizer.

Set \(\sigma\)=\(\lambda\)=A=H=Δ=1 and require \(h_{0}\)=\(h_{1}\)=h.

Require q and g to belong to C({[}0,1{]}) intersect \(C^{2}\)((0,1)), to be symmetric and lower bounded, with derivatives through order two bounded on compact subsets of (0,1). Require h in \(C^{1}\)({[}0,1{]}) and h(1)=g(0)=g(1).

For i in \{0,1\}, the reset-phase deficit \(D_{i}\)(r)=L(r)(1/2-\(a_{h}\)(0,\(z_{i}\)(r))) tends to 1/4 as r decreases to zero.

There exists \(r_{0}\) in (0,1/4) such that \(a_{h}\)(s,x)=0 whenever s is in {[}1/3,2/3{]} and x belongs to (0,\(r_{0}\)) union (1-\(r_{0}\),1).

There exists a rational \(p_{cert}\) in (0,1) such that \(P_{1/2}\)(\(\tau \leq\)2/3)\(\geq p_{cert}\).

Prove or refute that autonomous symmetric stopped-HJB data can generate a genuinely phase-dependent canonical feedback which has the statically inaccessible logarithmic coefficient 1/4 at reset phase, becomes identically zero in a fixed boundary neighborhood during a later positive-duration recurring phase slab, and consequently yields a quantitative finite-time exit witness.
\end{problem}

\clearpage
\section{P084: Weak-only convergence from atomic metric growth}
\label{problem:P084}

\begin{problem}
For the standard basis (\(e_{n}\)), D \(e_{n}=n^{-1} e_{n}\) and A(u,v)=(0,Dv). Thus \(ker(A)=\ell^2(N)\)⊕\{0\} is infinite-dimensional and ran(A) is nonclosed.

(w⊗w)z=⟨z,w⟩w.

\(M_{k}\)=I+∑\(_{j<k} a_{j}(w_{j}\)⊗\(w_{j}\)). The Bessel budget gives I\(\preccurlyeq M_{k}\preccurlyeq(1+C)\)I and \(M_{k+1}\succcurlyeq M_{k}\).

\(x_{k}\)⇀0 means ⟨\(x_{k},test_{vector}\)⟩\(\to\)0 for every \(test_{vector} \in\)H.

Failure of strong convergence to zero means that some \(\varepsilon >\)0 and subsequence \(k_{j}\) satisfy ∥\(x_{k_{j}}\)∥\(\geq \varepsilon\).

For a fixed metric value M occurring in the schedule, a frozen replay is \(y_{r+1}=(I+\mathrm{MA})^{-1} y_{r}\).

H is a separable infinite-dimensional real Hilbert space.

A:H\(\to\)H is bounded, self-adjoint, positive semidefinite, and maximally monotone, with nonempty infinite-dimensional kernel and nonclosed range.

The iterates are exact and satisfy \(x_{k+1}=J_{c_{k} M_{k} A} x_{k}\) with 0\(< c_{min} \leq c_{k} \leq c_{max} < \infty\).

Every \(M_{k}\) is bounded, self-adjoint, and uniformly positive definite, with mI\(\preccurlyeq M_{k}\preccurlyeq\)MI for fixed 0\(<\) m\(\leq\)M\(< \infty\).

There are \(η_{k} \geq\)0 with ∑\(_{k} η_{k} < \infty\) such that \((1+η_{k})^{-1} M_{k}\preccurlyeq M_{k+1}\) for every k.

There are no errors, relaxation, inertia, anchoring projections, or iteration-dependent operators.

Let \(H=\ell^2(N)\)⊕\(\ell^2(N)\) and A(u,v)=(0,Dv), where D \(e_{n}=n^{-1} e_{n}\). Consider exact iterations \(x_{k+1}=(I+M_{kA})^{-1} x_{k}\) with \(M_{0}\)=I and \(M_{k+1}=M_{k}+a_{k}(w_{k}\)⊗\(w_{k}\)), where \(a_{k} >\)0, ∥\(w_{k}\)∥=1, and the weighted atoms have a uniform Bessel bound. Thus the metrics are uniformly conditioned and satisfy A5 with \(η_{k}\)=0. No upper relative Loewner inequality, summable atomic-mass condition, relaxation, or iteration-dependent operator is allowed.

\(H=\ell^2(N)\)⊕\(\ell^2(N)\) and A(u,v)=(0,Dv), with D \(e_{n}=n^{-1} e_{n}\).

\(M_{0}\)=I and \(M_{k+1}=M_{k}+a_{k}(w_{k}\)⊗\(w_{k}\)) for positive \(a_{k}\) and unit vectors \(w_{k}\).

There is C\(< \infty\) such that ∑\(_{j\geq0} a_{j} \mid\)⟨z,\(w_{j}\)⟩\(\mid ^2 \leq\)C∥z∥\(^2\) for every z\(\in\)H.

\(c_{k}\)=1 for every k and \(η_{k}\)=0 verifies A5.

No condition ∑\(_{k} a_{k} < \infty\) or upper relative Loewner inequality may be imposed.

Determine whether weak-only convergence can occur for the canonical diagonal nonclosed-range operator when the proximal parameter is constant and the metric grows only through uniformly Bessel-bounded positive rank-one increments.
\end{problem}

\clearpage
\section{P085: Moment contraction with unstable growing epochs}
\label{problem:P085}

\begin{problem}
Set \(c=(1+η\lambda)^{(-1)}\), so \(prox_{ηψ}(y)\)=cy.

Define \(\rho_{1}(\mu,\lambda\),η)=\(\sup_{‖x‖=1}\) E{[}‖c(I-ηH)x‖\(^{(3/2)}\){]}.

For one epoch write \(x_{t}=K_{t} \widetilde{x}, K_{0}\)=I, \(B_{t}\)=c(I-\(ηH_{t}\)), and \(D_{t}=K_{t}\)-I. Then \(D_{t+1}=B_{tD,t}\)-cη(\(\lambda\)I+A), where A is that epoch's frozen snapshot Hessian.

For m\(\geq1, K_{m}\)=I-cη \(\sum_{j=0}^{m-1} B_{m-1} B_{m-2}\)\ldots \(B_{j+1}(\lambda\)I+A), with the product interpreted as I when j=m-1.

The dimension is two, q=3/2, u is uniform on \{e1,(e1+e2)/√(2)\}, and Z is independent of u with P(Z\(>\) t)=\(t^{(-7/4)}\) for t\(\geq\)1.

Each stochastic loss is \(f_{Z,u}(x)=x^{T}\) Hx/2 with \(H=\mu\) I+Z \(uu^{T}\), where \(\mu >\)0, and the composite term is ψ(x)=\(\lambda\)‖x‖\(^{2/2}\) with \(\lambda >\)0.

At every epoch, the snapshot Hessian \(A_{s}\) and all inner Hessians \(H_{s,t}\) are mutually independent copies of H and are independent of the past.

The recursion is exactly \(x_{s,0}=\widetilde{x}_{s}, v_{s,t}=H_{s,t}(x_{s,t}-\widetilde{x}_{s})+A_{s} \widetilde{x}_{s}, x_{s,t+1}=prox_{η ψ}(x_{s,t}\)-η \(v_{s,t}\)), and \(\widetilde{x}\)\emph{(s+1)=x}(s,\(m_{s}\)), with fixed η\(>\)0 and last-inner-iterate reporting.

The growing-epoch experiment uses \(m_{s}\)=s+1 and deterministic initialization \(\widetilde{x}_{0}\)=e2; the comparison experiment uses \(m_{s}\)=1 with the same law and parameters.

Only the two-dimensional Pareto-curvature proximal snapshot recursion of A1-A5 is considered. The Pareto exponent is 7/4, q=3/2, the two directions are e1 and (e1+e2)/√(2), the snapshot batch has size one, the stepsize is fixed, and each epoch reports its last inner iterate.

Use \(H=\mu\)I+\(Zuu^{T}\) with exactly the distribution in A1; no truncation, batching, or commuting replacement is part of the target.

With \(c=(1+η\lambda)^{(-1)}\), require \(\rho_{1}(\mu,\lambda\),η)=\(\sup_{‖x‖=1}\) E{[}‖c(I-ηH)x‖\(^{(3/2)}{]}<\)1.

For the instability experiment use \(m_{s}\)=s+1, \(\widetilde{x}_{0}\)=e2, and \(\widetilde{x}\)\emph{(s+1)=x}(s,\(m_{s}\)).

Failure means that some \(\varepsilon >\)0 satisfies \(limsup_{s \to \infty}\) P(‖\(\widetilde{x}_{s}\)‖\(> \varepsilon)>\)0.

Determine whether there exist \(\mu>0, \lambda\)>0, and η>0 in the model fixed by A1-A5 such that the one-step epoch map is a strict uniform 3/2-moment contraction while, for \(m_{s}\)=s+1 and \(\widetilde{x}_{0}\)=e2, the endpoint sequence fails to converge in probability to zero.
\end{problem}

\clearpage
\section{P086: The \(1/8\) deletion threshold for rank three}
\label{problem:P086}

\begin{problem}
\(K_{n}\) consists of real exact-rank-three matrices L=U\(\Sigma V^{T}\) whose singular values lie in {[}1,2{]}, whose left and right leverage scores are at most 6/n, and for which every unit a in \(\mathbb R^{3}\) gives at least ceil(3n/4) coordinates of Ua and Va with magnitude at least 1/(100 √(n)).

\(\Gamma\)⊆{[}n{]}\(\times\){[}n{]} is \(\alpha\)-admissible when every row and column contains at most floor(\(\alpha\)n) elements of \(\Gamma\).

\(\beta_{n}(\alpha\)) is the infimum of ‖\(P_{([n]\times[n]) minus (\Gamma_{1} union \Gamma_{2})}\)(L-\(L_{\prime}\))‖\(_{F}\)/‖L-\(L_{\prime}\)‖\(_{F}\) over distinct L,\(L_{\prime}\) in \(K_{n}\) and separately \(\alpha\)-admissible \(\Gamma_{1},\Gamma_{2}\).

\(\tau_{n}\)=inf\{\(\alpha\) in {[}0,1{]}:\(\beta_{n}(\alpha\))=0\}, with value +\(\infty\) if the set is empty.

\(\kappa_{n}\) is the infimum of \(\alpha\) for which distinct L,\(L_{\prime}\) in \(K_{n}\) and separately \(\alpha\)-admissible \(\Gamma_{1},\Gamma_{2}\) satisfy \(P_{([n]\times[n]) minus (\Gamma_{1} union \Gamma_{2})}\)(L-\(L_{\prime}\))=0.

For n=8m, repeat each phase \(θ_{j}\)=2πj/8 exactly m times and give every corresponding row of F the value (1,√(2) cos \(θ_{j}\),√(2) sin \(θ_{j}\))/√(n).

Dimensions satisfy n at least 32 and tend to \(\infty\); every clean matrix is real, n-by-n, and of exact rank three.

Each clean matrix has a singular-value decomposition L=U Sigma \(V^{T}\) with all singular values in {[}1,2{]} and every left and right leverage score at most 6/n.

Fix \(c_{s}\)=1/100 and \(\gamma_{s}\)=3/4. For every unit a in \(\mathbb R^{3}\), at least ceil(3n/4) coordinates of Ua and at least ceil(3n/4) coordinates of Va have magnitude at least 1/(100 √(n)).

For each fixed C\(>\)0 and deterministic schedule \(p_{n}\) in (0,1{]} with \(p_{n}\) at least C log(n)/n eventually, entries are observed independently with probability \(p_{n}\). Random-observation moduli use \(q=p_{n}\), full-observation moduli use q=1, empirical-density normalization is excluded, and the conjecture asks whether one absolute \(C_{0}\) makes the conclusion hold for every C at least \(C_{0}\); no sampling stability is assumed.

The two competing corruption supports are subsets of the observed set and are separately \(\alpha\)-admissible in every realized row and column; their locations and unbounded values may depend on the clean matrices and the observed set.

Real n-by-n exact-rank-three matrices satisfying A1-A3, dimensions n=8m tending to \(\infty\), complete observation with q=1, and two separately degree-capped deletion supports as authorized by A5. The lower bound ranges over all endpoint pairs in \(K_{n}\), not only phase-structured pairs. Bernoulli sampling, estimators, and noise are excluded.

Dimensions tend to \(\infty\) through n=8m with m at least 4.

Set \(\Omega\)={[}n{]}\(\times\){[}n{]} and q=1 throughout.

Each of \(\Gamma_{1}\) and \(\Gamma_{2}\) separately has at most floor(\(\alpha\)n) entries in every row and column.

The infimum defining \(\beta_{n}\) ranges over every distinct pair in \(K_{n}\) and includes normalized limits of coalescing pairs.

Distinguish \(\tau_{n}\), defined through \(\beta_{n}\)=0, from the finite-collision threshold \(\kappa_{n}\).

For complete observation in dimensions n divisible by eight, determine whether the finite-collision and zero-modulus thresholds of \(K_{n}\) converge to 1/8 and whether every fixed margin below 1/8 has a dimension-uniform positive trimmed secant modulus.
\end{problem}

\clearpage
\section{P087: The hard-safe signaling coefficient}
\label{problem:P087}

\begin{problem}
Set s\(\star(x)\)=2x/3 and u\(\star(x)\)=-x/3 for \(\mid\) x\(\mid \leq\)3/4; set s\(\star(x)\)=(x+1/4)/2 and u\(\star(x)\)=-(x-1/4)/2 for x\(\geq\)3/4, with the reflected formulas for x\(\leq\)-3/4.

\(V_{\mathrm{PO}}(η)\) and \(V_{\mathrm{FS}}\) are the infima specified by A5, and direct full-state integration gives \(V_{\mathrm{FS}}\)=43/384.

For η\(>\)0 with finite \(V_{\mathrm{PO}}(η)\), set \(\mathbb R(η)=(V_{\mathrm{PO}}(η)-V_{\mathrm{FS}})/η^{2}\).

Set \(C_{hyb}\)=7/24, the sum of the formal interior-overlap contribution 1/8 and outer exact-quantization contribution 1/6.

X0 is uniform on {[}-1,1{]}. Independently, N is uniform on {[}-η,η{]} for 0\(<\) η\(\leq\)3/4, and N=0 when η=0.

Controller 1 observes X0 and chooses U0, after which X1=X0+U0. Controller 2 observes Y=X1+N and chooses U1, after which X2=X1+U1 and the horizon ends.

The fixed bounds are \(\mid\) U0\(\mid \leq1/2, \mid\) U1\(\mid \leq1/2, \mid\) X1\(\mid \leq\)3/4, and \(\mid\) X2\(\mid \leq\)1/4.

A partial-information behavioral policy consists of Borel probability kernels K0(du0\(\mid\) x) specified for every x in {[}-1,1{]} and K1(du1\(\mid\) y) specified for every y in the real line. For every x, every u0 in the topological support of K0 given x, every n in {[}-η,η{]} including endpoints, y=x+u0+n, and every u1 in the topological support of K1 given y, all bounds in A3 must hold. Private randomization is allowed, and Controller 2 has no access to X0, U0, or Controller 1's randomization except through Y.

\(V_{\mathrm{PO}}(η)\) is the infimum of E{[}\(U0^{2}+U1^{2}+X2^{2}\){]} over policies admissible under A1-A4. \(V_{\mathrm{FS}}\) is the analogous infimum when Controller 2 observes X1 exactly and uses a pointwise Borel kernel specified for every X1, with all other bounds and supportwise requirements unchanged. The infimum of an empty admissible class is +\(\infty\), and the value gap is considered only when both values are finite.

The fixed two-stage scalar model of A1-A5 for sufficiently small positive η, with endpoint-inclusive supportwise safety and the complete class of randomized, asymmetric, discontinuous, nonmonotone, and η-dependent Borel behavioral policies.

Only η decreasing to zero through positive values for which the partial-information class is nonempty is considered.

No determinism, monotonicity, finite-range, continuity, or prescribed-cell restriction may be imposed on either controller.

The number 7/24 must be derived from the optimized problem, not assumed through a hybrid-policy ansatz.

Under A1-A5, prove or refute that the full-policy small-noise value gap satisfies lim as η decreases to zero of (\(V_{\mathrm{PO}}(η)-V_{\mathrm{FS}})/η^{2}\) = 7/24.
\end{problem}

\clearpage
\section{P088: Convergence of variable-kernel Bregman proximal gradients}
\label{problem:P088}

\begin{problem}
For a differentiable convex kernel q, \(D_{q}(u,v)\)=q(u)-q(v)-\(< \nabla\) q(v),u-v\(>\).

F=f+g on C.

\(Xstar=\operatorname*{arg\,min}_{x \in C}\) F(x).

\(x_{k+1}\) is the unique minimizer over x in C of g(x)+\(< \nabla\) f(\(x_{k}\)),x-\(x_{k} >+(1/\tau_{k})D_{h_{k}}\)(x,\(x_{k}\)).

C is a nonempty open convex subset of \(\mathbb R^{2}\), and all iterates and comparison minimizers in the claim lie in C.

The reference h and every \(h_{k}\) are proper closed Legendre functions with int(dom h)=int(dom \(h_{k}\))=C and are twice continuously differentiable on C. For each k and \(x_{k}\) in C, the displayed proximal-gradient subproblem has a unique minimizer \(x_{k+1}\) in C.

g:\(\mathbb R^{2} \to(-\infty,+\infty\){]} is proper lower-semicontinuous and convex, f:C\(\to \mathbb R\) is continuously differentiable and convex, F=f+g is proper on C and bounded below there, and X\(\star=\operatorname*{arg\,min}_{x \in C}\)F(x) is nonempty.

There is a finite constant L\(>\)0 such that L \(h_{k}\)-f and L \(h_{k}\)+f are convex on C for every k.

There are constants 0\(<\) m\(\leq\)M\(< \infty\) such that m \(D_{h}(u,v)\leq D_{h_{k}}(u,v)\leq\)M \(D_{h}(u,v)\) for every integer k\(\geq\)0 and every u,v in C.

The iteration starts from \(x_{0}\) in C, is exact, uses \(\tau_{k}\) in one fixed interval {[}\(\tau_{min},\tau_{max}\){]} with 0\(< \tau_{min} \leq \tau_{max} <\)1/L, and the Euclidean orbit (\(x_{k}\)) is bounded.

Finite-dimensional convex optimization on the common open domain C contained in \(\mathbb R^{2}\), with exact updates, uniformly bounded stepsizes, two-sided relative smoothness, and all-pairs comparison of every changing Bregman divergence with one fixed reference divergence. No summable kernel-variation condition or additional Euclidean conditioning may be assumed.

The positive target quantifies over every instance satisfying A1-A6 and every bounded exact orbit generated by that instance.

A refuting instance must verify A5 for every ordered pair u,v in C and every k, not merely along the orbit.

A refuting instance must have nonempty Xstar contained in C and every finite iterate must remain in C.

Under exactly A1-A6, decide whether every bounded exact variable-kernel Bregman proximal-gradient orbit converges in Euclidean norm to a single point of Xstar. If this universal assertion is false, exhibit a fully assumption-checked two-dimensional instance whose orbit does not converge to any point of Xstar.
\end{problem}

\clearpage
\section{P089: The endpoint stepsize under arbitrary delays}
\label{problem:P089}

\begin{problem}
x\(\star=(b_{1}+b_{2}+b_{3}\))/3.

Write \(p_{j}=(c_{j}^{x},c_{j}^{y}\)). At the start of round k, set \(u_{10}^{k}=p_{k-1}\) if k\(\geq\)1 and \(d_{k-1}\)=0 and set it to zero otherwise; set \(u_{20}^{k}=p_{k-2}\) if k\(\geq\)2 and \(d_{k-2}\)=1 and set it to zero otherwise; set \(u_{11}^{k}=p_{k-1}\) if k\(\geq\)1 and \(d_{k-1}\)=1 and set it to zero otherwise. Their receiver labels are determined by j modulo three. The slots \(u_{20}\) and \(u_{10}\), in that order when present, encode \(Q^{k}\){[}0{]}, while \(u_{11}\) encodes \(Q^{k}\){[}1{]}.

At the start of round k define \(L^{k}=(x^{k}\)-x\(\star 1,y^{k},z^{k}\)-x\(\star 1,h^{k}\)-x\(\star 1,e_{x}^{k},e_{y}^{k},u_{10}^{k},u_{20}^{k},u_{11}^{k}\)), in this order. The first six entries are three-vectors and the last three are paired scalar payloads, giving 24 scalar coordinates.

\(E_{k}\) is the maximum absolute value of the 24 scalar coordinates of \(L^{k}\).

Agents are indexed modulo three, with successor s(i)=i+1 and predecessor p(i)=i-1. The primal channel uses half self-weight and half of the stored predecessor reconstruction, while the tracking channel retains half of an active agent's post-delivery tracker mass and sends the other half to its successor.

The computing agent at round k is \(a_{k}\)=1+(k mod 3). Only \(a_{k}\) performs a primal and gradient-difference computation; packet deliveries do not count as activations.

At the end of round k, the active agent forms one paired packet \(P_{k}=(a_{k}\),s(\(a_{k}),c_{k}^{x},c_{k}^{y}\)) and chooses \(d_{k}\) in \{0,1\}; \(P_{k}\) is consumed at the start of round k+1+\(d_{k}\), before that round's computation. The delay word is an arbitrary element of \{0,1\} indexed by the nonnegative integers and need not be periodic. At the start of round k, \(Q^{k}\){[}0{]} and \(Q^{k}\){[}1{]} are multisets of packets scheduled for rounds k and k+1. Every packet in \(Q^{k}\){[}0{]} is delivered once and removed. After computation, the remaining queue is shifted and \(P_{k}\) is appended according to \(Q^{k+1}{[}0{]}=Q^{k}\){[}1{]} multiset-union \{\(P_{k}\) if \(d_{k}\)=0\} and \(Q^{k+1}\){[}1{]}=\{\(P_{k}\) if \(d_{k}\)=1\}. The paired components share \(d_{k}\), no packet is consumed in its formation round, and packets on each edge are delivered FIFO.

Let \(z_{r}^{k}\) be receiver r's stored reconstruction of its predecessor's primal state, \(h_{i}^{k}\) sender i's primal encoder reference, and \(e_{i}^{x,k}, e_{i}^{y,k}\) sender i's primal and tracking compression residuals. At the start of round k, for each receiver r set bar \(z_{r}^{k}=z_{r}^{k}\) plus the sum of \(c^{x}\) over packets in \(Q^{k}\){[}0{]} addressed to r, and bar \(y_{r}^{k}=y_{r}^{k}\) plus the corresponding sum of \(c^{y}\). After these packets are consumed, set \(z_{r}^{k+1}\)=bar \(z_{r}^{k}\) for every r, so every delivered primal increment persists in the stored reconstruction. For \(i=a_{k}\) set \(x_{i}^{k+1}=(x_{i}^{k}\)+bar \(z_{i}^{k})/2-\alpha\) bar \(y_{i}^{k}\) and \(y_{i}^{k+1}\)=bar \(y_{i}^{k}\)/2+{[}gradient \(f_{i}(x_{i}^{k+1}\))-gradient \(f_{i}(x_{i}^{k}\)){]}; for j not equal to i set \(x_{j}^{k+1}=x_{j}^{k}\) and \(y_{j}^{k+1}\)=bar \(y_{j}^{k}\). The active sender forms \(v_{i}^{x}=x_{i}^{k+1}-h_{i}^{k}+e_{i}^{x,k}, c_{k}^{x}=qv_{i}^{x}, e_{i}^{x,k+1}=v_{i}^{x}-c_{k}^{x}\), and \(h_{i}^{k+1}=h_{i}^{k}+c_{k}^{x}\), together with \(v_{i}^{y}\)=bar \(y_{i}^{k}/2+e_{i}^{y,k}, c_{k}^{y}=qv_{i}^{y}\), and \(e_{i}^{y,k+1}=v_{i}^{y}-c_{k}^{y}\). All inactive encoder references and residuals remain unchanged, and \(P_{k}\) is inserted into the shifted queue exactly as prescribed in A3.

Each objective is \(f_{i}(x)=(x-b_{i})^{2/2}\) on the real line, where the offsets \(b_{i}\) are arbitrary, and the target optimizer is x\(\star=(b_{1}+b_{2}+b_{3}\))/3.

The stepsize \(\alpha\) is constant and positive and q belongs to (0,1{]}. Initial primal states are arbitrary; \(y_{i}^{0}=x_{i}^{0}-b_{i}, z_{i}^{0}=x_{p(i)}^{0}, h_{i}^{0}=x_{i}^{0}\), both residuals are zero, and \(Q^{0}\){[}0{]} and \(Q^{0}\){[}1{]} are empty.

The claim concerns the horizontal endpoint slice \(\alpha\)=1/2, all q in {[}1/2,1), and every arbitrary admissible binary delay word. It is distinct from the established frozen-implies-switched result, whose stepsize scope excludes \(\alpha\)=1/2, and from the previous q=1 frozen-boundary candidate.

Fix \(\alpha\)=1/2 and let 1/2\(\leq\)q\(<\)1.

Uniformity is over every infinite delay word d in \{0,1\}\(^{N_{0}}\), without periodicity or frequency restrictions.

Convergence and failure are evaluated on the canonical lifted states generated from A6 initializations; an obstruction must affect a declared lifted coordinate.

The geometric constants may depend on q and need not remain bounded as q approaches 1.

For the recurrence fixed by A1-A6 with \(\alpha\)=1/2, every strict compression factor q in [1/2,1) yields delay-word-uniform geometric convergence of the complete initialized lifted error state, although the uncompressed endpoint q=1 has a reachable periodic failure.
\end{problem}

\clearpage
\section{P090: Strong convergence of Douglas--Rachford shadows}
\label{problem:P090}

\begin{problem}
For a positive-definite operator Q and operator C, \(J_{C}^{Q}=(I+Q^{-1}C)^{-1}\). For \(C=N_{U}\) this is the Q-orthogonal projection onto U.

At iteration k, \(u_{k}\) approximates \(J_{A}^{M_{k}} z_{k}, v_{k}\) approximates \(J_{B}^{M_{k}}(2u_{k} - z_{k}\)), and \(z_{k+1}=z_{k}+\lambda_{k}(v_{k} - u_{k}\)), as specified in A7.

The shadow converges strongly to p when \(\lim_{k\to\infty}\)‖\(u_{k} -\)p‖=0 in the ambient Hilbert norm.

H is a separable infinite-dimensional real Hilbert space.

U and V are closed linear subspaces of H, \(A=N_{U}, B=N_{V}\), and zer(A+B)=U\(\cap\)V is nonempty.

Each \(M_{k}\) is bounded, self-adjoint, and positive definite, and fixed constants 0\(<\) m\(\leq\)M\(< \infty\) satisfy mI\(\preccurlyeq M_{k}\preccurlyeq\)MI for every k.

There are numbers \(η_{k} \geq\)0 such that \(M_{k+1}\preccurlyeq(1+η_{k})M_{k}\) for every k and ∑\(_{k\geq0} η_{k} < \infty\).

The two metric-resolvent evaluations at iteration k have additive errors bounded in the ambient Hilbert norm by \(\varepsilon_{k}^{A}\) and \(\varepsilon_{k}^{B}\), where ∑\(_{k\geq0}(\varepsilon_{k}^{A}+\varepsilon_{k}^{B})< \infty\).

The relaxation parameters satisfy 0\(< \lambda_{min} \leq \lambda_{k} \leq \lambda_{max} <\)2 for every k.

Starting from arbitrary \(z_{0} \in\)H, choose \(u_{k}\) with ‖\(u_{k} - J_{A}^{M_{k}} z_{k}\)‖\(\leq \varepsilon_{k}^{A}\), choose \(v_{k}\) with ‖\(v_{k} - J_{B}^{M_{k}}(2u_{k} - z_{k}\))‖\(\leq \varepsilon_{k}^{B}\), and update \(z_{k+1}=z_{k}+\lambda_{k}(v_{k} - u_{k}\)); all norms and convergence claims use the ambient Hilbert norm.

Separable infinite-dimensional real Hilbert spaces, two closed linear subspaces, uniformly coercive variable metrics satisfying summable one-sided Loewner drift, ambient-norm summable resolvent errors, and relaxations uniformly contained in (0,2). No operator-norm metric-variation, compactness, angle-gap, weak-convergence, or commutativity assumption is added.

The problem quantifies over exactly the A1--A7 data and orbits, without strengthening A4 to summable operator-norm variation.

All error bounds and convergence conclusions use the original Hilbert norm.

A refuting orbit need only fail ambient-norm convergence to U\(\cap\)V; weak convergence is neither assumed nor required.

Metrics and subspaces need not commute or admit a common block diagonalization.

Decide whether every orbit satisfying exactly A1–A7 has an ambient-norm limit \(p\in U\cap V\). Equivalently, determine whether an admissible orbit can have a shadow sequence \(u_k\) that fails to converge in ambient norm to every member of \(U\cap V\).
\end{problem}

\clearpage
\section{P091: Alternating moment circuits on an elliptic curve}
\label{problem:P091}

\begin{problem}
For D=\{\(P_{1}\),\ldots,\(P_{N_{d}}\)\}, \(a_{D}=(a_{1}\),\ldots,\(a_{N_{d}}\)) is the unique kernel generator satisfying \(\sum_{i} a_{i}\) f(\(P_{i}\))=0 for every f in \(V_{d}\), normalized by \(\sum_{a_{i}>0} a_{i}=\sum_{a_{i}<0}(-a_{i}\))=1.

The atomic-weight margin is \(\omega_{d}(D)=\min_{i} \mid a_{i} \mid\).

If the points are in cyclic order, \(\delta(D)\) is the minimum arclength of a consecutive cyclic arc divided by the total arclength of E.

Define \(\mu_{D}=\sum_{a_{i}>0} a_{i} \delta_{P_{i}}\) and \(\nu_{D}=\sum_{a_{i}<0}(-a_{i}) \delta_{P_{i}}\). Alternation and even \(N_{d}\) give exactly \(r_{d}\) atoms in each probability measure, and the circuit identity gives equality of all \(V_{d}\) moments.

The canonical robust lift assigns \(\mu_{D}\) and \(\nu_{D}\) to two clique-local atomic measures with private coordinates zero, retains moments through degree \(2r_{d}\), and couples only their common \(V_{d}\) separator moments.

Put \(p_{d}=N_{d}\)-2 and \(\mathbb R_{d}(D)=\omega_{d}(D)/\delta (D)_{d}^{p}\). The predicate \(P_{d}\) is: \(\in f_{D \in A_{d}} \mathbb R_{d}(D)>\)0; there exists a sequence \(D_{k}\) in \(A_{d}\) with \(\delta(D_{k}\)) tending to zero and \(\sup_{k} \mathbb R_{d}(D_{k})< \infty\); and every D in \(A_{d}\) has the enumerated canonical flat-lift properties.

The correlative sparsity graph admits a finite clique tree satisfying the running-intersection property.

Each clique has a sufficiently high-order normalized truncated moment sequence satisfying its moment and localizing positivity constraints and an explicit flatness rank equality that yields a unique finitely atomic clique representing measure.

Every projected clique measure on a separator has at most r distinct atoms, all lying on the fixed oval E.

For each clique-tree edge, with d \(\geq\) 1, independent clique sequences are coupled by equality of the restrictions to E of all real polynomials of total degree at most d, and no higher-degree separator moments are implicitly identified merely because they are retained locally for flatness.

All clique-local feasible sets are compact, and the equality \(y^{2} = (1-x^{2}\))(2-x) together with -1 \(\leq\) x \(\leq\) 1 is imposed on each relevant separator projection.

For odd d, \(A_{d}\) consists of cyclically ordered sets of \(N_{d}\) distinct points whose \(V_{d}\) evaluation map has a one-dimensional left kernel generated by a vector with nonzero coefficients alternating in cyclic order. For even d, \(A_{d}\) consists of reduced all-oval zero divisors of nonzero real degree-d sections with their unique alternating residue dependence. Coefficients are normalized so each sign part has mass one. Constants may depend on d but not on D. The canonical lift has two cliques, private coordinates fixed to zero, local moments through degree \(2r_{d}\), and cross-clique coupling on exactly \(V_{d}\).

\(N_{d}\)=3d+1 for odd d and \(N_{d}\)=3d for even d, and every admissible circuit contains exactly \(N_{d}\) distinct cyclically ordered points.

The \(V_{d}\) evaluation kernel on D is one-dimensional and has a generator with nonzero coefficients alternating in cyclic order.

\(\delta(D)\) is the minimum intrinsic arclength between consecutive cyclic points of D divided by the total arclength of E.

The sparse lift consists of two cliques with separator variables (x,y), one private coordinate fixed to zero in each clique, atomic moments through degree \(2r_{d}\), and rank \(M_{r_{d}}\)=rank \(M_{r_{d}-1}=r_{d}\).

For \(p_{d}=N_{d}\)-2 and \(\mathbb R_{d}(D)=\omega_{d}(D)/\delta (D)_{d}^{p}, P_{d}\) is the conjunction of a positive uniform infimum for \(\mathbb R_{d}\), existence of a degenerating sequence on which \(\mathbb R_{d}\) is bounded above, and the explicitly enumerated canonical-lift conclusion for every D.

For each d\(\geq\)1, set \(N_{d}\)=3d+1 for odd d and \(N_{d}\)=3d for even d, \(r_{d}=N_{d}\)/2, and \(p_{d}=N_{d}\)-2. Let \(A_{d}\) be the admissible minimal alternating \(V_{d}\)-circuits specified below and \(\mathbb R_{d}(D)=\omega_{d}(D)/\delta (D)_{d}^{p}\). Decide the single predicate \(P_{d}\) for every d: (i) \(\in f_{D \in A_{d}} \mathbb R_{d}(D)\)>0; (ii) there is a sequence \(D_{k}\) in \(A_{d}\) with \(\delta(D_{k}\)) tending to zero and \(\sup_{k} \mathbb R_{d}(D_{k})<\infty\); and (iii) for every D in \(A_{d}\), its two normalized sign measures have exactly \(r_{d}\) atoms, agree on \(V_{d}\), obey the same weight lower bound, and their explicitly defined canonical two-clique lift satisfies A1-A5, rank \(M_{r_{d}}\)=rank \(M_{r_{d}-1}=r_{d}\) in both cliques, and has no global representing measure with both distinct separator marginals. Thus the positive assertion is exactly forall d \(P_{d}\) and the refutation assertion is exactly exists d not \(P_{d}\).
\end{problem}

\clearpage
\section{P092: Quantitative compactification of a Maxwell graph}
\label{problem:P092}

\begin{problem}
\(D_{z}(y,x)=h_{z}(y)-h_{z}(x)-h_{z}\)'(x)(y-x).

\[
\mathbb R_{z}(y)=f(y)-f(0)+D_{z}(y,0)
\]

\(A(y)=\partial_{z} D_{z}(y,0)\) evaluated at z=0.

\[
\begin{aligned}
s_z(x)&=h_z'(x)-h_z'(0),\\
f(y)+D_z(y,x)-f(0)-D_z(0,x)
&=\mathbb R_z(y)-s_z(x)y.
\end{aligned}
\]

For real v, \(v_{+}\)=max(v,0) and \(v_{-}\)=max(-v,0).

\(QEG_{q,\beta}\) holds if there exist \(\delta\in(0,\rho), \varepsilon>0, \gamma>0, L<\infty, c_+>0, c_->0\), and real \(b_+,b_-\) such that the signwise errors in
\[
\mathbb R_0(y)=c_+(y_+)^{2q}+c_-(y_-)^{2q}
+O(|y|^{2q+\beta})
\]
and
\[
A(y)=b_+(y_+)^q+b_-(y_-)^q+O(|y|^{q+\beta})
\]
are bounded by \(L\) with those powers; \(|\partial_{zz}D_z(y,0)|\leq L(|z|+|y|)^\beta\) on the local box; \(\Psi(v)=c_+(v_+)^{2q}+c_-(v_-)^{2q}+b_+(v_+)^q+b_-(v_-)^q\) has a unique nonzero global minimizer; and \(\mathbb R_a(y)\geq\gamma\) for \(0<a\leq\varepsilon\) and \(\delta\leq|y|\leq\rho\).

An admissible \(QEG_{q,\beta}\) instance is a pair (f,h) satisfying the five cited base assumptions and every clause of \(QEG_{q,\beta}\).

Z(a,s)=argmin over y in \(I_{\rho}\) of \(\mathbb R_{a}(y)\)-sy and V(a,s)=min over y in \(I_{\rho}\) of \(\mathbb R_{a}(y)\)-sy. The original localized proximal set is P(a,x)=Z(a,\(s_{a}(x)\)).

For \(a>0\) and \((a,s)\neq(0,0)\), set
\[
\begin{aligned}
r(a,s)&=a^{1/q}+|s|^{1/(2q-1)},\\
\alpha(a,s)&=\frac{a}{r(a,s)^q},&
\xi(a,s)&=\frac{s}{r(a,s)^{2q-1}}.
\end{aligned}
\]
and \(p(a,s)=(\alpha(a,s),\xi(a,s))\in K_q\).

\(C(v)=c_+(v_+)^{2q}+c_-(v_-)^{2q}, B(v)=b_+(v_+)^q+b_-(v_-)^q\), and \(G_p(v)=C(v)+\alpha B(v)-\xi v\) for \(p=(\alpha,\xi)\). Define \(\mathcal M(p)=\operatorname*{arg\,min}_{v\in\mathbb R}G_p(v)\) and \(\mu(p)=\min_{v\in\mathbb R}G_p(v)\).

\(\operatorname{Graph}\mathcal M\) is the set of \((p,v)\) with \(p\in K_q\) and \(v\in\mathcal M(p)\). Distances to \(\operatorname{Graph}\mathcal M\) and between parameter points are Euclidean.

\(X_q\) is the set of \(p\in K_q\) for which \(\mathcal M(p)\) contains at least two points.

For nonempty subsets S,T of the real line, exc(S,T)=sup over u in S of inf over v in T of \(\mid\) u-v\(\mid\).

\(BCG_{\beta}\) holds at \(p\star\) with profile J if for every \(r_{n}\) decreasing to zero and \(p_{n}=(\alpha_{n},xi_{n}\)) in \(K_{q}\) converging to \(p\star\), the functions \(r_{n}^{(-\beta)}\)\{\(r_{n}^{(-2q)}{[}\mathbb R\)\_(\(\alpha_{n} r_{n}^{q})(r_{n}\) v)-\(xi_{n} r_{n}^{(2q)}\)v{]}-\(G_{p_{n}}(v)\)\} converge locally uniformly in v to J(v).

For an admissible instance and p\(\star=(\alpha_{\star},xi_{\star}\)), Clust(\(p\star\)) is the set of v for which there exist \(r_{n}\) decreasing to zero and \(y_{n}\) in Z(\(\alpha_{\star} r_{n}^{q},xi_{\star} r_{n}^{(2q-1)}\)) such that \(y_{n}/r_{n}\) tends to v.

The ambient space is the real line, the anchor center and kernel state are (x,z) = (0,0), \(\lambda\) = 1, and minimization is localized to a fixed interval {[}-\(\rho,\rho\){]}.

The objective f is proper and lower semicontinuous, is finite at zero with f(0) = 0, and has zero in its limiting subdifferential at zero.

Relative to \(h_{0}\), f is relatively prox-regular at zero for the zero subgradient, and its least local modulus is \(r\star\) = 1, so \(\lambda r\star\) = 1 exactly; no strict subcritical margin may be assumed.

The map (z,y) to \(h_{z}(y)\) is jointly C2 near (0,0); each \(h_{z}\) extends to a supercoercive Legendre function; the y-curvatures have common positive local lower and upper bounds; \(h_{0}(y) = y^{2/2}\); and affine normalization of \(h_{z}\) is mathematically immaterial.

The localized anchor problem f(y) + \(D_{h_{0}}(y,0)\) has zero as its unique minimizer in {[}-\(\rho,\rho\){]}, with zero in the interval's interior.

Scalar minimization over \(I_{\rho}={[}-\rho,\rho\){]}, positive kernel states a tending to zero, and centers represented by the affine-invariant tilt \(s_{a}(x)\). The target is joint in state and center and is mathematically distinct from the established fixed-center endpoint rate.

Only positive kernel states a tending to zero are considered.

Center dependence is represented by \(s_{a}(x)=h_{a}\)`(x)-\(h_{a}\)'(0).

Uniform control through a Maxwell transition is measured by distance to the graph of the limiting argmin correspondence.

Fiberwise control is asserted only when the compact parameter is at distance at least \(\omega(r)\) from the Maxwell set, where \(\omega(r)/r^{\beta}\) tends to \(\infty\).

Deterministic Maxwell branch selection is asserted only under \(BCG_{\beta}\).

The target concerns all joint state-center paths, Maxwell switching, and branch realizability, not only s=0.

The positive and refutation predicates explicitly define every sequence, set, estimate, limit, and failure alternative used in their verification.

Assume \(QEG_{q,\beta}\). Prove the uniform value and argmin-graph estimates, the off-Maxwell fiber estimate, the \(BCG_{\beta}\) detuned branch-selection and value-correction limits, and the two-well Maxwell classification and exact branch-realization assertions stated in \(TARGET_{QUANTITATIVE,MAXWELL,GRAPH}\). Every comparison object and quantified negation is defined below, so neither the positive nor refutation target relies on an omitted predicate.
\end{problem}

\clearpage
\section{P093: A condition-corrected four-cycle constant}
\label{problem:P093}

\begin{problem}
For every admitted instance define \(t=\lambda/\kappa\); membership in the class indexed by t requires \(\lambda/\kappa\)=t.

For a fixed admitted instance I and \(\rho >\)0, let R\(\rho(I;t,η)\) be the supremum of r/q over its admissible updates with \(x_{k}\) and \(x_{k+1}\) in the radius-\(\rho\) ball centered at \(\bar{x}\). Define \(Q(I;t,η)=limsup_{\rho\downarrow0}\)R\(\rho(I;t,η)\), and define Q₄,K\(\star(t,η)=\sup_{I}\) Q(I;t,η), where I ranges over fixed admitted instances with \(\lambda/\kappa\)=t and cond₂(DT(\(\bar{x}))\leq\)K.

Set a=1-η\(^2\). For B\(\in\)B₄,2D(K) and a unit vector h\(\in\)R\(^2\) define \(\Phi_{t,η}(B,h)\)=‖h+(t/a)Bh‖-(ηt/a)‖Bh‖.

Define m₄,2D(K,t,η)=inf\{\(\Phi_{t,η}(B,h)\): B\(\in\)B₄,2D(K), ‖h‖=1\}.

For an admitted instance put A=DT(\(\bar{x}\)) and \(B=\kappa\)A. Instancewise differentiation of A3 gives \(s_{min}(B)\geq\)1, while cond₂(B)=cond₂(A)\(\leq\)K.

H is a real Hilbert space.

T:H⇒H is maximal monotone and \(S=T^{-1}(0)\) is nonempty.

There are a declared constant \(\kappa >\)0 and a neighborhood U of a reference zero such that dist(x,S)\(\leq \kappa\) dist(0,T(x)) for every x\(\in\)U; \(\kappa\) is a certified bound defining the operator class and is not assumed to be the least possible modulus.

The proximal parameter \(\lambda\) is a fixed positive scalar.

Each tested update selects \(v_{k+1} \in\)T(\(x_{k+1}\)) and an arbitrary error \(e_{k+1}\) satisfying \(x_{k}-x_{k+1}=\lambda v_{k+1}+e_{k+1}\) and ‖\(e_{k+1}\)‖\(\leq\)η‖\(x_{k}-x_{k+1}\)‖ for a fixed η\(\in\){[}0,1).

Every local one-step assertion is restricted to updates with \(x_{k+1} \in\)U; no multi-step convergence assertion is authorized unless all graph points at which metric subregularity is used remain in U.

All finite-dimensional real Hilbert spaces and fixed A1--A6 instances for which T is single-valued and continuously Fréchet differentiable near an isolated reference zero \(\bar{x}\), A=DT(\(\bar{x}\)) is invertible with cond₂(A)\(\leq\)K, and the local graph is four-cyclically monotone. The localization limit is taken separately for each fixed instance before the supremum over instances.

There is a neighborhood V⊆U of an isolated reference zero \(\bar{x}\) on which T is single-valued and C¹, A=DT(\(\bar{x}\)) is invertible, and cond₂(A)=‖A‖/\(s_{min}(A)\leq\)K.

For all x₁,x₂,x₃,x₄\(\in\)V, with x₅=x₁, one has \(\Sigma_{i=1}^{4}\) ⟨T(\(x_{i}),x_{i}-x_{i+1}\)⟩\(\geq\)0.

Every instance entering Q₄,K\(\star(t,η)\) satisfies \(\lambda/\kappa\)=t, cond₂(DT(\(\bar{x}))\leq\)K, and the declared η.

For each fixed instance, take the limsup as the localization radius decreases to zero before taking the supremum over different operators and neighborhoods.

For each fixed instance, the localized supremum uses A5 updates with \(x_{k}\) and \(x_{k+1}\) in the shrinking ball, q=dist(\(x_{k}\),S)\(>\)0, and r=dist(\(x_{k+1}\),S)\(>\)0.

Fix K\(\geq\)1, t>0, and η\(\in(0,1)\). For each fixed admitted finite-dimensional A1–A6 instance I, take its localized worst-case ratio Q(I;t,η) before taking the supremum Q₄,K*(t,η) over instances with \(\lambda/\kappa\)=t and cond₂(DT(\(\bar{x}))\leq\)K. With m₄,2D(K,t,η) defined by the planar derivative optimization below, the positive predicate is Q₄,K*(t,η)=1/m₄,2D(K,t,η) for every K\(\geq\)1, t>0, and η\(\in(0,1)\). Its logical refutation predicate is the existence of K\(\geq\)1, t>0, and η\(\in(0,1)\) for which Q₄,K\(\star(t,η)\)≠1/m₄,2D(K,t,η).
\end{problem}

\clearpage
\section{P094: A four-step dimension jump}
\label{problem:P094}

\begin{problem}
Let θ = (a, b, c) be the unique element of Θ and define \(h_{bdca}\) = (b, 3/2, c, a).

For an admissible instance I, generate \(x_{0}\) through \(x_{4}\) using \(h_{bdca}\) and A3-A4, and set \(E_{4}(I)\) = {[}f(\(x_{4}\))-f(\(x\star\)){]}/(L \(D^{2}\)).

\(W_{4}(h_{bdca}\)) = sup\{\(E_{4}(I)\): I ranges over the full positive finite-dimensional instance domain\}.

\(W_{4}^{(1)}(h_{bdca}\)) = sup\{\(E_{4}(I)\): I ranges over the dimension-one instance domain\}.

\(P_{dim}\) is the proposition \(W_{4}^{(1)}(h_{bdca}) \leq 1/32< W_{4}(h_{bdca}\)).

L\(>\)0 and D\(>\)0, and the ambient dimension d may be any positive finite integer.

f:\(\mathbb R^{d} \to \mathbb R\) is convex, differentiable, and L-smooth, attains its minimum at \(x\star\), and the initial point satisfies ‖\(x_{0}-x\star\)‖ \(\leq\) D for at least one minimizer.

For a horizon n, h = (\(h_{0}\), \ldots, \(h_{n-1}\)) is any real scalar schedule fixed before observing gradients, and \(x_{t+1} = x_{t}-(h_{t}\)/L) \(\nabla\) f(\(x_{t}\)). Zero and negative entries are not excluded a priori.

The method uses only the current gradient and the fixed scalar schedule: no momentum, memory variables, auxiliary sequences, line search, or adaptivity.

For each h in \(\mathbb R^{n}, W_{n}(h)\) is the supremum of {[}f(\(x_{n}\))-f(\(x\star\)){]}/(L \(D^{2}\)) over the class in A1-A4 and all allowed finite dimensions, and \(V_{n} = \in f_{h \in \mathbb R^{n}} W_{n}(h)\); neither the infimum nor any supremum is assumed to be attained.

The problem concerns one exact algebraic four-step schedule. It compares the complete one-dimensional smooth-convex class with the unrestricted union of positive finite dimensions. It does not ask for the exact value of \(W_{4}(h_{bdca}\)), the global minimax value \(V_{4}\), a schedule atlas, or a permutation-orbit classification.

The schedule is exactly \(h_{bdca}\) = (b, 3/2, c, a) in this order.
\(W_{4}^{(1)}\) ranges over every one-dimensional convex differentiable L-smooth objective satisfying A1-A4, not merely quadratics or clipped quadratics.
\(W_{4}\) retains the A5 supremum over all positive finite dimensions.
Both comparisons use the exact rational threshold 1/32.
A rank-two attaining or feasible interpolation datum does not by itself prove a dimension gap.

Under A1-A5, let (a, b, c) be the unique real triple in the stated isolating box satisfying \(a^{2} = 2, b^{2}\)+a*b-2-2*a = 0, and \(c^{2}\) = 2*a+2*b+3, and fix \(h_{bdca}\) = (b, 3/2, c, a). Let \(W_{4}^{(1)}(h_{bdca}\)) be the supremum defining \(W_{4}(h_{bdca}\)) restricted to ambient dimension d = 1. Decide whether \(W_{4}^{(1)}(h_{bdca}) \leq 1/32<W_{4}(h_{bdca}\)).
\end{problem}

\clearpage
\section{P095: Polynomial arcs in three-state selection regions}
\label{problem:P095}

\begin{problem}
Set h(y)=y(y-1)(y-2), \(L_{i}(y)\)=product over j in I with j different from i of (y-j)/(i-j), and f(x,y)=sum over i in I of \(ell_{i} (x)L_{i}\)(y). Then the feasible states are exactly I and \(f(x,i)=ell_{i}(x)\).

\(\mathbb R_{i}\)=\{x:\(ell_{i}(x)< ell_{j}(x)\) for every j different from i\}, and \(C_{i}\) is the closure of \(\mathbb R_{i}\).

\(T_{i}\) consists of w in \(\mathbb R^{2}\) for which there exist \(x_{k}\) in \(C_{i}\) and positive \(s_{k}\) tending to zero such that \(x_{k}\) tends to xbar and (\(x_{k}\)-xbar)/\(s_{k}\) tends to w.

\(D_{i}^{N}\) is the set of unit limits \(\gamma(t)\)/norm(\(\gamma(t)\)) as t decreases to zero, where \(\gamma(0)\)=xbar, both coordinates of \(\gamma\) are polynomials of degree at most N, \(\gamma(t)\) is nonzero, and \(\gamma(t)\) belongs to \(\mathbb R_{i}\) for every sufficiently small positive t.

For E contained in the unit circle, Rad(E)=\{0\} union \{r e:r\(>\)0 and e belongs to E\}.

The upper objective F and all lower-level objective and constraint data are twice continuously differentiable near the reference parameter and the finitely many reference lower solutions.

At the reference parameter xbar, the lower problem has finitely many global optima and one or more specified associated KKT complementarity tuples at each optimum, and its feasible sets are uniformly level-bounded near xbar. This premise does not assert that the specified tuples exhaust all multiplier tuples.

For every specified reference KKT tuple and every adjacent complementarity face, the corresponding face-restricted KKT generalized equation is uniformly strongly regular on a specified neighborhood, with one common regularity modulus. Strict complementarity is not assumed.

There are pairwise disjoint neighborhoods \(U_{i}\) of the reference lower optima, a common compact localization K, and \(\gamma >\)0 such that all nearby global lower minimizers lie in K and every nearby feasible y in K outside the union of the \(U_{i}\) has lower objective at least \(\gamma\) larger than the minimum lower objective attained over feasible points in that union. This premise does not assert KKT-tuple coverage or branch exhaustiveness.

Let x=(u,v) range near xbar=(0,0) in \(\mathbb R^{2}\) and let d be an integer at least two. The lower feasible set is \{0,1,2\}, represented by h(y)=y(y-1)(y-2)=0. For i in I=\{0,1,2\}, the lower trace \(ell_{i}\) is a real polynomial of total degree at most d, \(ell_{0}(0)=ell_{1}(0)=ell_{2}\)(0), and every pairwise difference \(ell_{i}-ell_{j}\) is nonzero. The lower objective is the Lagrange interpolation polynomial f(x,y)=sum over i of \(ell_{i} (x)L_{i}\)(y), where \(L_{i}(j)\) equals one if i=j and zero otherwise. The upper objective is identically zero. This fixed system satisfies A1-A4. Only exact polynomial access to lower-selection regions is tested; generalized-gradient identities already established in the quartic special case are not restated.

The parameter space is \(\mathbb R^{2}\), the lower feasible set is exactly \{0,1,2\}, and the three lower traces have total degree at most the variable integer d\(\geq\)2.

All three lower traces agree at the origin, and \(ell_{i}-ell_{j}\) is a nonzero polynomial for every distinct i and j.

An accessing arc must be polynomial in its scalar parameter, remain in the strict selection region for every sufficiently small positive parameter, and have each coordinate of degree at most \(d^{2}\).

For planar polynomial lower traces of arbitrary degree d, determine whether every tangent direction of every incident strict three-state selection region is approximable by directions of exact polynomial arcs of coordinate degree at most \(d^{2}\). The fixed three-state equality-constrained lower problem makes branch exhaustiveness automatic, so the only unresolved issue is the effective geometry of two simultaneous polynomial inequalities.
\end{problem}

\clearpage
\section{P096: First variation of exit risk under observation error}
\label{problem:P096}

\begin{problem}
For η\(\geq\)0, F(η) is the supremum of \(P_{x0}(\tau_{D} \leq\)T) over the nonanticipative error sequences authorized by A4 with norm at most η at every update.

\(η_{geom}\)=(1/4)min\{\(w_{col}\),r0,\(\rho_{D}\)\}, with the quantities specified in SC4.

For 0\(<\) η\(< η_{geom}, \mathbb R(η)\)=(F(η)\(-\)F(0))/η.

For Borel B⊂D, P(\(X_{kΔ} \in\)B,\(\tau_{D} >\) kΔ)=∫\(_{B} p_{k}(x)\)dx under exact observations.

\(Q_{k}(x,a)\) is the probability of exit by T when \(X_{kΔ}\)=x, action a is held on {[}kΔ,(k+1)Δ), and exact observations are used at all later updates.

\[
\begin{aligned}
L={}&\sum_{k=1}^{N-1}\int_{S_{\mathrm{reg}}}p_k(z)\\
&\quad\cdot\left|Q_k(z,a^+(z))-Q_k(z,a^-(z))\right|\\
&\quad\mathrm dH^{d-1}(z).
\end{aligned}
\]

\(η_{c}\)=sup\{η\(\geq\)0:F(η)\(\leq \alpha\)\}, with values in {[}0,\(\infty\){]}; endpoint safety is the separate predicate \(η_{c} < \infty\) and F(\(η_{c})\leq \alpha\).

D is a bounded C3 domain with positive reach and bounded principal curvatures; x0 is a fixed point of D with recorded clearance \(\delta\)0=dist(x0,\(\partial\)D)\(>\)0.

U is compact; b and \(\Sigma\) are bounded and uniformly Lipschitz in the state on a neighborhood of the closure of D, uniformly over u; and every progressively measurable U-valued control gives a unique strong solution up to \(\tau_{D}\).

Bounded continuous running cost \(\ell\), terminal cost g, and exit cost q define the fixed H-horizon stopped exit-cost problem. It is stipulated, rather than inferred from the preceding regularity, that this problem admits a Borel time-zero minimizing selector \(\mu_{H}\) on D. A deterministic tie-breaking rule and a fixed boundary action extend \(\mu_{H}\) to the closure of D; measurability is not interpreted as continuity or robustness.

On the filtered probability space carrying W, let \(E_{k}\) be measurable with respect to the completed pre-decision history generated by \(X_{0}\), W up to kΔ, and \(E_{0}\) through \(E_{k-1}\), with norm(\(E_{k})\leq\)η. Set \(Y_{k}=X_{kΔ}+E_{k}\). Let \(\Pi_{D}\) be a fixed Borel nearest-point selector onto the closure of D, equal to the unique metric projection in the reach tube and using deterministic tie-breaking elsewhere. The implemented control is \(u_{t}=\mu_{H}(\Pi_{D}(Y_{k}\))) for t in {[}kΔ,(k+1)Δ)\(\cap\){[}0,T{]} until exit, followed by a fixed cemetery action.

For exact observations \(E_{k}\)=0, the same zero-order-held controller satisfies the pointwise certificate \(P_{x0}(\tau_{D} \leq\)T)\(\leq \alpha\)-s for fixed T, \(\alpha\) in (0,1), and s\(>\)0. No uniform certificate over other initial or reachable replanning states is assumed.

The parameters \(\varepsilon \geq\)0 and 0\(<\) Δ\(\leq\)H\(\leq\)T are explicit, and the boundary-normal covariance n(x)\(^{T} \Sigma(x)\Sigma(x)^{T}\)n(x) is retained as a local quantity; degenerate normal diffusion remains in scope.

One fixed stopped diffusion and zero-order-held controller satisfying A1--A6 is considered. The horizon is T=NΔ with fixed finite N\(\geq\)2, and the effective covariance \(\varepsilon \Sigma \Sigma ^{T}\) is uniformly elliptic. The selector is piecewise constant across finitely many compact C\(^2\) interior hypersurfaces, agrees on each regular interface with one of its two adjacent traces, is constant in a boundary collar, and has only an exceptional junction set whose η-neighborhood has volume o(η). The initial state is separated from the switching set. F remains defined for every η\(\geq\)0, while the right asymptotic is restricted to 0\(<\) η\(< η_{geom}\).

T=NΔ for a fixed integer N\(\geq\)2, and the same selector is sampled at kΔ for k=0,\ldots,N\(-\)1.

For every action used by \(\mu_{H}\), b(\(\cdot\),a) and \(\Sigma\) are \(C^{1,\beta}\) on a neighborhood of the closure of D, and \(\lambda_{eff}\) I\(\leq \varepsilon \Sigma(x)\Sigma(x)^{T} \leq \Lambda_{eff}\) I for fixed 0\(< \lambda_{eff} \leq \Lambda_{eff}\).

There are finitely many Borel action cells with pairwise disjoint interiors whose union is the closure of D. Their entire interior switching set is \(S_{reg} \cup\)J, where \(S_{reg}\) is a finite union of compact C\(^2\) hypersurface pieces, every point of \(S_{reg}\) has exactly two adjacent open cells, \(\mu_{H}\) on \(S_{reg}\) equals one of the two adjacent action traces, and the exceptional compact set J satisfies Leb(\{x\(\in\)D:dist(x,J)\(\leq\)r\})=o(r) as r\(\downarrow\)0.

Record \(w_{col} >\)0 such that \(\mu_{H}\) is constant on \{x\(\in\)D:dist(x,\(\partial\)D)\(< w_{col}\)\}; set r0=dist(x0,\(S_{reg} \cup\)J)\(>\)0, let \(\rho_{D}\) be the reach supplied by A1, and define \(η_{geom}\)=(1/4)min\{\(w_{col}\),r0,\(\rho_{D}\)\}.

For the effectively uniformly elliptic finite-update subclass of A1–A6 whose selector has a geometrically tame finite action partition, determine whether F(η)=F(0)+Lη+o(η) as η\(\downarrow\)0. The coefficient L sums, over positive replanning times and regular switching interfaces, the nominal killed-state density times the absolute jump in conditional exit risk caused by forcing either adjacent action for the next holding interval.
\end{problem}

\clearpage
\section{P097: Attainment of the Maxwell distortion threshold}
\label{problem:P097}

\begin{problem}
Set \(\varepsilon_{k}=2^{(-k)}, x_{k}=\varepsilon_{k}\), h(t)=t log t, f(t)=-h(t)/2, and \(\lambda\)=1.

For differentiable g, \(D_{g}(y,x)\)=g(y)-g(x)-g'(x)(y-x).

For a profile A, let \(g_{A}\) be any twice differentiable function with \(g_{A}\)'\,'(s)=A(s)/s; its affine normalization is immaterial.

\(Lip_{log}(A)\) is the least \(\beta\) such that \(\mid\) log A(s)-log A(t)\(\mid \leq \beta \mid\) log s-log t\(\mid\){} for all s,t in L.

\(\Gamma(A)\) is the infimum over s in I, s not equal to 1, of the minimum of \(D_{g_{A}}(s,1)/D_{h}(s,1)\) and \(D_{g_{A}}(1,s)/D_{h}(1,s)\).

\(Psi_{A,p}(s)=g_{A}(s)\)-h(s)/2-ps on J.

A profile is Maxwell-feasible when it satisfies \(SC_{EXACT,SUPPORT,PROFILE}, \Gamma(A)\geq\)13/20, and there exist p, u, and v satisfying \(SC_{SEPARATED,MAXWELL,PAIR}\) with \(Psi_{A,p}(u)=Psi_{A,p}(v)=\min_{s \in J} Psi_{A,p}(s)\).

\(\beta_{\star}\) is the infimum of \(Lip_{log}(A)\) over all Maxwell-feasible profiles A, with \(\beta_{\star}=\infty\) if none exists.

Set \(h_{k}\)'`(\(\varepsilon_{k}\) s)=A(s)/(\(\varepsilon_{k}\) s) on \(\varepsilon_{k}\) L, set \(h_{k}\)'`(t)=1/t outside that layer, and match h and h' on the left exterior component. Endpoint matching and the two moment conditions give a globally C2 Legendre kernel equal to h on both exterior components.

The ambient space is finite-dimensional, and h and every \(h_{k}\) are Legendre functions with the same nonempty open convex interior domain \(\Omega\).

Each kernel is twice continuously differentiable and has positive-definite Hessian on \(\Omega\), while \(h_{k}\) converges to h in C1 on every fixed compact subset of \(\Omega\); no uniform Hessian or divergence comparison is assumed near the boundary.

The objective f is fixed, proper, lower-semicontinuous, and prox-bounded relative to the reference kernel h.

There are attentive graph points (\(x_{k},v_{k}\)) with \(v_{k}\) in the limiting subdifferential of f at \(x_{k}, x_{k}\) approaching the boundary of \(\Omega\), shrinking neighborhoods \(U_{k}\), and a constant r such that f(z) is at least f(x)+\(<\) v,z-x\(>-rD_{h}(z,x)\) for every sufficiently large k and every authorized attentive graph point (x,v) and z in \(U_{k}\).

The proximal parameter \(\lambda\) is fixed and positive with \(\lambda\) r\(<\)1, and all localized subproblems under consideration remain inside \(\Omega\).

Use exactly A1-A5 in dimension one with \(\Omega=(0,\infty\)), h(t)=t log t, \(\lambda\)=1, f=-h/2, \(\varepsilon_{k}=2^{(-k)}\), and \(x_{k}=\varepsilon_{k}\). Generate \(h_{k}\) from one positive continuous normalized curvature profile A on L={[}1/4,2{]} satisfying endpoint matching and both exact-support moments. Require \(\Gamma(A)\geq\)13/20 on I={[}1/2,3/2{]} and two global minimizers in \(J_{0}\)={[}49/64,55/64{]} separated by at least 1/32. Minimize \(Lip_{log}(A)\) over these profiles.

Set h(t)=t log t, f(t)=-h(t)/2, \(\lambda=1, \varepsilon_{k}=2^{(-k)}\), and \(x_{k}=\varepsilon_{k}\).

The profile A is positive and continuous on L, satisfies A(1/4)=A(2)=1, \(\in t_{L}(A-1)\) ds=0, and \(\in t_{L}(A-1)\) ds/s=0.

The bidirectional anchored divergence functional \(\Gamma(A)\) is at least 13/20.

The two global minimizers u and v lie in \(J_{0}\)={[}49/64,55/64{]} and satisfy v-u\(\geq\)1/32.

Determine whether the minimum log-coordinate distortion needed for an exact-support boundary layer to create a separated Maxwell pair under a strict bidirectional anchored divergence margin is finite and attained.
\end{problem}

\clearpage
\section{P098: The nine-round delay threshold}
\label{problem:P098}

\begin{problem}
Set \(X_{k}=x_{1}^{k}+x_{2}^{k}, U_{k}=x_{1}^{k}-x_{2}^{k}, Y_{k}=y_{1}^{k}+y_{2}^{k}\), and \(V_{k}=y_{1}^{k}-y_{2}^{k}\), with all negative-index sender states equal to zero.

Define \(q_{+}(\lambda)=2lambda^{10}-\lambda^{9}-1, q_{-}(\lambda)=2lambda^{10}-\lambda^{9}+1, F_{+}(\lambda,\alpha)=q_{+} (\lambda)^{2}\)+4alpha \(\lambda^{18}(\lambda\)-1), and \(F_{-}(\lambda,\alpha)=q_{-} (\lambda)^{2}\)+4alpha \(\lambda^{18}(\lambda\)-1). The fixed optimizer mode \(\lambda\)=1 is removed from \(F_{+}\) when testing stability.

For θ in Dphase set u(θ)=2cos(θ)-1+cos(9theta), v(θ)=2sin(θ)-sin(9theta), and g(θ)=2u(θ)v(θ)(cos(θ)-1)-(u\((θ)^{2}\)-v\((θ)^{2}\))sin(θ). Define \(\theta_{9}\) as the unique zero of g in Dphase for which -{[}u(θ)+i v(θ){]}\(^{2}\)/{[}4(exp(i θ)-1){]} is positive real.

\(\alpha_{9}\)={[}u\((\theta_{9})^{2}\)+v\((\theta_{9})^{2}\){]}/{[}8sin(\(\theta_{9}\)/2){]}, numerically about 0.1095702253.

\(E_{k}=\sum_{i} \mid x_{i}^{k}\)-x\(\star\mid{}^{2}+\sum_{i} \mid y_{i}^{k} \mid{}^{2}+\mid \sum_{i} y_{i}^{k}-\sum_{i} \nabla f_{i}(x_{i}^{k})\mid{}^{2}\), where x\(\star=(b_{1}+b_{2}\))/2.

Fix integers n at least 2, d at least 1, H at least 1, and B at least 0; real constants 0 \(<{} \mu \leq\) L, 0 \(<\){} a \(\leq\) 1/2, and 0 \(<\){} q \(\leq\) 1. Each deterministic \(f_{i}\) on \(\mathbb R^{d}\) is differentiable, L-smooth, and \(\mu\)-strongly convex, and \(f=(1/n)sum_{i} f_{i}\) has the unique minimizer \(x\star\).

At round k, \(A_{k}=(a_{ij}^{k}\)) is row-stochastic and \(C_{k}=(c_{ij}^{k}\)) is column-stochastic, where entry (i,j) multiplies information sent from j to i. Define the active graph \(G_{k}\) so that, for i not equal to j, arc j\(\to\)i is present exactly when both \(a_{ij}^{k}\) and \(c_{ij}^{k}\) are positive. Thus the two operators have identical positive off-diagonal supports. Every diagonal and every positive off-diagonal entry of either operator is at least a. The union of every H consecutive active graphs is strongly connected. The operators may have different weights and need not be doubly stochastic.

For each active cross-arc j\(\to\)i at round k, sender j forms \(d_{ij},x^{k}\)=Q(\(x_{j}^{k}-s_{ij},x^{k}\)) and \(d_{ij},y^{k}\)=Q(\(y_{j}^{k}-s_{ij},y^{k}\)), immediately adds these increments to its corresponding sender memories, and enqueues one timestamped pair. Every enqueued message is delivered exactly once after an integer delay in {[}0,B{]}, independently of whether its link remains active, through a per-link FIFO channel. At each round, messages are formed first; all messages due that round are then delivered in timestamp order and added to the receiver memories before the state update. A duplicate or timestamp no newer than the last accepted timestamp is ignored. There are no losses, and sender memories on inactive links remain unchanged.

One fixed deterministic compressor Q with Q(0)=0 is used throughout a run. Its intrinsic quality is \(q_{Q}\)=1-sup over nonzero v of ‖Q(v)-v‖\(^{2}\)/‖v‖\(^{2}\), and admissibility requires \(q_{Q}\) in {[}q,1{]}. Quality is determined by this realized supremum rather than by a declared nonsharp label.

Choose arbitrary initial \(x_{i}^{0}\), set \(y_{i}^{0}=\nabla f_{i}(x_{i}^{0}\)), and initialize every cross-link sender and receiver reconstruction to zero, every last accepted timestamp to -1, and every queue to empty. For a constant \(\alpha\) in (0,1/L{]}, after round-k deliveries set \(r_{ii},x^{k}=x_{i}^{k}\) and \(r_{ii},y^{k}=y_{i}^{k}\), then update \(x_{i}^{k+1}=\sum_{j} a_{ij}^{k} r_{ij},x^{k}-\alpha y_{i}^{k}\) and \(y_{i}^{k+1}=\sum_{j} c_{ij}^{k} r_{ij},y^{k}+\nabla f_{i}(x_{i}^{k+1})-\nabla f_{i}(x_{i}^{k}\)). For fixed n and B the stored state is finite-dimensional; no bounded-trajectory premise is imposed.

Fix n=2, d=1, \(L=\mu\)=1, H=1, B=9, a=1/2, and q=1. These constraints force \(A_{k}=C_{k}=J=(1/2)11^{T}\) and Q(v)=v. Every cross-link message is assigned delay exactly nine. Quantify over all objectives \(f_{i}(x)=(x-b_{i})^{2/2}+c_{i}\), all initial primal states, and the initialization prescribed by A5.

Every cross-link message formed at round k is delivered at round k+9.

Set n=2, H=1, a=1/2, and q=1.

The target identifies the endpoint of the connected convergence interval issuing from \(\alpha\)=0 and makes no claim that instability persists at every larger stepsize.

In the two-agent scalar complete-averaging identity-compressed cell with every cross-link message delayed exactly nine rounds, prove or refute that the first loss of uniform exact convergence occurs at the explicit disagreement-mode value \(\alpha_{9}\) defined below.
\end{problem}

\clearpage
\section{P099: The sum problem in bidual defect one}
\label{problem:P099}

\begin{problem}
\(J_{X}(x)\) is the element of X** defined by \(J_{X}(x)\)(x\(\star) = <\) x, x\(\star>\).

(A+B)(y) = \{astar+bstar: astar in A(y), bstar in B(y)\}.

gra(A+B) = \{(y, astar+bstar): y in X, astar in A(y), bstar in B(y)\}.

(x, xstar) is monotonically related to gra(A+B) when \(<\) x-y, xstar-astar-bstar\(>{} \geq\) 0 for every y in X, astar in A(y), and bstar in B(y).

X is a real Banach space and its canonical image \(J_{X}(X)\) has codimension one in X\textbf{, i.e.~dim(X}/\(J_{X}(X)\)) = 1.

A and B are maximal monotone operators from X to subsets of \(X\star\).

dom(A) intersect int(dom(B)) is nonempty.

Universal over real Banach spaces whose canonical image has codimension one in the bidual and over all maximal monotone pairs satisfying the stated interior-domain condition; no full-domain, subdifferential, normal-cone, representability, or type assumption is added.

The interior in A3 is the norm interior in X, and all duality pairings are the canonical pairings.
A and B range over all maximal monotone operators allowed by A2 rather than only source-listed special classes.
A refutation must exhibit a point monotonically related to gra(A+B) but not in gra(A+B), not merely a failure of a proposed proof route.

Let X be a real Banach space with dim(X\(\star*/J_{X}(X)\)) = 1. Let A, B:X \(\to 2^{(X\star)}\) be maximal monotone and suppose dom(A) intersect int(dom(B)) is nonempty. Determine whether A+B is maximal monotone.
\end{problem}

\clearpage
\section{P100: The directed-cycle penalty}
\label{problem:P100}

\begin{problem}
Let \(D=\operatorname{diag}(1,2,1)\) and \(z_\star=0\).  A
tracker-consistent initial state satisfies \(y_0=Dx_0\).  For the trajectory
generated by \(I\) and \(\alpha>0\), define
\[
\begin{aligned}
E_k^\alpha
&=\left(\lVert x_k\rVert_2^2
  +\alpha^2\lVert y_k\rVert_2^2\right)^{1/2},\\
\mathcal R_k(I,\alpha)
&=\sup_{\substack{x_0\in\mathbb R^3,\ y_0=Dx_0\\E_0^\alpha>0}}
  \frac{E_k^\alpha}{E_0^\alpha}.
\end{aligned}
\]
Set
\[
\begin{aligned}
S(I)=\{\alpha>0:\;&
\sup_{k\geq0}\mathcal R_k(I,\alpha)<\infty,\\
&\lim_{k\to\infty}\mathcal R_k(I,\alpha)=0\}.
\end{aligned}
\]
Also set
\[
\begin{aligned}
T_{1/4}(I,\alpha)
&=\inf\left\{K\in\mathbb N_0:
\sup_{k\geq K}\mathcal R_k(I,\alpha)\leq\frac14\right\},\\
N_{\mathrm{CYCLE}}(\eta)
&=\sup_{I\in\mathcal D_{\mathrm{CYCLE}}(\eta)}
  \inf_{\alpha\in S(I)}T_{1/4}(I,\alpha).
\end{aligned}
\]
The last two quantities equal \(+\infty\) when their defining set is empty.
For \(I\in\mathcal D_{\mathrm{CYCLE}}(\eta)\), let
\[
M_j(\alpha)=
\begin{pmatrix}
A_j&-\alpha I_3\\
D(A_j-I_3)&B_j-\alpha D
\end{pmatrix}.
\]
The period matrix is
\[
P_I(\alpha)=M_1(\alpha)M_0(\alpha).
\]

The permitted initial states are (\(x_{0},Dx_{0}\)), and their period-boundary reachable space is the smallest \(P_{I}(\alpha\))-invariant subspace containing all such states.

There are n\(\geq\)3 agents with scalar objectives \(f_{i}(z)=h_{i} (z-r_{i})^{2/2}\), where 1\(\leq h_{i} \leq\)2. Their average has the unique minimizer \(z\star\), and \(\nabla\) F(x)=(\(h_{i}(x_{i}-r_{i}))_{i=1} ^{n}\).

For a constant \(\alpha >\)0, the algorithm is \(x_{k+1}=A_{k} x_{k}-\alpha y_{k}\) and \(y_{k+1}=B_{k} y_{k}+\nabla\) F(\(x_{k+1})-\nabla\) F(\(x_{k}\)), with tracker-consistent initialization \(y_{0}=\nabla\) F(\(x_{0}\)).

Each \(A_{k}\) is row-stochastic and each \(B_{k}\) is column-stochastic. Both have positive diagonal entries, and for i≠j, \((A_{k})_{ij} >\)0 if and only if \((B_{k})_{ij} >\)0 if and only if the directed edge j\(\to\)i is present at time k. Every positive entry of either matrix is at least η, where 0\(<\) η\(\leq\)1/2.

For a fixed integer H\(\geq\)1, the union of the directed graphs in every H consecutive times is strongly connected.

An admissible instance I consists of the objective data and matrix schedule satisfying A1-A4, but not an initialization. For each I and \(\alpha, E_{k}^{\alpha}, \mathbb R_{k}, T_{\varepsilon}\), S(I), and \(N_{\mathrm{GT}}\) are defined as in the claim shape, with \(\mathbb R_{k}\) taking the supremum over all tracker-consistent \(x_{0}\) having \(E_{0}^{\alpha} >\)0 and \(N_{\mathrm{GT}}\) taking the supremum over all admissible instances. An instance with S(I) empty makes \(N_{\mathrm{GT}}\) infinite. No finiteness or heterogeneity-control property of the chosen normalization is assumed.

All two-periodic row-stochastic and column-stochastic matrix pairs satisfying A1-A5 with h=(1,2,1), r=(0,0,0), and the exact phase supports. Every active edge weight and complementary diagonal entry is at least η, and the stepsize is optimized over the explicitly defined set S(I).

At even times the only off-diagonal edge is 1\(\to\)2; at odd times the only off-diagonal edges are 2\(\to\)3 and 3\(\to\)1. The matrix sequence repeats with period two.

Fix h=(1,2,1) and r=(0,0,0), so \(z\star\)=0 and tracker-consistent initialization is \(y_{0}\)=diag\((1,2,1)x_{0}\).

For each matrix instance, minimize \(T_{1/4}\)(I,\(\alpha\)) over every \(\alpha\) in S(I), using only tracker-consistent initial states.

Fix \(n=3\), \(H=2\), \(D=\operatorname{diag}(1,2,1)\), and
\(r=(0,0,0)\), with two-periodic supports
\[
E_0=\{1\to2\},
\qquad E_1=\{2\to3,3\to1\}.
\]
Determine whether there exist \(\eta_0\in(0,1/2]\) and constants
\(0<c_{\mathrm{lower}}\leq c_{\mathrm{upper}}<\infty\) such that
\[
c_{\mathrm{lower}}\eta^{-1}
\leq N_{\mathrm{CYCLE}}(\eta)
\leq c_{\mathrm{upper}}\eta^{-1}
\qquad (0<\eta\leq\eta_0).
\]
\end{problem}

\endgroup

\end{document}